\documentclass{article}
\usepackage[english]{babel}

\usepackage[nottoc]{tocbibind}

\usepackage[authoryear,round,longnamesfirst]{natbib}
\usepackage{animate}
\usepackage{fontawesome5}
\usepackage[toc]{appendix}
\usepackage{tocloft}
\usepackage{setspace}
\usepackage[table]{xcolor}
\usepackage{mdframed}
\definecolor{algobg}{RGB}{245,245,245}  

\usepackage{amsthm}
\usepackage{booktabs}
\usepackage{amsmath}
\usepackage{float}
\newtheorem{lemma}{Lemma}
\newtheorem{theorem}{Theorem}
\usepackage{amssymb}
\usepackage{vmargin}
\usepackage{graphicx}
\usepackage{subcaption}
\setmarginsrb{2.5 cm}{2.5 cm}{2.5cm}{2.5 cm}{.45 cm}{.45 cm}{.45 cm}{.45 cm}
\usepackage{minted}
\usepackage{caption}
\usepackage{float}

\newfloat{listing}{htbp}{lop}
\floatname{listing}{Listing}

\setminted{
    fontsize=\small,
    baselinestretch=1.1,
    linenos=true,
    breaklines=true,
    breakanywhere=true,
    frame=lines,
    framesep=2mm
}

\usepackage[hidelinks]{hyperref}

\title{Optimization and Regularization Under Arbitrary Objectives%
\thanks{For full functionality of the animated figures presented in this document, please view this PDF in \textbf{Adobe Acrobat Reader}. Other PDF viewers (including browser-based
viewers such as Chrome, Edge, and Preview on macOS) may display static images only.}}
\author{
  Jared N. Lakhani\\
  \textit{Department of Statistical Sciences, University of Cape Town}\\
  \texttt{lkhjar001@myuct.ac.za}
  \and
  Etienne Pienaar\\
  \textit{Department of Statistical Sciences, University of Cape Town}\\
  \texttt{etienne.pienaar@uct.ac.za}
}
\date{}
\newif\iffoo
\footrue

\begin{document}
\maketitle

\begin{abstract}
    This study investigates the limitations of applying Markov Chain Monte Carlo (MCMC) methods to arbitrary objective functions, focusing on a two-block MCMC framework which alternates between Metropolis-Hastings and Gibbs sampling. While such approaches are often considered advantageous for enabling data-driven regularization, we show that their performance critically depends on the sharpness of the employed likelihood form. By introducing a sharpness parameter and exploring alternative likelihood formulations proportional to the target objective function, we demonstrate how likelihood curvature governs both in-sample performance and the degree of regularization inferred by the training data. Empirical applications are conducted on reinforcement learning tasks: including a navigation problem and the game of tic-tac-toe. The study concludes with a separate analysis examining the implications of extreme likelihood sharpness on arbitrary objective functions stemming from the classic game of blackjack, where the first block of the two-block MCMC framework is replaced with an iterative optimization step. The resulting hybrid approach achieves performance nearly identical to the original MCMC framework, indicating that excessive likelihood sharpness effectively collapses posterior mass onto a single dominant mode.
    \end{abstract}

\section{Introduction}
In supervised learning regimes, modern machine learning models can be viewed as flexible classes of non-linear basis functions that are capable of approximating complex relationships between inputs and outputs \citep{vapnik1998statistical, hastie2009elements}. Neural networks, kernel methods, and other parametric architectures provide rich sets of possible input-output relationships within which a predictive mapping can be learned from data \citep{bishop2006pattern,shaleyshwartz2014understanding}. Training such models typically involves selecting a particular functional form from this class and estimating its parameters by minimizing an objective function that quantifies the discrepancy between model predictions and observed outcomes \citep{vapnik1991principles}. This objective is most commonly defined through a loss function, such as mean squared error or cross-entropy, and optimization is performed with respect to the model parameters using gradient-based methods.\\

However, the expressive power that makes modern machine learning models attractive also introduces significant statistical challenges. In particular, many commonly used models are overparameterized, meaning they contain far more parameters than are strictly required to interpolate the training data \citep{hastie2009elements}. While this overparameterization enables near-perfect in-sample fit, it simultaneously increases the risk of poor generalization to unseen data. As a result, controlling model complexity becomes essential. Regularization provides a principled mechanism for addressing this issue by augmenting the objective function with a penalty that discourages overly complex parameter configurations \citep{bishop2006pattern}. By constraining the effective capacity of the model, regularization enforces a trade-off between fitting the training data and achieving robust out-of-sample performance.
\\

In supervised learning settings, regularization is naturally integrated into gradient-based optimization frameworks, and its effects are well understood both from statistical learning theory and from a Bayesian perspective \citep{bishop2006pattern, hastie2009elements}. However, many learning problems of practical interest fall outside the supervised learning paradigm: in particular, reinforcement learning tasks, where agents learn by interacting with an environment and receive feedback in the form of rewards, rather than labeled input-output pairs \citep{sutton2018reinforcement}. In such settings, the objective function may be noisy, discontinuous, or non-differentiable with respect to the model parameters.  This motivates the notion of an arbitrary objective: an objective that need not be differentiable with respect to the environment or directly tied to observed data. It only requires that the objective function assigns high values to desirable behaviour and low values to undesirable behaviour. These arbitrary objectives render conventional gradient-based optimization techniques inapplicable or unreliable, motivating the use of alternative optimization strategies.
\\

One class of approaches capable of handling such arbitrary objectives is based on stochastic search and sampling methods. Evolutionary algorithms, such as genetic algorithms, explore the parameter space through population-based updates without relying on gradient information \citep{holland1975adaptation}. Similarly, Markov Chain Monte Carlo (MCMC) methods can be adapted for optimization by reformulating objective functions as probability distributions and sampling from said distributions, as in simulated annealing (SA) \citep*{kirkpatrick1983optimization, geyer1995annealing}. In particular, SA reformulates a cost function $C(x)$ into a Boltzmann-like distribution, $\pi_{\beta}(x) \propto \exp\!\left(-\beta C(x)\right)$, where $\beta = 1/T$ denotes the inverse temperature. At high temperature (small $\beta$), the Markov chain explores the state space broadly, whereas at low temperature (large $\beta$) it increasingly concentrates around global minima of the cost function.\\

Among sampling-based approaches, MCMC methods are often presented as appealing due to their flexibility and their supposed ability to allow regularization to be inferred from data \citep{gelman2006prior, robert2004monte, bishop2006pattern}. However, longstanding discussions in the MCMC and Bayesian computation literature emphasize that such interpretations rely critically on the probabilistic meaning of the target distribution and the manner in which likelihoods are specified or scaled \citep{geyer1995annealing, robert2004monte}. In particular, following the perspective articulated by \citet*{neal2001annealed}, when MCMC targets are not grounded in a well-defined generative model, changes to likelihood sharpness or scaling modify the distribution being sampled, making it unclear whether the procedure should be interpreted as posterior inference or as a form of stochastic optimization.\\

This study extends the discussion by examining the shortcomings of applying MCMC to arbitrary objectives, with particular emphasis on two-block MCMC (alternating Metropolis-Hastings (MH) and Gibbs sampling), a method often presented as advantageous for allowing the training set to infer regularization. Rather than simply exponentiating an arbitrary objective as in SA, the study investigates alternative likelihood formulations deliberately shaped to remain proportional to the arbitrary objective function. This shifts MCMC from a sampling paradigm towards a mode-seeking algorithm, paralleling the perspective of SA. In addition to altering the likelihood specification, the study explicitly modulates the sharpness of the likelihood form through a sharpness parameter. The analysis highlights that the sharpness of the likelihood form plays a decisive role in both determining in-sample performance and the strength of regularization inferred by the training set.\\

To illustrate these points, the study applies two-block MCMC to reinforcement learning tasks: specifically a navigation problem and tic-tac-toe. Beforehand, the study investigates the role of regularization on out-of-sample performance for solutions obtained from a genetic algorithm (GA), and additionally compares these results to solutions obtained by random search (RS).\\

The study concludes by illustrating the implications of increased likelihood sharpness, demonstrating this through reinforcement learning tasks associated with blackjack. Here, the study simplifies the two-block MCMC by replacing the first block with an iterative optimization procedure and compares this hybrid approach to the original scheme. The resulting near-identical performance demonstrates that increasing likelihood sharpness ultimately collapses posterior mass onto the dominant mode, echoing the observation of \cite{kirkpatrick1983optimization} that lower temperatures (that is, higher likelihood sharpness $\beta$) increasingly concentrate samples around the global minima of the cost function.\\

Furthermore, this study makes three primary contributions. First, it provides a conceptual clarification of how MCMC behaves when applied to arbitrary objectives, showing that such procedures should often be interpreted as stochastic optimization methods rather than as Bayesian posterior inference. Second, it demonstrates empirically that the sharpness of the likelihood formulation acts as an implicit regularization control, and directly influences both in-sample and out-of-sample performance. Thirdly, the study shows that increasing likelihood sharpness causes two-block MCMC to collapse toward deterministic, mode-seeking behaviour, echoing the mechanism underlying SA. Together, these findings clarify what MCMC-based procedures are effectively accomplishing in arbitrary objective settings. Additionally, the study also offers practical guidance on how to implement two-block MCMC effectively in these settings, where the study highlights the importance of adaptive MCMC procedures to improve mixing and stabilizing performance.

\section{Literature Review}
\subsection{Overparameterization and regularization}
Neural networks, along with various other machine learning architectures, constitute a class of overparameterized models: models with more parameters than training samples. As a consequence of this overparameterization, such models exhibit a propensity to achieve near-perfect fit on the training data yet require the incorporation of regularization techniques to ensure satisfactory generalization to previously unseen data. This concept of overparameterization can be formally understood through the Vapnik-Chervonenkis (VC) Generalization Bound, which relates in-sample and out-of-sample errors. With probability $1- \delta$ we have: $E_{out}(g) \leq E_{in}(g) + \sqrt{\frac{8}{N}\ln\left(\frac{4(2N)^{d_{VC}}+1}{\delta}\right)}$ \footnote{$E_{out}$ and $E_{in}$ denote the out-of-sample and in-sample errors respectively, $g$ denotes the final chosen hypothesis function, $N$ denotes the number of samples and $\delta$ denotes the tolerance.}, wherein the VC dimension $d_{VC}$ serves as a measure of model complexity. Specifically, an increase in the complexity of the hypothesis space corresponds to a higher VC dimension, which in turn, induces an expansion of the generalization gap thereby increasing the risk of overfitting, as stated in \cite{vapnik1991principles}.\\

Regularization, as formally expounded in \cite{hastie2009elements}, is a fundamental technique in statistical learning that introduces a penalty term to the model’s objective function, thereby discouraging excessive model complexity. One widely adopted regularization approach is L2 regularization, also known as ridge regularization, wherein the squared magnitude of the model coefficients (or weights) is penalized. Unlike L1 regularization, which promotes sparsity by driving certain coefficients exactly to zero, L2 regularization instead constrains the magnitude of all parameters, thereby ensuring a controlled reduction in model complexity without entirely eliminating any particular parameter. In the context of fitting a single-ouptut neural network, if we wanted to minimize the mean square error objective of $\sum_{i = 1}^N (y_i - a_1(i)^L)^2$ \footnote{Here, $y_i$ denotes the $i^{th}$ outcome of the $N$ observations, where $a_1(i)^L$ represents the sole node of the $i^{th}$ observation in the output layer $L$. Full neural net architecture is described in \ref{app: NN}.}, we would include the constraint $\sum_{j, k, l} (w_{kj}^l)^2 \leq \tau$. \footnote{Here, $w_{kj}^l$ denotes the $kj$-th weight linking the $k$-th node in layer $l$ - 1 and the $j$-th node in layer $l$.}, where $\tau$ would act as a means to constrain the parameters of the model to be closer to zero, thus lessening model complexity. Furthermore, the model fitting procedure becomes a constrained optimization problem for which we can re-write the penalized objective as:
\begin{equation}
   \sum_{i = 1}^N (y_i - a_1(i)^L)^2 + \nu \sum_{j, k, l} (w_{kj}^l)^2, \label{L2 Loss}
\end{equation}
whereby increasing our Lagrangian multiplier $\nu$ would be equivalent to decreasing $\tau$, thereby enforcing a stricter penalty on the model parameters.
\\

Given that regularization introduces an additional penalty term to the objective function, the resulting optimization problem remains differentiable with respect to the model parameters, thereby necessitating the use of gradient-based optimization techniques such as gradient descent. Specifically, gradient descent iteratively updates the model parameters in the direction of the negative gradient of the objective function, ensuring convergence to an optimal solution under appropriate step-size selection and convexity conditions, as described in \cite{amari1993backpropagation}. Furthermore, the choice of regularization strength, governed by the parameter $\nu$, critically influences the trade-off between model complexity and generalization. As such, determining an optimal value for $\nu$ requires an empirical approach, typically employing a validation set to assess model performance under varying degrees of regularization. This validation set approach, entails training the model with multiple candidate values of $\nu$ and subsequently evaluating its predictive performance on held-out validation data. The value of $\nu$ that minimizes the validation error is then selected, ensuring that the model is sufficiently complex to replicate nuances of the underlying pattern in the data, but also not too complex such that the model simply ``recalls'' what has been seen in the data. \\

From a Bayesian perspective, regularization can be interpreted as placing prior distributions on the parameter space of an overparameterized model. Specifically, L2 regularization corresponds to assuming a normal (Gaussian) prior on the parameters. 
\begin{proof}
We assume our observed response $y_i \vert \mathbf{w}, \mathbf{a}(i)^0 \sim \mathcal{N}(a_1(i)^L, \sigma_y^2)$ and are independent from all other $N$ observations. That is, we assume our observations follow a Gaussian distribution with mean equal to the sole node of the $i^{th}$ observation in the output layer $L$, with some variance $\sigma_y^2$. Here, the vector $\mathbf{w}$ denotes the $R$ number of weights and biases of a single-output neural network, with vector of inputs $\mathbf{a}(i)^0$ denoting the input nodes (on the $0^{th}$ layer) for the $i^{th}$ observation.\\

Furthermore, we assume $\mathbf{w} \sim \mathcal{N}(\mathbf{0}, \sigma_w^2\mathbf{I}_R)$. That is, the vector of $R$ independent weights and biases is multivariate Gaussian distributed with mean of $\mathbf{0}$ with some covariance matrix $\sigma_w^2 \mathbf{I}_R$. \\

Using Bayes rule we have:\\
\begin{align*}
    p(\mathbf{w} \mid \mathcal{D}) &= \frac{p(\mathcal{D} \mid \mathbf{w}) p(\mathbf{w})}{p(\mathcal{D})} \nonumber \\
    &\propto p(\mathcal{D} \mid \mathbf{w}) p(\mathbf{w}) \\
    & \propto \left[\prod_{i = 1}^N \mathcal{N}(y_i; \mathbf{a}(i)^1, \sigma_y^2) \right]\mathcal{N}(\mathbf{w}; \mathbf{0}, \sigma_w^2 \mathbf{I}) \\
    & \propto \prod_{i = 1}^N \mathcal{N}(y_i; \mathbf{a}(i)^1, \sigma_y^2) \prod_{i = 1}^R\mathcal{N}(w_i; 0, \sigma_w^2),\nonumber
\end{align*}
where, given a dataset $\mathcal{D}$, $p(\mathcal{D} \mid \mathbf{w})$ is the likelihood, representing how well the parameters explain the observed data; $p(\mathbf{w})$ is the prior, encoding our beliefs about the parameters before seeing the data; and $p(\mathbf{w} \mid \mathcal{D})$ is the parameter posterior, the updated distribution of the parameters after observing the data.\\

Now taking the negative log probability of the parameter posterior:
\begin{align*}
    -\text{log}[ p(\mathbf{w} \mid \mathcal{D}) ] &\propto -\sum_{i = 1}^N \text{log} \left[\mathcal{N}(y_i; \mathbf{a}(i)^1, \sigma_y^2)\right] -\sum_{i = 1}^R \text{log} \left[\mathcal{N}(w_i; 0, \sigma_w^2)\right] \nonumber \\
    & \propto \frac{1}{2\sigma_y^2}\sum_{i = 1}^N \left(y_i - a_1(i)^L\right)^2 + \frac{1}{2\sigma_w^2} \sum_{i = 1}^R w_i^2 \nonumber \\
    & \propto\sum_{i = 1}^N \left(y_i - a_1(i)^L\right)^2 + \nu \sum_{i = 1}^R w_i^2, 
\end{align*}
whereby $\nu$ controls the strength of the regularization as in the L2 penalized loss function in Equation \ref{L2 Loss}. 
\end{proof}
Clearly, minimizing the L2 penalized loss function is equivalent to maximizing the posterior distribution of the parameters under a Gaussian prior, which corresponds to Maximum a Posteriori (MAP) estimation.

\subsection{Optimization beyond supervised learning}
In the context of reinforcement learning (RL), an agent learns to perform tasks by interacting with an environment, making sequential decisions, and receiving feedback in the form of rewards, as described in \cite{StanfordCS234LectureNotes}. The objective function, in this case, is to maximize the cumulative reward of the agent within the environment. Unlike supervised learning, gradients cannot be computed directly with respect to labeled data, as RL models are not trained on explicit input-output pairs. As a result, the objective is generally not expressed as a differentiable loss function evaluated on a fixed dataset, and gradients cannot be obtained by straightforward backpropagation through data. This motivates the notion of an arbitrary objective: an objective that need not be differentiable with respect to the environment or directly tied to observed data. It only requires that the objective function assigns high values to desirable behaviour and low values to undesirable behaviour.\\

Although reinforcement learning objectives are generally arbitrary and not derived from likelihood-based models, there exist special cases in which such objectives can be expressed as expectations with respect to a parameterized probability distribution. When an objective takes the form\footnote{Here, $f(x)$ denotes a reward function and $p_\theta(x)$ is a distribution over outcomes induced by parameters $\theta$.}
\[
J(\theta) = \mathbb{E}_{x \sim p_\theta(x)}[f(x)],
\]
gradient-based optimization becomes feasible. In particular, the likelihood-ratio identity yields
\[
\nabla_\theta J(\theta)
= \mathbb{E}_{x \sim p_\theta(x)}\!\left[f(x)\,\nabla_\theta \log p_\theta(x)\right],
\]
allowing gradients of the expected objective to be computed without requiring differentiability of the reward function. This result enables the use of gradient-based optimization methods even when the underlying objective is non-supervised, noisy, or non-differentiable with respect to the environment.\\

When arbitrary objectives cannot be expressed in a form amenable to gradient estimation, rendering standard gradient-based optimization inapplicable, alternative optimization methods capable of solving arbitrary objectives must be employed. Genetic algorithms (GAs) are a prominent class of such methods. They simulate the evolutionary process by initiating a population of random candidate solutions and employing selection mechanisms to iteratively recombine and mutate these candidates. This process yields new populations with potentially enhanced fitness (objective function values) relative to previous populations. Now given sufficient iterations, this process systematically favours traits (parameter values) that should enhance performance in achieving the objective function. As a result, we obtain an evolutionary algorithm in which the population of solutions gradually evolves toward an optimal solution.\\

While GAs provide a biologically inspired mechanism for navigating complex search spaces, it is important to contextualize their performance against simpler baseline strategies. In particular, random search (RS) offers a natural point of comparison due to its algorithmic simplicity and lack of heuristic bias. RS operates by sampling candidate solutions uniformly at random from the feasible domain, evaluating their objective values, and retaining the best-performing solution observed over a fixed number of trials. Despite its simplicity and lack of adaptive guidance, RS has demonstrated competitive performance in a range of settings, particularly when the objective landscape is noisy, discontinuous, or lacks exploitable structure as detailed in \cite{bergstra2012random}. This makes RS an appropriate benchmark for assessing whether a GA is truly needed, or if a GA's performance gains are only marginal relative to an unguided search.

\subsection{Regularization in reinforcement learning}
While regularization is well understood and formally grounded in supervised learning, its role in RL and other arbitrary-objective settings is considerably less explicit. In RL, learning proceeds through interaction with an environment rather than from fixed input-output pairs, and the objective is typically defined as the maximization of cumulative reward. As a result, standard notions of generalization and overfitting are less directly applicable, and regularization is often introduced implicitly or heuristically rather than derived from a probabilistic framework.\\

One class of approaches introduces regularization explicitly at the parameter level. For example, weight decay (e.g., L$2$ regularization) is commonly applied to arbitrary objectives as a means of controlling model complexity. In this setting, the regularization parameter $\lambda$ is not inferred from the data, but is instead treated as a fixed hyperparameter chosen externally by the practitioner. Typical selection procedures include heuristic tuning, grid search, or adherence to values used in prior studies as expounded in \citet*{mnih2016asynchronous} and \citet*{henderson2018deep}. As noted by \cite{henderson2018deep}, empirical performance in deep reinforcement learning can be highly sensitive to such implementation choices, including the strength of L$2$ regularization. \\

In cases where the arbitrary objective, stemming from a particular RL task, can be expressed in a form amenable to gradient estimation\footnote{Entropy regularization only becomes operational when gradient information is available, as its effect relies on computing how the entropy of the policy changes with respect to the parameters $\theta$. Without access to $\nabla_\theta \mathcal{H}(\pi_\theta(\cdot \mid s))$, the regularization term cannot provide directional guidance for optimization and thus cannot be meaningfully enforced.}, entropy regularization is commonly used to encourage stochastic policies and promote exploration as stated in \cite{williams1991function} and \cite{mnih2016asynchronous}. Entropy regularization augments the reinforcement learning objective $J(\theta)$ by adding an expected entropy term\footnote{Let $\pi_\theta(a \mid s)$ denote a parameterized policy, which assigns a probability to each action $a$ given a state $s$, where $\theta$ represents the policy parameters. The entropy of the policy at state $s$ is defined as
\[
\mathcal{H}(\pi_\theta(\cdot \mid s)) = - \sum_a \pi_\theta(a \mid s)\,\log \pi_\theta(a \mid s),
\]
which quantifies the uncertainty of the action distribution.},
\[
J_{\text{reg}}(\theta)
= J(\theta) + \lambda\,\mathbb{E}_s\!\left[\mathcal{H}(\pi_\theta(\cdot \mid s))\right],
\]
where $\lambda > 0$ controls the strength of the regularization. This modification biases learning toward policies that retain randomness, discouraging premature convergence to deterministic behavior and smoothing the optimization landscape.\\

Closely related are trust-region methods, which impose a Kullback-Leibler (KL) divergence penalty\footnote{The KL divergence is \[
\mathrm{KL}\!\left(\pi_{\theta_{\text{old}}}(\cdot \mid s)\,\|\,\pi_{\theta}(\cdot \mid s)\right)
= \sum_{a} \pi_{\theta_{\text{old}}}(a \mid s)\,
\log \frac{\pi_{\theta_{\text{old}}}(a \mid s)}{\pi_{\theta}(a \mid s)} .
\]
} between successive policy updates, effectively regularizing the learning process by restricting abrupt changes in policy space. In particular, Trust Region Policy Optimization (TRPO) enforces an explicit constraint on the KL divergence between the old and updated policies, ensuring that each update remains within a local trust region and preventing overly aggressive changes in policy space as conducted in \cite{schulman2015trust}. Proximal Policy Optimization (PPO) adopts a computationally simpler approximation to this idea, replacing the hard KL constraint with a clipped surrogate objective or an adaptive KL penalty that discourages large deviations from the previous policy as carried out in \cite{schulman2017proximal}. In both cases, the KL divergence acts as a regularizer that stabilizes learning by restricting abrupt policy updates. While these techniques improve stability and empirical performance, the strength of regularization is again, externally specified and not inferred from the data.
\\

A second strand of work introduces regularization implicitly through randomness in the learning process rather than through explicit penalty terms. In this setting, stochasticity is used to stabilize learning and prevent overly confident policies. For example, \cite{gal2016dropout} show that dropout acts as a form of regularization by randomly disabling neurons during training. By temporarily ignoring all weights connected to the dropped neurons, dropout prevents the network from relying too heavily on any single set of parameters, thereby helping to control overfitting. Another example of stochastic regularization is provided by \cite{fortunato2017noisy}, who add small random perturbations directly to the neural network parameters, producing a slightly modified but consistent policy that improves exploration. \\

Additionally, during genetic algorithm optimization, stochastic regularization is enforced by evaluating candidate solutions under randomized environments: whereby the environment configuration is resampled at each iteration to promote robustness and reduce overfitting to any single environment realization, as demonstrated in \cite{tobin2017domain} and \cite{peng2018sim}. This variability discourages solutions that exploit specific environment realizations and instead favours parameter configurations that perform robustly across perturbations, thereby acting as an implicit form of regularization, as discussed in \cite{hansen2021towards}. In all these implicit approaches, regularization arises as a side effect of stochasticity rather than being explicitly imposed or inferred from data.

\subsection{MCMC for optimization} \label{sec: mcmc for optimization}

In the framework of Bayesian Neural Networks, the parameters of the neural network are treated as random variables rather than fixed values. This means they follow a probability distribution, reflecting our uncertainty about their true values. Given a dataset $\mathcal{D}$, our goal is to infer the posterior distribution over the parameters $\boldsymbol{\theta}$, which is given by Bayes' theorem: $p(\boldsymbol{\theta} \mid \mathcal{D})  \propto p(\mathcal{D} \mid \boldsymbol{\theta}) p(\boldsymbol{\theta})$. Since the parameter posterior is typically intractable due to high-dimensional parameter spaces, that is, it either has a complex or unknown form, as stated in \cite{dobson2018introduction}, MCMC methods may be employed to approximate it. MCMC generates samples from the parameter posterior by constructing a Markov chain whose stationary distribution is the true posterior. In this way, the utilization of MCMC aims to provide the entire posterior distribution $p(\boldsymbol{\theta} \mid \mathcal{D})$ and not just the MAP estimate; corresponding to $\operatorname{argmax}_{\boldsymbol{\theta}}p(\boldsymbol{\theta}\mid \mathcal{D})$ which is the mode of the posterior. According to \cite{dobson2018introduction}, Markov chains simplify complex problems since the next sample in the chain depends solely on the previous sample: $p(\boldsymbol{\theta}^{(j)} = \mathbf{a} \vert \boldsymbol{\theta}^{(j-1)}, \boldsymbol{\theta}^{(j-2)}, \ldots, \boldsymbol{\theta}^{(0)}) = p(\boldsymbol{\theta}^{(j)} = \mathbf{a} \vert \boldsymbol{\theta}^{(j-1)})$.\\

One of the earliest approaches to applying MCMC to arbitrary objective functions was achieved through SA, where one reformulates optimization as sampling from a distribution where good solutions have high probability. That is, the method reformulates a cost function $C(x)$ into a Boltzmann-like distribution, $\pi_{\beta}(x) \propto \exp\!\left(-\beta C(x)\right)$, where $\beta = 1/T$ denotes the inverse temperature. At high temperature (small $\beta$), the Markov chain explores the state space broadly, whereas at low temperature (large $\beta$) it increasingly concentrates around global minima of the cost function. The algorithm proceeds by running a standard MH sampler targeting $\pi_{\beta}$ and gradually increasing $\beta$ according to a cooling schedule. In the limit as $T \to 0$, the chain places its mass on the global minima of $C(x)$. This approach was popularized in physics and combinatorial optimization by \cite{kirkpatrick1983optimization} and rigorously studied in the context of Bayesian image analysis by \cite{geman1984stochastic}, who also established theoretical convergence guarantees under logarithmic cooling.\\

Subsequent work generalized SA by embedding temperature schedules more explicitly within MCMC frameworks to improve exploration of complex, multimodal objective landscapes. Parallel tempering, also known as Metropolis-coupled MCMC, runs multiple Markov chains in parallel at different temperatures, each targeting a tempered version of the objective-induced distribution, and periodically proposes swaps between chains, as in \cite{geyer1991markov}. This mechanism allows high-temperature chains to explore broadly while low-temperature chains concentrate probability mass near optimal solutions, thereby mitigating premature convergence to local optima. Closely related approaches include simulated tempering, which treats the temperature itself as a stochastic variable within the Markov chain, as carried out by \cite{marinari1992simulated}. Additionally, annealed MCMC schemes formalize temperature schedules into the sampling procedure, causing the Markov chain to progressively focus on high-quality solutions, as conducted by \cite{geyer1995annealing}. Collectively, these methods retain the core idea of optimization via sampling from a sequence of tempered distributions whose mass progressively concentrates on optimal or near-optimal solutions as the temperature is lowered.

\subsection{Two-block MCMC methods}
Two-block MCMC methods are widely used in hierarchical Bayesian models to jointly infer model parameters and regularization-related hyperparameters from data. In this framework, the unknowns are partitioned into two blocks: typically the primary model parameters and a set of latent variables or hyperparameters governing prior strength, noise variance, or smoothness. Conditional on one block, the other can often be sampled efficiently, leading to alternating Gibbs or Metropolis-Hastings updates. This approach has been extensively applied in Bayesian regression and inverse problems, where regularization parameters such as variance components are treated as random variables and inferred from the data rather than fixed \textit{a priori} \citep{gelman2006prior, robert2007bayesian}. For example, Hierarchical Gaussian models commonly use two-block MCMC to alternate between sampling regression coefficients and their associated precision parameters, allowing the data to determine the effective strength of shrinkage, as described in \cite{banerjee2014hierarchical}. Similar two-block MCMC schemes also appear in Bayesian variable selection, where latent inclusion indicators and regression coefficients are sampled in separate blocks, allowing model complexity, measured by the number of active predictors, to be determined by the data rather than fixed \textit{a priori}, as proposed by \cite{george1997approaches}.\\

Two-block MCMC methods have also played a central role in Bayesian image analysis and inverse problems, where regularization is essential to counteract ill-posedness. In these settings, one block typically consists of the latent signal or image, while the second block contains hyperparameters controlling smoothness or noise precision. \cite{geman1984stochastic} popularized this approach in Bayesian image restoration, using alternating updates to infer both the image and its regularization strength. Subsequent work in inverse problems has employed similar schemes to infer spatial smoothness and noise levels directly from data, as in \cite{kaipio2005statistical}. Importantly, several authors have noted that as likelihoods become increasingly sharp, two-block MCMC transitions from genuine posterior sampling toward mode-seeking behaviour, closely resembling SA, as explained by \cite{neal2001annealed}. This observation highlights that while two-block MCMC is often framed as fully Bayesian inference, its practical behavior, and the degree to which regularization is truly inferred, depends critically on the relative concentration of the likelihood and prior.

\section{Technical Preliminaries}
This section marks the beginning of the methodological framework employed in this study. In particular, we detail one of the fundamental MCMC algorithms: the Metropolis-Hastings (MH) algorithm as illustrated by \cite{hastings1970monte} and \cite{metropolis1953equation}. The section also presents the adaptive MCMC procedures employed in this study which improve chain mixing and overall sampling efficiency.
\subsection{Metropolis-Hastings}
Given the current state of $\boldsymbol{\theta} \in \mathbb{R}^S$, that is $\boldsymbol{\theta}^{(j)}$, the MH algorithm proposes a new value $\boldsymbol{\theta}^*$ obtained from $\boldsymbol{\theta}^* = \boldsymbol{\theta}^{(j)} + \mathbf{Q}$. Subsequently, $\boldsymbol{\theta}^*$ is accepted as the new value in the Markov chain under the following acceptance criterion:
\begin{align}
\boldsymbol{\theta}^{(j+1)} =
\begin{cases}
\boldsymbol{\theta}^*, & \text{if } U < \alpha, \\
\boldsymbol{\theta}^{(j)}, & \text{otherwise.}
\end{cases} \label{eq: compare to U}
\end{align}
Now the vector $\mathbf{Q}$ denotes drawn values from a proposal density (usually $\mathbf{Q}\sim \mathcal{N}(\mathbf{0}, \sigma_{Q}^2 \mathbf{I}_S)$), and $U$ is a drawn value from a uniform distribution between 0 and 1, that is $U \sim \mathcal{U}(0,1)$. Furthermore, $\alpha$ is the acceptance probability given by:
\begin{align}
    \alpha &= \min \left( 
    \frac{p(\boldsymbol{\theta}^*\mid \mathcal{D})}{p(\boldsymbol{\theta}^{(j)}\mid \mathcal{D})} \cdot 
    \frac{Q(\boldsymbol{\theta}^{(j)}\mid \boldsymbol{\theta}^* )}{Q(\boldsymbol{\theta}^{*}\mid \boldsymbol{\theta}^{(j)})}, 1 
    \right) \nonumber \\
    &= \min \left( 
    \frac{p(\mathcal{D} \mid \boldsymbol{\theta}^*) p(\boldsymbol{\theta}^*)}{p(\mathcal{D} \mid 
    \boldsymbol{\theta}^{(j)}) p(\boldsymbol{\theta}^{(j)})} \cdot 
    \frac{Q(\boldsymbol{\theta}^{(j)}\mid \boldsymbol{\theta}^* )}{Q(\boldsymbol{\theta}^{*}\mid \boldsymbol{\theta}^{(j)})}, 1 
    \right). \nonumber
\end{align}
If the proposal density is symmetric, then $\alpha$ simplifies to
\begin{align}
    \alpha &=  \min \left( 
    \frac{p(\mathcal{D} \mid \boldsymbol{\theta}^*) p(\boldsymbol{\theta}^*)}{p(\mathcal{D} \mid 
    \boldsymbol{\theta}^{(j)}) p(\boldsymbol{\theta}^{(j)})}, 1 
    \right).  \label{eq: alpha}
\end{align}
For the present study we impose symmetry by assuming Gaussian priors for our parameters $\boldsymbol{\theta} \in \mathbb{R}^S$, that is, $\boldsymbol{\theta} \sim \mathcal{N}(\mathbf{0}, \sigma_{\theta}^2 \mathbf{I}_S)$ (as elaborated in Section \ref{sec: the obj}). Here, the likelihood ratio $\frac{p(\mathcal{D} \mid \boldsymbol{\theta}^*)}{p(\mathcal{D} \mid \boldsymbol{\theta}^{(j)})}$ reflects how much more likely the proposed parameter $\boldsymbol{\theta}^*$ is, compared to the current parameter $\boldsymbol{\theta}^{(j)}$, in explaining the observed data $\mathcal{D}$. Since the likelihood function quantifies the plausibility of the data given $\boldsymbol{\theta}$, the MH algorithm accepts proposed moves with higher likelihoods more readily, while still allowing occasional transitions to lower-likelihood regions to ensure proper exploration of the posterior $p( \boldsymbol{\theta} \mid \mathcal{D} )$. Additionally, the prior ratio $\frac{p( \boldsymbol{\theta}^*)}{p(\boldsymbol{\theta}^{(j)})}$ contributes to exploration by favouring moves toward regions of the parameter space that are more consistent with prior beliefs, especially when the likelihood offers little guidance such as being relatively flat. Furthermore, one may view the the prior ratio as a means to ensure acceptance is not solely driven by proposals which increase the likelihood. In terms of terminology used later, we refer to a likelihood-driven sampler as one in which the likelihood ratio dominates the prior ratio in the acceptance probability expression of Equation \ref{eq: alpha}.

\subsubsection{\texorpdfstring{$\sigma_\theta^2$}{sigma theta squared}: A parameter with a hyperprior} \label{sec: 2-sample}
We now assume the variance of the prior of $\boldsymbol{\theta}$, $\sigma_\theta^2$, to not be fixed, but rather having its own distribution. Instead of sampling the entire parameter vector $\boldsymbol{\Lambda} = \left[\boldsymbol{\theta}', \sigma_\theta^2 \right]' \in \mathbb{R}^{S+1}$ in a joint MH framework (as illustrated in \ref{app:JointMH}), we split the parameters into groups (or blocks) and sample each block conditionally on the others, leveraging their conditional distributions. Hence, under the two-block MCMC framework, the algorithm effectively samples from the joint posterior $p(\boldsymbol{\theta}, \sigma_\theta^2 \mid \mathcal{D})$ by alternately drawing from the conditionals $p(\boldsymbol{\theta}\mid   \sigma_\theta^2, \mathcal{D})$ and $ p(\sigma_\theta^2 \mid \mathcal{D}, \boldsymbol{\theta})$. To reiterate, we know that since $ p(\boldsymbol{\theta}\mid  \sigma_\theta^2, \mathcal{D}) = \frac{p(\boldsymbol{\theta},  \sigma_\theta^2\mid \mathcal{D})}{p( \sigma_\theta^2\mid \mathcal{D})}$, when sampling $\boldsymbol{\theta}$ from $p(\boldsymbol{\theta} \mid \sigma_\theta^2, \mathcal{D})$, $\sigma_\theta^2$ is treated as fixed. This is because it is conditioned on the value sampled in the previous step of the sampler, which results in $p(\boldsymbol{\theta}\mid \sigma_\theta^2, \mathcal{D})  \propto p(\boldsymbol{\theta},  \sigma_\theta^2\mid \mathcal{D})$. The same logic may be used to conclude $p(\sigma_\theta^2\mid\boldsymbol{\theta},  \mathcal{D})  \propto p(\boldsymbol{\theta},  \sigma_\theta^2\mid \mathcal{D})$, since when sampling from $p(\sigma_\theta^2\mid\boldsymbol{\theta},  \mathcal{D})$, $\boldsymbol{\theta}$ is treated as fixed. The two-block MCMC framework can simplify the sampling process, particularly when the joint proposal distribution is complex or high-dimensional. Additionally, it can improve mixing in cases where there is strong posterior correlation between $\boldsymbol{\theta}$ and $\sigma_\theta^2$, which can hinder the efficiency of joint updates. In this context, adequate mixing refers to the sampler's ability to explore the parameter space effectively: typically indicated by reduced autocorrelation between successive samples (for example, between $\boldsymbol{\theta}^{(j)}$ and $\boldsymbol{\theta}^{(j+1)}$ for all $j$), thereby promoting more reliable convergence to the stationary distribution. We outline the algorithm for the two alternating blocks as follows:

\noindent
\begin{mdframed}[backgroundcolor=algobg, linecolor=gray!40, linewidth=0.6pt]

\textbf{Block 1}: Sampling $\boldsymbol{\theta} \mid \sigma_\theta^2, \mathcal{D}$  
We assume the prior $\boldsymbol{\theta} \mid \sigma_\theta^2 \sim \mathcal{N}(\mathbf{0}, \sigma_\theta^2 \mathbf{I}_S)$ and a symmetric proposal density $\mathbf{Q}_\theta \sim \mathcal{N}(\boldsymbol{\theta}^{(j)}, \sigma_{Q_\theta}^2 \mathbf{I}_S)$. Hence, the acceptance probability for the first block is:
\begin{align*}
    \alpha_\theta &= \min \left( 
    \frac{p\left(\boldsymbol{\theta}^{*}\mid \left(\sigma_\theta^2\right)^{(j)},  \mathcal{D}\right)}{p\left(\boldsymbol{\theta}^{(j)}\mid\left(\sigma_\theta^2\right)^{(j)},  \mathcal{D}\right)} \cdot 
    \frac{Q(\boldsymbol{\theta}^{(j) }\mid \boldsymbol{\theta}^{*} )}{Q\left(\boldsymbol{\theta}^{*}\mid \boldsymbol{\theta}^{(j)}\right)}, 1 
    \right) \nonumber \\
    &= \min \left( 
    \frac{p\left(\mathcal{D} \mid \boldsymbol{\theta}^{*},\left(\sigma_\theta^2\right)^{(j)}\right) p\left(\boldsymbol{\theta}^{*}\mid \left(\sigma_\theta^2\right)^{(j)}\right)}{p(\mathcal{D} \mid 
    \boldsymbol{\theta}^{(j)}, \left(\sigma_\theta^2\right)^{(j)}) p\left(\boldsymbol{\theta}^{(j)}\mid \left(\sigma_\theta^2\right)^{(j)}\right)} \cdot 
    \frac{Q\left(\boldsymbol{\theta}^{(j)}\mid \boldsymbol{\theta}^{*} \right)}{Q\left(\boldsymbol{\theta}^{*}\mid \boldsymbol{\theta}^{(j)}\right)}, 1 
    \right) &\text{Bayes Theorem} \nonumber \\
    &= \min \left( 
    \frac{p\left(\mathcal{D} \mid \boldsymbol{\theta}^{*}\right) p\left(\boldsymbol{\theta}^{*}\mid\left(\sigma_\theta^2\right)^{(j)}\right)}{p\left(\mathcal{D} \mid 
    \boldsymbol{\theta}^{(j)}\right) p\left(\boldsymbol{\theta}^{(j)}\mid \left(\sigma_\theta^2\right)^{(j)}\right)}, 1 
    \right) &\text{Likelihood not dependent on $\sigma_\theta^2$} \\
    & = \min \left( 
    \frac{p\left(\mathcal{D} \mid \boldsymbol{\theta}^{*}\right) \cdot \frac{1}{ \sqrt{ \left(2\pi \left(\sigma_{\theta}^2\right)^{(j)} \right)^S}} \text{exp}\left( -\frac{1}{2\left(\sigma_{\theta}^2\right)^{(j)}} \lVert \boldsymbol{\theta}^* \rVert^2 \right)} {p\left(\mathcal{D} \mid 
    \boldsymbol{\theta}^{(j)}\right)  \cdot \frac{1}{\sqrt{\left(2\pi \left(\sigma_{\theta}^2\right)^{(j)}\right)^S}} \text{exp}\left( -\frac{1}{2\left(\sigma_{\theta}^2\right)^{(j)}} \lVert \boldsymbol{\theta}^{(j)} \rVert^2 \right)}, 1 
    \right). \nonumber
\end{align*}
Hence, taking the $\log$, we obtain:
\begin{align}
\log \left(\alpha_\theta\right) &= \min \Bigg( \log\left( p\left(\mathcal{D} \mid \boldsymbol{\theta}^{*}\right) \right) - \frac{1}{2\left(\sigma_\theta^2\right)^{(j)}} \lVert \boldsymbol{\theta}^* \rVert^2 - \log \left(p\left(\mathcal{D} \mid \boldsymbol{\theta}^{(j)}\right)\right) + \frac{1}{2\left(\sigma_\theta^2\right)^{(j)}}  \lVert \boldsymbol{\theta}^{(j)} \rVert^2, 0 \Bigg), \label{eq: log alpha}
\end{align}
after which, using Equation \ref{eq: compare to U} to ascertain $\boldsymbol{\theta}^{(j+1)}$.
\end{mdframed}

\begin{mdframed}[backgroundcolor=algobg, linecolor=gray!40, linewidth=0.6pt]
\textbf{Block 2: Sampling $\sigma_\theta^2 \mid \boldsymbol{\theta}, \mathcal{D}$}  
We assume the hyperprior $\sigma_\theta^2 \sim \text{Inv-Gamma}(a, b)$. Since $\sigma_\theta^2$ does not appear in the likelihood, the data $\mathcal{D}$ provides no additional information about $\sigma_\theta^2$ beyond what $\boldsymbol{\theta}$ already does. Hence, conditioning on $\mathcal{D}$ does not change the distribution of $\sigma_\theta^2 \mid \boldsymbol{\theta}$:
\begin{align*}
   p(\sigma_\theta^2 \mid \boldsymbol{\theta}, \mathcal{D}) & \propto p(\mathcal{D} \mid \boldsymbol{\theta}, \sigma_\theta^2 )\cdot p(\sigma_\theta^2 \mid \boldsymbol{\theta}) \nonumber \\
    & = p(\mathcal{D} \mid \boldsymbol{\theta} )\cdot p(\boldsymbol{\theta} \mid \sigma_\theta^2) \cdot p(\sigma_\theta^2)  &\text{Likelihood independent of $\sigma_\theta^2$}\nonumber \\
    & \propto p(\boldsymbol{\theta} \mid \sigma_\theta^2) \cdot p(\sigma_\theta^2) &\text{Likelihood constant for fixed $\boldsymbol{\theta}$} \nonumber\\
    & \propto \frac{1}{\sqrt{(2\pi\sigma_{\theta}^2)^S}} \exp\left(-\frac{\lVert \boldsymbol{\theta} \rVert^2}{2{\sigma}_\theta^2}
    \right) \cdot \frac{b^a}{\Gamma(a)}(\sigma_\theta^2)^{-(a+1)} \exp{\left(-\frac{b}{\sigma_\theta^2}\right)} \\
    & \propto (\sigma_\theta^2)^{-(a + \frac{S}{2} +1)} \exp{\left(- \frac{b + \frac{\lVert \boldsymbol{\theta} \rVert^2}{2}} {\sigma_\theta^2}  \right)}. \nonumber
\end{align*}
\end{mdframed}
Hence $\left(\sigma_\theta^2\right)^{(j+1)}\mid \boldsymbol{\theta}^{(j+1)}, \mathcal{D} \sim \text{Inv-Gamma}\left(a + \frac{S}{2}, b + \frac{\lVert \boldsymbol{\theta}^{(j+1)} \rVert^2}{2} \right)$: a distribution from which we can sample directly. Therefore, the MH algorithm is not required for this block. We choose $a, b \approx 0$  in order to specify a nearly uninformative hyperprior of $\sigma_\theta^2$: meaning that the prior exerts minimal influence on the posterior $p(\sigma_\theta^2 \mid \boldsymbol{\theta})$, allowing the observed data to primarily determine the inference. This approach avoids imposing strong assumptions on $\sigma_\theta^2$ and reflects prior ignorance about its scale.\\

To compute the MAP estimate $\boldsymbol{\hat{\theta}}^{MAP}= \operatorname*{argmax}_{\boldsymbol{\theta}} p( \boldsymbol{\theta} \mid \mathcal{D})$, we merely identify the mode of the posterior distribution $p(\boldsymbol{\theta } \mid \mathcal{D})$. Likewise, ${\hat{\theta}}_i^{MAP}= \operatorname*{argmax}_{{\theta}_i} p( {\theta_i} \mid \mathcal{D})$, which is merely the mode of the marginal posterior $p({\theta_i } \mid \mathcal{D})$, which implicitly integrates out both the remaining components $\boldsymbol{\theta}_{-i}$ and the dispersion parameter $\sigma_{\theta}^2$. In this context, each component, $\theta_i$ for $i = 1, \ldots, S$, has a marginal posterior distribution given by $p(\theta_i \mid \mathcal{D}) = \int  p(\boldsymbol{\theta}, \sigma_{\theta}^2 \mid \mathcal{D}) d\boldsymbol{\theta}_{-i} d\sigma_{\theta}^2$. \\

In hierarchical Bayesian modeling, assigning a prior distribution to $\sigma_{\theta}^2$ allows the training data to inform the appropriate degree of regularization. Section \ref{sec: the obj} elucidates $\sigma_{\theta}^2 \propto \frac{1}{\nu}$, therefore smaller $\sigma_{\theta}^2$ values imply stronger shrinkage toward zero, while larger values allow greater flexibility. By endowing $\sigma_{\theta}^2$ with a hyperprior, we integrate over uncertainty in the regularization strength rather than fixing it arbitrarily. Consequently, the resulting MAP estimates of the parameters inherently reflect an optimal degree of regularization, effectively “baking in” regularization informed by the training data.\\

Thus, in a hierarchical Bayesian framework, the training data plays a dual role: it informs the marginal posterior distributions of the model parameters $\boldsymbol{\theta}$, $p( {\theta_i} \mid \mathcal{D})$ for $i  = 1, \ldots ,S$,  while simultaneously guiding the level of regularization through inference on the dispersion parameter $\sigma_{\theta}^2$.

\subsection{Adaptive Metropolis-Hastings} \label{sec: adaptive MH}
In standard MH algorithms, the proposal distribution remains fixed throughout sampling. In contrast, adaptive MCMC methods dynamically tune aspects of the proposal distribution using information gathered during earlier iterations. This adaptive strategy aims to improve mixing by better matching the geometry of the stationary distribution, as expounded in \cite{roberts1997weak}.\\

We previously stated that the proposal distribution for $\boldsymbol{\theta} \in \mathbb{R}^S$ at current iteration $j$ is $\mathbf{Q}_\theta \sim \mathcal{N}(\boldsymbol{\theta}^{(j)}, \boldsymbol{\Sigma_j}) \text{ where } \boldsymbol{\Sigma_j}= \sigma_{Q_\theta}^2 \mathbf{I}_S$: a multivariate normal random walk centered at the current state of $\boldsymbol{\theta}$, $\boldsymbol{\theta}^{(j)}$, with isotropic covariance matrix $\sigma_{Q_\theta}^2 \mathbf{I}_S$. During the burn-in phase only, we adapt $\boldsymbol{\Sigma_j}$ based on previously accepted proposals of $\boldsymbol{\theta}$. By limiting adaptation to the burn-in period, we avoid violating the diminishing adaptation and ergodicity conditions required for convergence to the correct stationary distribution in fully adaptive MCMC algorithms, as explained by \cite{roberts2007coupling}. Additionally, we may still utilise Equation \ref{eq: alpha} to compute acceptance probabilities, $\alpha_\theta$, as our covariance matrices used will all be equivalent after the burn-in phase. That is $\boldsymbol{\Sigma_{j}} = \boldsymbol{\Sigma_{j+1}}$ for $j > \text{burn-in}$, giving rise to symmetric proposal densities.\\

Per \cite{haario2001adaptive}, the covariance matrix at iteration $j$ is given as:
\begin{align}
        \boldsymbol{\Sigma_j}& = \text{Cov} \left( f\left(\boldsymbol{\theta}^{(1)}, \boldsymbol{\theta}^{(2)}, \ldots, \boldsymbol{\theta}^{(j)} \right)\right) + \epsilon\mathbf{I} \nonumber, 
\end{align}
where $\boldsymbol{\Sigma_j}$ is the empirical covariance matrix of a function of the first $j$ samples and $\epsilon \mathbf{I}_S$ is a small positive-definite matrix (for example, $\epsilon = 10^{-6}$) added to maintain numerical stability. The proposal then becomes $\mathbf{Q}_\theta \sim \mathcal{N}(\boldsymbol{\theta}^{(j)}, s^2\boldsymbol{\Sigma_j})$, where $s^2 = \frac{2.38}{S}$. \cite{roberts1997weak} showed that for efficient exploration of high-dimensional target distributions, the acceptance rate should be around $0.234$, and the optimal step size (variance) scales as $s^2 = \frac{2.38}{S}$. \\

To reduce the correlation between the proposals, we employ:
\begin{align}
f\left( \boldsymbol{\theta}^{(1)}, \boldsymbol{\theta}^{(2)}, \ldots, \boldsymbol{\theta}^{(j)} \right) = 
\begin{cases}
\left( 
\boldsymbol{\theta}^{(j - \delta\Delta)},\,
\boldsymbol{\theta}^{(j - \delta\Delta + \delta)},\,
\ldots,\,
\boldsymbol{\theta}^{(j - \delta\Delta + (\Delta - 1)\delta)},\,
\boldsymbol{\theta}^{(j)} 
\right), & \text{if } j > \delta \Delta, \\[1em]
\left( 
\boldsymbol{\theta}^{(j - \lfloor \frac{j}{\delta} \rfloor \delta)},\,
\boldsymbol{\theta}^{(j - \lfloor \frac{j}{\delta} \rfloor \delta + \delta)},\,
\ldots,\,
\boldsymbol{\theta}^{(j - \lfloor \frac{j}{\delta} \rfloor \delta + (\lfloor \frac{j}{\delta} \rfloor - 1)\delta)},\,
\boldsymbol{\theta}^{(j)} 
\right), & \text{if } \delta < j \leq \delta \Delta,
\\
\left( \boldsymbol{\theta}^{(1)}, \boldsymbol{\theta}^{(2)}, \ldots, \boldsymbol{\theta}^{(j)} \right),
 & \text{if } j \leq \delta,
 \end{cases} \label{eq: autocorr sigma} 
\end{align} 
with $\boldsymbol{\theta}^{(1)}$ randomly drawn from the multivariate normal distribution  $\mathcal{N}\left(\mathbf{0}_{S\times 1}, \sigma_{\text{Init}}^2\mathbf{I}_S \right)$. Here, $\delta$ serves as a stride parameter that ensures non-consecutive $\boldsymbol{\theta}$ values are used in the empirical covariance calculation, thereby mitigating autocorrelation between proposals. The window size parameter $\Delta$ is purely curated to ease computation of the empirical covariances at each iteration (its inclusion is optional in which case one would omit the first case of Equation \ref{eq: autocorr sigma}).
\subsection{Adaptive scaling}
Previously, it was noted that according to the optimal scaling theory of \citet{roberts1997weak}, the optimal proposal scale for high-dimensional target distributions is given by $s^2 = \frac{2.38}{S}$. However, in order to directly control the average acceptance rate $\alpha$ toward its theoretical optimum of approximately $0.234$, we adopt an alternative strategy for scaling the covariance matrix $\boldsymbol{\Sigma_j}$ during sampling. Specifically, during the burn-in phase, the scaling factor $s^2$ is adaptively updated at each iteration $j$ according to the observed acceptance rate $\alpha_j$:
\begin{align}
    (s^2)^{(j+1)} = (s^2)^{(j)} \times \exp\left( \gamma_j \cdot (\alpha_j - 0.234) \right) ,\nonumber
\end{align}
where $\gamma_j$ is a sequence of diminishing adaptation rates. In accordance with the theoretical guarantees for ergodicity of adaptive MCMC methods established by \citet{roberts2007coupling}, we choose $\gamma_j =\frac{1}{j^{\kappa}}$ for some $0.5 <\kappa < 1$. This choice satisfies the standard conditions $\sum_j \gamma_j = \infty$ and $\sum_j \gamma_j^2 < \infty$ as per \cite{roberts2007coupling} ensuring that the adaptation is both diminishing and stable. While these conditions are necessary for ensuring ergodicity in fully adaptive MCMC, in our implementation, adaptation is restricted to the burn-in phase. This pragmatic restriction alleviates these theoretical concerns.

\section{\texorpdfstring{The Objective $\&$ Regularization}{The Objective and Regularization}}
\label{sec: the obj}
This section demonstrates the necessity for the likelihood function, $p(\mathcal{D} \mid \boldsymbol{\theta})$, to be proportional to (or a monotonically increasing function of) the objective function being maximized. Consider the arbitrary objective, for a given parameter configuration $\boldsymbol{\theta} \in  \mathbb{R}^S$, denoted as $ \operatorname*{argmax}_{\boldsymbol{\theta}} \text{Obj}\left(\boldsymbol{\theta} \right)$. By including L$2$ regularization, the L$2$ penalized objective becomes:
\begin{align}
     \operatorname*{argmax}_{\boldsymbol{\theta}} \left( \text{Obj}(\boldsymbol{\theta}) -  \nu \sum_{i = 1}^S \theta_i^2 \right) .\label{eq: l2 obj}
\end{align}
Under the assumption that the likelihood $p(\mathcal{D} \mid \boldsymbol{\theta})$ is monotonic increasing with respect to $\text{Obj}\left(\boldsymbol{\theta} \right)$, and assuming Gaussian priors for our parameters $\boldsymbol{\theta} \in \mathbb{R}^S$, that is $\boldsymbol{\theta} \sim \mathcal{N}(\mathbf{0}, \sigma_\theta^2 \mathbf{I}_S)$, using Bayes rule we have:
\begin{align*}
    p(\boldsymbol{\theta} \mid \mathcal{D}) &= \frac{p(\mathcal{D} \mid \boldsymbol{\theta}) p(\boldsymbol{\theta})}{p(\mathcal{D})} \nonumber \\
    &\propto p(\mathcal{D} \mid \boldsymbol{\theta}) p(\boldsymbol{\theta}) \\
    & \propto p(\mathcal{D} \mid \boldsymbol{\theta}) \mathcal{N}(\boldsymbol{\theta}; \mathbf{0}, \sigma_\theta^2 \mathbf{I}) \\
    & \propto p(\mathcal{D} \mid \boldsymbol{\theta}) \prod_{i = 1}^S\mathcal{N}(\theta_i; 0, \sigma_\theta^2). \nonumber
\end{align*}
Now taking the log probability of the parameter posterior:
\begin{align*}
    \text{log}[ p(\boldsymbol{\theta} \mid \mathcal{D}) ] &\propto  \log\left[ p(\mathcal{D} \mid \boldsymbol{\theta})\right] 
  + \sum_{i = 1}^S \text{log} \left[\mathcal{N}(\theta_i; 0, \sigma_\theta^2)\right] \nonumber \\
    & \propto \log\left[p(\mathcal{D} \mid \boldsymbol{\theta})\right] - \frac{1}{2\sigma_\theta^2} \sum_{i = 1}^S \theta_i^2. \nonumber
\end{align*}
Now since we assumed the likelihood $p(\mathcal{D} \mid \boldsymbol{\theta})$ to be proportional to $\text{Obj}\left(\boldsymbol{\theta} \right)$, that is,  $p(\mathcal{D} \mid \boldsymbol{\theta}) \propto \text{Obj}\left(\boldsymbol{\theta} \right)$, we have:
\begin{align}
    \text{log}[ p(\boldsymbol{\theta} \mid \mathcal{D}) ] &\propto\text{Obj}\left(\boldsymbol{\theta} \right)-  \nu \sum_{i = 1}^S \theta_i^2 \nonumber,
\end{align}
whereby $\nu\propto \frac{1}{\sigma_\theta^2}$ controls the strength of the regularization as in the L$2$ penalized objective in Equation \ref{eq: l2 obj}. Clearly, $\operatorname*{argmax}_{\boldsymbol{\theta}} p( \boldsymbol{\theta} \mid \mathcal{D})$, is equivalent to the L$2$ penalized objective in Equation \ref{eq: l2 obj}, that is, $\hat{\boldsymbol{\theta}}^{MAP}=\operatorname*{argmax}_{\boldsymbol{\theta}} p( \boldsymbol{\theta} \mid \mathcal{D})=\operatorname*{argmax}_{\boldsymbol{\theta}} \left( \text{Obj}\left(\boldsymbol{\theta} \right)- \frac{1}{2\sigma_\theta^2} \sum_{i = 1}^S \theta_i^2\right)$. We may say that the L$2$ penalization constraint depends on fixed $\sigma_\theta^2$, the number of coefficients, $S$, and their magnitudes $\theta_i$, written as function $f(\sigma_\theta^2, S, \theta_i)$. \\

However, within the context of the two-block MCMC scheme, since $\sigma_\theta^2$ is sampled at each iteration, it is no longer the case that the MAP estimate $\hat{\boldsymbol{\theta}}^{MAP}$ is equivalent to $\operatorname*{argmax}_{\boldsymbol{\theta}} \left( \text{Obj}(\boldsymbol{\theta}) - \nu \sum_{i=1}^S \theta_i^2 \right)$ with $\nu \propto \frac{1}{\sigma_\theta^2}$, as this equivalence only holds when $\sigma_\theta^2$ is fixed across all iterations. Rather, at each iteration $j$, we induce a shrinkage factor $\nu ^{(j)} \propto \frac{1}{(\sigma_\theta^2)^{(j)}}$ on the proposal $\boldsymbol{\theta}^*$, as shown in Equation \ref{eq: log alpha}.  Nevertheless, we may still characterize the amount of regularization associated with the MAP estimates by examining the marginal posterior $p(\sigma_\theta^2 \mid \mathcal{D})$. This distribution reflects the posterior uncertainty about the degree of shrinkage, that is, how much regularization the data supports. For example, we may utilise the mean of $p(\sigma_\theta^2 \mid \mathcal{D})$ to reflect an ``effective ridge penalty'' or have the posterior intervals to show uncertainty in the amount of regularization inferred.\\

This two-block MCMC framework also results in our marginal prior for $\boldsymbol{\theta}$ to not be Gaussian as before. Since we assume priors: $\boldsymbol{\theta}\mid \sigma_\theta^2 \sim \mathcal{N}(\mathbf{0}, \sigma_\theta^2\mathbf{I}_S)$ and $\sigma_\theta^2 \sim \text{Inv-Gamma}(a, b)$, we have:
\begin{align}
    p(\boldsymbol{\theta}) &= \int_{0}^\infty p(\boldsymbol{\theta}, \sigma_\theta^2) \ d \sigma_\theta^2 \nonumber \\
    & = \int_{0}^\infty p(\boldsymbol{\theta}\mid \sigma_\theta^2) p( \sigma_\theta^2)  \ d \sigma_\theta^2 \nonumber \\
    & = \int_{0}^\infty (2\pi\sigma_\theta^2)^{(-\frac{S}{2})} \exp\left(-\frac{1}{2\sigma_\theta^2} \lVert\boldsymbol{\theta} \rVert^2 \right) \frac{b^a}{\Gamma(a)} (\sigma_\theta^2) ^{-(a+1)}  \exp\left( -\frac{b}{\sigma_\theta^2} \right) \ d\sigma_\theta^2 \nonumber \\
    & =  \frac{b^a}{\Gamma(a)} (2\pi)^{-(\frac{S}{2})}\int_{0}^\infty (\sigma_\theta^2 )^{-(a+ \frac{S}{2}+1)} \exp\left( -\frac{1}{2\sigma_\theta^2} \left(\lVert\boldsymbol{\theta} \rVert^2+2b \right)\right)d\sigma_\theta^2. \nonumber
\end{align}
Using $\int_{0}^\infty x^{-(\alpha+1)}\exp(-\frac{\beta}{x}) \ dx =\beta^{-\alpha}  \Gamma(\alpha)$, we have the following:
\begin{align}
    p(\boldsymbol{\theta})&= \frac{\Gamma(a + \frac{S}{2})}{\Gamma(a)} (2\pi)^{-(\frac{S}{2})} b^a \left(b + \frac{1}{2}\lVert\boldsymbol{\theta} \rVert^2  \right)^{-(a+\frac{S}{2})} \nonumber \\
    & = \frac{\Gamma\left(\frac{2a+ S}{2}\right)}{\Gamma\left(\frac{2a}{2}\right) (2a)^{\frac{S}{2}} \pi^{\frac{S}{2}} \left(\frac{b}{a} \right)^{\frac{S}{2}}} \left( 1+ \frac{1}{2a}\boldsymbol{\theta}'\left(\frac{b}{a}\mathbf{I}_S\right)^{-1} \boldsymbol{\theta}  \right)^{-\left(\frac{2a+S}{2}\right)}. \nonumber
\end{align}
Hence, $\boldsymbol{\theta} \sim t_S\left(\mathbf{0}, \frac{b}{a}\mathbf{I}_S, 2a \right)$. That is, the marginal prior on $\boldsymbol{\theta}$ is multivariate Student-$t$ with $2a$ degrees of freedom, location vector $\mathbf{0}$, and scale matrix $\frac{b}{a}\mathbf{I}_S$, rather than multivariate Gaussian. Accordingly:
\begin{align}
    \text{log}[ p(\boldsymbol{\theta} \mid \mathcal{D}) ] &\propto\text{Obj}\left(\boldsymbol{\theta} \right)- \left(a +\frac{S}{2}\right) \log\left(b +\frac{1}{2}\sum_{i = 1}^S \theta_i^2\right). \nonumber
\end{align}
Hence $\hat{\boldsymbol{\theta}}^{MAP} =\operatorname*{argmax}_{\boldsymbol{\theta}} \left( \text{Obj}\left(\boldsymbol{\theta} \right)- \left(a +\frac{S}{2}\right) \log\left(b +\frac{1}{2}\sum_{i = 1}^S \theta_i^2\right)\right)$. Constraining $\text{Obj}\left(\boldsymbol{\theta} \right)$ with a penalty of the form $\log\!\big(f(\sum_{i=1}^S \theta_i^2)\big)$, rather than the quadratic penalty $\sum_{i=1}^S \theta_i^2$ that arises under a Gaussian marginal prior, highlights the difference between the two. Both behave similarly for small coefficients $\theta_i$, but the Student-$t$ prior imposes much weaker shrinkage in the tails, applying almost no penalization to large coefficients relative to the Gaussian case. Previously, under a fixed Gaussian prior, the penalization constraint for L$2$ regularization was a function $f(\sigma_\theta^2, S, \theta_i)$. In the two-block MCMC framework, since $\sigma_\theta^2$ itself is treated hierarchically via the hyperparameters $(a,b)$, the penalization constraint is now a function $f(a,b,S,\theta_i)$.

\section{The Likelihood} \label{sec: the likelihood}
We emphasize that, because our objectives are arbitrary, no concrete assumptions about the data-generating process can be made. In other words, there is no well-defined probabilistic model for the data. Conventionally, the likelihood is derived from the joint density of the data samples $\{x_i\}_{i=1}^n$ given the parameters $\boldsymbol{\theta}$: $p(\mathcal{D}\mid \boldsymbol{\theta}) = f(x_1,\ldots,x_n \mid \boldsymbol{\theta}) = \prod_{i=1}^n f(x_i \mid \boldsymbol{\theta}) \ \  (\text{for } x_i \text{ i.i.d.})$. In contrast, in our setting, we only have an objective function $\text{Obj}(\boldsymbol{\theta})$ that we aim to maximize. Hence, the study utilises likelihoods (more accurately described as pseudo-likelihoods, although we use the terms interchangeably throughout the study) which are purposely fabricated to be proportional to $\text{Obj}(\boldsymbol{\theta})$, as prescribed in Section \ref{sec: the obj}, to ensure 
$\hat{\boldsymbol{\theta}}^{{MAP}} = \operatorname*{argmax}_{\boldsymbol{\theta}} \left( \text{Obj}(\boldsymbol{\theta}) - f\left(\sum_{i = 1}^S \theta_i^2\right) \right)$. Additionally, the likelihoods are tempered: that is, constructed to allow control over their sharpness with respect to $\text{Obj}(\boldsymbol{\theta})$. This tempering increases the sensitivity of the likelihood to changes in the objective function values, since $p(\mathcal{D} \mid \boldsymbol{\theta}) \propto \text{Obj}(\boldsymbol{\theta})$. An increased sharpness may also be interpreted as a greater concentration of mass around the modal regions of their densities. Consequently, it translates into a more pronounced influence of the likelihood on the conditional posterior distribution $p(\boldsymbol{\theta} \mid \sigma_{\theta}^2, \mathcal{D})$, since $p(\boldsymbol{\theta} \mid \sigma_{\theta}^2, \mathcal{D}) \propto p(\mathcal{D} \mid \boldsymbol{\theta}) \cdot p(\boldsymbol{\theta} \mid \sigma_{\theta}^2)$. Within the context of the two-block MCMC framework described in Section~\ref{sec: 2-sample}, increased likelihood sharpness implies that the sampler becomes more likelihood-driven, thereby diminishing the influence of the prior ratio in exploring the conditional posterior.\\

In this light, the pseudo-likelihoods employed in the study are not intended to represent rigorous data-generating models. Instead, they are used within the MCMC framework, not to sample from a full posterior, but to facilitate optimization over the parameter space. The MCMC algorithm is thus repurposed as a mode-seeking procedure, targeting the mode of the conditional posterior $p(\boldsymbol{\theta} \mid \sigma_{\theta}^2, \mathcal{D})$. That is to say, to concentrate samples around a dominant mode of $p(\boldsymbol{\theta} \mid \sigma_{\theta}^2, \mathcal{D})$. It suffices that the proposal mechanism is guided by likelihood functions that monotonically increase with the objective of interest, biasing the sampling process toward regions of high posterior density. These regions necessarily correspond to high-likelihood (thus high-valued objective) areas since $p(\boldsymbol{\theta} \mid \sigma_{\theta}^2, \mathcal{D}) \propto p(\mathcal{D} \mid \boldsymbol{\theta}) \cdot p(\boldsymbol{\theta} \mid \sigma_{\theta}^2)$. As such, the requirement for a fully specified likelihood linking the data to the model becomes less critical.\\

We emphasise now, for the purposes of convergence in MCMC, it is often more desirable for the sampler to be likelihood-driven rather than prior-driven. Equation \ref{eq: log alpha} further illustrates this concept. In the limiting case of a negligible likelihood, it simplifies to:
\begin{align*}
\log \left(\alpha_\theta\right) &= \min \Bigg( \frac{1}{2\left(\sigma_\theta^2\right)^{(j)}}\left( \lVert \boldsymbol{\theta}^{(j)} \rVert^2 - \lVert \boldsymbol{\theta}^{*} \rVert^2 \right), 0 \Bigg), \nonumber
\end{align*}
indicating that when the likelihood is approximately flat, that is, $\log\left( p\left(\mathcal{D} \mid \boldsymbol{\theta}^{*}\right) \right)\approx \log \left(p\left(\mathcal{D} \mid \boldsymbol{\theta}^{(j)}\right)\right)$, proposals that reduce $\lVert \boldsymbol{\theta}^{*} \rVert^2$ are more likely to be accepted, leading to an indefinite contraction of $\boldsymbol{\theta}$ towards zero. Although a balanced contribution between prior and likelihood is theoretically desirable, the practical impossibility of pre-specifying this balance motivates us to ensure that the likelihood is sufficiently dominant in shaping the conditional posterior $p(\boldsymbol{\theta} \mid \sigma_{\theta}^2, \mathcal{D})$.
\subsection{\texorpdfstring{Likelihood dominance $\&$ implications}{Likelihood dominance and implications}}
In the context of the two-block MCMC scheme as in Section \ref{sec: 2-sample}, where sampling is performed from the conditional posterior 
$p(\boldsymbol{\theta} \mid \sigma_\theta^2, \mathcal{D}) \propto p(\mathcal{D} \mid \boldsymbol{\theta}) \cdot p(\boldsymbol{\theta} \mid \sigma_\theta^2)$, 
it is important to note that the mode of conditional $p(\boldsymbol{\theta} \mid \sigma_\theta^2, \mathcal{D})$ need not coincide exactly with the mode of the marginal $p(\boldsymbol{\theta} \mid \mathcal{D})$. This is because the former is directly influenced by the specific value of $\sigma_\theta^2$. That is, $\sigma_\theta^2$ directly determines the shape of the prior $p(\boldsymbol{\theta} \mid \sigma_\theta^2)$. We note that:
\begin{align}
    p(\boldsymbol{\theta} \mid \mathcal{D}) 
    &= \int p(\boldsymbol{\theta}, \sigma_\theta^2 \mid \mathcal{D}) \, d\sigma_\theta^2 \nonumber \\
    &= \int p(\boldsymbol{\theta} \mid \sigma_\theta^2, \mathcal{D}) \cdot p(\sigma_\theta^2 \mid \mathcal{D}) \, d\sigma_\theta^2. \label{eq: integral marg cond}
\end{align}
This expression implies that the marginal posterior $p(\boldsymbol{\theta} \mid \mathcal{D})$ is a weighted average of the conditionals $p(\boldsymbol{\theta} \mid \sigma_\theta^2, \mathcal{D})$, where the weights come from the marginal $p(\sigma_\theta^2 \mid \mathcal{D})$. In other words, to obtain $p(\boldsymbol{\theta} \mid \mathcal{D})$, one averages over the uncertainty in $\sigma_\theta^2$.\\

There are two main scenarios where the mode of the marginal posterior $p(\boldsymbol{\theta} \mid \mathcal{D})$ will approximately coincide with the mode of the conditional posterior $p(\boldsymbol{\theta} \mid \sigma_\theta^2, \mathcal{D})$. First, if the marginal $p(\sigma_\theta^2 \mid \mathcal{D})$ has low variance (that is, it is sharply peaked around a single value), then the integral in Equation \ref{eq: integral marg cond} is dominated by a narrow range of $\sigma_\theta^2$, effectively treating $\sigma_\theta^2$ as nearly constant. Second, if the likelihood $p(\mathcal{D} \mid \boldsymbol{\theta})$ is highly informative (that is, dominates the conditional $p(\boldsymbol{\theta} \mid \sigma_\theta^2, \mathcal{D})$), then it largely determines the shape of the conditional, making it sharply peaked in roughly the same region of $\boldsymbol{\theta}$ regardless of the specific value of $\sigma_\theta^2$. In this case, all the conditional posteriors in the integral of Equation \ref{eq: integral marg cond} are peaked in the same region, and so the resulting marginal $p(\boldsymbol{\theta} \mid \mathcal{D})$ will also be peaked there.\\

Furthermore, given the assumption that the likelihood  $p(\mathcal{D} \mid \boldsymbol{\theta})$ is sufficiently dominant, we have $\operatorname*{argmax}_{\boldsymbol{\theta}} p( \boldsymbol{\theta} \mid \sigma_\theta^2, \mathcal{D})  \approx \boldsymbol{\hat{\theta}}^{MAP}= \operatorname*{argmax}_{\boldsymbol{\theta}} p( \boldsymbol{\theta} \mid \mathcal{D})$. Additionally, it may be argued that through likelihood dominance, since $p(\boldsymbol{\theta} \mid \sigma_\theta^2, \mathcal{D}) \propto p(\mathcal{D} \mid \boldsymbol{\theta}) \cdot p(\boldsymbol{\theta} \mid \sigma_\theta^2)$, we may have $\operatorname*{argmax}_{\boldsymbol{\theta}} p( \boldsymbol{\theta} \mid \sigma_\theta^2, \mathcal{D}) \approx \operatorname*{argmax}_{\boldsymbol{\theta}} p(\mathcal{D} \mid \boldsymbol{\theta}) =  \boldsymbol{\hat{\theta}}^{MLE}$. Furthermore, due to proportionality $p(\mathcal{D} \mid \boldsymbol{\theta}) \propto \text{Obj}(\boldsymbol{\theta})$, $\boldsymbol{\hat{\theta}}^{MLE} = \operatorname*{argmax}_{\boldsymbol{\theta}}\text{Obj}(\boldsymbol{\theta})$. Hence, altogether, we may argue that, given sufficient likelihood dominance, 
$\operatorname*{argmax}_{\boldsymbol{\theta}} p( \boldsymbol{\theta} \mid \sigma_\theta^2, \mathcal{D})  \approx \boldsymbol{\hat{\theta}}^{MAP}= \operatorname*{argmax}_{\boldsymbol{\theta}} p( \boldsymbol{\theta} \mid \mathcal{D}) \approx \boldsymbol{\hat{\theta}}^{MLE} = \operatorname*{argmax}_{\boldsymbol{\theta}} p(\mathcal{D} \mid \boldsymbol{\theta}) = \operatorname*{argmax}_{\boldsymbol{\theta}} \text{Obj}(\boldsymbol{\theta})$. That is, the mode of the conditional $p(\boldsymbol{\theta} \mid \sigma_\theta^2, \mathcal{D})$ will necessarily lie in a region where $\text{Obj}(\boldsymbol{\theta})$ attains high values. This imples that, when the prior is severely undermined, two-block MCMC merely serves as an optimization technique to seek out $\operatorname*{argmax}_{\boldsymbol{\theta}} \text{Obj}(\boldsymbol{\theta})$, where negligible regularization is inferred to the MAP estimates.

\section{Experimental Problems with Arbitrary Objectives}
This section outlines the reinforcement learning problems with arbitrary objective functions used in the study. Each experimental setup includes a description of how the environment is encoded and how a neural network is employed as a means of control. Within each of the subsequent sections, the study examines the effects of regularization, comparing the performance of a GA and RS under varying pre-specified regularization strengths $\nu$. Subsequently, the results obtained using two-block MCMC are presented, along with a discussion of caveats. \\

For the first experimental setup, a navigation problem, the study evaluates the impact of different pseudo-likelihood forms on MCMC performance: considering both in-sample and out-of-sample results, as well as the inferred regularization strength. In the second experimental setup, tic-tac-toe, the analysis focuses on how likelihood sharpness and the initial variance parameter, $\sigma_{\text{Init}}^2$, influence MCMC performance. Finally, the third experimental setup of S-$17$ blackjack is used to investigate the performance of a hybrid algorithm relative to the original two-block MCMC approach employed throughout the study.\\

Furthermore, a neural network with two hidden layers and three nodes per layer was used (see \ref{sec: overall specs}), with hyperbolic tangent activation functions in both layers. The network was intentionally kept small to control model complexity, since the aim of this study is to examine how regularization behaves and to assess the shortcomings of two-block MCMC, rather than to maximize predictive performance. The 3-3 structure provides enough nonlinearity to represent reinforcement learning policies while avoiding excessive flexibility that could confound the interpretation of inferred regularization. The hyperbolic tangent function was chosen because it is smooth, zero-centred, and bounded, which helps keep the learning dynamics stable and makes the effects of shrinkage-based regularization easier to interpret. As such, these architectural choices are meant to provide a simple and stable baseline that isolates the mechanisms under investigation.

\subsection{A navigation problem}
In what follows, the study employs a navigation problem to assess how different pseudo-likelihood formulations influence both in-sample and out-of-sample performance, as well as the level of regularization inferred. Specifically, we construct a navigation task in which a neural network serves as the control policy governing movement for a drone that needs to escape a ``forest'' of moving obstacles. This environment is adapted from a simpler configuration presented in \cite{pienaar2024evolutionary} and is deliberately chosen to assess MCMC performance using a continuous environment (the actions of the drone agent occur in a bounded space which is encoded as continuous values).\\ 

Formally, consider $T$ drones navigating within a two-dimensional circular arena. The arena is defined as the annular region bounded by an inner radius $R_{inner}$ and an outer radius $R_{outer}$. Each drone aims to escape this arena within $K$ steps, where each step has a fixed length $\delta$. A drone is considered to have crashed if it comes within a distance $R_{crash}$ of any of the $J$ orbiting obstacles. Upon crashing, the drone ceases further navigation. An illustration of the navigation problem is displayed in Figure \ref{fig: drone move}.

\iffoo
\begin{figure}[H]
\centering
\animategraphics[autoplay,loop,width=0.62\linewidth]{16}{Drones/frame_}{000}{214}
\caption{Illustration of the navigation problem: $T$ drones (black dots) attempting to escape the annular arena while avoiding the $J$ orbiting obstacles (larger green disks) (displaying $\boldsymbol{\hat{\theta}}^{\text{GA},\text{(II)}}_{\nu = 4\times 10^{-6}}$ evaluated on an arbitrary out-of-sample initialization).}
\label{fig: drone move}
\end{figure}
\else
\begin{figure}[H]
\centering

\begin{subfigure}{0.32\linewidth}
    \centering
    \includegraphics[width=\linewidth]{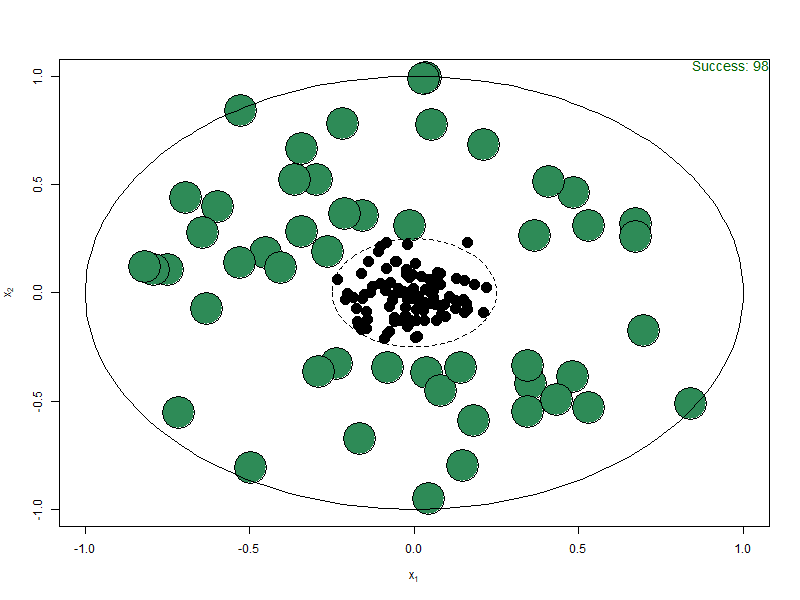}
    \caption{$k = 0$}
\end{subfigure}
\hfill
\begin{subfigure}{0.32\linewidth}
    \centering
    \includegraphics[width=\linewidth]{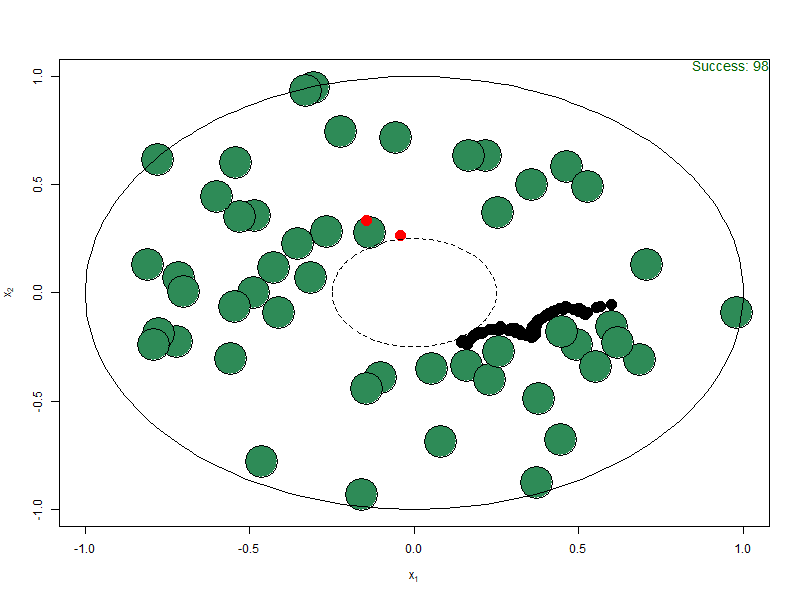}
    \caption{$k = 40$}
\end{subfigure}
\hfill
\begin{subfigure}{0.32\linewidth}
    \centering
    \includegraphics[width=\linewidth]{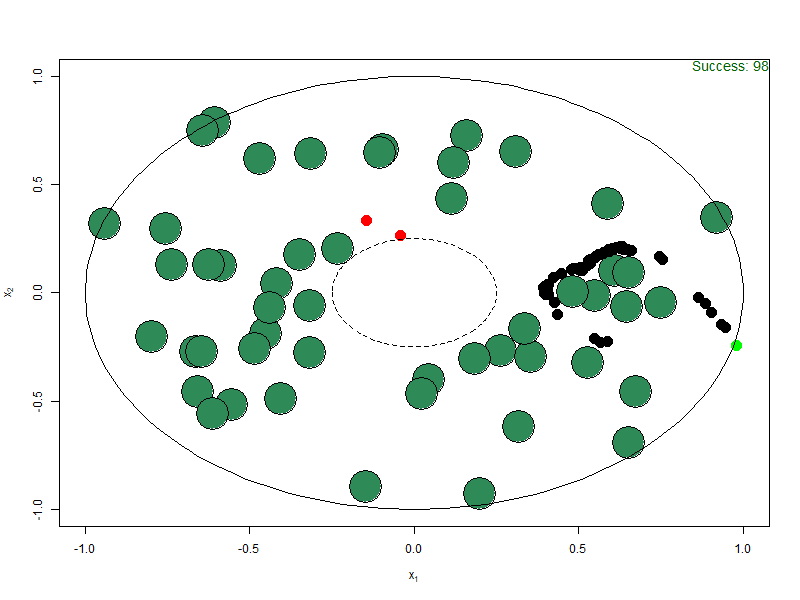}
    \caption{$k = 80$}
\end{subfigure}

\vspace{0.3cm}

\begin{subfigure}{0.32\linewidth}
    \centering
    \includegraphics[width=\linewidth]{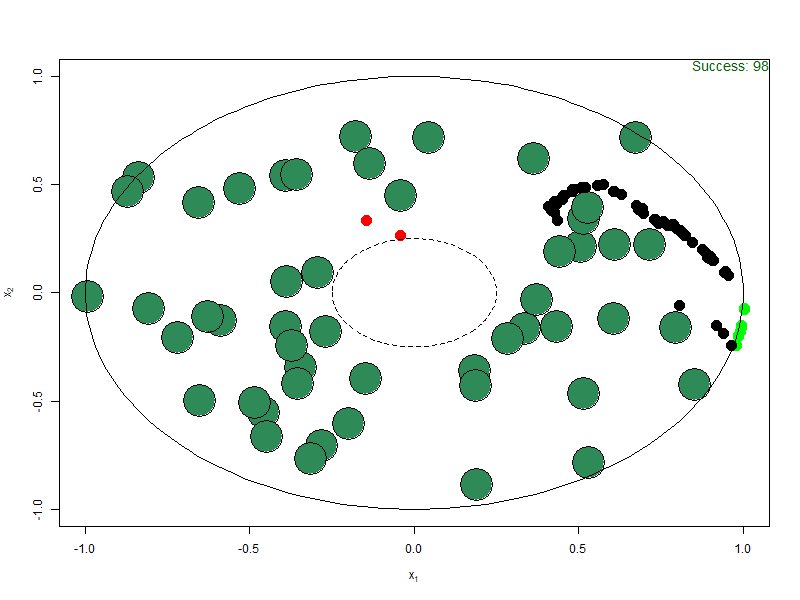}
    \caption{$k = 120$}
\end{subfigure}
\hfill
\begin{subfigure}{0.32\linewidth}
    \centering
    \includegraphics[width=\linewidth]{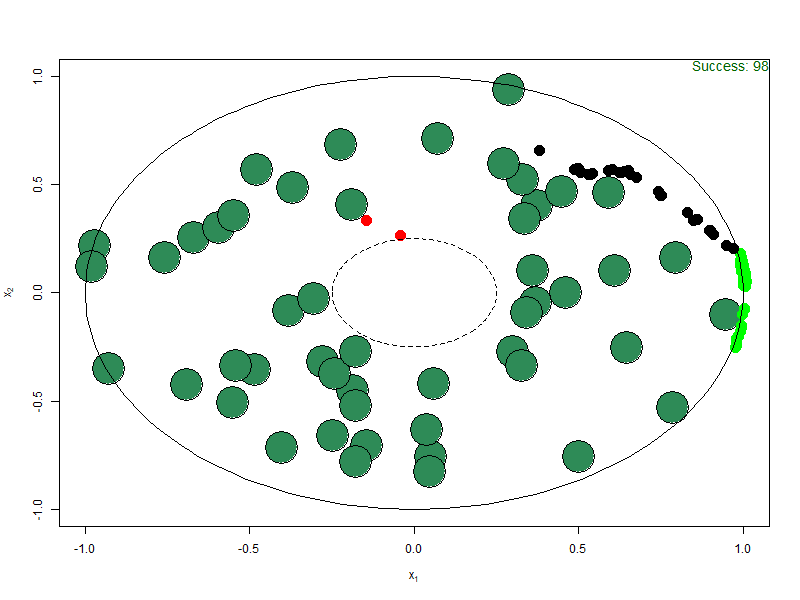}
    \caption{$k = 160$}
\end{subfigure}
\hfill
\begin{subfigure}{0.32\linewidth}
    \centering
    \includegraphics[width=\linewidth]{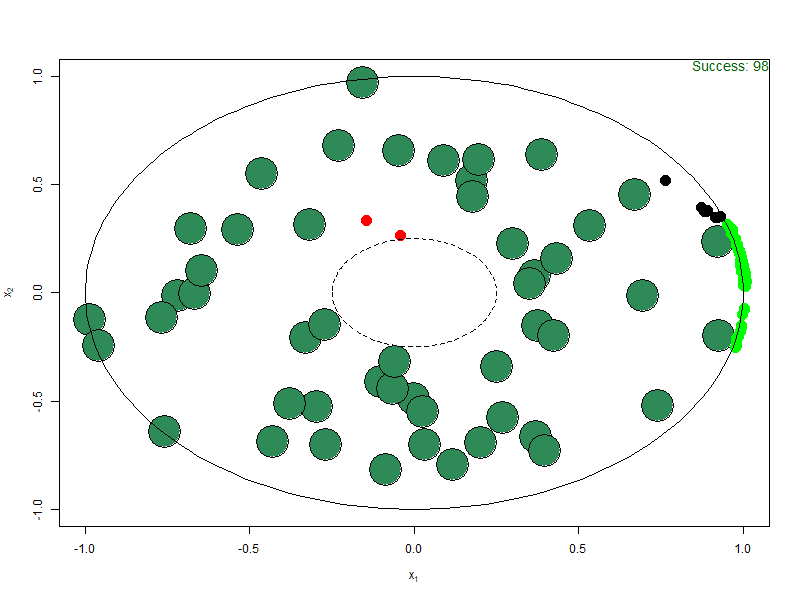}
    \caption{$k = 200$}
\end{subfigure}

\caption{Illustration of the navigation problem: $T$ drones (black dots) attempting to escape the annular arena while avoiding the $J$ orbiting obstacles (larger green disks) (snapshots of 
$\boldsymbol{\hat{\theta}}^{\text{GA},\text{(II)}}_{\nu = 4\times 10^{-6}}$
evaluated on an arbitrary out-of-sample initialization).}
\label{fig: drone move}
\end{figure}
\fi

\subsubsection{Encoding}
\begin{figure}[H]
    \centering
    \includegraphics[width=0.5\linewidth]{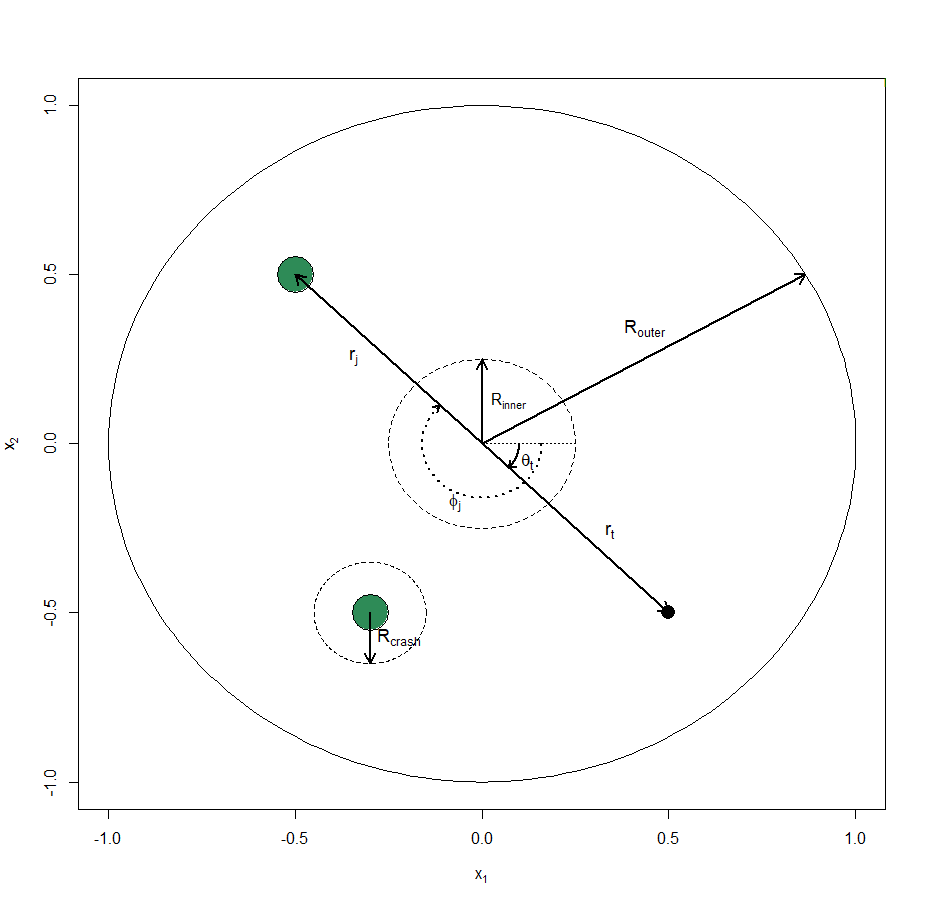}
    \caption{Diagram illustrating the mathematical notation of the navigation problem (noting an orbiting obstacle as a large green disk and drone as a black dot).}
    \label{fig: drone_diag}
\end{figure}

We represent the annular arena by the region [$R_{inner} \cos(\theta), R_{outer} \cos(\theta)$] $\times$  [$R_{inner} \sin(\theta), R_{outer} \sin(\theta)$] $\in \mathbb{R}^2$ for $\theta\in [0, 2\pi)$. The $J$ obstacles orbit the arena with unique angular frequencies, $\omega_j$, and orbital radii, $r_j$ where the $j^{th}$ obstacle has a coordinate at the $k^{th}$ iteration: $\mathbf{o}_j^{(k)}  = (o_{j1}^{(k)}, o_{j2}^{(k)}): j = 1, 2, \hdots, J$ where $o_{j1}^{(k)} = r_{j}\cos(\omega_{j}\cdot k+ \phi_j)$ and $o_{j2}^{(k)} = r_{j}\sin(\omega_{j}\cdot k+ \phi_j)$ for angular frequencies, $\omega_{j} = 2\pi U( \frac{1}{P_{upper}} , \frac{1}{P_{lower}})$, phase shifts, $\phi_j = 2\pi U(0, 1)$ and orbital radii, $r_{j} = U(R_{inner}+R_{crash},R_{outer})$. We draw the initial $T$ drones at coordinates $\mathbf{x}_t^{(0)} = (x_{t1}^{(0)}, x_{t2}^{(0)}): t = 1, 2, \hdots, T$ where $x_{t1}^{(0)} = r_{t} \cos(\theta_{t})$ and $x_{t2}^{(0)} = r_{t} \sin(\theta_{t})$ for $\theta_{t} = 2\pi U(0, 1)$ and $r_{t} = U(0, R_{inner})$. Figure \ref{fig: drone_diag} provides a schematic illustration of the annular arena and associated notation.

\subsubsection{The game state}
After each movement made by an obstacle, the game state must be evaluated to determine whether a collision with a drone has occurred. Similarly, following each drone movement, the game state must again be assessed to establish whether the drone has either succeeded or failed. A failure is defined as the drone either (i) coming within a distance $R_{crash}$ of an obstacle, signifying a crash, or (ii) failing to reach the boundary of the circular arena within $K$ permitted steps. A success is defined as the drone reaching the edge of the arena. Consequently, the game state is evaluated twice during each time step $k$: once after the obstacles move and once after the drones move. If neither a success nor failure condition is met, the drone proceeds to the next step. We encode this as:

\begin{align}
    s_t^k &= 
    \begin{cases}
        +1, & \text{if } \sqrt{x_{t1}^{(k)} + x_{t2}^{(k)} } \geq R_{{outer}}, \\    
        -1, & \text{if } \exists j \in \{1, 2, \dots, J\} \text{ such that } \|\mathbf{x}_t - \mathbf{o}_j\| \leq R_{{crash}}, \\
        0, &\text{otherwise}. \nonumber
    \end{cases}
\end{align}
where $s_t^k$ denotes the games status for drone $t$ at step $k$ while $k \leq K$. Now if $s_t^k \in \{-1, 1\}$ for $k\leq K$ then the navigation for that particlar $t^{th}$ drone ends. Otherwise, the game continues until $k > K$ and the status is recorded as $-1$ for that particlar $t^{th}$ drone.

\subsubsection{Control: Drone movement}
During each step, we must move each drone by changing $(x_{t1}^{(k)}, x_{t2}^{(k)})$ by at most $\delta$ for at most $K$ steps. Hence, the game-updating equation follows: 

$$\mathbf{x}_t^{(k+1)} = \mathbf{x}_t^{(k)} + \mathbf{ct}(\mathbf{x}_t^{(k)}, \mathbf{o}_j^{(k)},  \boldsymbol{\theta})\delta_t,$$ where $\mathbf{ct}(\mathbf{x}_t^{(k)}, \mathbf{o}_j^{(k)}, \boldsymbol{\theta}) \in [-1, 1]^2$ is some control vector for some parameter configuration $\boldsymbol{\theta} \in \mathbb{R}^S$. Now, the interface between a model and the navigation is undergone through this control vector for which $$\mathbf{ct}: (\mathbf{x}_t^{(k)}, \mathbf{o}_j^{(k)} ,\boldsymbol{\theta}) \rightarrow \mathbf{model} \left(\boldsymbol{\Omega}\left(\mathbf{x}_t^{(k)}, \mathbf{o}_j^{(k)} , \boldsymbol{\phi}\left(\boldsymbol{\theta}_1\right)\right), \boldsymbol{\theta}_2 \right) \xrightarrow{\sigma_L(.)}\left(\tilde{x}_{t1}^{(k)}, \tilde{x}_{t2}^{(k)}\right) \in [-1, 1]^2,$$ where $\boldsymbol{\theta}_1 \subset \boldsymbol{\theta}, \boldsymbol{\theta}_2 \subseteq \boldsymbol{\theta}$ and $\boldsymbol{\theta}$ is fixed throughout the navigation period and all $t = 1, 2, \hdots, T$ drones move according to this $\boldsymbol{\theta}$. Additionally, $\left(\tilde{x}_{t1}^{(k)}, \tilde{x}_{t2}^{(k)}\right)$ represents the additions to the $x$ and $y$ coordinate of the $t^{th}$ drone for the $k^{th}$ step such that $\mathbf{x}_t^{(k+1)} = \mathbf{x}_t^{(k)} + \mathbf{\tilde{x}}_t^{(k)}\delta_t$. Furthermore, when the $t^{{th}}$ drone is deemed terminal, we set $\delta_t = 0$, resulting in no positional update, that is, $\mathbf{x}_t^{(k+1)} = \mathbf{x}_t^{(k)}$
for the $k^{th}$ iteration in which terminality is detected. Otherwise, $\delta_t = \delta$. In other words, if the $t^{th}$ drone has reached a terminal state, its coordinates $\mathbf{x}_t^{(k)}$ are still passed through to the control vector $\mathbf{ct}$. Now in the framework of using a neural network as our model, we define $\boldsymbol{\Omega}: \left(\mathbf{x}_t^{(k)}, \mathbf{o}_j^{(k)} , \boldsymbol{\phi}\left(\boldsymbol{\theta}_1\right)\right) \rightarrow \mathbf{a}(t)^0 \in \mathbb{R}^{d_0}$ which signifies the vector of input nodes for the $t^{th}$ drone, where $\boldsymbol{\phi}$ is a user-defined mapping. Additionally, $\boldsymbol{\theta}_2$ are the weights and biases of the neural network, $\mathbf{w} \in \mathbb{R}^R$, where $R\leq S$, and $\sigma_L(.)$ represents the hyperbolic tangent activation function applied at the output layer, allowing movement of the $t^{th}$ drone in all directions. 
\paragraph{Feature engineering}
For effective manoeuvring of a drone around ``clumps'' of obstacles, we propose to set the neural net input, $\boldsymbol{\Omega}\left(\mathbf{x}_t^{(k)}, \mathbf{o}_j^{(k)} ,  \boldsymbol{\phi}\left(\boldsymbol{\theta}_1\right)\right)$, for the $t^{th}$ drone to merely the sum of all reciprocal Euclidean distances between said drone and all obstacles within a specified radius $R_{detection}$ of the drone. Indeed, this approach has two key benefits: first, the input value increases when there are many obstacles within the drone's vicinity. Second, the input value becomes larger the closer an obstacle is to the drone. In this way, the input can quantify the level of caution the $t^{th}$ drone should exercise. This is actually critical for the generalization ability of the agent, since we want the agent to not simply learn the coordinates of objects in a particular presentation of a game, but some measure of spatial proximity to objects in general.\\

Hence we set $\boldsymbol{\phi} \left(\boldsymbol{\theta}_1 \right) = R_{detection}$ and $\boldsymbol{\Omega}: \left(\mathbf{x}_t^{(k)}, \mathbf{o}_j^{(k)} ,  R_{detection} \right) \rightarrow {a}(t)^0 \in \mathbb{R}^{d_0 = 1}$, that is, a single input node is created for each $t^{th}$ drone with:
\begin{align}
    \boldsymbol{\Omega}(\mathbf{x}_t^{(k)}, \mathbf{o}_j^{(k)} , R_{detection}) &= 
    \begin{cases}
        \sum_{j^* = 1}^{J^*}\frac{1}{ \|\mathbf{x}_t^{(k)} - \mathbf{o}_{j^*}\|  - R_{crash}} & \text{ if } J^*>0, \\
        \frac{1}{ \min_{1 \leq j \leq J} \left\| \mathbf{x}_t^{(k)} - \mathbf{o}_{j} \right\|
  - R_{crash}} & \text{ if } J^*=0.
    \end{cases}\label{eq: rdetection}
\end{align} 
where we subtract $R_{crash}$ here as $\|\mathbf{x}_t^{(k)} - \mathbf{o}_{j^*}\|  - R_{crash}$ represents the true distance the $t^{th}$ drone may come to an obstacle without crashing. Now, $\|\mathbf{x}_t^{(k)} - \mathbf{o}_{j^*}\|$  denotes the euclidean distance between the $t^{th}$ drone and the $j^{*^{th}}$ obstacle within radius $R_{detection}$ of the drone and there exists $J^*\leq J$ obstacles within the $t^{th}$ drone's ``detection'' circle. Furthermore, if $J^* = 0$ (that is, if there are no trees within radius $R_{detection}$ from the $t^{th}$ drone) we return the reciprocal of the minimum euclidean distances between the $t^{th}$ drone and all the $J$ obstacles. We observe that, during the optimization process, if it is estimated that $\hat{R}_{{detection}} < R_{{crash}}$, the first case in Equation \ref{eq: rdetection} becomes futile in controlling the movement of non-terminal drones. For example, if $J^* = 1$ (that is, if there is a single obstacle within radius $R_{{detection}}$ of the $t^{\text{th}}$ drone) then the drone would have already collided with that obstacle, rendering it terminal. Consequently, in such scenarios, the only viable control feedback for non-terminal drones is obtained from the second case of Equation \ref{eq: rdetection}, which governs behavior when no obstacles are detected within $R_{detection}$ of the $t^{th}$ drone. We refer to this configuration, with the entirety of Equation \ref{eq: rdetection} dictating the input node, as  $\overset{(\text{I})}{\mathbf{model}}$ with input node $\overset{(\text{I})}{a}(t)^0$ and weights ${\boldsymbol{\theta}}_2^{(\text{I})}$ embedded in ${\boldsymbol{\theta}}^{(\text{I})}$. Now to ensure a meaningful $R_{detection}$ is estimated during the optimization process, we introduce a second model, namely $\overset{(\text{II})}{\mathbf{model}}$, which replaces the second case in Equation \ref{eq: rdetection} with $0 \text{ if } J^* = 0$, to create input node $\overset{(\text{II})}{a}(t)^0$ with weights ${\boldsymbol{\theta}}_2^{(\text{II})}$ as a subset of ${\boldsymbol{\theta}}^{(\text{II})}$.\\

Lastly, since we assume Gaussian priors for our parameters $\boldsymbol{\theta} \in \mathbb{R}^S$, that is $\boldsymbol{\theta} \sim \mathcal{N}(\mathbf{0}, \sigma_{\theta}^2 \mathbf{I}_S)$, where for our case we have $\boldsymbol{\theta}  = \left[\boldsymbol{\theta_1}, \boldsymbol{\theta_2} \right]^{'}$ for which $\boldsymbol{\theta_1}  = \theta_1 \in \mathbb{R}$ and $\boldsymbol{\theta_2}  = \mathbf{w} \in \mathbb{R}^{R}$, we must ensure $\boldsymbol{\phi}\left(\boldsymbol{\theta}_1 ={\theta}_1 \right) = R_{detection} \geq 0$. Accordingly, we define $\boldsymbol{\phi}=\phi$ as a logistic function with a scaling factor $sf$ to ensure that the radius remains positive while being constrained by an upper bound equal to $sf$. More formally, define $\boldsymbol{\phi}=\phi$ as a mapping $\phi : \theta_1 \xrightarrow{\frac{sf}{1+\exp(-\theta_1)}} R_{detection}$, where the upper bound $sf$ is a user-specified parameter determined by the geometry of the problem.

\subsubsection{An arbitrary objective} \label{sec: arb obj drone}
Consider now an arbitrary objective where, for a given parameter configuration $\boldsymbol{\theta} \in  \mathbb{R}^S$ and $T$ drones, we record the relative frequency of successes. That is, we evaluate the arbitrary objective:
\begin{align}
     \operatorname*{argmax}_{\boldsymbol{\theta}}\text{Obj}\left( \boldsymbol{\theta}\right) & = \operatorname*{argmax}_{\boldsymbol{\theta}} \frac{1}{T} \sum_{t = 1}^T \mathbb{I}({s}_t^K(\boldsymbol{\theta}) = + 1) \nonumber \\
    &= \operatorname*{argmax}_{\boldsymbol{\theta}}  \frac{1}{T}k(\boldsymbol{\theta}) \nonumber.
\end{align}
where ${s}_t^K(\boldsymbol{\theta})$ represents the success status of the $t^{th}$ drone after K steps, dependent on our parameter vector $\boldsymbol{\theta}$. Again, we distinguish this as an arbitrary objective since it is not differentiable with respect to the unknown parameter vector and we're imposing a simple goal-based objective of ``win as often as possible'', hoping that maximizing said objective corresponds to the behaviour we want to illicit. \\

By including L$2$ regularization, the L$2$ penalized objective becomes:
\begin{align}
       \operatorname*{argmax}_{\boldsymbol{\theta}} \left( \frac{1}{T}k(\boldsymbol{\theta}) -  \nu \sum_{i = 1}^S \theta_i^2 \right) .\label{eq: bin obj}
\end{align} 

\subsubsection{Likelihoods} \label{sec: likelihoods}
The subsequent section introduces three distinct likelihood formulations, more accurately described as pseudo-likelihoods as previously explained in Section \ref{sec: the likelihood}. The former two are loosely motivated by the fact that for count-based objectives (i.e., objectives quantifying the number of successes), it is natural to construct a likelihood based on the binomial distribution, which defines a probability mass function for a fixed success probability and number of trials. However, as discussed in Section \ref{sec: the obj}, it is necessary that these likelihoods be proportional to the objective function, that is, $p(\mathcal{D} \mid \boldsymbol{\theta} ) \propto \text{Obj}(\boldsymbol{\theta})$. Accordingly, we restructure these likelihoods to ensure they align with this notion. Additionally, Section~\ref{sec: the likelihood} indicates that our likelihoods are tempered; hence, we incorporate a likelihood sharpness parameter $\beta \in \mathbb{R}^+$ to enable control over this.
 \\
 
More broadly, the pseudo-likelihood formulations in this section inherently exhibit differing sharpness (without explicitly inducing sharpness through the parameter $\beta$). Therefore, this section aims to facilitate a discussion on how different likelihoods, with inherently different sharpnesses, can induce distinct behaviours within the two-block MCMC framework.

\paragraph{Binomial-based likelihood} \label{sec: Bin-Based Lik}
We model the likelihood $p(\mathcal{D}  \mid \boldsymbol{\theta})$ as the the probability of observing $k(\boldsymbol{\theta}) = \sum_{t = 1}^T \mathbb{I}({s}_t^K(\boldsymbol{\theta}) = + 1) \leq T$ successes for $T$ total drones given a success probability of $p_{\boldsymbol{\theta}} \approx \frac{k(\boldsymbol{\theta})}{T}$, that is $k(\boldsymbol{\theta})\sim BIN(T,p_{\boldsymbol{\theta}})$ for $k(\boldsymbol{\theta}) \in \{0,1, \ldots,  T\}$.  We note that the number of successes $k(\boldsymbol{\theta})$ is dependent on the specific parameter $\boldsymbol{\theta}$ used, and $p_{\boldsymbol{\theta}}$ is obtained empirically, hence, rendering our likelihood a simulation-based likelihood as follows:
\begin{align}
    p(\mathcal{D} \mid \boldsymbol{\theta}) &= \binom{T}{k(\boldsymbol{\theta})} p_{\boldsymbol{\theta}}^{k(\boldsymbol{\theta})} (1 - p_{\boldsymbol{\theta}})^{T-k(\boldsymbol{\theta})} \nonumber \\
    & = \binom{T}{k(\boldsymbol{\theta})} \left(\frac{k(\boldsymbol{\theta})}{T}\right)^{k(\boldsymbol{\theta})} \left(1 - \left(\frac{k(\boldsymbol{\theta})}{T}\right)\right)^{T-k(\boldsymbol{\theta})}. \label{eq: likelihood drone og}
\end{align}
We emphasize that this is not a classical likelihood function in the strictest sense, as both the observed outcome $k(\boldsymbol{\theta})$ and the estimated success probability $p_{\boldsymbol{\theta}}$ are derived from the same data. This creates a circularity in which the probability of the outcome, $\boldsymbol{\theta}$, is conditioned on a parameter, $p_{\boldsymbol{\theta}}$,  that itself depends on the outcome. We view this pseudo-likelihood as a proxy for how well $\boldsymbol{\theta}$ explains the observed outcomes by quantifying the plausibility of $k(\boldsymbol{\theta})$ under a binomial model with $p_{\boldsymbol{\theta}} \approx \frac{k(\boldsymbol{\theta})}{T}$. \\

Now, one can not assume that $p(\mathcal{D} \mid \boldsymbol{\theta})$ in Equation \ref{eq: likelihood drone og} is monotonic increasing with respect to the number of successes $k(\boldsymbol{\theta})$. In fact, $p(\mathcal{D} \mid \boldsymbol{\theta})$ in Equation \ref{eq: likelihood drone og} exhibits a sole minimum when $k(\boldsymbol{\theta}) = \frac{T}{2}$, Theorem \ref{thm: bin} in conjunction with Lemma \ref{lem: trigamma} clarifies this. 


\begin{lemma} $\frac{1}{n} > \psi^{(1)}(n+1)$ \text{for} $n > 0$. \label{lem: trigamma}

\begin{proof}
    By \cite{guo2015sharp}, we know $\psi^{(1)}(x) < \frac{1}{x+ \frac{1}{2}} + \frac{1}{x^2}$ for $x > 0$. Hence $\psi^{(1)}(n+1) < \frac{1}{(n+1)+ \frac{1}{2}} + \frac{1}{(n+1)^2} < \frac{1}{n+1} + \frac{1}{(n+1)^2}$ for $n > 0$. We show $\frac{1}{n} > \psi^{(1)}(n+1)$ by contradiction. Consider:
    \begin{align}
        \frac{1}{n} & \leq \frac{1}{n+1} + \frac{1}{(n+1)^2} \nonumber \\
        \therefore 1 & \leq \frac{n^2 + 2n}{n^2 +2n +1} \nonumber \\
        \therefore 1 & \leq 0 \nonumber
    \end{align}
    Which is a contradiction, implying $\frac{1}{n} > \frac{1}{n+1} + \frac{1}{(n+1)^2} > \psi^{(1)}(n+1)$ for $n > 0$.
\end{proof}
\end{lemma} 

\begin{theorem} $f(x) = \binom{T}{x}(\frac{x}{T})^{x} (1 - \frac{x}{T})^{T-x}$ exhibits a sole minimum at $x = \frac{T}{2}$ for $x \in [0, T]$ with $T \in (0, \infty)$, with $f(x)$ being symmetric about $x = \frac{T}{2}$. \label{thm: bin}

\begin{proof}
Since $\lim_{x \to 0^+} f(x) = 1$ and $\lim_{x \to T^-} f(x) = 1$, we know $f(x)$ is continuous on the closed interval $[0, T]$. Consider:
    \begin{align}
       l(x) &= \log[f(x)] = \log\left(T!\right) - \log\left(x!\right) - \log\left[(T-x)!\right] + x \log\left(\frac{x}{T}\right) + (T-x) \log\left(\frac{T-x}{T}\right). \nonumber
    \end{align}
    Using $\log\left(n!\right) = \log\left[\Gamma(n+1)\right]$ we have:
    \begin{align}
        l(x) &= \log\left[\Gamma(T+1)\right] - \log\left[\Gamma(x+1)\right] - \log\left[\Gamma(T-x+1)\right] + x \log\left(\frac{x}{T}\right) + (T-x) \log\left(\frac{T-x}{T}\right). \nonumber
    \end{align} 
    Taking the first derivative with respect to $x$ and using the digamma function $\psi(n) = \frac{d}{dn} \log\left[\Gamma(n)\right]$:
    \begin{align}
        l'(x) &= -\psi(x+1) + \psi(T-x+1) + \log\left(\frac{x}{T-x}\right). \nonumber
    \end{align}
    Since $l'(\frac{T}{2}) = -\psi(\frac{T}{2}+1) + \psi(\frac{T}{2}+1) + \log\left(1\right) = 0$, we know $f(x)$ has a critical point at $x = \frac{T}{2}$. We now show $f''(x) > 0$ on the interval $(0, T)$ implying $f(x)$ is convex and has most one minimum on this interval. We use the trigamma function $\frac{d \psi(n)}{dn} = \psi^{(1)}(n)$ and consider:
    \begin{align}
        l''(x) &=  -\psi^{(1)}(x+1) - \psi^{(1)}(T-x+1) +\frac{1}{x} + \frac{1}{T-x}. \nonumber
    \end{align}
    By Lemma \ref{lem: trigamma}, since $\frac{1}{x} >  \psi^{(1)}(x+1)$ and $\frac{1}{T-x} > \psi^{(1)}(T-x+1)$, we have $l''(x) > 0$ on $x \in (0, T)$. Additionally, since:
    \begin{align}
         f(T-x) &= \binom{T}{T-x} \left(\frac{T-x}{T}\right)^{T-x} \left(1 - \frac{T-x}{T}\right)^{T-(T-x)} \nonumber \\
        &= \binom{T}{x} \left(\frac{x}{T}\right)^{x} \left(1 - \frac{x}{T}\right)^{T-x} \nonumber \\
        &= f(x). \nonumber
    \end{align}
    we have shown that $f(x)$ is symmetric about $x = \frac{T}{2}$.
    \end{proof}
\end{theorem}

To ensure a monotonic increasing likelihood $p(\mathcal{D} \mid \boldsymbol{\theta})$ on interval $k(\boldsymbol{\theta}) \in \{0,1, \ldots,  T\}$, we propose artificially changing the structure of the likelihood given in Equation \ref{eq: likelihood drone og}. Since Theorem \ref{thm: bin} implies that Equation \ref{eq: likelihood drone og} is monotonic decreasing on $k(\boldsymbol{\theta}) \in \{0,1, \ldots, \frac{T}{2} \}$, after which being monotonic increasing on $k(\boldsymbol{\theta}) \in \{\frac{T}{2}, \frac{T}{2}+1, \ldots,  T\}$, to enforce monotonicity over the entire interval, we define a piecewise function $h(x)$ as follows:
\begin{align}
    h(x) &= 
    \begin{cases}
        g(x), & \text{if } x < \frac{T}{2}, \\    
        f(x), & \text{if } x \geq \frac{T}{2}. \nonumber
    \end{cases}
\end{align}
where $f(x) = \binom{T}{x} (\frac{x}{T})^{x} (1 - \frac{x}{T})^{T-x}$ for $x \in \{0,1, \ldots,  T \}$ represents the original binomial-based likelihood with $g(x) =  a x $ merely being a linear function where scaling factor $a = \frac{2}{T}f\left( \frac{T}{2}\right)$ has a dual function of ensuring continuity at $x = \frac{T}{2}$ as well as ensuring $g(x) \geq 0$ for $x \in \{0, 1, \ldots, \frac{T}{2}\} $. Additionally, one could make the argument that any monotonic increasing function $g(x)$ could be used, satisfying  $g(x) \geq 0$ for $x \in \{0, 1, \ldots, \frac{T}{2}\}$ as well as the continuity constraint of $g(\frac{T}{2}) = f(\frac{T}{2})$. This notion is further elaborated on in Section \ref{sec: gx forms}. \\

To allow for sharper likelihoods, specifically, to increase the rate of change of $h(x)$
over the domain, we introduce a sharpness parameter $\beta \in \mathbb{R}^+$. This parameter modulates the steepness of the likelihood function by exponentiating it directly. That is, we define $ \big[ h(x) \big]^\beta$, where larger values of $\beta$ yield a more pronounced increase in likelihood as a function of $k(\boldsymbol{\theta})$, effectively sharpening the likelihood surface, further elucidated in Section \ref{sec: beta}. We exponentiate $h(x)$ by $\beta$, rather than scale it via $\beta \cdot h(x)$, because our focus is on amplifying the steepness of the log-likelihood, which plays a central role in acceptance probability computations (as seen in Equation \ref{eq: log alpha}). Specifically, taking the logarithm of $\left[h(x)\right]^\beta$ yields $\beta \cdot\log \left[h(x)\right]$, thereby linearly scaling the log-likelihood. Alternatively, the parameter $\beta$ can be interpreted as a means of amplifying the likelihood ratio in Equation \ref{eq: alpha}, yielding the modified expression $\left(\frac{p(\mathcal{D} \mid \boldsymbol{\theta}^*)}{p(\mathcal{D} \mid \boldsymbol{\theta}^{(j)})}\right)^\beta$. Increasing $\beta$ makes the Markov chain more inclined to accept proposed solutions $\boldsymbol{\theta}^*$ that yield higher objective values, given the proportionality $p(\mathcal{D} \mid \boldsymbol{\theta}) \propto \text{Obj}(\boldsymbol{\theta})$, in effect, making the MCMC sampler more likelihood-driven. Thus, the prior ratio plays a reduced role in the acceptance step.\\

\begin{figure}[H]
    \centering
    \begin{minipage}{0.45\textwidth}
            \centering
        \includegraphics[width=\linewidth]{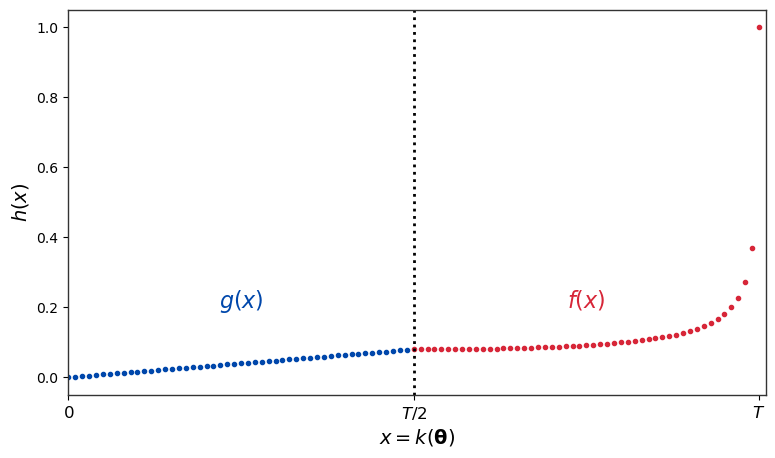}
        \caption{$h(x)$ on interval $[0, 1]$ for $x =k(\boldsymbol{\theta}) \in \{0, 1, \ldots, T\}$.}
        \label{fig: hx}
    \end{minipage}
    \hfill
    \begin{minipage}{0.45\textwidth}
                \centering
        \includegraphics[width=\linewidth]{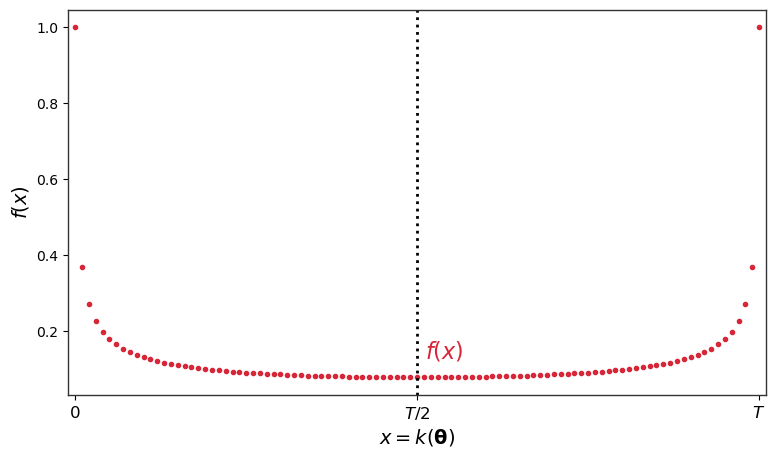}
        \caption{$f(x)$ on interval $[0, 1]$ for $x = k(\boldsymbol{\theta}) \in \{0, 1, \ldots, T\}$}.
        \label{fig: fx}
    \end{minipage}
\end{figure}

\begin{figure}[H]
    \centering
        \includegraphics[width=0.45\linewidth]{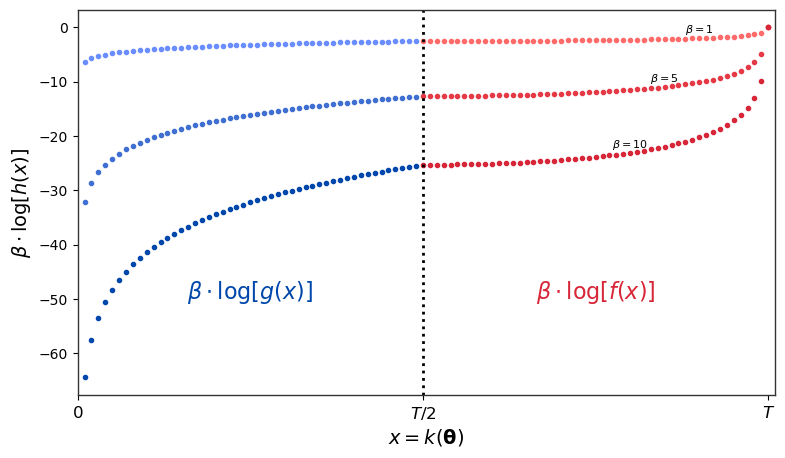}
        \caption{$\beta \cdot \log \left[ h(x) \right]$ for $x =k(\boldsymbol{\theta}) \in \{0, 1, \ldots, T\}$ for various $\beta$.}
        \label{fig: log likelihood bin}
    \end{figure}

Furthermore, for $\boldsymbol{\theta} \in \mathbb{R}^S$, $k(\boldsymbol{\theta}) \in \{0, 1, \ldots, T\}$ and sharpness $\beta \in \mathbb{R}^+$, we have our new likelihood as:
\begin{align}
    p(\mathcal{D} \mid \boldsymbol{\theta}) &= \big[h \left(k(\boldsymbol{\theta})  \right)\big]^\beta. \label{eq: likelihood bin}
\end{align}
Clearly, $\operatorname*{argmax}_{\boldsymbol{\theta}} p( \boldsymbol{\theta} \vert \mathcal{D})$, would now be equivalent to maximizing the L2 penalized objective in Equation \ref{eq: bin obj} due to the monotonic increasing nature of $h(x)$ in Equation \ref{eq: likelihood bin}, as displayed in Figure \ref{fig: hx} and compared to the original likelihood of Equation \ref{eq: likelihood drone og}, in Figure \ref{fig: fx}. Additionally, Figure \ref{fig: log likelihood bin} illustrates the scaled log-likelihood, $\beta \cdot \log \left[ h(x) \right]$, highlighting how increasing the sharpness parameter $\beta$ amplifies the curvature of the log-likelihood. This results in a steeper surface, thereby enhancing the sensitivity of the likelihood to changes in $x = k(\boldsymbol{\theta})$.

\subparagraph{Rebuttals against different $g(x)$ forms:} \label{sec: gx forms}
It is important to re-emphasize that the modified likelihood consisting of $h \left( k(\boldsymbol{\theta})  \right)  $ is constructed primarily to facilitate efficient sampling in an MCMC context, where the goal is to identify the mode of the conditional posterior, $p(\boldsymbol{\theta} \mid \sigma_\theta^2, \mathcal{D})$. Since the mode necessarily lies in the high-likelihood (high-valued objective) region (due to $p(\boldsymbol{\theta} \mid \sigma_\theta^2, \mathcal{D}) \propto p(\mathcal{D} \mid \boldsymbol{\theta}) \cdot p(\boldsymbol{\theta} \mid \sigma_\theta^2)$), specifically in the upper half of the domain $k(\boldsymbol{\theta}) \in \{\frac{T}{2}, \frac{T}{2}+1, \ldots, T\}$, where the original likelihood function $f \left(k(\boldsymbol{\theta})  \right) $ is already monotonic increasing, our primary interest lies in accurately sampling from this latter half. The lower half $ k(\boldsymbol{\theta})\in \{0, 1, \ldots, \frac{T}{2}\}$, where the original likelihood is decreasing, serves primarily as a transitional region that we wish to exit efficiently during sampling. As such, the choice of $g \left(k(\boldsymbol{\theta})  \right) $ on this interval can be quite flexible: it need only be monotonic increasing and continuous at $k(\boldsymbol{\theta}) = \frac{T}{2}$. For this reason, we adopt a simple linear form $g\left(k(\boldsymbol{\theta})  \right) =a \cdot k(\boldsymbol{\theta}) $, which accelerates the sampler's movement through low-likelihood regions and thus enhances convergence toward the high-likelihood regions that contribute meaningfully to the conditional posterior mode. This design choice is well justified given that our objective is not full posterior sampling, but rather efficient localization of the posterior mode (achieved by deliberately concentrating samples around it).

\subparagraph{Sharpness $\beta \in \mathbb{R}^+$:} \label{sec: beta} 
In the context of the two-block MCMC framework, wherein a pseudo-likelihood is modulated by a sharpness parameter $\beta$, this parameter plays a critical role in shaping the behaviour of the Markov chain by governing the peakedness of the likelihood. Since $\beta$ exponentiates $p(\boldsymbol{\theta} \mid \mathcal{D})$, not only is the likelihood's rate of change across the domain increased, but the density itself becomes sharper in the sense that mass is more strongly concentrated around its mode. This is because exponentiating the likelihood amplifies higher values and suppresses lower ones, thereby steepening the posterior landscape. Specifically, a high value of $\beta$ accentuates the pseudo-likelihood, yielding a sharply peaked conditional posterior $p(\boldsymbol{\theta} \mid \sigma_\theta^2, \mathcal{D})$, given that $p(\boldsymbol{\theta} \mid \sigma_\theta^2, \mathcal{D}) \propto p(\mathcal{D} \mid \boldsymbol{\theta}) \cdot p(\boldsymbol{\theta} \mid \sigma_\theta^2)$, and thus concentrates samples around the mode. This enhances exploitation by focusing the chain on high-likelihood regions. However, in the presence of multimodality, a large $\beta$ may cause the chain to become trapped in a single dominant mode, hindering exploration of other dominant modes, as proposals that move away from the current region can receive extremely low acceptance probabilities $\alpha_{\theta}$. Conversely, a low $\beta$ reduces the sharpness of the likelihood, flattening the conditional posterior and enabling the chain to move more freely across the parameter space, thereby promoting exploration of multiple modes at the expense of slower convergence to high-likelihood regions. The choice of $\beta$ therefore embodies a trade-off between exploration and exploitation in the MCMC process.\\

An alternative perspective on the role of the sharpness parameter $\beta$ is that it controls the number and prominence of modes in the conditional posterior. For example, a low $\beta$ flattens the likelihood, allowing multiple regions of the parameter space to ``compete" for conditional posterior mass and thereby inducing multimodality. In contrast, a high $\beta$ sharpens the conditional posterior, collapsing it onto dominant modes and potentially suppressing minor alternatives.

\paragraph{Beta-based likelihood}
Invoking the identity $n! = \Gamma(n + 1)  \text{ for } n \in \mathbb{W}$, Equation \ref{eq: likelihood drone og} may be written as such:

\begin{align}
  p(\mathcal{D} \mid \boldsymbol{\theta})&= \frac{\Gamma\left( T+1\right)}{\Gamma\left(k(\boldsymbol{\theta})+1\right)\cdot \Gamma\left(T - k(\boldsymbol{\theta})+1\right)} \left(\frac{k(\boldsymbol{\theta})}{T}\right)^{k(\boldsymbol{\theta})} \left(1 - \left(\frac{k(\boldsymbol{\theta})}{T}\right)\right)^{T-k(\boldsymbol{\theta})}, \nonumber
\end{align}
for $k(\boldsymbol{\theta}) \in \{0,1, \ldots,  T\}$. Yet, since we would like to model the proportion of successes $\frac{1}{T}k(\boldsymbol{\theta}) \in [0, 1]$, yet retain the constraint of the original likelihood  $ p(\mathcal{D} \mid \boldsymbol{\theta}) \in [0, 1]$ (better seen as $f(x)$ in Figure \ref{fig: fx}), we model the likelihood as such:
\begin{align}
    p(\mathcal{D} \mid \boldsymbol{\theta}) 
    &= \frac{\Gamma\left( 2\right)}{\Gamma\left(\frac{1}{T}k(\boldsymbol{\theta})+1\right)\cdot \Gamma\left(2 - \frac{1}{T}k(\boldsymbol{\theta})\right)} \left(\frac{k(\boldsymbol{\theta})}{T}\right)^{\frac{1}{T}k(\boldsymbol{\theta})}  \left(1 - \left(\frac{k(\boldsymbol{\theta})}{T}\right)\right)^{1-\frac{1}{T}k(\boldsymbol{\theta})}, \label{eq: likelihood drone beta}
\end{align} 
for $\frac{1}{T}k(\boldsymbol{\theta}) \in \{0, \frac{1}{T},\frac{2}{T}, \ldots, 1 \}$. Rescaling Equation \ref{eq: likelihood drone og}, using the proportion of successes, $\frac{1}{T} k(\boldsymbol{\theta})\in [0,1]$, instead of the number of successes, $k(\boldsymbol{\theta})$, resembles a binomial distribution for a single trial, $T = 1$. Now, as before, since Theorem \ref{thm: bin} implies that Equation \ref{eq: likelihood drone beta} is monotonic decreasing on $\frac{1}{T} k(\boldsymbol{\theta})\in[0, \frac{1}{2})$, after which being monotonic increasing on $ \frac{1}{T}k(\boldsymbol{\theta}) \in[\frac{1}{2}, 1]$, to enforce monotonicity over the entire interval, we define a piecewise function $h(x)$ as follows:
\begin{align}
    h(x) &= 
    \begin{cases}
        g(x), & \text{if } x < \frac{1}{2}, \\    
        f(x), & \text{if } x \geq \frac{1}{2}, \nonumber
    \end{cases}
\end{align}
where $f(x) = \frac{\Gamma(2)}{\Gamma(x+1)\Gamma(2 - x)} (x)^{x} (1 - x)^{1-x}$ for $x \in [0, 1]$ represents the original likelihood in Equation \ref{eq: likelihood drone beta}. Furthermore, we note that $f(x)$ is proportional to the density function of a $\text{Beta}(\alpha, 3 - \alpha)$ distribution scaled by $\frac{1}{2}$ where $\alpha =x+ 1$. Now $g(x) =  a x $ is a linear function with scaling factor $a = 2f\left( \frac{1}{2}\right)$. 
Furthermore, for $\boldsymbol{\theta} \in \mathbb{R}^S$, $\frac{1}{T}k(\boldsymbol{\theta}) \in [0, 1]$ and sharpness $\beta \in \mathbb{R}^+$, we have our new likelihood as:
\begin{align}
   p(\mathcal{D} \mid \boldsymbol{\theta}) &=  \left[ h\left(  \frac{1}{T}k(\boldsymbol{\theta})  \right)\right]^\beta. \label{eq: likelihood drone beta new}
\end{align}

\begin{figure}[H]
    \centering
    \begin{minipage}{0.45\textwidth}
            \centering
        \includegraphics[width=\linewidth]{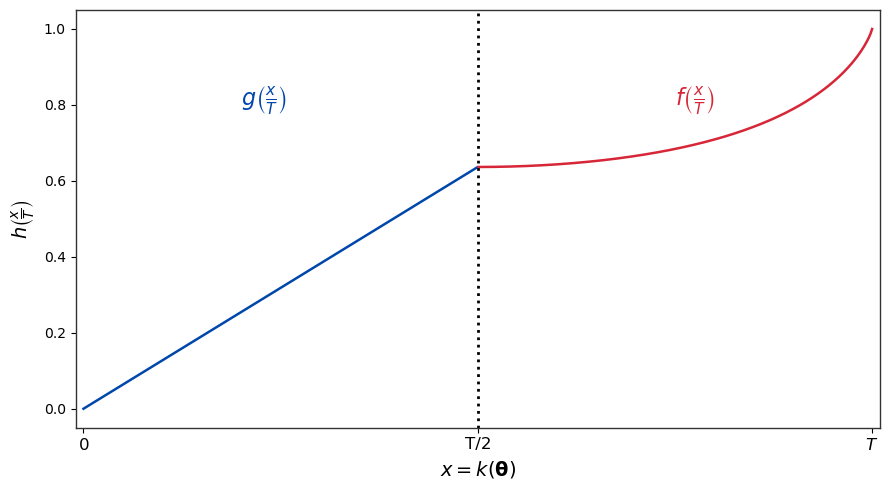}
        \caption{$\beta \cdot h\left(\frac{x}{T}\right)$ on interval $[0, \beta]$ for $x =k(\boldsymbol{\theta}) \in \{0, 1, \ldots, T\}$ for various $\beta$.}
        \label{fig: hx_beta}
    \end{minipage}
    \hfill
    \begin{minipage}{0.45\textwidth}
                \centering
        \includegraphics[width=\linewidth]{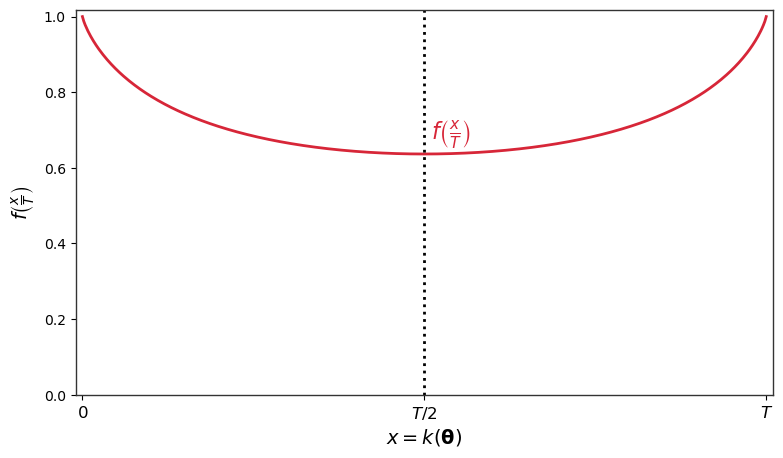}
        \caption{$f\left(\frac{x}{T}\right)$ on interval $[0, 1]$ for $x =k(\boldsymbol{\theta}) \in \{0, 1, \ldots, T\}$}
        \label{fig: fx_beta}
    \end{minipage}
\end{figure}

\begin{figure}[H]
    \centering
        \includegraphics[width=0.45\linewidth]{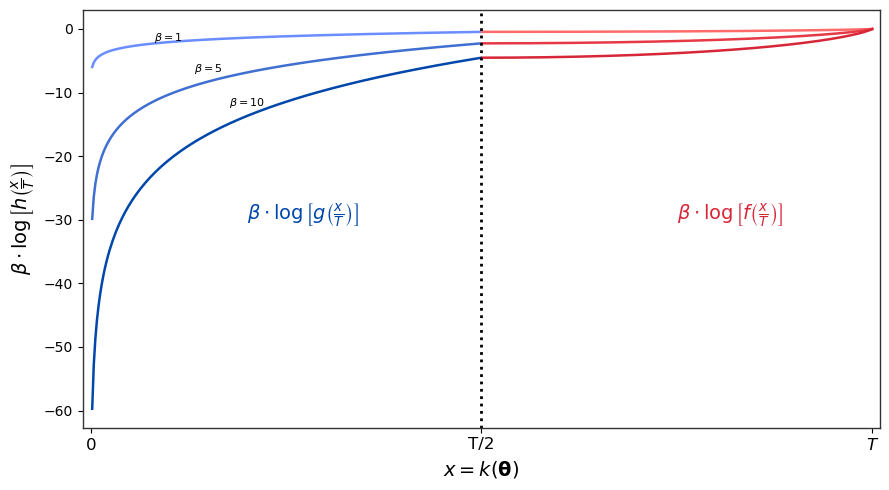}
        \caption{$\beta \cdot \log \left[ h(x) \right]$ for $x =k(\boldsymbol{\theta}) \in \{0, 1, \ldots, T\}$ for various $\beta$.}
        \label{fig: log likelihood beta}
    \end{figure}

Figure \ref{fig: hx_beta} illustrates the re-scaled likelihood function (beta-based) from Equation \ref{eq: likelihood drone beta new}, while Figure \ref{fig: fx_beta} displays the original likelihood from Equation \ref{eq: likelihood drone beta}. Additionally, Figure \ref{fig: log likelihood beta} illustrates the scaled log-likelihood, $\beta \cdot \log \left[ h(x) \right]$ for various $\beta$.

\begin{figure}[H]
    \centering
    \begin{minipage}{0.45\textwidth}
            \centering
        \includegraphics[width=\linewidth]{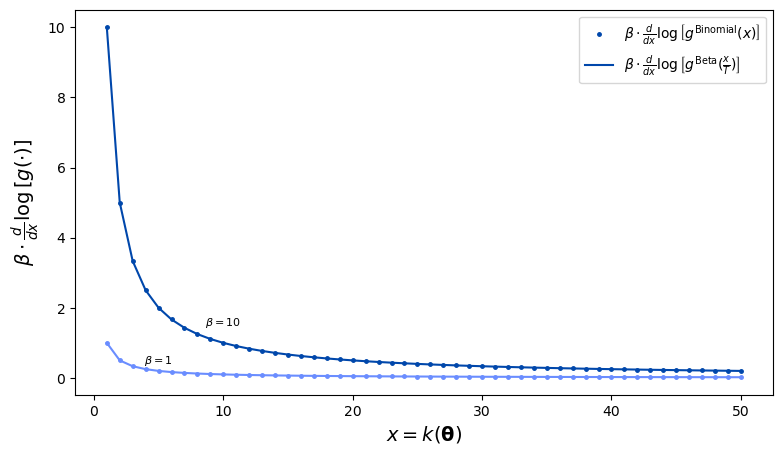}
        \caption{$\beta \cdot \frac{d}{dx} \log \left[ g\left(\cdot\right)\right]$ on interval $[0, \beta]$ for $x =k(\boldsymbol{\theta}) \in \{0, 1, \ldots, T\}$ for various $\beta$.}
        \label{fig: der_gx}
    \end{minipage}
    \hfill
    \begin{minipage}{0.45\textwidth}
                \centering
        \includegraphics[width=\linewidth]{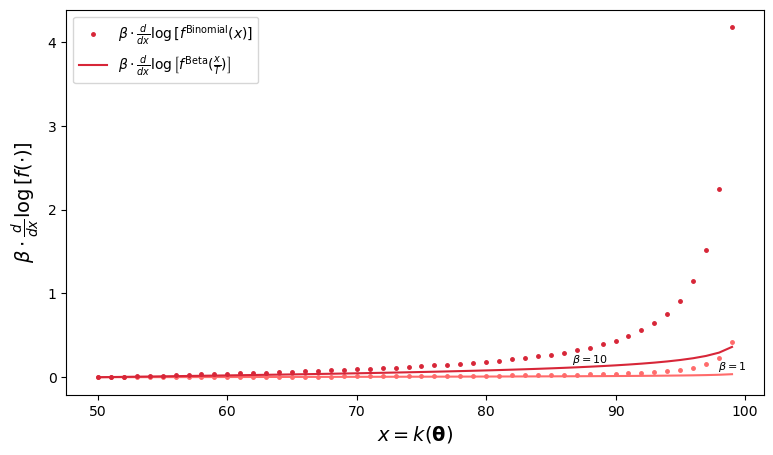}
        \caption{$\beta \cdot \frac{d}{dx} \log \left[ f\left(\cdot\right)\right]$ for $x =k(\boldsymbol{\theta}) \in \{0, 1, \ldots, T\}$}
        \label{fig: der_fx}
    \end{minipage}
\end{figure}

A key distinction between the beta-based likelihood in Equation \ref{eq: likelihood drone beta new} and the binomial-based likelihood in Equation \ref{eq: likelihood bin} (see Section \ref{sec: Bin-Based Lik}) lies in the differing rates of change across the support $k(\boldsymbol{\theta})  \in \{0, 1 \ldots, T\}$. Specifically, the binomial-based log-likelihood exhibits a relatively flat profile across much of the support, followed by a pronounced increase in steepness at higher values of $k(\boldsymbol{\theta})$. This behavior is visually evident in Figure \ref{fig: der_fx}, where the gradient of the binomial-based log-likelihood clearly exceeds that of the beta-based log-likelihood for large $k(\boldsymbol{\theta})$. In the context of the MCMC algorithm, such a steep ascent corresponds to a more sharply peaked posterior $p(\boldsymbol{\theta} \mid \sigma_\theta^2, \mathcal{D})$, hence acceptance into higher likelihood regions, $k(\boldsymbol{\theta}) \in \{\frac{T}{2} +1, \frac{T}{2} +2 ,  \ldots,  T\}$, are likely to be accelerated when using the binomial-based likelihood. It is also worthwhile to note that the derivative of both binomial and beta-based log-likelihoods for domain $k(\boldsymbol{\theta}) \in \{0, 1, \ldots, \frac{T}{2} \}$ are equal, that is, $\beta \cdot \log \left[ \frac{d}{dx}g^\mathrm{Binomial}(x) \right] = \beta \cdot \log \left[ \frac{d}{dx}g^\mathrm{Beta}(\frac{x}{T}) \right] = \beta \cdot \frac{1}{x}$ as shown in Figure \ref{fig: der_gx}. This similarity implies that, from an MCMC perspective, transitions out of lower-likelihood regions, $k(\boldsymbol{\theta}) \in \{0, 1, \ldots, \frac{T}{2} \}$, into the higher likelihood regions, $k(\boldsymbol{\theta}) \in \{\frac{T}{2} +1, \frac{T}{2} +2 ,  \ldots,  T\}$, are likely to proceed with similar efficiency under both likelihoods. It is important to note, however, that the derivatives depicted reflect infinitesimal changes in the log-likelihood. In contrast, the MH algorithm typically evaluates differences over finite, and often substantial, differences in log-likelihood values. As such, our derivative plots provide interpretive value primarily in scenarios where the proposed $\boldsymbol{\theta}$ yield objective function values of comparable magnitude.

\paragraph{Exponential-based likelihood}
Now to reiterate, in the context of this work, our primary objective is not to perform full Bayesian inference via MCMC, but rather to identify the mode of the conditional $p( \boldsymbol{\theta} \mid \sigma_\theta^2, \mathcal{D})$. In such cases, the MCMC algorithm is not employed for its traditional role in posterior sampling, but rather as a stochastic optimization tool that facilitates a guided random search over the parameter space. From this perspective, strict adherence to the exact posterior structure is unnecessary. It suffices that the proposal mechanism is guided by a function that monotonically increases with the objective of interest, thereby biasing the random walk toward high-likelihood (or high-valued objective) regions. In treating MCMC as a mode-seeking algorithm, the requirement for an explicit, well-defined likelihood linking $\boldsymbol{\theta}$ to the data becomes less critical.\\

Being such, we utilize an exponential function as said monotonic increasing function as a substitute for a well-defined likelihood where we define $h(x) = \exp (x)$ for $x >0$, analogous to the procedure used in SA. Hence for $\boldsymbol{\theta} \in \mathbb{R}^S$, $\frac{1}{T}k(\boldsymbol{\theta}) \in [0, 1]$ and sharpness $\beta \in \mathbb{R}^+$, we have our new likelihood as:
\begin{align}
    p(\mathcal{D} \mid \boldsymbol{\theta}) &=  \left[ h\left(  \frac{1}{T}k(\boldsymbol{\theta})  \right) \right]^\beta \nonumber \\
    & = \exp\left(\beta\cdot \frac{1}{T}k(\boldsymbol{\theta})   \right).\label{eq: likelihood drone exp}
\end{align}

\begin{figure}[H]
    \centering
    \begin{minipage}{0.45\textwidth}
        \includegraphics[width=\linewidth]{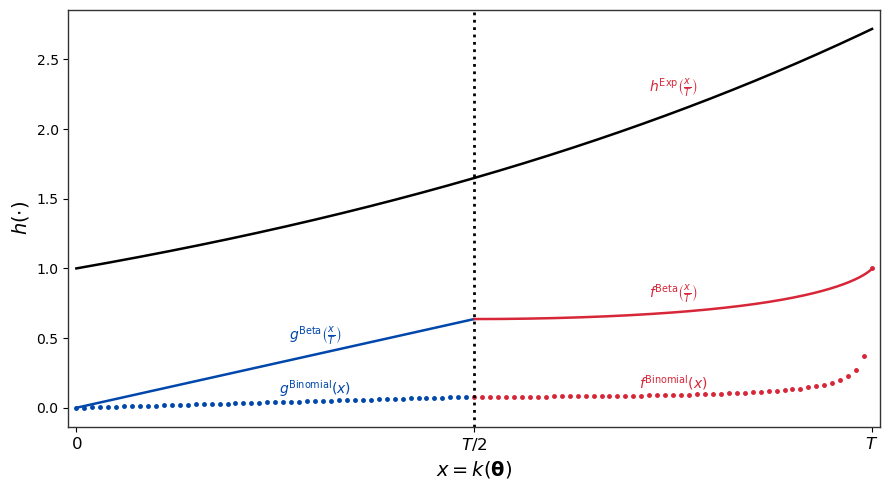}
        \caption{$h^{\mathrm{Binomial}}\left(x \right),  h^{\mathrm{Beta}}\left(\frac{x}{T} \right)$ and $ h^{\mathrm{Exp}}\left(\frac{x}{T} \right)$  for $x =k(\boldsymbol{\theta}) \in \{0, 1, \ldots, T\}$.}
        \label{fig: exp}
    \end{minipage}
    \hfill
    \begin{minipage}{0.45\textwidth}
                \centering
        \includegraphics[width=\linewidth]{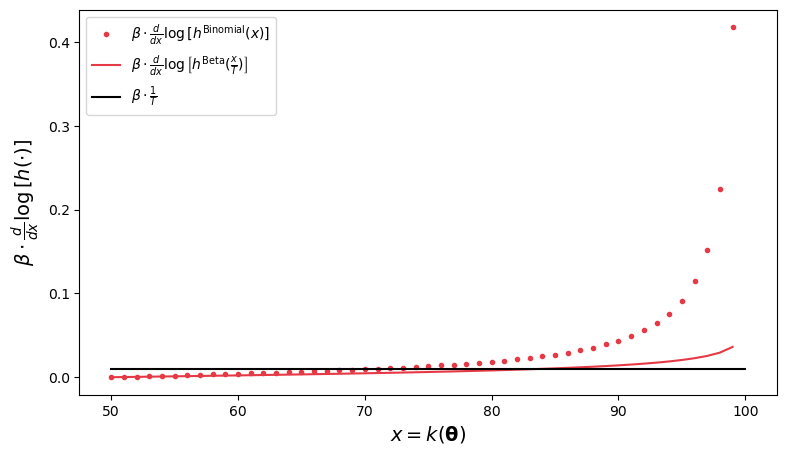}
        \caption{$\beta \cdot \frac{d}{dx} \log \left[ h\left(\cdot\right)\right]$ for $x =k(\boldsymbol{\theta}) \in \{\frac{T}{2}, \frac{T}{2}+1, \ldots, T\}$ for $\beta = 1$.}
        \label{fig: der_exp}
    \end{minipage}
\end{figure}

We observe in Figure \ref{fig: exp} a comparison between the exponential likelihood defined in Equation \ref{eq: likelihood drone exp} and the alternative likelihood formulations given in Equations \ref{eq: likelihood bin} and \ref{eq: likelihood drone beta new}. Given that the MH algorithm bases proposal acceptance on the ratio of likelihoods, more precisely, the difference in log-likelihoods, the absolute magnitude of the likelihood function is of limited relevance. Instead, the relative rate of change, as illustrated in Figure \ref{fig: der_exp}, offers more informative insight into how the Markov chain is guided through the parameter space.  \\

As illustrated in Figure \ref{fig: der_exp}, the derivative of the exponential-based log-likelihood, given by $\beta \cdot \frac{1}{T}$, exceeds the gradients of the alternative log-likelihoods over much of the domain $k(\boldsymbol{\theta}) \in \left\{ \frac{T}{2}, \frac{T}{2}+1, \ldots, T \right\}$. However, this dominance progressively diminishes as $k(\boldsymbol{\theta})$ increases. In contrast, the derivative of the binomial-based log-likelihood surpasses those of the alternatives at higher values of $k(\boldsymbol{\theta})$. In this regard, the binomial-based formulation may be considered advantageous, as its steeper gradient in the upper region of the domain can facilitate more dynamic transitions toward higher-valued objective areas, potentially reducing the risk of the Markov chain becoming trapped in local optima.

\subsubsection{Results: Effects of regularization} \label{sec: effects of reg drone}
\begin{figure}[H]
    \centering
    \begin{subfigure}[t]{0.48\linewidth}
        \centering
        \includegraphics[width=\linewidth]{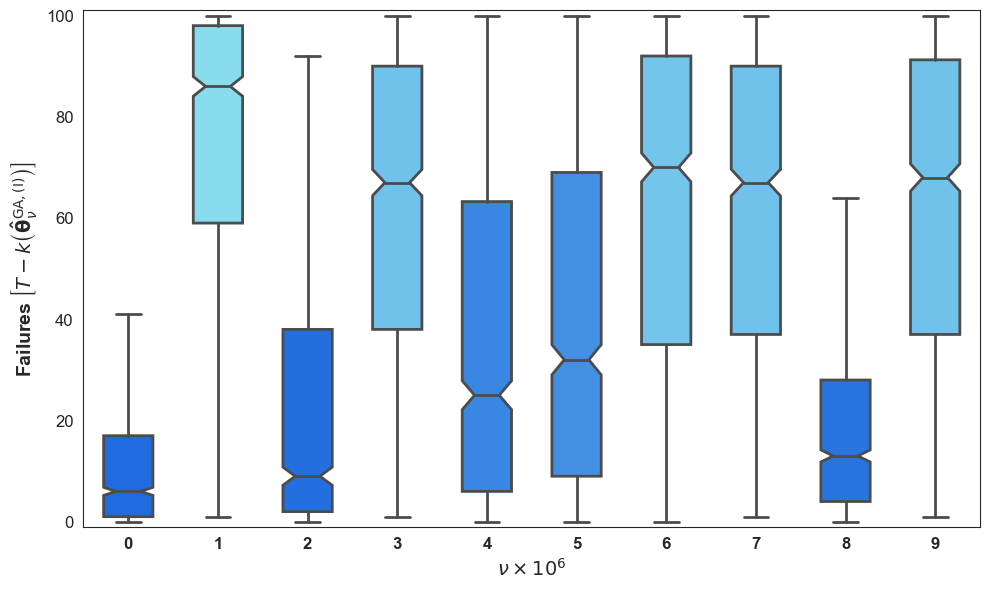}
        \caption*{$\overset{(\text{I})}{\mathbf{model}}$}
    \end{subfigure}
    \hfill
    \begin{subfigure}[t]{0.48\linewidth}
        \centering
        \includegraphics[width=\linewidth]{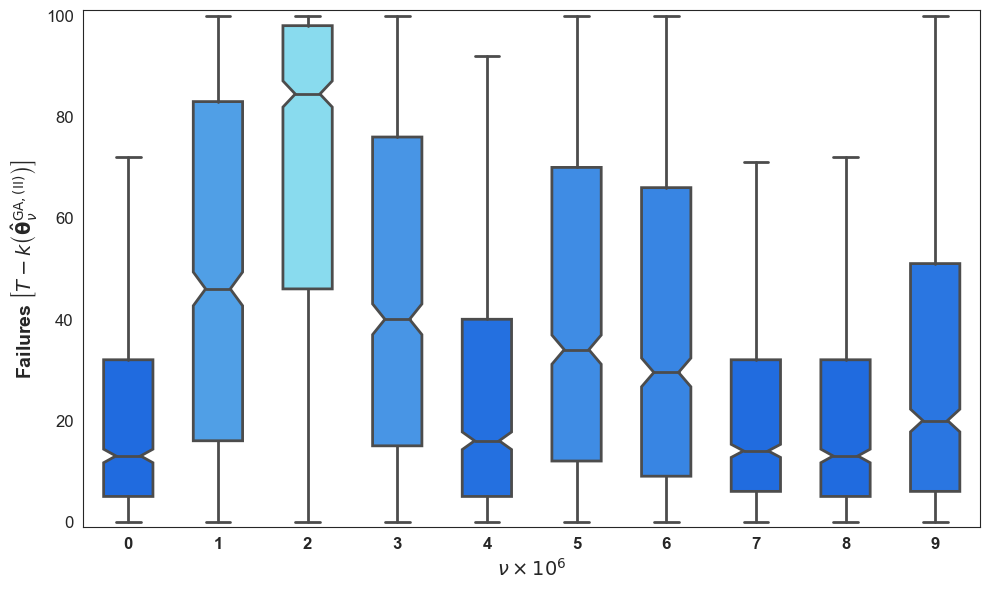}
        \caption*{$\overset{(\text{II})}{\mathbf{model}}$}
    \end{subfigure}
    \captionsetup{justification=centering}
    \caption{Boxplots illustrating distributions of failures, $T - k\left(\boldsymbol{\hat{\theta}}^{\text{GA}}_{\nu}\right)$, against varying $\nu$ obtained through 1000 test initializations for $T = 100$ for $\overset{(\text{I})}{\mathbf{model}}$ (left) and $\overset{(\text{II})}{\mathbf{model}}$ (right).}
    \label{fig: failure dist}
\end{figure}

To preliminarily evaluate the effect of regularization on the performance of 
$\boldsymbol{\theta}$, we assess generalization under a given regularization strength $\nu$. Specifically, we apply the estimator $\boldsymbol{\hat{\theta}}^{\text{GA}}_{\nu}$, obtained via a GA with $L_{2}$ regularization, to a newly initialized environment, defined by a distinct set of $J$ obstacle coordinates at the $k^{\text{th}}$ iteration, $\mathbf{o}_j^{(k)}$ for $j = 1, 2, \ldots, J$. These coordinates are governed by a new set of angular frequencies $\omega_j$, phase shifts $\phi_j$, and orbital radii $r_j$. Additionally, a new initialization is also characterized by $T$ new initial drone coordinates, $\mathbf{x}_t^{(0)}$ for $t = 1, 2, \hdots, T$. We dictate our training initialization by using seed value $\omega_0^{\text{Train}}$, with $1000$ test initializations governed by seed values  $\{\omega_j^{\text{Test}} \}_{j=1}^{1000}$. More formally, our game-updating equation becomes $\mathbf{x}_t^{(k+1)} = \mathbf{x}_t^{(k)} + \mathbf{ct}(\mathbf{x}_t^{(k)}, \boldsymbol{\hat{\theta}}^{\text{GA}}_{\nu})\delta_t$ for which $\forall j, k: \mathbf{o}_j^{(k)} \cap {\mathbf{o}_j^{(k)}}^{\text{Train}} = \emptyset$ and $\forall t: \mathbf{x}_t^{(0)} \cap {\mathbf{x}_t^{(0)}}^{\text{Train}} = \emptyset$. Figure \ref{fig: failure dist} illustrates the distributions of failures, where the number of failures are defined as $T - k\left(\boldsymbol{\hat{\theta}}^{\text{GA}}_{\nu}\right)$, for $1000$ of these initializations against varying regularization strengths $\nu$ for both $\overset{(\text{I})}{\mathbf{model}}$ and $\overset{(\text{II})}{\mathbf{model}}$. Table \ref{tab: success dist} displays the median and mean of the success distributions, where the number of successes are defined as $k\left(\boldsymbol{\hat{\theta}}^{\text{GA}}_{\nu} \right)$, as well as displaying $\hat{R}_{detection}$ which is subsumed in $\boldsymbol{\hat{\theta}}^{\text{GA}}_{\nu}$. \\

With respect to $\overset{(\text{I})}{\mathbf{model}}$, Table \ref{tab: success dist} indicates that improved out-of-sample performance is achieved when the optimization procedure estimates $\hat{R}_{{detection}} < R_{{crash}} = 0.05$ (see \ref{app: nav specs} for additional specifications). This suggests that only the second case of Equation \ref{eq: rdetection} is necessary to govern the behavior of non-terminal drones in order to enhance out-of-sample performance. In contrast, for $\overset{(\text{II})}{\mathbf{model}}$, Table \ref{tab: success dist} suggests that improved out-of-sample performance is attained when the estimated detection radius satisfies $\hat{R}_{{detection}} \approx 0.07$. These notions are further supported by the results in Table \ref{tab: success dist large}.
\\

Furthermore, no consistent pattern in drone success rates on the out-of-sample set is observed as $\nu$ varies, as illustrated in Figure \ref{fig: failure dist}. Instead, performance appears to depend primarily on whether $\hat{R}_{{detection}}$ is estimated to be greater or less than the crash radius $R_{{crash}} = 0.05$ under $\overset{(\text{I})}{\mathbf{model}}$, or whether $\hat{R}_{{detection}} \approx 0.07$ under $\overset{(\text{II})}{\mathbf{model}}$. The results suggest that there are many plausible solutions that yield high objective values on the in-sample set, although only a subset of these generalize well to the out-of-sample set.

\begin{table}[H]
\centering
\resizebox{\textwidth}{!}{%
        \begin{tabular}{ccccccccc}
            & \multicolumn{4}{c}{$\overset{(\text{I})}{\mathbf{model}}\left(\overset{(\text{I})}{a}(t)^0, {\boldsymbol{\hat{\theta}}}_{2, \nu}^{\text{GA}, (\text{I})} \right)$} & \multicolumn{4}{c}{$\overset{(\text{II})}{\mathbf{model}}\left(\overset{(\text{II})}{a}(t)^0, {\boldsymbol{\hat{\theta}}}_{2, \nu}^{\text{GA}, (\text{II})} \right)$} \\
         \cmidrule(lr){2-5}  \cmidrule(lr){6-9} 
            & \multicolumn{1}{c}{In-Sample} & \multicolumn{2}{c}{Out-of-Sample} &  & \multicolumn{1}{c}{In-Sample} & \multicolumn{2}{c}{Out-of-Sample} \\
         \cmidrule(lr){2-2}  \cmidrule(lr){3-4} \cmidrule(lr){6-6}  \cmidrule(lr){7-8} 
             $\nu \times 10^{6}$ & $k\left(\boldsymbol{\hat{\theta}}^{\text{GA}, \text{(I)}}_{\nu}\right)$ &  $\tilde{k}\left(\boldsymbol{\hat{\theta}}^{\text{GA}, \text{(I)}}_{\nu}\right)$ &  $\overline{k}\left(\boldsymbol{\hat{\theta}}^{\text{GA}, \text{(I)}}_{\nu}\right)$ & $\hat{R}_{detection}$ & $k\left(\boldsymbol{\hat{\theta}}^{\text{GA}, \text{(II)}}_{\nu}\right)$ &  $\tilde{k}\left(\boldsymbol{\hat{\theta}}^{\text{GA}, \text{(II)}}_{\nu}\right)$ & $\overline{k}\left(\boldsymbol{\hat{\theta}}^{\text{GA}, \text{(II)}}_{\nu}\right)$ & $\hat{R}_{detection}$ \\
            \midrule
            0 & 100 & $\mathbf{94.00}$  & $\mathbf{87.58}$ & $\mathbf{0.0350}$ &100 & $\mathbf{87.00}$ & $\mathbf{78.67}$ & $\mathbf{0.0726}$ \\
            1  & 99 & 14.00 & 24.86 & 0.1993  & 98 & 54.00 & 50.61 & 0.1056\\
            2 & 99 & $\mathbf{91.00}$ & $\mathbf{75.69}$   & $\mathbf{0.0094}$ &100 & 15.50 & 29.77 & 0.1206  \\
            3 &99 & 33.00 & 37.06 & 0.1238 & 98 & 60.00 & 54.6 & 0.0983\\
            4 &99 &  75.00 & 63.61 & 0.0048 & 100 & 84.00 & 73.90 & 0.1075\\
            5 &100 & 68.00 &59.55 &0.0063 & 99 & 66.00 & 58.56 & 0.0962  \\
            6 &98 & 30.00  & 36.64 &0.1481 & 98 & 70.50 & 61.09 & 0.1016 \\
            7 &98 &33.00 &37.29 & 0.1309 &100 & $\mathbf{86.00}$ & $\mathbf{78.47}$ & $\mathbf{0.0721}$\\
            8 &100 & $\mathbf{87.00}$ & $\mathbf{79.97}$ & $\mathbf{0.0136}$ &100 & $\mathbf{87.00}$ & $\mathbf{79.34}$ & $\mathbf{0.0724}$   \\
            9 &98 & 32.00 & 36.51 & 0.1249 &100 & 80.00 & 69.20 & 0.1161 \\
            \bottomrule
        \end{tabular}
        }
        \captionsetup{justification=centering}
        \caption{Number of successes $k\left(\boldsymbol{\hat{\theta}}^{\text{GA}}_{\nu}\right)$ for the in-sample initialization, median ($\tilde{k}$) and mean ($\bar{k}$) for the distributions of successes on the $1000$ test initializations and estimated $\hat{R}_{detection}$, against varying $\nu$ for $T = 100$ and $R_{crash}  =0.05$ using both $\overset{(\text{I})}{\mathbf{model}}$ and $\overset{(\text{II})}{\mathbf{model}}$.}
        \label{tab: success dist}
\end{table}

As shown in Table \ref{tab: success dist large}, there is a clear trend of decreasing in-sample performance as the regularization strength increases across both models. This behaviour is consistent with underfitting resulting from excessive regularization. That is, the models become overly constrained. Furthermore, the results suggest that the use of the GA may not be essential for achieving improved performance on the in-sample set, as the RS\footnote{Just as the GA utilises $M = 1000$ generations of population $N = 100$, the RS likewise performs $M\cdot N$ iterations.} (used only for $\overset{(\text{II})}{\mathbf{model}}$) performs comparably to the GA at low values of $\nu$. This observation implies that the fine-tuning capabilities of the GA offer limited benefit when the model is only weakly regularized. Under such conditions, the GA's exploitation mechanisms appear to be inconsequential, with its role in refining existing parent solutions rendered largely unnecessary. Interestingly, at moderate regularization levels, the RS appears to estimate $\hat{R}_{{detection}} \approx 0.07$ more effectively than the GA. However, this outcome reinforces the notion that multiple plausible solutions exist which yield high objective values on the in-sample set, and that the RS may have coincidentally identified ones that generalize well. Nevertheless, Table \ref{tab: success dist} demonstrates that the GA is indeed capable of consistently discovering this advantageous value of $\hat{R}_{{detection}}$.

\begin{table}[H]
\centering
\resizebox{\textwidth}{!}{%
        \begin{tabular}{cccccccccc}
            & \multicolumn{3}{c}{$\overset{(\text{I})}{\mathbf{model}}\left(\overset{(\text{I})}{a}(t)^0, {\boldsymbol{\hat{\theta}}}_{2, \nu}^{\text{GA}, (\text{I})} \right)$} & \multicolumn{3}{c}{$\overset{(\text{II})}{\mathbf{model}}\left(\overset{(\text{II})}{a}(t)^0, {\boldsymbol{\hat{\theta}}}_{2, \nu}^{\text{GA}, (\text{II})} \right)$}  & \multicolumn{3}{c}{$\overset{(\text{II})}{\mathbf{model}}\left(\overset{(\text{II})}{a}(t)^0, {\boldsymbol{\hat{\theta}}}_{2, \nu}^{\text{RS}, (\text{II})} \right)$}\\
         \cmidrule(lr){2-4}  \cmidrule(lr){5-7} \cmidrule(lr){8-10} 
            & \multicolumn{1}{c}{In-Sample} & \multicolumn{1}{c}{OOS} &  & \multicolumn{1}{c}{In-Sample} & \multicolumn{1}{c}{OOS} & &  \multicolumn{1}{c}{In-Sample} & \multicolumn{1}{c}{OOS}\\
         \cmidrule(lr){2-2}  \cmidrule(lr){3-3} \cmidrule(lr){5-5}  \cmidrule(lr){6-6} \cmidrule(lr){8-8} \cmidrule(lr){9-9} 
             $\nu$ & $k\left(\boldsymbol{\hat{\theta}}^{\text{GA}, \text{(I)}}_{\nu}\right)$ &  $\overline{k}\left(\boldsymbol{\hat{\theta}}^{\text{GA}, \text{(I)}}_{\nu}\right)$ & $\hat{R}$ & $k\left(\boldsymbol{\hat{\theta}}^{\text{GA}, \text{(II)}}_{\nu}\right)$ & $\overline{k}\left(\boldsymbol{\hat{\theta}}^{\text{GA}, \text{(II)}}_{\nu}\right)$ & $\hat{R}$ & $k\left(\boldsymbol{\hat{\theta}}^{\text{RS}, \text{(II)}}_{\nu}\right)$ & $\overline{k}\left(\boldsymbol{\hat{\theta}}^{\text{RS}, \text{(II)}}_{\nu}\right)$ & $\hat{R}$  \\
            \midrule
            $0.000001$ & 99 & 24.86  & $0.1993$ & 98 & 50.61 & 0.1056 & 100 & $\textbf{77.51}$ & $\textbf{0.0666}$  \\
            $0.00001$ & 98 & 37.54 & 0.1362 & 100 & 68.29 & 0.0847 & 98 & $\textbf{79.55}$ & $\textbf{0.0668}$ \\
            $0.0001$ & 100 & $\textbf{75.04}$ & $\textbf{0.0070}$  &100 & 42.63 & 0.2843 & 96 & $\textbf{78.36}$ & $\textbf{0.0779}$ \\
            $0.001$ &96 & 31.27& 0.1463 & 98 & 25.30 & 0.0942 & 70 & 18.68 & 0.0375 \\
            $0.01$ & 74 & 18.62 & 0.5863 & 74 & 18.61 & 0.5406 & 21 & 19.45 & 0.9914 \\
            $0.1$ &62 & 17.63 & 0.4640 & 73 & 18.14 & 0.4550  & 8 & 18.62 & 0.1608 \\
            $1$ &8  & 12.15  & 0.4403 & 49 & 13.57 & 0.5855 & 70 & 18.68 & 0.1757 \\
            \bottomrule
        \end{tabular}
        }
        \captionsetup{justification=centering}
        \caption{Number of successes $k\left(\boldsymbol{\hat{\theta}}^{\text{GA}}_{\nu}\right)$ for the in-sample initialization, mean ($\bar{k}$) for the distributions of successes on the $1000$ test initializations (OOS) and estimated $\hat{R}_{detection}$, against varying $\nu$ for $T = 100$ and $R_{crash}  =0.05$ using both $\overset{(\text{I})}{\mathbf{model}}$ and $\overset{(\text{II})}{\mathbf{model}}$.}
        \label{tab: success dist large}
\end{table}

\paragraph{Response curves}
To further evaluate the role of regularization and to examine how the control vector integrates drone state information and environmental elements to output drone evasive manoeuvres, we analyze the response curves shown in Figure \ref{fig: response drone}. For a given input $a^0 \in \mathbb{R}$, the model yields the output $\mathbf{\tilde{x}} = (\tilde{x}_1, \tilde{x}_2)$ using $\overset{(\text{I})}{\mathbf{model}}\left(\overset{(\text{I})}{a^0}, {\boldsymbol{\hat{\theta}}}_{2, \nu}^{\text{GA}, (\text{I})} \right)$ and $\overset{(\text{II})}{\mathbf{model}}\left(\overset{(\text{II})}{a^0}, {\boldsymbol{\hat{\theta}}}_{2, \nu}^{\text{GA}, (\text{II})} \right)$. It is worth noting that values of $a^0 < 0$ are feasible, as such cases arise only when a given drone has entered a terminal state. Furthermore, no distinguishable pattern appears to emerge as $\nu$ varies; each response curve exhibits a unique shape, with no visually discernible structure that consistently correlates with improved performance as reflected in Table \ref{tab: success dist}.

\iffoo
\begin{figure}[H]
    \centering
    \begin{minipage}{0.48\linewidth}
        \centering
        \animategraphics[controls, autoplay,loop,width=\linewidth]{1}{Response_drones_model1/}{0}{9}
        \caption*{$\overset{(\text{I})}{\mathbf{model}}$}
    \end{minipage}
    \hfill
    \begin{minipage}{0.48\linewidth}
        \centering
        \animategraphics[controls, autoplay,loop,width=\linewidth]{1}{Response_drones_model2/}{0}{9}
        \caption*{$\overset{(\text{II})}{\mathbf{model}}$}
    \end{minipage}
    \captionsetup{justification=centering}
    \caption{Response Curves of $\tilde{\mathbf{x}} = (\tilde{x}_1, \tilde{x}_2)$ given $\overset{(\text{I})}{a^0}$ using $\overset{(\text{I})}{\mathbf{model}}\left(\overset{(\text{I})}{a^0}, {\boldsymbol{\hat{\theta}}}_{2, \nu}^{\text{GA}, (\text{I})} \right)$ (left) and $\tilde{\mathbf{x}} = (\tilde{x}_1, \tilde{x}_2)$ given $\overset{(\text{II})}{a^0}$ using $\overset{(\text{II})}{\mathbf{model}}\left(\overset{(\text{II})}{a^0}, {\boldsymbol{\hat{\theta}}}_{2, \nu}^{\text{GA}, (\text{II})} \right)$ (right) for various $\nu$.}
    \label{fig: response drone}
\end{figure}
\else

\begin{figure}[H]
\centering
\begin{subfigure}[t]{0.33\linewidth}
\includegraphics[width=\linewidth]{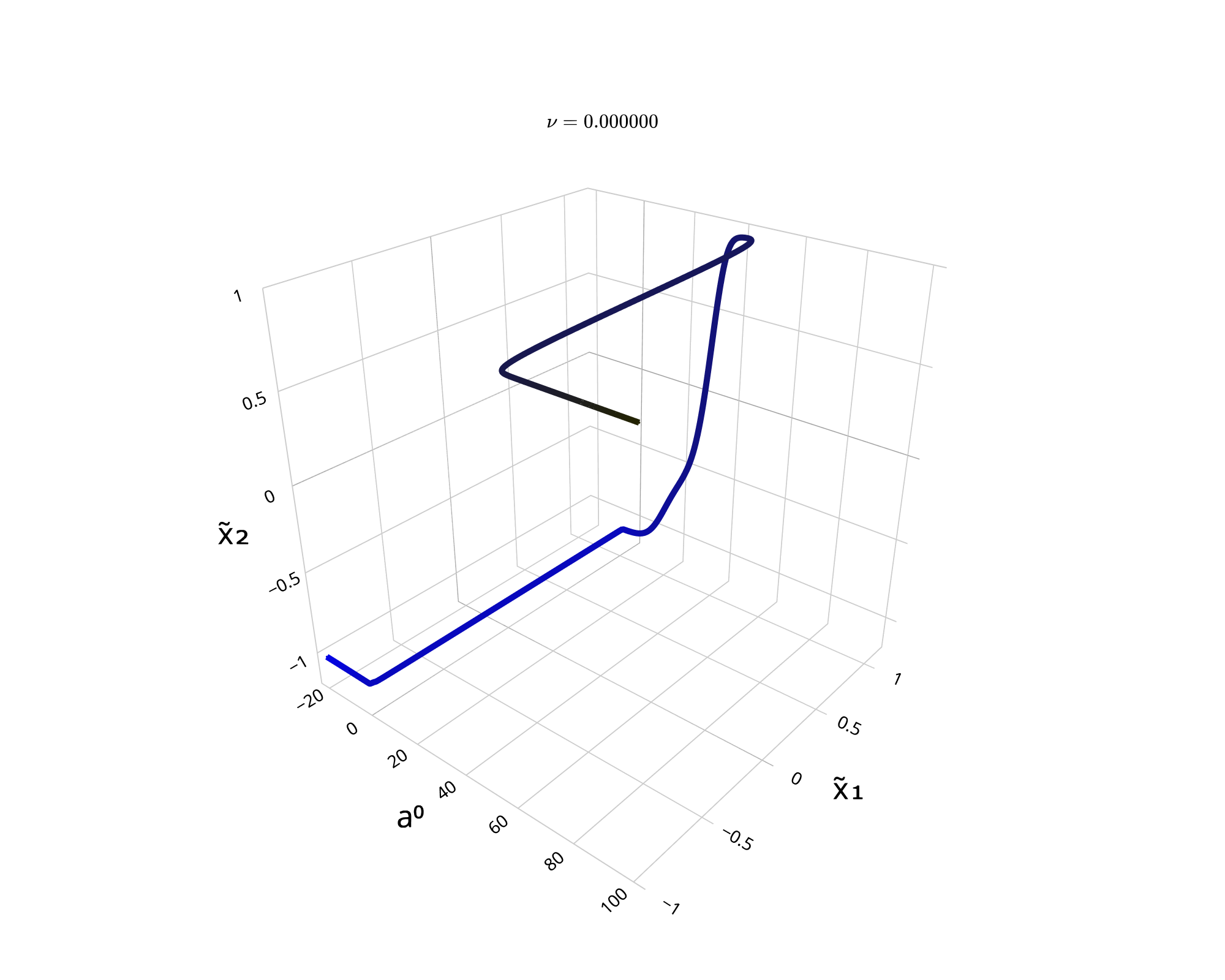}
\end{subfigure}\hfill
\begin{subfigure}[t]{0.33\linewidth}
\includegraphics[width=\linewidth]{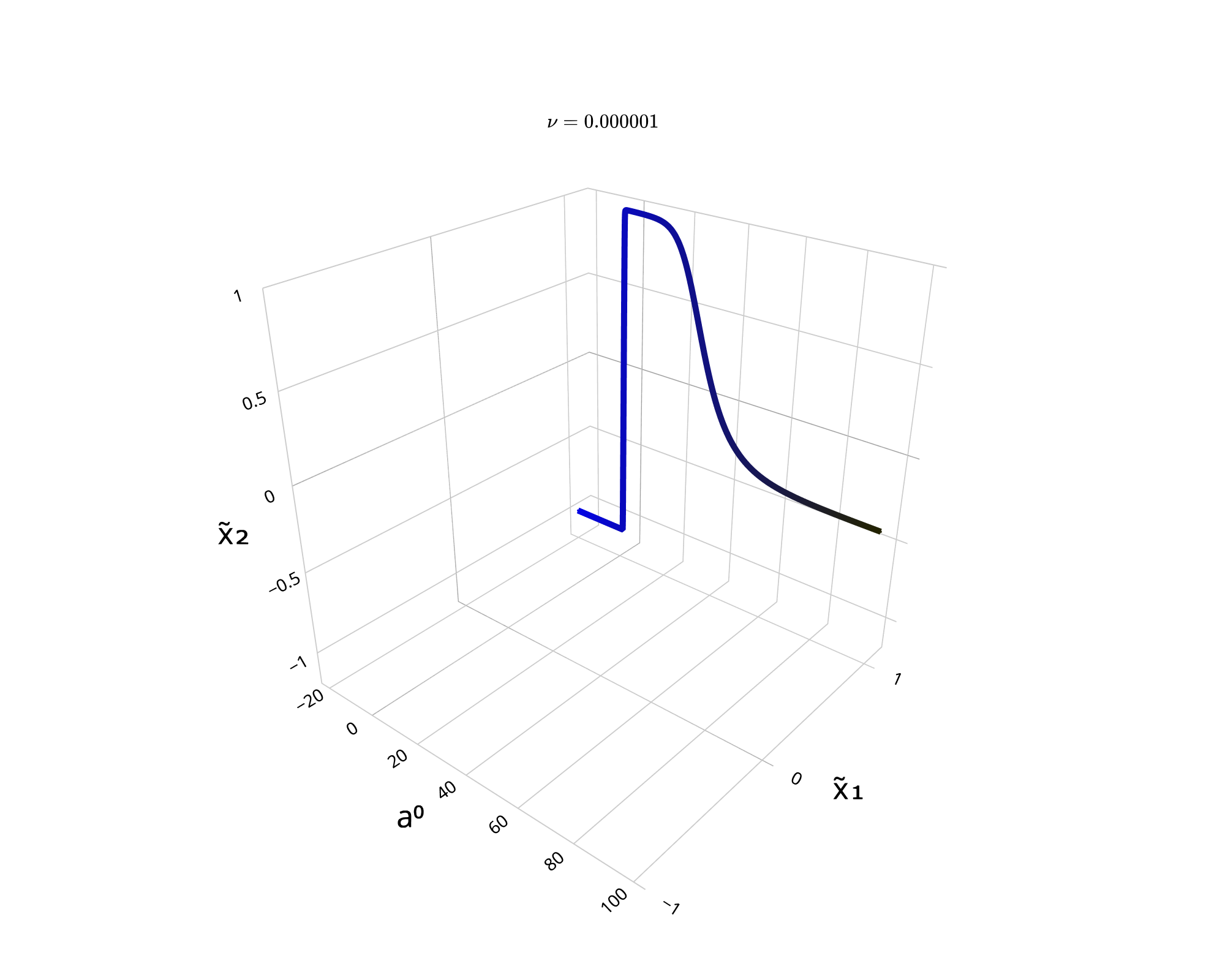}
\end{subfigure}\hfill
\begin{subfigure}[t]{0.33\linewidth}
\includegraphics[width=\linewidth]{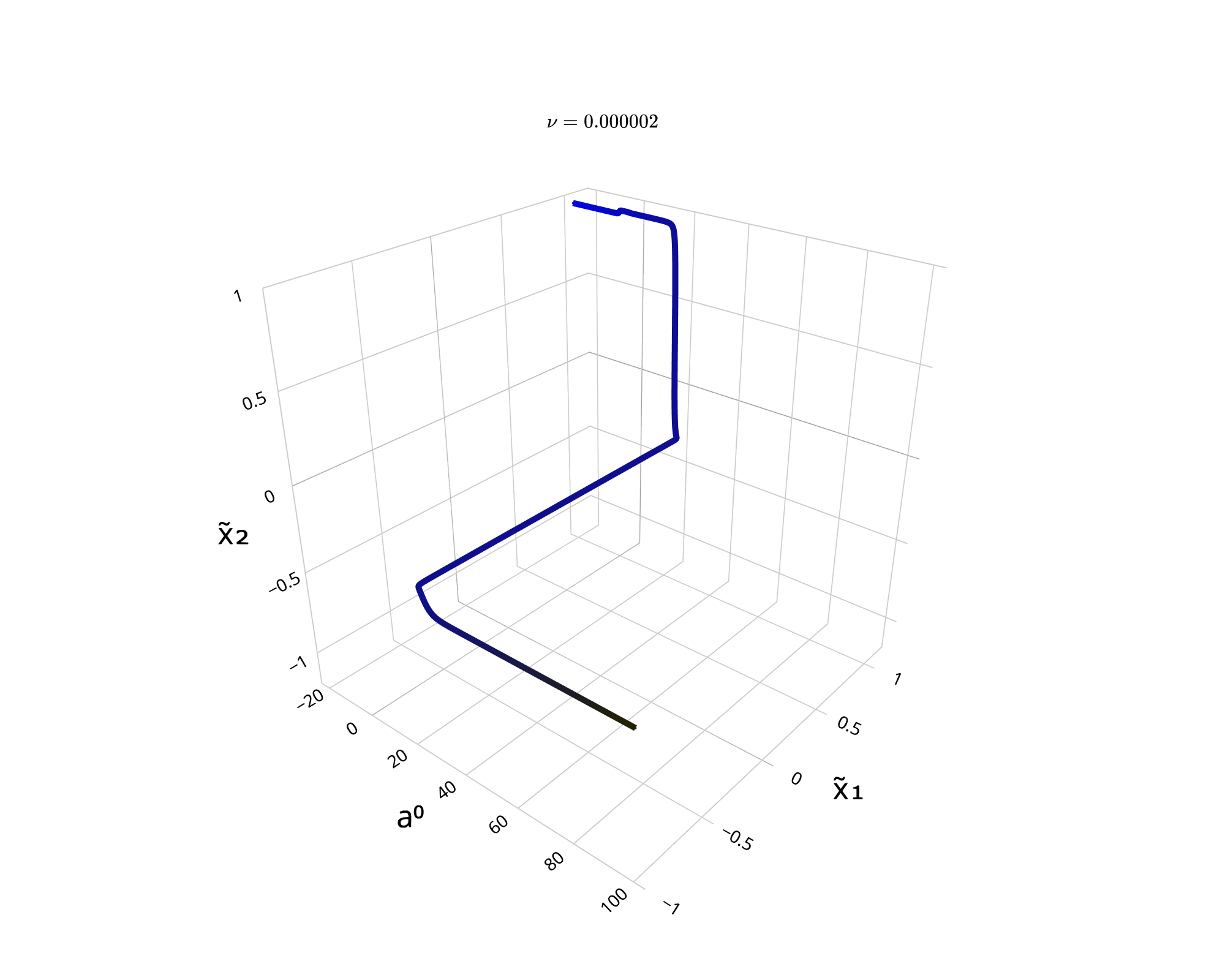}
\end{subfigure}

\vspace{0.3cm}

\begin{subfigure}{0.33\linewidth}
\includegraphics[width=\linewidth]{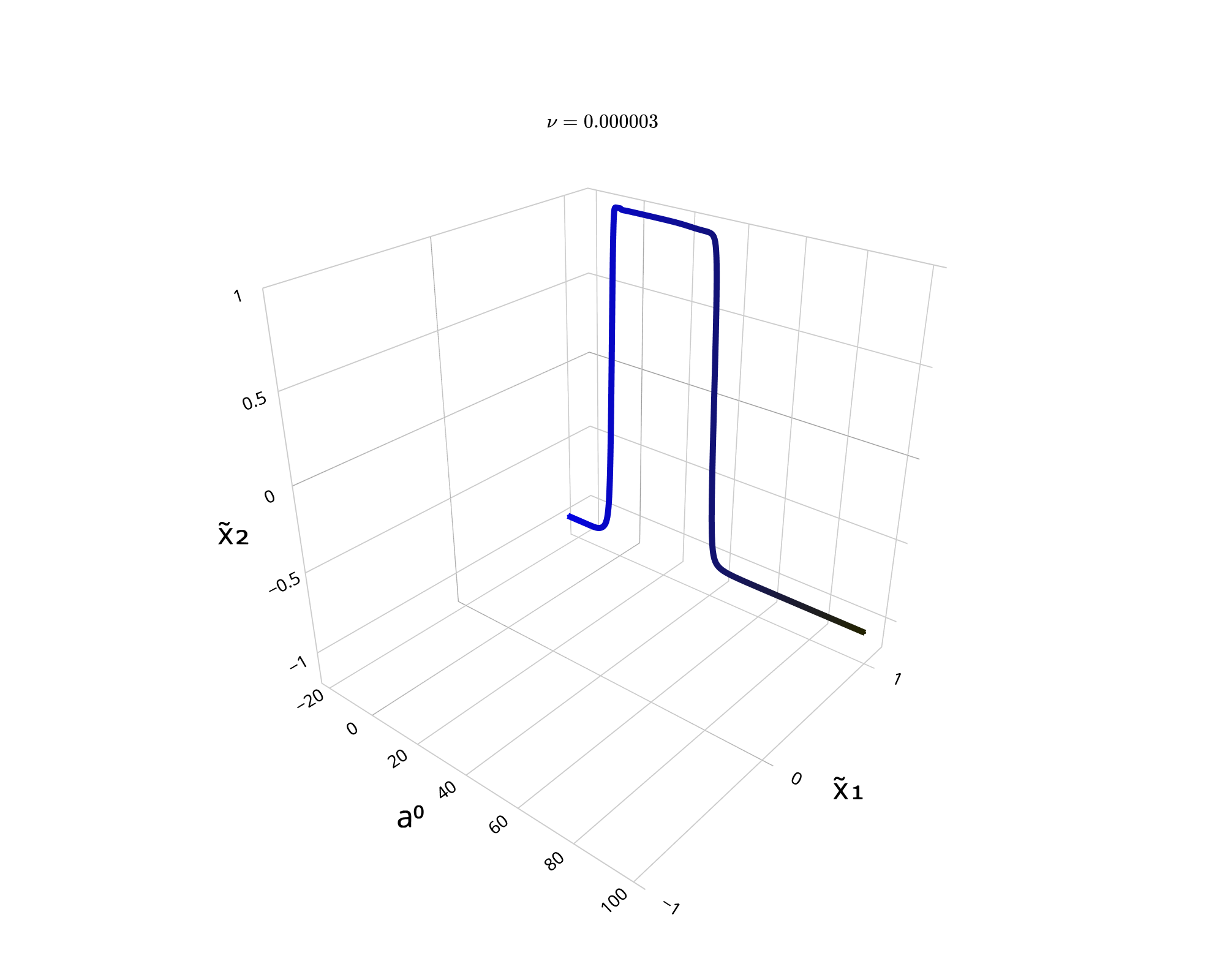}
\end{subfigure}\hfill
\begin{subfigure}{0.33\linewidth}
\includegraphics[width=\linewidth]{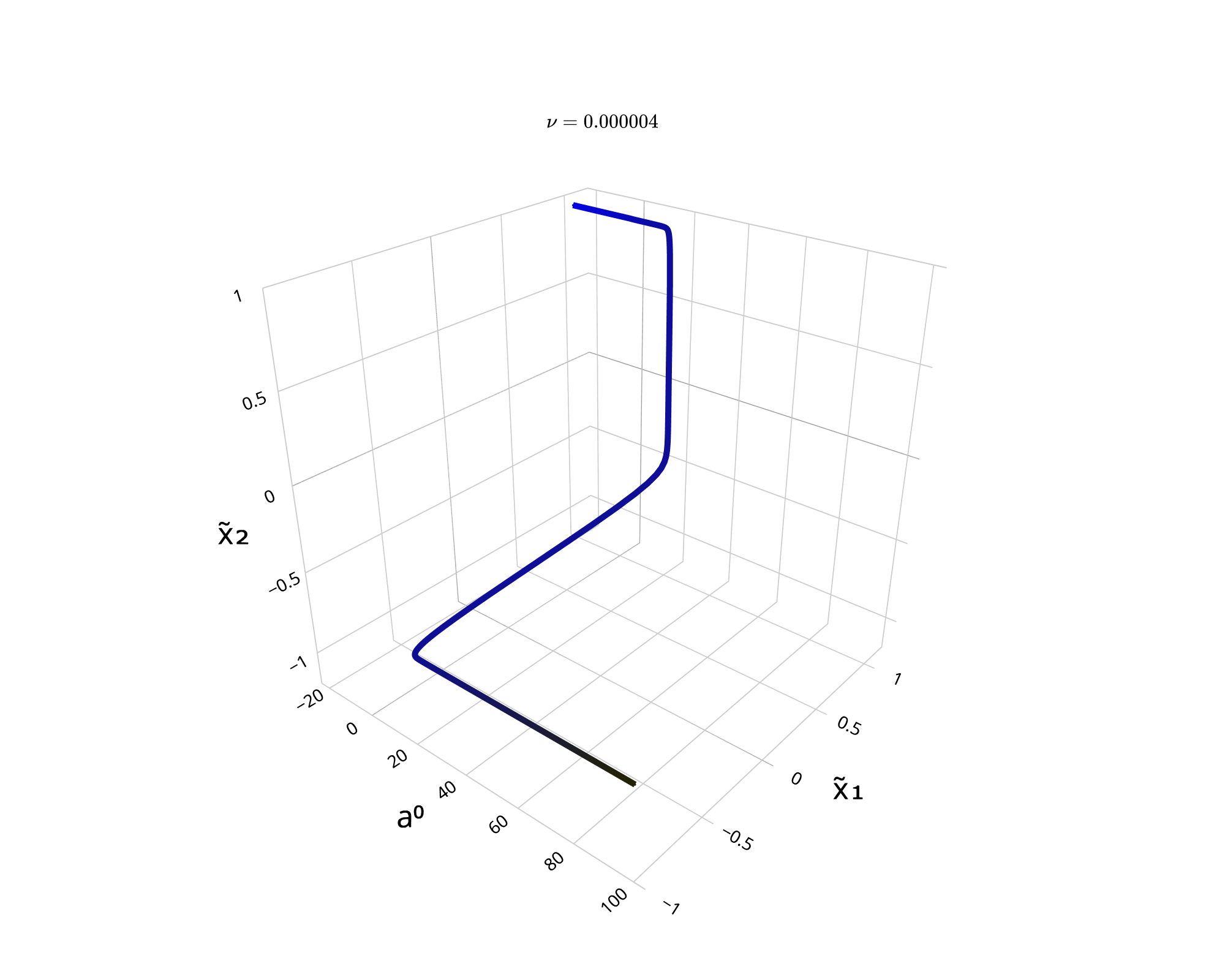}
\end{subfigure}\hfill
\begin{subfigure}{0.33\linewidth}
\includegraphics[width=\linewidth]{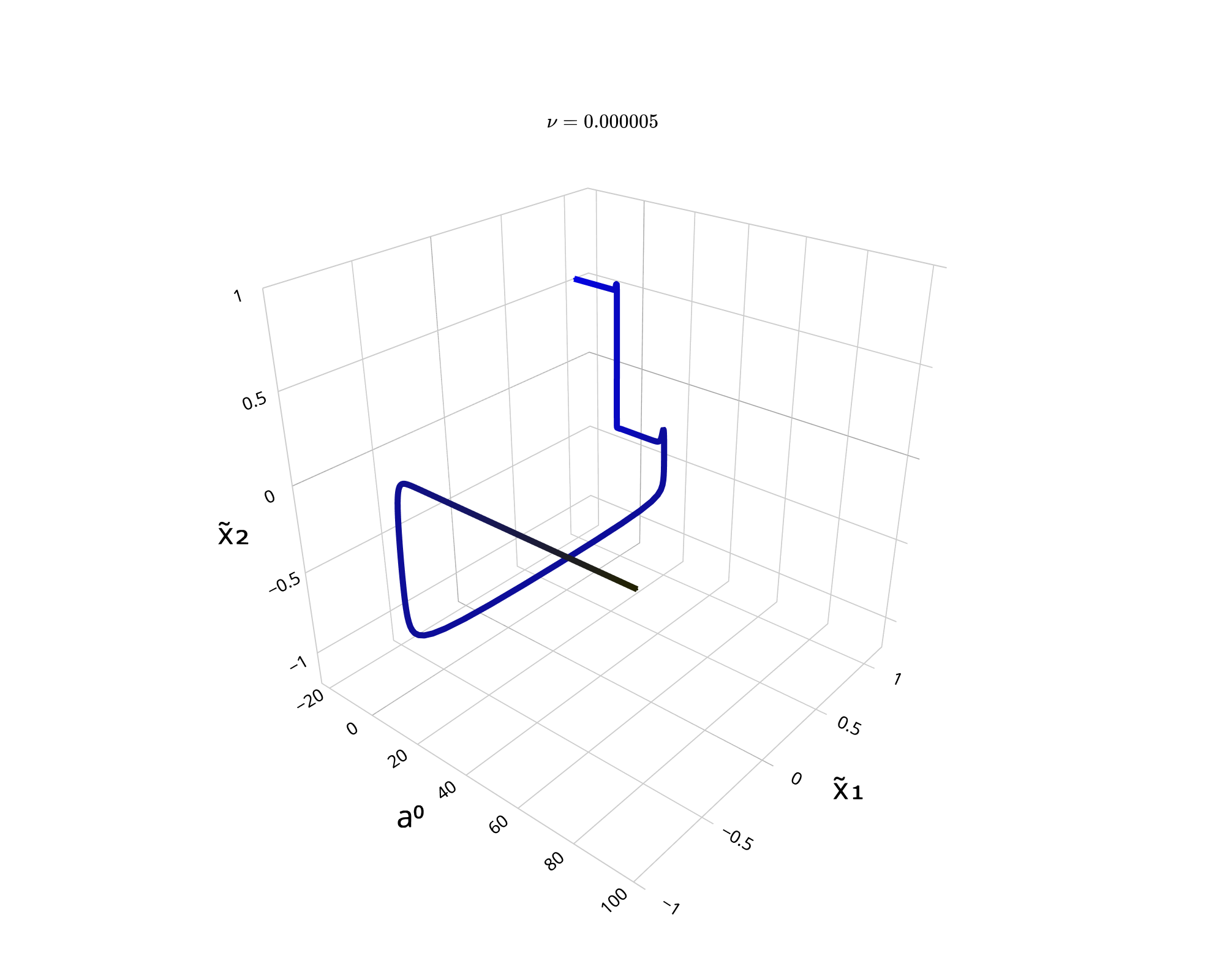}
\end{subfigure}
\caption*{$\overset{(\text{I})}{\mathbf{model}}$}
\vspace{0.35cm}

\begin{subfigure}{0.33\linewidth}
\includegraphics[width=\linewidth]{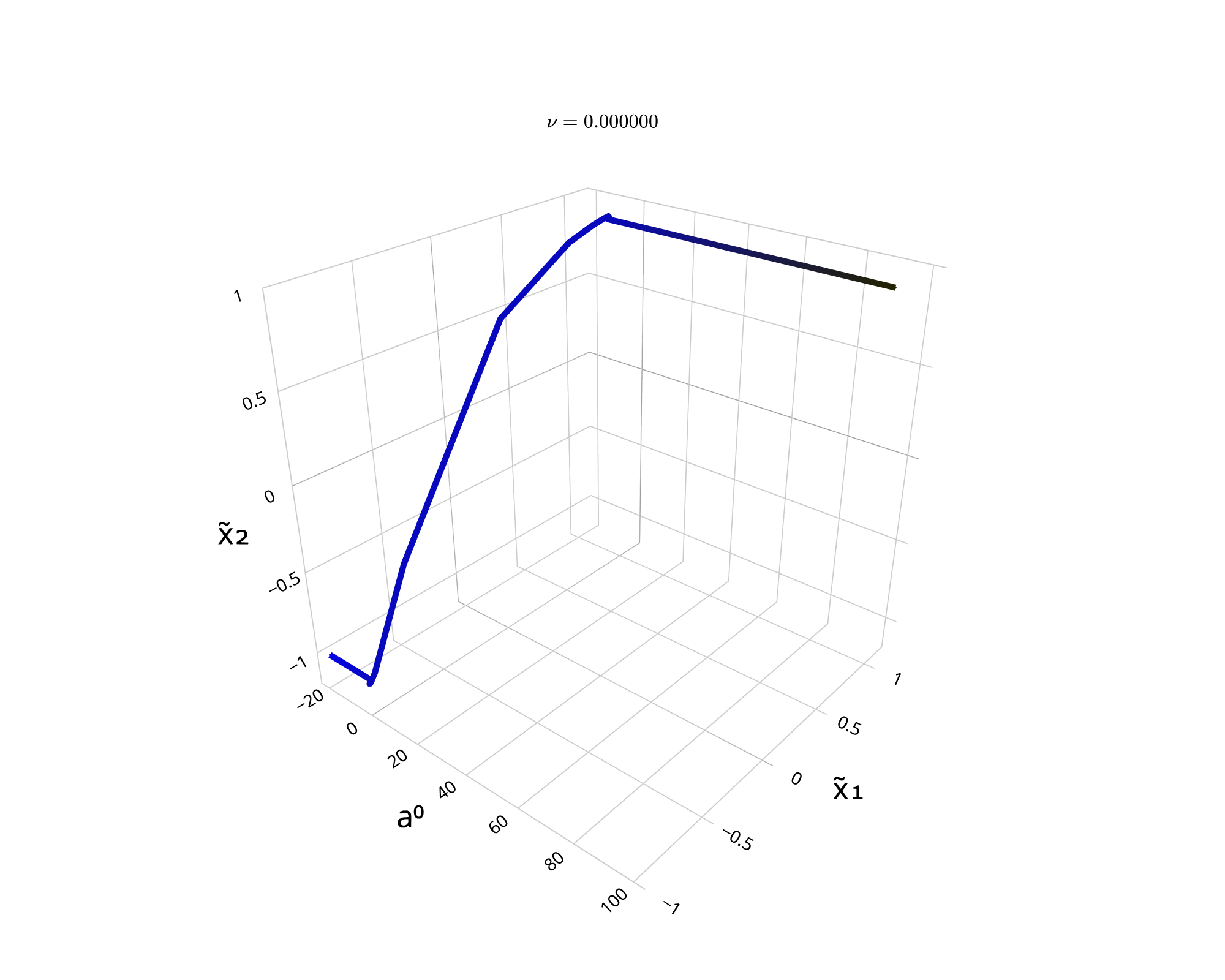}
\end{subfigure}\hfill
\begin{subfigure}{0.33\linewidth}
\includegraphics[width=\linewidth]{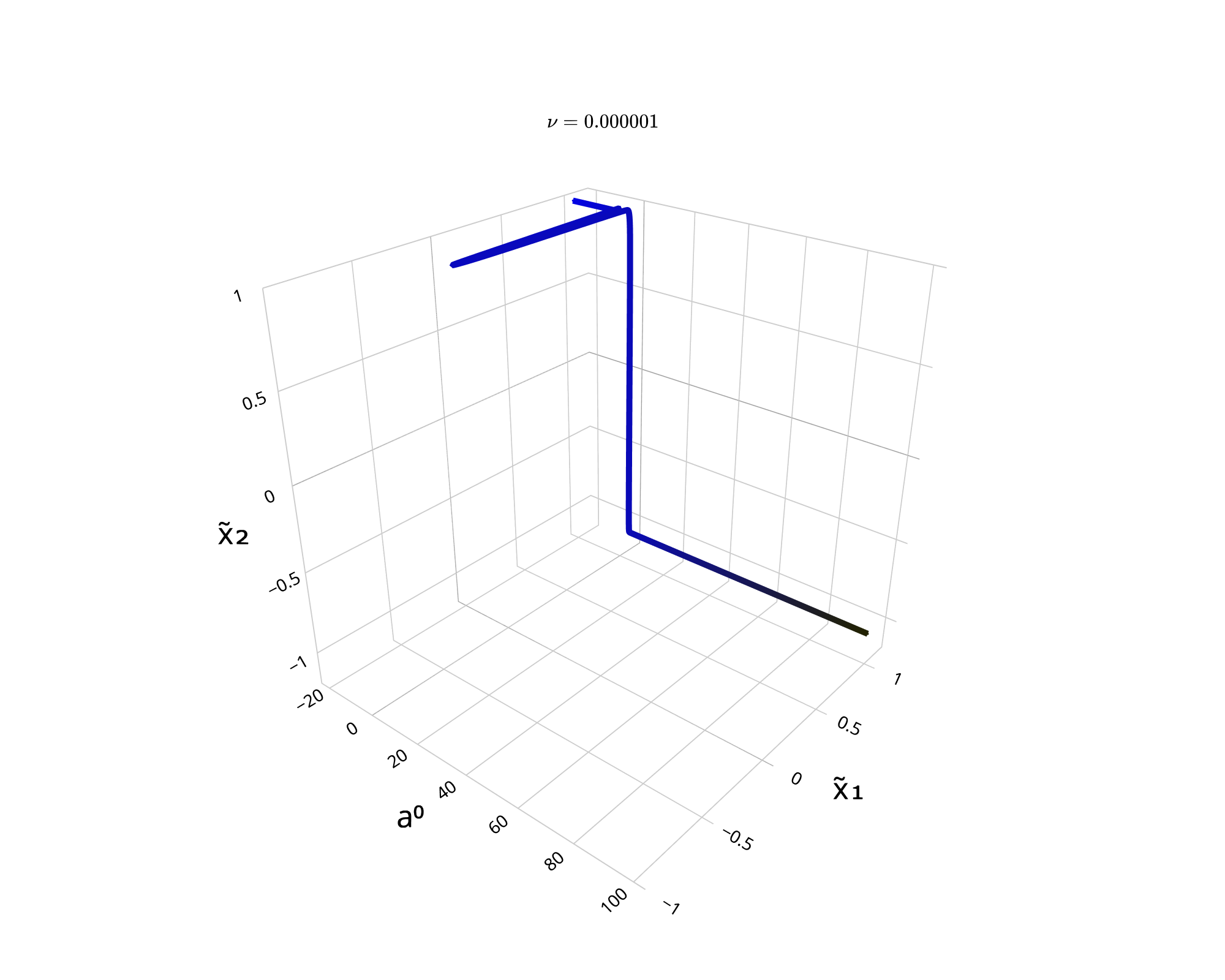}
\end{subfigure}\hfill
\begin{subfigure}{0.33\linewidth}
\includegraphics[width=\linewidth]{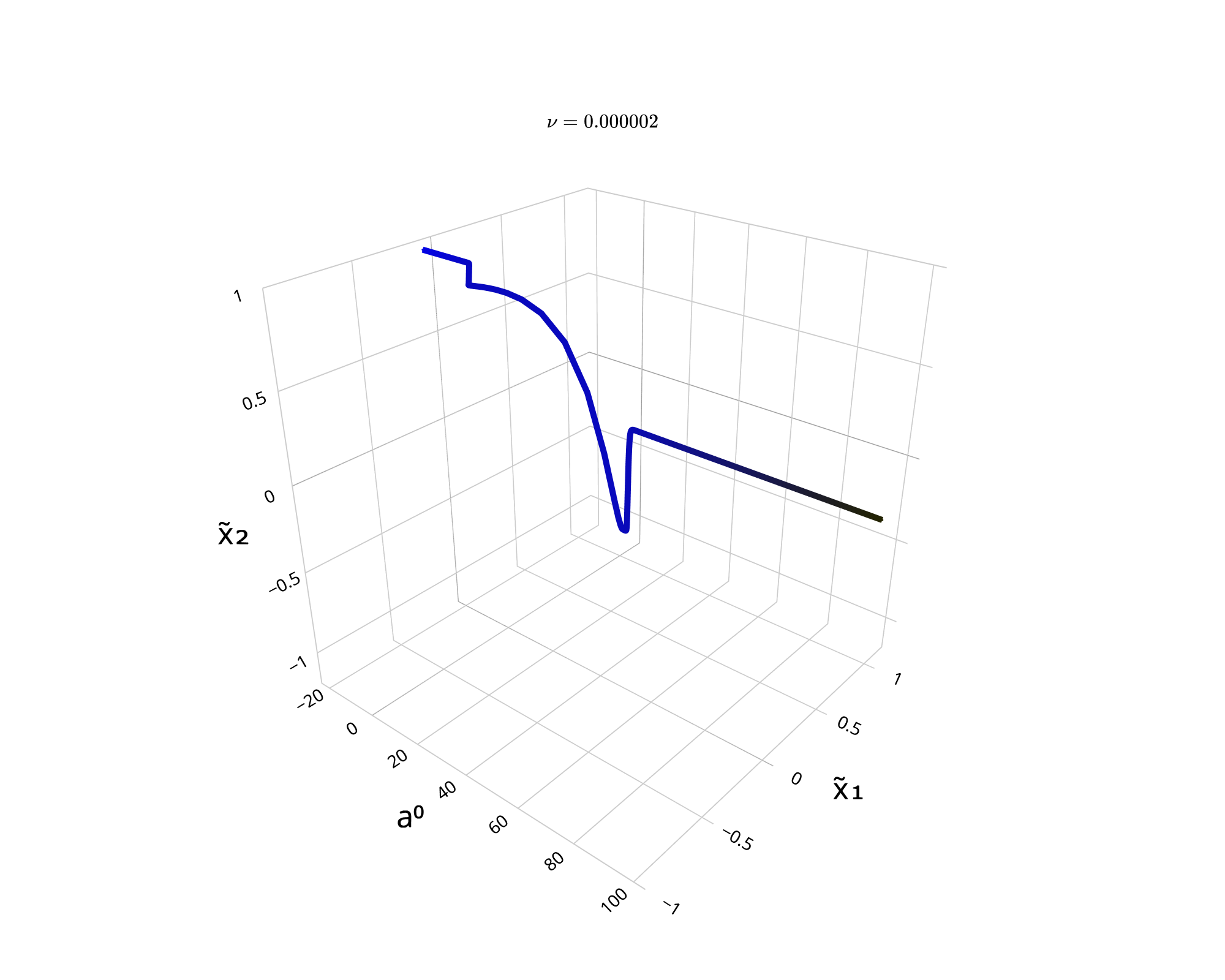}
\end{subfigure}

\vspace{0.35cm}

\begin{subfigure}{0.33\linewidth}
\includegraphics[width=\linewidth]{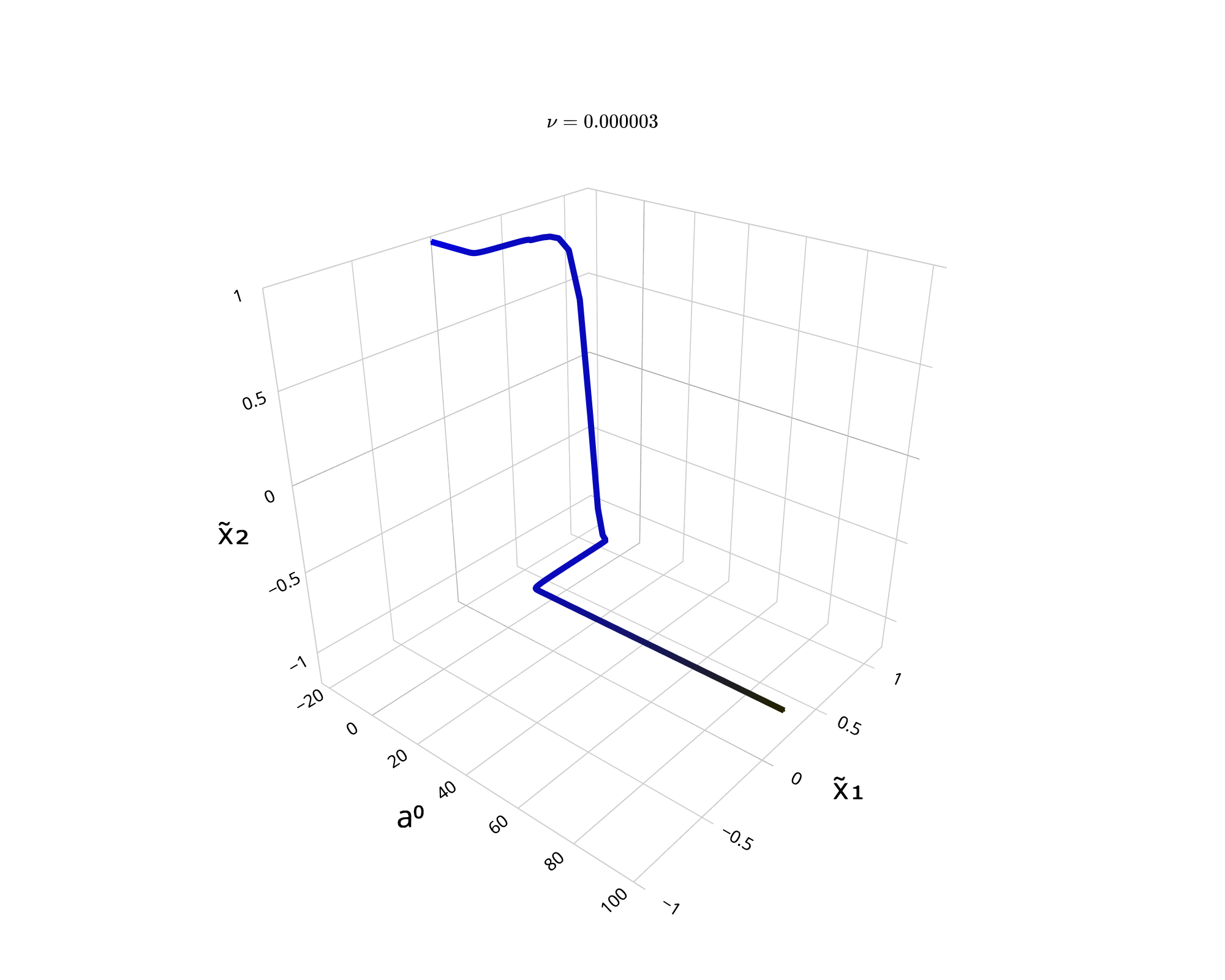}
\end{subfigure}\hfill
\begin{subfigure}{0.33\linewidth}
\includegraphics[width=\linewidth]{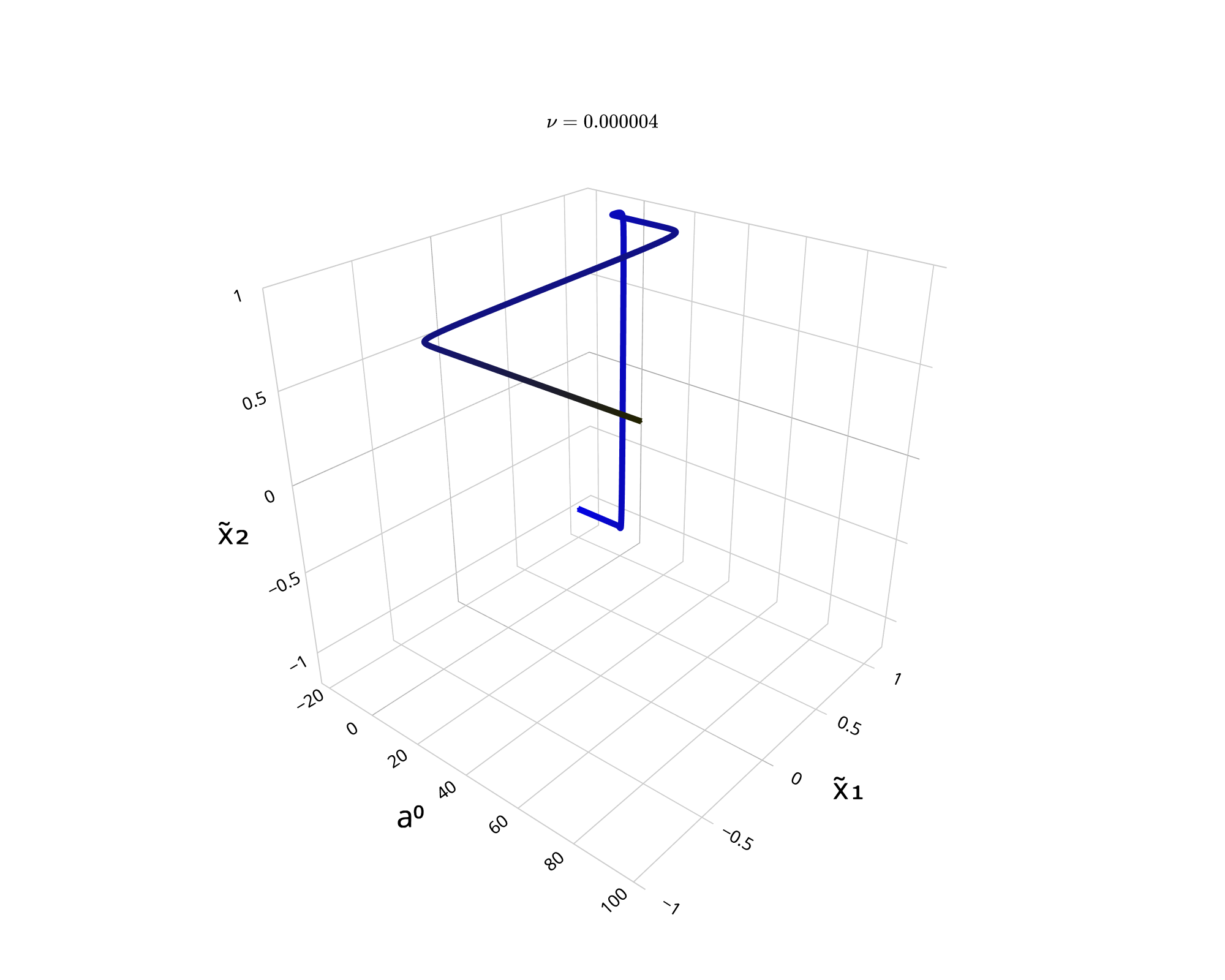}
\end{subfigure}\hfill
\begin{subfigure}{0.33\linewidth}
\includegraphics[width=\linewidth]{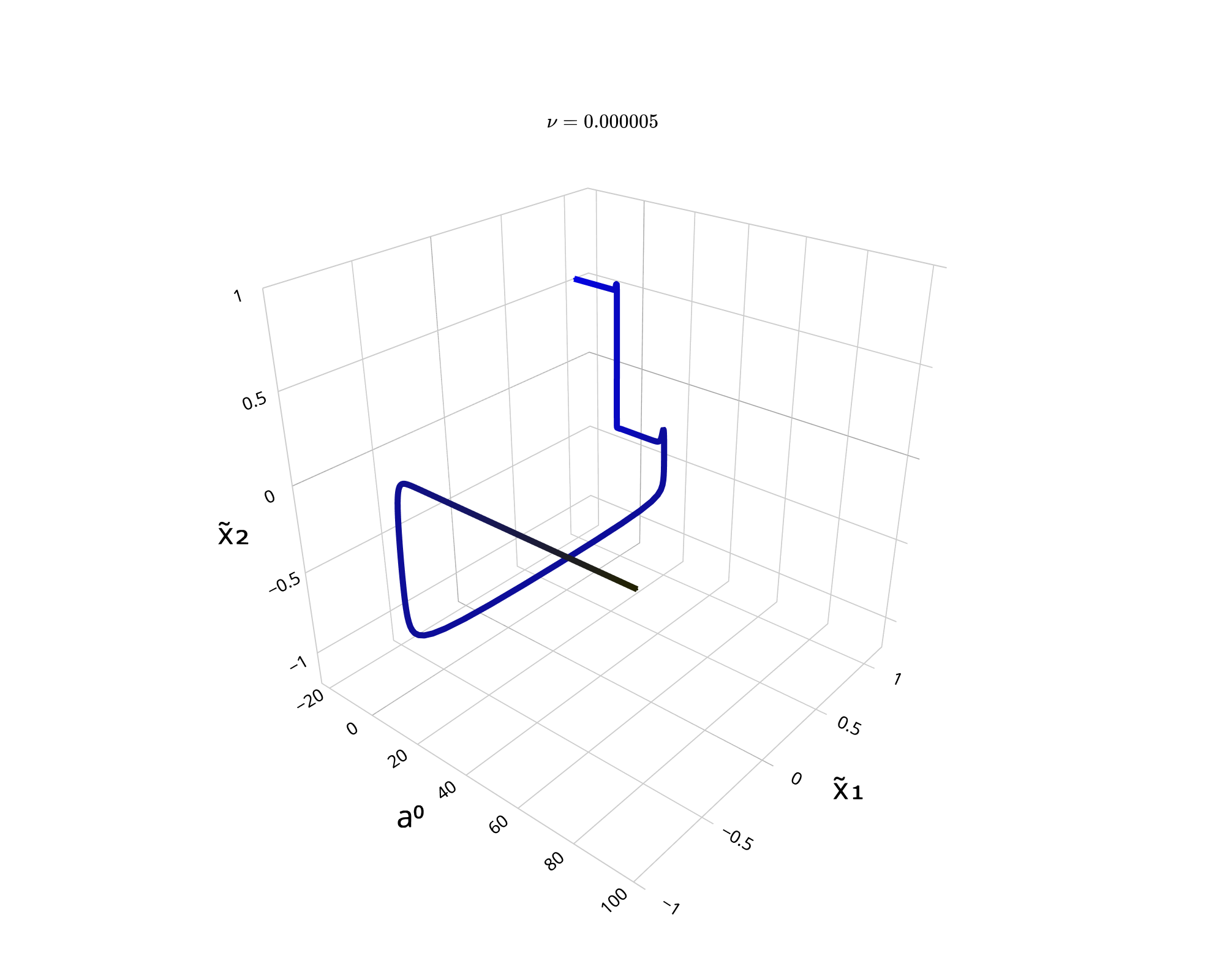}
\end{subfigure}
\caption*{$\overset{(\text{II})}{\mathbf{model}}$}
\captionsetup{justification=centering}
    \caption{Response curves of $\tilde{\mathbf{x}} = (\tilde{x}_1, \tilde{x}_2)$ given $\overset{(\text{I})}{a^0}$ using $\overset{(\text{I})}{\mathbf{model}}\left(\overset{(\text{I})}{a^0}, {\boldsymbol{\hat{\theta}}}_{2, \nu}^{\text{GA}, (\text{I})} \right)$ (top) and $\tilde{\mathbf{x}} = (\tilde{x}_1, \tilde{x}_2)$ given $\overset{(\text{II})}{a^0}$ using $\overset{(\text{II})}{\mathbf{model}}\left(\overset{(\text{II})}{a^0}, {\boldsymbol{\hat{\theta}}}_{2, \nu}^{\text{GA}, (\text{II})} \right)$ (bottom) for various $\nu$.}
    \label{fig: response drone}
\end{figure}
\fi

\subsubsection{Results: MCMC} \label{sec: drones mcmc}
We aim to compare the behaviour of the three likelihood formulations discussed previously: the binomial-based likelihood (Equation \ref{eq: likelihood drone og}), the beta-based likelihood (Equation \ref{eq: likelihood drone beta new}), and the exponential-based likelihood (Equation \ref{eq: likelihood drone exp}), in terms of their impact on the MCMC results obtained. To ensure a fair comparison, we evaluate each likelihood using the same fixed value of the sharpness parameter, specifically $\beta = 20$, selected arbitrarily. It is important to note that each pseudo-likelihood induces values of differing magnitudes when evaluated at the same $k(\boldsymbol{\theta})$ (this notion is illustrated in Figure \ref{fig: exp}). Consequently, we refrain from reporting the corresponding values of the conditional posterior $p(\boldsymbol{\theta} \mid \sigma_\theta^2, \mathcal{D})$ (since $p(\boldsymbol{\theta} \mid \sigma_\theta^2, \mathcal{D}) \propto p(\mathcal{D} \mid \boldsymbol{\theta}) \, p(\boldsymbol{\theta} \mid \sigma_\theta^2)$ and the scale of the likelihood $p(\mathcal{D} \mid \boldsymbol{\theta})$ varies considerably across formulations). Furthermore, we employ $\overset{(\text{II})}{\mathbf{model}}$ in this section.\\

Table \ref{tab: success dist mcmc} presents the results, from which it is evident that the binomial-based likelihood yields the best in-sample solution. However, none of the likelihood formulations appear to generalize well to the out-of-sample set of initializations. This is most likely attributable to the estimation of the parameter $R_{detection}$. As discussed in Section \ref{sec: effects of reg drone}, the out-of-sample performance of a given solution appears to be highly sensitive to the value estimated for this parameter. \\

Additionally, the multivariate effective sample sizes (ESS) reported in Table \ref{tab: success dist mcmc} provide evidence of satisfactory mixing. High ESS values indicate that the sampler is generating effectively independent samples, that is, with low autocorrelation, which are representative of the target posterior distribution. All reported ESS values exceed the commonly accepted threshold of 100 (as recommended by \cite{vehtari2021rank}, Section~4), which supports the claim of efficient exploration. This performance is not coincidental: as discussed in Section \ref{sec: adaptive MH}, the sampler was deliberately tuned using an adaptive MH scheme to encourage good mixing. Furthermore, Table \ref{tab: success dist mcmc} suggests that the exponential-based likelihood exhibits the best mixing properties, as indicated by the ESS. That said, there is no clear evidence of poor mixing in the other likelihood formulations.\\

Figure \ref{fig: inv-gammas drones} illustrates the poor convergence behaviour of the beta- and exponential-based likelihoods, as evidenced by the plots of $\lVert \boldsymbol{\theta}^{(j)} \rVert^2$ across MCMC iterations. The top panel displays all $100{,}000$ MCMC iterations. Their trajectories exhibit an initial phase of high variability and gradual decline, followed by a transition into a region of relative stability. This behavior suggests that the sampler undergoes a prolonged adaptation phase before reaching a region of the parameter space where $\lVert \boldsymbol{\theta}^{(j)} \rVert^2$ fluctuates around a stable value. The prolonged descent and subsequent stabilization imply delayed convergence for the beta- and exponential-based likelihood types, necessitating a longer burn-in period to ensure samples are drawn from the stationary distribution. We note, however, that the binomial-based likelihood exhibits relatively good convergence. For the beta- and exponential-based likelihoods, a burn-in of $60{,}000$ iterations was applied, motivated by the point at which $\lVert \boldsymbol{\theta}^{(j)} \rVert^2$ appears to stabilize, whereas a shorter burn-in of $20{,}000$ iterations was used for the binomial-based likelihood. The corresponding marginal distributions, $p(\sigma_\theta^2 \mid \mathcal{D})$, are shown in the bottom panel of Figure \ref{fig: inv-gammas drones}.\\  

Despite the different convergence dynamics, the marginal distributions, $p(\sigma_\theta^2 \mid \mathcal{D})$, across all likelihood types appear to be consistent with an inverse-gamma form. Given that the conditional posterior $\sigma_\theta^2 \mid \boldsymbol{\theta}, \mathcal{D} \sim \text{Inv-Gamma}\left(a + \tfrac{S}{2},\; b + \tfrac{\lVert \boldsymbol{\theta} \rVert^2}{2} \right)$ with $a, b \approx 0$ (known from Section \ref{sec: 2-sample}), we expect that if $\lVert \boldsymbol{\theta}^{(j)} \rVert^2$ fluctuates around a constant value $c$, then the marginal distribution $\sigma_\theta^2 \mid \mathcal{D} \sim \text{Inv-Gamma}\left(a + \tfrac{S}{2},\; b + \tfrac{c}{2} \right)$. That is, it should also retain the inverse-gamma form with approximately constant shape and scale parameters. Regarding the level of regularization inferred from the training set, where regularization is represented via the dispersion parameter $\sigma_\theta^2 \propto \tfrac{1}{\nu}$, we may draw meaningful conclusions from the marginals of $\sigma_\theta^2 \mid \mathcal{D}$ shown in Figure \ref{fig: inv-gammas drones}. Specifically, the variation in these distributions across different likelihood types suggests that each likelihood inherently induces a different degree of regularization, which is subsequently reflected in the MAP estimates of the parameters.\\

Furthermore, we may attribute the superior in-sample performance of the binomial-based likelihood to its sharper increase at high values of $k(\boldsymbol{\theta})$, as we previously alluded to in Figure \ref{fig: der_exp}. We propose that the likelihood function can be interpreted as a mechanism for modulating the structural complexity of the conditional posterior. Specifically, in the case of flatter likelihoods, the influence of the likelihood on the posterior is weak, resulting in a diffuse posterior landscape that may support multiple regions with comparable acceptance probabilities $\alpha_\theta$. This, in turn, can induce a multimodal posterior structure. In such a regime, the sampler is encouraged to explore broadly, often traversing disconnected or competing modes across the parameter space. However, this extensive exploration may lead the Markov chain to eventually converge to a mode that is not aligned with a high-value region of the objective function, given the presence of multiple modes. In contrast, a more sharply peaked likelihood suppresses minor modes and concentrates posterior mass around a dominant region, thereby sharpening the posterior and promoting unimodality. We posit that this behaviour underlies the relatively better performance observed with the binomial-based likelihood on the in-sample set, in contrast to the beta- and exponential-based alternatives, which likely converge to sampling around one of these minor modes. \\

Now, to potentially improve the in-sample performance using the beta- and exponential-based likelihoods, one could increase their likelihood sharpness parameter $\beta$, thereby inducing a more concentrated conditional posterior. We do not pursue this adjustment in the present analysis however.

\begin{table}[H]
\centering
        \begin{tabular}{ccccccc}
            & \multicolumn{1}{c}{In-Sample} & \multicolumn{2}{c}{Out-of-Sample} &   \\
         \cmidrule(lr){2-2}  \cmidrule(lr){3-4} 
            \bottomrule
             Likelihood Type & $k\left(\boldsymbol{\hat{\theta}}^{\text{GA}, \text{(II)}}_{\nu}\right)$ &  $\tilde{k}\left(\boldsymbol{\hat{\theta}}^{\text{GA}, \text{(II)}}_{\nu}\right)$ &  $\overline{k}\left(\boldsymbol{\hat{\theta}}^{\text{GA}, \text{(II)}}_{\nu}\right)$ & $\hat{R}_{detection}$ & ESS \\
            \midrule
            Binomial-based & 100 & 13.00 & 24.45 & 0.3912 & 179.8550\\
            Beta-based & 50 & 8.00 & 12.72 & 0.5175 & 170.0205  \\
            Exponential-based & 52 & 12.00 & 17.25 & 0.5601   &  317.2764 \\ 
            \bottomrule
        \end{tabular}
        \captionsetup{justification=centering}
        \caption{Number of successes $k\left(\boldsymbol{\hat{\theta}}^{\text{GA}}_{\nu}\right)$ for the in-sample initialization, summary statistics for the distributions of successes on the $1000$ test initializations and estimated $\hat{R}_{detection}$, for the three likelihood types.}
        \label{tab: success dist mcmc}
\end{table}

\iffoo
\begin{figure}[H]
    \centering
\animategraphics[controls,autoplay,loop,width=0.6\textwidth]{1}{Drones_MCMC/}{1}{3}
\caption{$ \lVert \boldsymbol{\theta}^{(j)} \rVert^2$ for $j  = 1, \ldots 100, 000$ (no burn-in) in the top panel, with distribution of marginal $\sigma_\theta^2 \mid \mathcal{D}$ (using $20,000$ burn-in for binomial-based and $60,000$ burn-in for beta- and exponential-based) for the three likelihood types in the bottom panel (all plots use the same scale).}
\label{fig: inv-gammas drones}
\end{figure}
\else

\begin{figure}[H]
    \centering
\begin{subfigure}{0.49\textwidth}
    \centering
    \includegraphics[width=\linewidth]{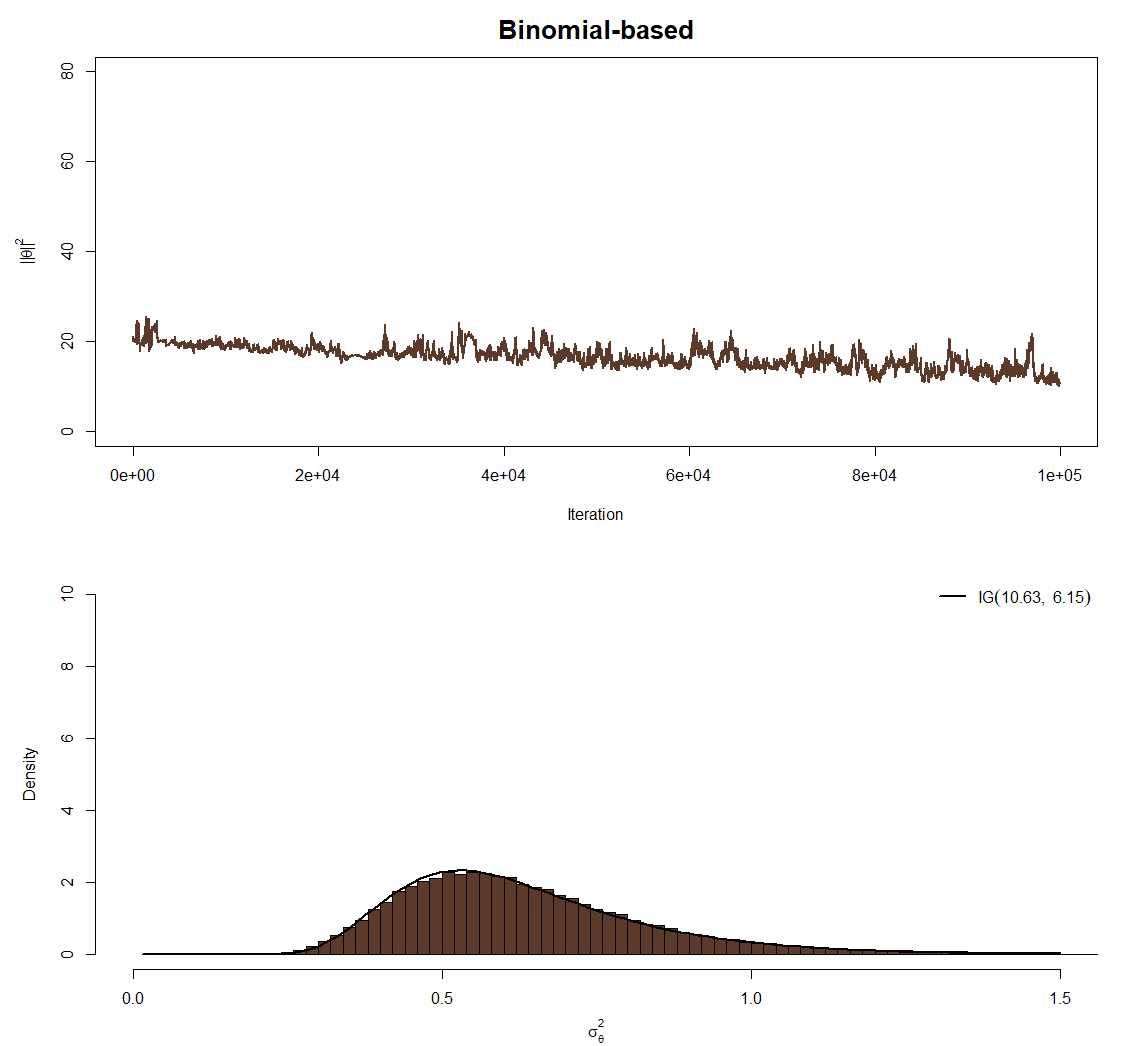}
\end{subfigure}
\hfill
\begin{subfigure}{0.49\textwidth}
    \centering
    \includegraphics[width=\linewidth]{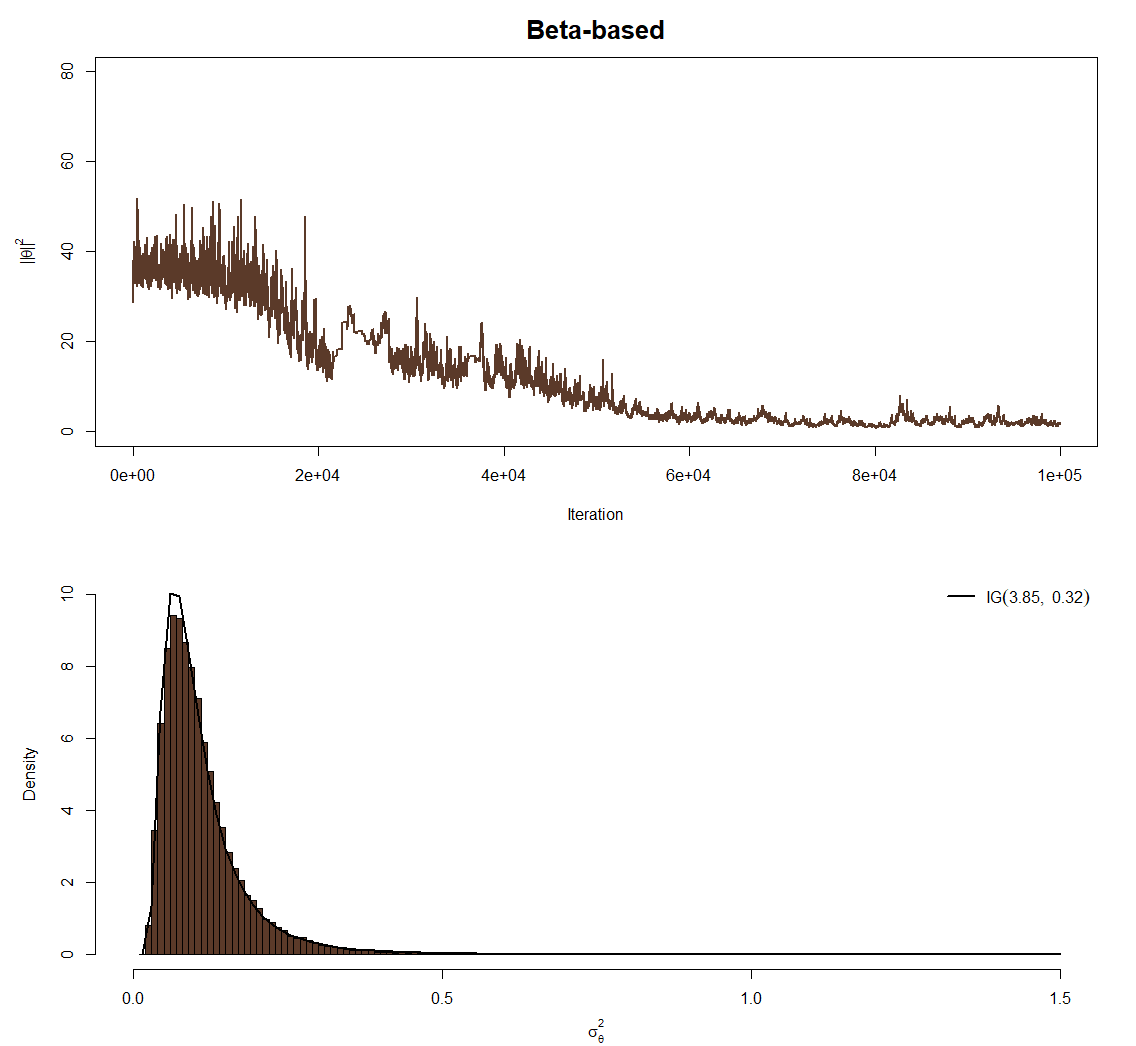}
\end{subfigure}

\vspace{0.35cm}
\begin{subfigure}{0.49\textwidth}
    \centering
    \includegraphics[width=\linewidth]{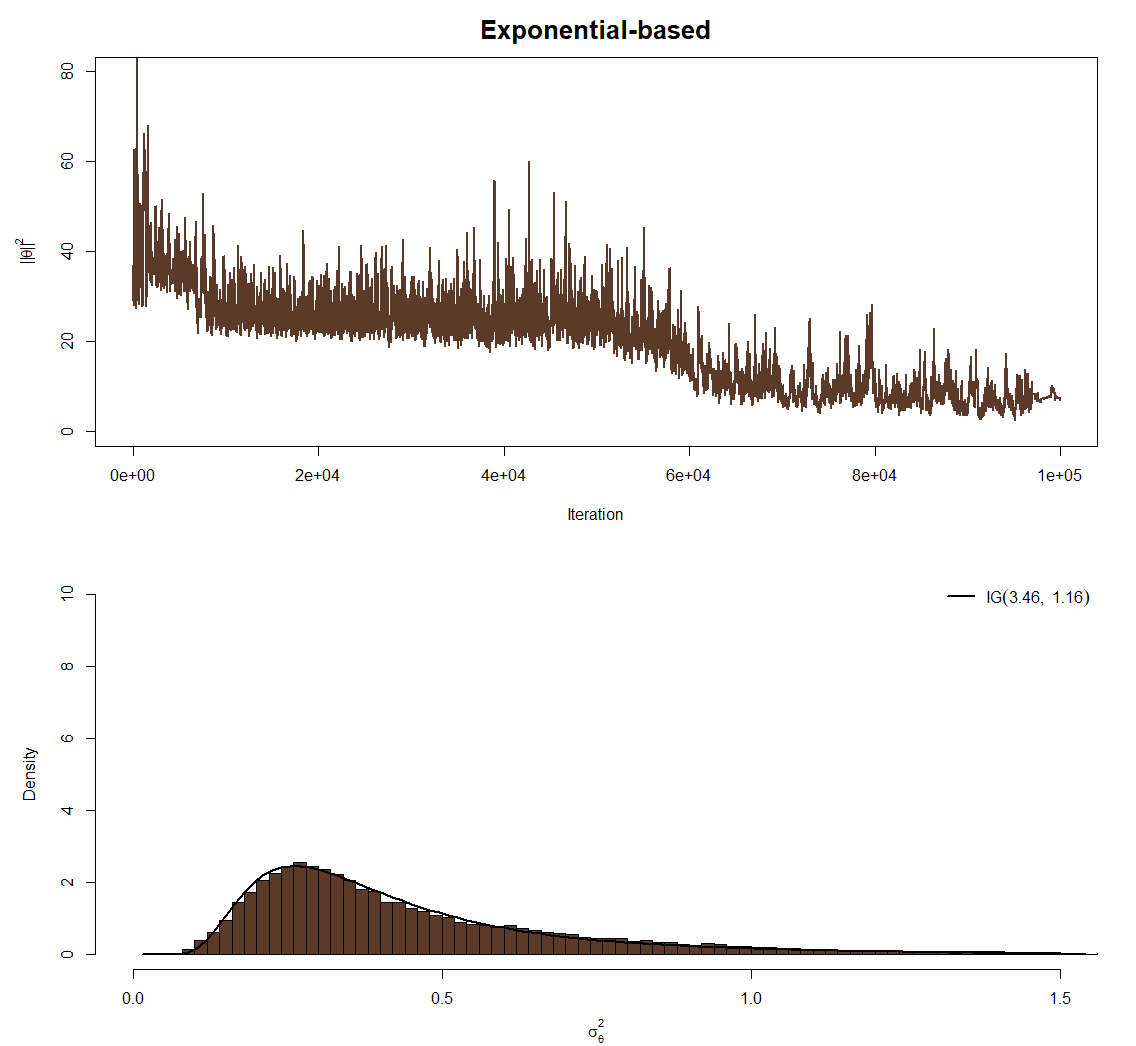}
\end{subfigure}
\caption{$ \lVert \boldsymbol{\theta}^{(j)} \rVert^2$ for $j  = 1, \ldots 100, 000$ (no burn-in) in the top panels, with distribution of marginal $\sigma_\theta^2 \mid \mathcal{D}$ (using $20,000$ burn-in for binomial-based and $60,000$ burn-in for beta- and exponential-based) for the three likelihood types in the bottom panels (all plots use the same scale).}
\label{fig: inv-gammas drones}
\end{figure}
\fi

\subsection{The Tic-Tac-Toe problem}
In what follows, the study employs a tic-tac-toe reinforcement learning problem to examine how increasing likelihood sharpness (including initial variance $\sigma_{\text{Init}}^2$) affects both in-sample and out-of-sample performance, as well as the level of regularization inferred. Specifically, a neural network serves as the control policy governing the player’s decisions at each turn, competing against an opponent whose moves are randomly assigned. This environment is deliberately chosen to evaluate MCMC performance in a discrete, stochastic setting with a discrete-valued arbitrary objective.\\ 

More specifically, the environment corresponds to the classic tic-tac-toe game: a two-player, deterministic, turn-based game in which the player and opponent alternately place their respective tokens $X$ and $O$ on a $3 \times 3$ grid. The objective of the player is to be the first to align three of their tokens consecutively in a row, column, or diagonal, and likewise for the opponent. \iffoo An animated illustration of tic-tac-toe is displayed in Figure \ref{fig: xo move}.
\begin{figure}[H]
    \centering
      \animategraphics[autoplay,loop,width=0.62\linewidth]{4}{XO_gif/frame_}{001}{278}
\caption{Illustration of tic-tac-toe (displaying $\boldsymbol{\hat{\theta}}^{\text{GA},\text{(II)}}_{\nu = 1\times 10^{-6}}$ evaluated on arbitrary out-of-sample games).}
\label{fig: xo move}
\end{figure}
\fi
\subsubsection{Encoding}
Mathematically, the $3 \times 3$ game board is represented by a matrix $\mathbf{M}_{3 \times 3}$, where an $X$ token is encoded as $-1$, an $O$ token as $+1$, and empty cells as $0$. For convenience, the matrix $\mathbf{M}$ is vectorized by row into a single 9-dimensional vector $\mathbf{m}$, thereby providing a compact representation of the board state.  In the configuration considered, the player is assumed to make the first move, assigned as the $O$ player.\\

Furthermore, we define a game of tic-tac-toe to be complete at time $k$, where the player places the $O$ token on the grid at discrete times $\tau_k^{(p)} = 1, \ldots, T_k^{(p)} \leq 5$ (since the player moves first), and the opponent places the $X$ token on the grid at discrete times $\tau_k^{(o)} = 1, \ldots, T_k^{(o)} \leq 4$,  while the overall time index for the $k^{\text{th}}$ game is given by $\tau_k = 1, \ldots, T_k \leq 9$. Hence $\mathbf{m}^{\tau_k - 1}$ would be the current board state before the player or opponent places their token at time $\tau_k$. Furthermore, we define the set of opponent's decisions/actions at the conclusion of the $k^{th}$ game in sequence $\mathcal{O}_k = \Big[ a_{\tau_k^{(o)} = 1}, \ldots, a_{\tau_k^{(o)} = T_k^{(o)}} \Big]'$ for any action $a \in \{1, 2\ldots, 9\}$ corresponding to available board positions. Furthermore, we define the set of opponent's actions for $K$ games as matrix $\boldsymbol{\mathcal{O}}_K = \Big[\mathcal{O}_1, \ldots, \mathcal{O}_K     \Big]$.

\subsubsection{The game's outcome}
At each turn of the game, the current state must be evaluated to determine whether the game has reached a terminal condition: namely, a win, loss, or draw, or whether play should continue. This involves computing the sums of each row, column, and diagonal of $\mathbf{M}$ to check for a winning configuration. Specifically, a sum of $-3$ or $3$ indicates a win for the $X$ or $O$ player, respectively.\\

We may evaluate the game state by transposing and post-multiplying $\mathbf{m}$ by a state-matrix:
\begin{align}
    \mathbf{S} & = 
    \begin{bmatrix}
        1 & 0 & 0 & 1 & 0 & 0 & 1 & 0\\
        1 & 0 & 0 & 0 & 1 & 0 & 0 & 0 \\
        1 & 0 & 0 & 0 & 0 & 1 & 0 & 1 \\
        0 & 1 & 0 & 1 & 0 & 0 & 0 & 0 \\
        0 & 1 & 0 & 0 & 1 & 0 & 1 & 1 \\
        0 & 1 & 0 & 0 & 0 & 1 & 0 & 0 \\
        0 & 0 & 1 & 1 & 0 & 0 & 0 & 1 \\
        0 & 0 & 1 & 0 & 1 & 0 & 0 & 0 \\
        0 & 0 & 1 & 0 & 0 & 1 & 1 & 0 \\
    \end{bmatrix}_{9 \times 8} \nonumber
\end{align}
hence, if any one of the $8$ entries in the game-state evaluation $\mathbf{m}'\mathbf{S}$ is equal to $-3$, $X$ has won. Likewise, if any of the entries is equal to $+3$, $O$ has won.\\

To assess whether a draw has occurred, it is necessary to verify that each of the three rows, three columns, and both diagonals of $\mathbf{M}$ contains at least one $X$ and one $O$. This ensures that no player can achieve a winning alignment in any direction. To do this, let $\mathbf{m}_+ = \left[ \mathbb{I}(m_i =+ 1) \right]_{i=1}^9$ where $\mathbb{I}(\cdot)$ is the indicator function, applied element-wise to the entries of $\mathbf{m}$. Hence, $\mathbf{m}_+$ is a 9-dimensional binary vector with ones at positions where the corresponding entries of $\mathbf{m}$ are equal to $+1$, and zeros elsewhere. Similarly, we define the 9-dimensional binary vector $\mathbf{m}_-$ analogously to indicate the positions of entries equal to $-1$ in $\mathbf{m}$. A draw has then occured if $(\mathbf{v}_+ )_{1\times 8}'(\mathbf{v}_{-})_{8\times 1} = 8$ where 8-dimensional vector $\mathbf{v}_+ = \left[ \mathbb{I}\left( \left(\mathbf{m}_{+}^{'} \mathbf{S}\right)_i > 0 \right) \right]_{i=1}^8$ and $\mathbf{v}_{-}= \left[ \mathbb{I}\left( \left(\mathbf{m}_{-}^{'} \mathbf{S}\right)_i > 0 \right) \right]_{i=1}^8$.\\

We denote $\rho_{\tau_k}$ to denote the value of the winning token or zero (for a draw) if the game is terminal after the player or opponent has placed their token at time $\tau_k = 1, 2 \ldots, T_k$ for the $k^{th}$ tic-tac-toe game. Hence:
\begin{align}
    \rho_{\tau_k} &= 
    \begin{cases}
        +1, & \text{if } \exists\, i \in \{1, \ldots, 8\} \text{ such that } \left({\left(\mathbf{m}^{\tau_k }\right)}' \mathbf{S}\right)_i = +3, \\    
        -1, & \text{if } \exists\, i \in \{1, \ldots, 8\} \text{ such that } \left({\left(\mathbf{m}^{\tau_k }\right)}' \mathbf{S}\right)_i = -3, \\
        0, & \text{if } \left(\mathbf{v}_{+}^{\tau_k } \right)' \left(\mathbf{v}_{-}^{\tau_k} \right)= 8, \\
        \text{NULL}, &\text{if game is not terminal}. \nonumber
    \end{cases}
\end{align}

\subsubsection{Control}
We control player decisions/actions at time $\tau_k^{(p)} = 1, \ldots, T_k^{(p)} \leq 5$ for the $k^{th}$ tic-tac-toe game  through the means of the control vector $\mathbf{ct}\left(\mathbf{m}^{\tau_k - 1},\boldsymbol{\theta}^{\text{Decision}}\right) \in \mathcal{A}_{\tau_k} \subseteq \{1, 2\ldots, 9\}$, where $\mathcal{A}_{\tau_k}$ represents the subset of available board positions on $\mathbf{m}^{\tau_k - 1}$ before the player opts to play at time $\tau_k$ for some parameter configuration $\boldsymbol{\theta}^{\text{Decision}}\in \mathbb{R}^R$. Naturally, once a grid position is occupied by a token, it becomes unavailable for subsequent moves by either the player or the opponent. Hence $\mathcal{A}_{\tau_k} = \left\{ i \in \{1, \ldots, 9\} : m_i^{\tau_k - 1} = 0 \right\}$. The player action selection is probabilistic and derived from a softmax distribution over logits. Hence, for $\ell_a$ being the logit score for any valid action $a \in \mathcal{A}_{\tau_k}$, the probability of selecting that action is:
\begin{align}
    \sigma_L(a \mid \mathbf{m}^{\tau_k - 1},\boldsymbol{\theta}^{\text{Decision}}) = \frac{\exp(\ell_a)}{\sum_{a' \in \mathcal{A}_{\tau_k}}\exp(\ell_{a'})} \nonumber.
\end{align}
The selected action $a^*$ corresponds to the action with the highest probability, that is, $a^* = \operatorname*{argmax}_{a\in \mathcal{A}_{\tau_k}} \sigma_L(a \mid \mathbf{m}^{\tau_k - 1},\boldsymbol{\theta}^{\text{Decision}})$. Invalid actions (i.e., $a \notin \mathcal{A}_{\tau_k}$) are assigned $\ell_a = -\infty$, ensuring a zero probability is attributed to that specific invalid action. Now the interface between a model and the player action is undergone through this control vector for which $\mathbf{ct}: (\mathbf{m}^{\tau_k - 1}, \boldsymbol{\theta}^{\text{Decision}}) \rightarrow \mathbf{model} \left(\boldsymbol{\Omega}\left(\mathbf{m}^{\tau_k - 1}\right),  \boldsymbol{\theta}^{\text{Decision}} \right) \xrightarrow{\sigma_L(.)} a^* \in \mathcal{A}_{\tau_k} $ where $\boldsymbol{\theta}^{\text{Decision}}$ is fixed throughout all $k = 1, 2, \ldots, K$ tic-tac-toe games for all player turns at times $\tau_k^{(p)}  =1, \ldots, T_k^{(p)} \leq 5$, and all player decisions are based on this fixed parameterization $\boldsymbol{\theta}^{\text{Decision}}$. In the framework of using a neural network as our model, we define $\boldsymbol{\Omega}:\mathbf{m}^{\tau_k - 1} \rightarrow \mathbf{a}^0 \in \mathbb{R}^{d_0}$ which signifies the vector of input nodes for times $\tau_k^{(p)}  =1, \ldots, T_k^{(p)} \leq 5$. Furthermore, $\boldsymbol{\theta}^\text{Decision}$ are the weights and biases of the neural network, $\mathbf{w}^{\text{Decision}} \in \mathbb{R}^R$. 

\paragraph{Feature engineering}
We construct the feature vector using two inputs: the current board state before the player places a token at time $\tau_k$, $\mathbf{m}^{\tau_k - 1}$, as well as the game-state evaluation, ${\left(\mathbf{m}^{\tau_k - 1}\right)}' \mathbf{S}$. The former encodes the spatial configuration of tokens on the board capturing positional information essential to the learning process. The latter provides a structured summary of token alignments across rows, columns, and diagonals, serving as a higher-level representation that facilitates the identification of a win or loss of the player. \\

While the raw board configuration, $\mathbf{m}^{\tau_k - 1}$, contains all information required to determine the optimal action, relying solely on this representation would require the neural network to implicitly learn the deterministic game logic governing winning, losing, and blocking configurations. This introduces unnecessary complexity into the learning problem. Hence, to mitigate this, we additionally include the game-state evaluation, ${\left(\mathbf{m}^{\tau_k - 1}\right)}' \mathbf{S}$, which provides an explicit summary of token alignments across rows, columns, and diagonals. This representation highlights strategically salient patterns, such as imminent wins or threats, thereby reducing the burden on the neural network to rediscover deterministic game logic. Therefore, our $1^{st}$ set of input nodes are defined as
$\mathbf{a}_{1}^0 = \mathbf{m}^{\tau_k - 1}$ which represents the board state at time $\tau_k-1$ and the $2^{nd}$ set of input nodes is given by $\mathbf{a}_{2}^0 ={\left(\mathbf{m}^{\tau_k - 1}\right)}' \mathbf{S}$ which denotes the game-state evaluation at time $\tau_k-1$.

\subsubsection{The arbitrary objective}
Consider an arbitrary objective where, for a given parameter configuration $\boldsymbol{\theta} = \boldsymbol{\theta}^{\text{Decision}}   \in  \mathbb{R}^{R}$ and after playing $K$ number of tic-tac-toe games, we count the number of times the player's token $O$ (encoded as $+1$) won the game, denoted as $\sum_{k = 1}^K \mathbb{I}\left(\rho_{T_k}\left( \boldsymbol{\theta}\right) = +1 \right) \in \{0, 1, \ldots K \}$. Hence:
\begin{align}
    \operatorname*{argmax}_{\boldsymbol{\theta}}\text{Obj} \left( \boldsymbol{\theta}\right) & = \operatorname*{argmax}_{\boldsymbol{\theta}} \frac{1}{K}\sum_{k = 1}^K \mathbb{I}\left(\rho_{T_k}\left( \boldsymbol{\theta}\right) = +1 \right)\nonumber. 
\end{align}
By including L2 regularization, our L2 penalized objective becomes:
\begin{align}
     \operatorname*{argmax}_{\boldsymbol{\theta}} \left(\frac{1}{K}\sum_{k = 1}^K \mathbb{I}\left(\rho_{T_k}\left( \boldsymbol{\theta}\right) = +1 \right) -  \nu \lVert\boldsymbol{\theta} \rVert^2 \right)
     \label{eq: xo obj}.
\end{align} 
Now congruent to Section \ref{sec: the obj}, $\operatorname*{argmax}_{\boldsymbol{\theta}} p( \boldsymbol{\theta} \mid \mathcal{D})$ must be equivalent to maximising the objective function in Equation \ref{eq: xo obj}. This is achieved by ensuring the likelihood $p(\mathcal{D} \mid \boldsymbol{\theta})$ is monotonic increasing with respect to $\sum_{k = 1}^K \mathbb{I}\left(\rho_{T_k}\left( \boldsymbol{\theta}\right) = +1 \right)$, that is,  $p(\mathcal{D} \mid \boldsymbol{\theta}) \propto \sum_{k = 1}^K \mathbb{I}\left(\rho_{T_k}\left( \boldsymbol{\theta}\right) = +1 \right)$.

\paragraph{Exponential-based likelihood}
Section \ref{sec: the likelihood} elucidated that when MCMC is employed primarily as a mode-seeking algorithm, that is, the mode of the conditional $p( \boldsymbol{\theta} \mid \sigma_\theta^2, \mathcal{D})$, rather than for full Bayesian inference, the necessity of an explicit and well-defined likelihood function linking the parameters $\boldsymbol{\theta}$ to the observed data becomes less critical. In such settings, it suffices to employ any monotonically increasing transformation of the objective function to guide the proposal mechanism of the MH algorithm, thereby biasing the random walk toward regions of high-likelihood (high-valued objective) regions to sample around a dominant mode of the conditional.\\

Accordingly, we adopt an exponential transformation as the chosen monotonic function, serving as a surrogate for the traditional likelihood, to facilitate efficient exploration of high-valued objective regions in the parameter space. Hence for $\boldsymbol{\theta} \in \mathbb{R}^{R}$, $\sum_{k = 1}^K \mathbb{I}\left(\rho_{T_k}\left( \boldsymbol{\theta}\right) = +1 \right)  \in \{0, 1, \ldots, K \}$ hence $\frac{1}{K}\sum_{k = 1}^K \mathbb{I}\left(\rho_{T_k}\left( \boldsymbol{\theta}\right) = +1 \right) \in [0, 1] $ and sharpness $\beta \in \mathbb{R}^+$, we have our new likelihood as:

\begin{align}
    p(\mathcal{D} \mid \boldsymbol{\theta}) &=  \exp \left(\beta \cdot \frac{1}{K} \sum_{k = 1}^K \mathbb{I}\left(\rho_{T_k}\left( \boldsymbol{\theta}\right) = +1 \right)   \right),
    \label{eq: likelihood xo exp}
\end{align}
with the log of conditional posterior being, noting the prior $\boldsymbol{\theta} \mid \sigma_\theta^2 \sim \mathcal{N}(\mathbf{0}, \sigma_\theta^2 \mathbf{I}_S)$ :
\begin{align}
\log\left(p(\boldsymbol{\theta} \mid \sigma_\theta^2, \mathcal{D})\right) 
& \propto  \log\left(p( \mathcal{D} \mid \boldsymbol{\theta} )\right) + \log\left(p(\boldsymbol{\theta} \mid \sigma_\theta^2)\right)  \nonumber \\
& \propto \beta \cdot \frac{1}{K} \sum_{k = 1}^K \mathbb{I}\left(\rho_{T_k}\left( \boldsymbol{\theta} \right) = +1 \right) 
- \frac{1}{2\sigma_\theta^2} \lVert \boldsymbol{\theta} \rVert^2  - \frac{S}{2}\log\left( 2\pi \sigma_\theta^2 \right)  \label{eq: posterior xo exp}.
\end{align}

\subsubsection{Results: Effects of regularization} \label{sec: eor xo}
To train the tic-tac-toe agent, we simulate $K = 100$ tic-tac-toe games in which the model learns to play against an opponent whose behavior is governed by a random decision policy. Specifically, the opponent selects among the available (i.e., unoccupied) grid positions uniformly at random when placing its $X$ token. As a result, the trajectory of each game, and by extension, the opponent's decision-making process, is contingent on a random seed for which we use seed values $\{\omega_i^{\text{Train}} \}_{i=1}^{100}$ corresponding to the $K = 100$ tic-tac-toe games. Now since the player's decisions, determined by the solution $\boldsymbol{\hat{\theta}}_\nu$, directly influences the set of random opponent decision sequences $\boldsymbol{\mathcal{O}}_{100}^{\text{Train}}$ for the $K = 100$ games (as the opponent may only place their $X$ token on unoccupied grid positions), we cannot assume that the set of random opponent decision sequences remains fixed across all solutions. For example, the solution $\boldsymbol{\hat{\theta}}_{\nu_1}$ will likely induce a different $\boldsymbol{\mathcal{O}}_{100}^{\text{Train}}$ than $\boldsymbol{\hat{\theta}}_{\nu_2}$ for $\nu_1 \neq \nu_2$, even when each of the $k =1 , \ldots, K$ tic-tac-toe games are initialized with the same seed. Consequently, the best we can do to ensure that the random opponent’s behavior is both deterministic and reproducible across different solutions, is to control it via a fixed seed, in an attempt to allow for controlled evaluation and consistent comparison of the agent’s performance across training runs. We may extend this notion further by observing that during the optimization process, each candidate solution dictates the set of random opponent decision sequences. As a result, the effective optimization surface to be maximized is not fixed but changes across iterations under this framework.\\

Furthermore, we define the test set as consisting of $K = 10{,}000$ simulated tic-tac-toe games, each initialized by a distinct seed value $\omega_i^{\text{Test}}$ such that $\omega_i^{\text{Test}} \neq \omega_j^{\text{Train}}$ for all $i, j$.  However, this condition alone does not ensure that all random opponent decision sequences in the test set, $\mathcal{O}_k^{\text{Test}}$ for $k \in \{1,\ldots, 10000 \}$, are disjoint from those encountered during training, $\mathcal{O}_k^{\text{Train}}$ for $k \in \{1,\ldots, 100 \}$. Additionally, to prevent the inflation or degradation of performance due to repeated identical random opponent decision sequences in the test set, we enforce that $\mathcal{O}_k^{\text{Test}} \neq \mathcal{O}_l^{\text{Test}}$ for all $k, l \in \{1, \ldots, 10000\}$ with $k \neq l$. That is, all random opponent decision sequences used during testing are mutually distinct. Hence, to guarantee that all test games are genuinely out-of-sample (and unique), we iteratively cycle through candidate seed values $\omega_j^{\text{Test}}$ until we obtain a collection of $K = 10{,}000$ test games whose random opponent decision sequences are distinct from those observed in the $K = 100$ training games (as well as being distinct from each other). Hence $j \geq 10,000$ for our out-of-sample test seed values $\omega_j^{\text{Test}}$. In doing so, we ensure that the training and test environments are disjoint, thereby enabling a valid assessment of the agent’s generalization performance to previously unseen opponent behaviors. As before, however, since the random opponent decision sequences are governed by the player's decisions, controlled by $\boldsymbol{\hat{\theta}}_\nu$, each $\boldsymbol{\hat{\theta}}_\nu$ would give rise to a different set of random opponent decision sequences  $\boldsymbol{\mathcal{O}}_{10,000}^{\text{Test}}$, hence rendering the $K = 10,000$ out-of-sample games to be somewhat different across solutions derived.  
\\

To evaluate the impact of the regularization strength $\nu$, we apply the estimator $\boldsymbol{\hat{\theta}}^{\text{GA}}_\nu$ to the test set and assess both in-sample and out-of-sample performance across a range of $\nu$ values, as reported in Table \ref{table: xo for a_1^0, a^0}. We denote by $\boldsymbol{\hat{\theta}}^{\text{GA},\text{(I)}}_\nu$ the solution obtained from $\overset{(\text{I})}{\mathbf{model}}$, which solely uses $\mathbf{a}_1^0$ as its feature vector. Likewise, $\boldsymbol{\hat{\theta}}^{\text{GA},\text{(II)}}_\nu $ corresponds to $\overset{(\text{II})}{\mathbf{model}}$, which incorporates the full feature vector $\mathbf{a}^0_{(9+8) \times 1} = \left[\left(\mathbf{a}_1^0\right)', \left(\mathbf{a}_2^0\right)'\right]'$. Furthermore, we also apply the estimator $\boldsymbol{\hat{\theta}}^{\text{RS}}_\nu$ on the test set in order to establish a baseline against which the performance of the GA can be compared (just for $\overset{(\text{II})}{\mathbf{model}}$).\\

As shown in Table \ref{table: xo for a_1^0, a^0 large}, there is a clear trend of decreasing performance, both in-sample and out-of-sample, as the regularization strength increases for both models. This behavior is consistent with underfitting due to excessive regularization, that is, the model becomes overly constrained. Moreover, the results suggest that the use of a GA is necessary to achieve improved performance on the in-sample set, as it consistently outperforms RS at low values of $\nu$. However, this pattern does not persist across all regularization strengths: at high regularization levels, RS appears to yield better in-sample solutions than the GA at times. This observation implies that the fine-tuning capability of the GA is most beneficial when the model is not heavily constrained. That is, under such conditions, the GA's exploitation properties appear to play a critical role in refining existing parent solutions.\\

Additionally, we observe, that in the absence of regularization ($\nu = 0$), $\overset{(\text{II})}{\mathbf{model}}$ slightly outperforms $\overset{(\text{I})}{\mathbf{model}}$ in both the in-sample and out-of-sample sets. Nonetheless, we refrain from making general claims regarding comparative performance across varying values of $\nu$ between the two models, as the two models differ in complexity with respect to their number of input nodes: $\overset{(\text{I})}{\mathbf{model}}$ utilizing  $\mathbf{a}_1^0$ as its feature vector and $\overset{(\text{II})}{\mathbf{model}}$ utilizing  $\mathbf{a}^0$ as its feature vector. Consequently, a given value of $\nu$ cannot be interpreted as exerting the same regularization effect (capturing the goal of making a model less complex) across both models, and direct comparisons of regularization magnitudes should be treated with caution. We conjecture, however, that the best out-of-sample performance may be attained at regularization strengths corresponding to $\nu \in [10^{-6}, 10^{-4}]$ for both models, where there seems to be an effective balance between in-sample performance and out-of-sample generalization.\\

Importantly, Table \ref{table: xo for a_1^0, a^0 large} also demonstrates that both models significantly outperform a baseline agent governed by purely random decision-making in cases where the model is not overly constrained, that is, when $\nu$ is not excessively large. This baseline agent (playing as the first mover) also faces a random opponent for $K = 10,000$ tic-tac-toe games all of which have distinct random opponent decision sequences ($\mathcal{O}_k \neq \mathcal{O}_l$ for all $k, l \in \{1, \ldots, 10000\}$ with $k \neq l$). The random baseline achieves a normalized win percentage of only $57.16\%$, which is consistently exceeded by the learned agents under moderate regularization strengths, shown in both Table \ref{table: xo for a_1^0, a^0 large} and \ref{table: xo for a_1^0, a^0}. \\

Now to investigate the local sensitivity of model performance to small variations in regularization strength, Table \ref{table: xo for a_1^0, a^0} reports in-sample and out-of-sample performance across finely spaced values of $\nu$ for both $\overset{(\text{I})}{\mathbf{model}}$ and $\overset{(\text{II})}{\mathbf{model}}$. The results reveal a non-monotonic relationship between performance and $\nu$: small increases in regularization do not uniformly degrade performance and, in certain cases, even yield improvements, particularly in out-of-sample performance.

\begin{table}[H]
    \centering
    \begin{tabular}{ccccccc}
       & \multicolumn{2}{c}{$\overset{(\text{I})}{\mathbf{model}} \left(\mathbf{a}_1^0, \boldsymbol{\hat{\theta}}^{\text{GA},\text{(I)}}_\nu\right)$} & \multicolumn{2}{c}{$\overset{(\text{II})}{\mathbf{model}} \left(\mathbf{a}^0, \boldsymbol{\hat{\theta}}^{\text{GA},\text{(II)}}_\nu\right)$} & \multicolumn{2}{c}{$\overset{(\text{II})}{\mathbf{model}} \left(\mathbf{a}^0, \boldsymbol{\hat{\theta}}^{\text{RS},\text{(II)}}_\nu\right)$} \\
         \cmidrule(lr){2-3} \cmidrule(lr){4-5} \cmidrule(lr){6-7}  
        $\nu$  & In-Sample & Out-of-Sample & In-Sample & Out-of-Sample  & In-Sample & Out-of-Sample   \\ 
        \hline
        $0.000001$ &97 &61.72  & $99$ & $83.92$ & 94 & 67.38 \\
        $0.00001$ &99 & 73.78 & $99$ & $74.84$ & 96 & 83.16 \\
        $0.0001$   &97 & 82.96 & $98$ & $66.96$ & 97 & 62.98\\
        $0.001$   &92 & 76.28 & $93$ & $59.96$ & 68 & 59.46\\
        $0.01$    &82 &77.58 & $85$ & $59.36$ & 45 & 47.64 \\
        $0.1$     &57 &45.92 & $74$ & $69.10$ & 77 & 46.92\\
        $1$        &44 &47.82 & $52$ & $46.04$ & 67 & 54.10\\
        \hline
    \end{tabular}
    \caption{The normalized number of $O$ wins, as a percentage $\left(\frac{100}{K} \sum_{k = 1}^K \mathbb{I}\left(\rho_{T_k}\left( \boldsymbol{\theta}\right) = +1 \right)  \right)$ for in-sample ($K = 100$) and out-of-sample ($K = 10, 000$) sets across regularization strengths $\nu$ using $\overset{(\text{II})}{\mathbf{model}} \left(\mathbf{a}^0, \boldsymbol{\hat{\theta}}^{\text{GA},\text{(II)}}_\nu\right)$ and $\overset{(\text{II})}{\mathbf{model}} \left(\mathbf{a}^0, \boldsymbol{\hat{\theta}}^{\text{RS},\text{(II)}}_\nu\right)$.}
    \label{table: xo for a_1^0, a^0 large}
\end{table}

\begin{table}[H]
    \centering
    \begin{tabular}{ccccc}
       & \multicolumn{2}{c}{$\overset{(\text{I})}{\mathbf{model}} \left(\mathbf{a}_1^0, \boldsymbol{\hat{\theta}}^{\text{GA},\text{(I)}}_\nu\right)$} & \multicolumn{2}{c}{$\overset{(\text{II})}{\mathbf{model}} \left(\mathbf{a}^0, \boldsymbol{\hat{\theta}}^{\text{GA},\text{(II)}}_\nu\right)$} \\
         \cmidrule(lr){2-3}  \cmidrule(lr){4-5} 
        $\nu$ &  In-Sample & Out-of-Sample& In-Sample & Out-of-Sample   \\ 
        \hline
        $0.0000$ & $98$ & $73.62$ & $99$ & $77.86$\\
        $0.0001$ & $97$ & $82.96$ & $98$ & $66.96$\\
        $0.0002$ & $94$ & $64.68$  & $97$ & $76.82$\\
        $0.0003$  & $98$ & $65.32$ & $95$ & $77.46$ \\
        $0.0004$ & $96$ & $77.20$ & $97$ & $73.72$\\
        $0.0005$ & $95$ & $78.58$ & $95$ & $73.74$\\
        $0.0006$ & $97$ & $63.00$ & $95$ & $79.50$\\
        $0.0007$ & $95$ & $58.96$ & $95$ & $72.86$\\
        $0.0008$ & $95$ & $73.20$ & $91$ & $62.00$\\           
        $0.0009$ & $96$ & $73.74$ & $94$ & $75.40$\\
        \hline
    \end{tabular}
    \caption{The normalized number of $O$ wins, as a percentage $\left(\frac{100}{K} \sum_{k = 1}^K \mathbb{I}\left(\rho_{T_k}\left( \boldsymbol{\theta}\right) = +1 \right)  \right)$ for in-sample ($K = 100$) and out-of-sample ($K = 10, 000$) sets across regularization strengths $\nu$ using $\overset{(\text{I})}{\mathbf{model}} \left(\mathbf{a}_1^0, \boldsymbol{\hat{\theta}}^{\text{GA},\text{(I)}}_\nu\right)$ and $\overset{(\text{II})}{\mathbf{model}} \left(\mathbf{a}^0, \boldsymbol{\hat{\theta}}^{\text{GA},\text{(II)}}_\nu\right)$.}
    \label{table: xo for a_1^0, a^0}
\end{table}

\subsubsection{Results: MCMC} \label{sec: xo mcmc}
We exclusively employ Equation \ref{eq: likelihood xo exp} as the likelihood function in the MH algorithm presented in Section \ref{sec: 2-sample}, in this section. Although one could reasonably argue for the use of alternative likelihoods, such as the binomial and beta-based forms presented in Section \ref{sec: likelihoods}, we refrain from doing so here as a comprehensive comparison among these three likelihood formulations has already been conducted in Section \ref{sec: drones mcmc}. The present section is dedicated solely to illustrating how increasing the sharpness of the likelihood (through the parameter $\beta \in \mathbb{R}^+$) may influence the results. Furthermore, we employ $\overset{(\text{II})}{\mathbf{model}}$ in this section.\\

Illustrated in Table \ref{table: xo mcmc betas} are the normalized number of $O$ wins, expressed as a percentage
$\left( \frac{100}{K} \sum_{k = 1}^K \mathbb{I}\left(\rho_{T_k}\left( \boldsymbol{\theta} \right) = +1 \right) \right)$, computed over $100{,}000$ total MCMC iterations, of which the first $20{,}000$ were discarded as burn-in. We observe a clear trend: as the sharpness parameter $\beta \in \mathbb{R}^+$ increases, the proportion of in-sample $O$ wins tends to improve. Section \ref{sec: beta} alluded to this phenomemon: the parameter $\beta$ can be interpreted as a means of amplifying the likelihood ratio in Equation \ref{eq: alpha}, yielding the modified expression $\left(\frac{p(\mathcal{D} \mid \boldsymbol{\theta}^*)}{p(\mathcal{D} \mid \boldsymbol{\theta}^{(j)})}\right)^\beta$. Increasing $\beta$ makes the Markov chain more inclined to accept proposed solutions $\boldsymbol{\theta}^*$ that yield higher objective values, given the proportionality $p(\mathcal{D} \mid \boldsymbol{\theta}) \propto \text{Obj}(\boldsymbol{\theta})$, thereby making the MCMC sampler more likelihood-driven.\\

Furthermore, as elucidated in Section \ref{sec: the likelihood}, it was justified that the MCMC sampler should be made more likelihood-driven, as failure to do so could result in an indefinite contraction of $\boldsymbol{\theta}$ toward zero. This phenomenon is corroborated by Figure \ref{fig: inv-gammas xo}, which depicts the trajectory of $\lVert \boldsymbol{\theta}^{(j)} \rVert^2$ across the $80{,}000$ post-burn-in iterations. For small values of $\beta$, we observe that the norm $\lVert \boldsymbol{\theta}^{(j)} \rVert^2$ exhibits a slow, monotonic decline across iterations: a strong indication that the Markov chain remains in its transient phase and has not yet reached stationarity. We posit that, this monotonic decay may reflect more than just delayed convergence. At low values of $\beta$, the pseudo-likelihood, and by extension the conditional posterior $
p(\boldsymbol{\theta} \mid  \sigma_\theta^2, \mathcal{D})
$ in Equation \ref{eq: posterior xo exp}, becomes too diffuse to meaningfully constrain the parameter space. As a result, the data exerts minimal influence over the proposed solution $\boldsymbol{\theta}^{(j)}$, and the conditional posterior is effectively dominated by the prior $p(\boldsymbol{\theta} \mid \sigma_\theta^2)$. In this setting, the likelihood becomes inconsequential, and the sampling dynamics are driven almost entirely by the prior structure. Consequently, the MH acceptance probabilities $\alpha_\theta$ favour proposals $\boldsymbol{\theta}^{(j)}$ that reduce the norm $\lVert \boldsymbol{\theta}^{(j)}\rVert^2$: as is evident by Equation \ref{eq: posterior xo exp}, where a reduction in $\lVert \boldsymbol{\theta}^{(j)}\rVert^2$ results in larger conditional posterior values thereby guiding the Markov chain to search in areas where low $\lVert \boldsymbol{\theta}^{(j)}\rVert^2$ values are obtained. This behavior is a direct reflection of what the conditional posterior, being flat and prior-dominated, is prescribing. The sampler is ``doing its job": in the absence of strong likelihood information, the proposals $\boldsymbol{\theta}^{(j)}$ are simply contracting toward the origin under the influence of the prior. \\

Additionally, the multivariate effective sample sizes (ESS) reported in Table \ref{table: xo mcmc betas} provide evidence of satisfactory mixing. All reported ESS values exceed the commonly accepted threshold of 100 (as recommended by \cite{vehtari2021rank}, Section~4), which supports the claim of efficient exploration. However, it is important to emphasize that ESS is a meaningful diagnostic only after convergence has been attained. In particular, for small values of $\beta$, we observe, via the continued drift in $\lVert \boldsymbol{\theta}^{(j)} \rVert^2$, that the chain remains in a transient phase, and thus has not yet fully converged to its stationary distribution. Consequently, while we report ESS values at the end of the $80,000$ usable MCMC iterations for completeness, we interpret them with caution in the low-$\beta$ regime.\\

Furthermore, for low values of $\beta$, where the MCMC samples exhibit non-stationary behavior in $\lVert \boldsymbol{\theta}^{(j)} \rVert^2$, the resulting marginal distribution of $\sigma_\theta^2 \mid \mathcal{D}$ deviates from an inverse-gamma form. Since the conditional posterior $\sigma_\theta^2 \mid \boldsymbol{\theta}, \mathcal{D} \sim \text{Inv-Gamma}\bigl(a + \tfrac{S}{2},\, b + \tfrac{\lVert \boldsymbol{\theta} \rVert^2}{2} \bigr)$ where $a, b \approx 0$, we know that if $\lVert \boldsymbol{\theta}^{(j)} \rVert^2$ fluctuates around some constant $c$, the marginal $\sigma_\theta^2 \mid \mathcal{D}$ should also be inverse-gamma distributed with constant shape and scale parameters. That is, if $\lVert \boldsymbol{\theta}^{(j)} \rVert^2 \approx c$, then $\sigma_\theta^2 \mid \mathcal{D} \sim \text{Inv-Gamma}\bigl(a + \tfrac{S}{2},\, b + \tfrac{c}{2} \bigr)$ as illustrated in Figure \ref{fig: inv-gammas xo}.\\

Now, Table \ref{table: xo mcmc betas} suggests that, in order to obtain solutions yielding in-sample performance comparable to that achieved by traditional optimization methods such the GA in Section \ref{sec: eor xo}, the likelihood sharpness parameter $\beta$ must be increased to sufficiently concentrate the conditional posterior $p(\boldsymbol{\theta} \mid \sigma_\theta^2, \mathcal{D})$. This ensures that the sampler is more decisively guided by the pseudo-likelihood, that is, made sufficiently likelihood-driven, resulting in proposed solutions $\boldsymbol{\theta}^{(j)}$ whose norms $\lVert \boldsymbol{\theta}^{(j)} \rVert^2$ stabilize across MCMC iterations. Such stabilization indicates convergence toward a dominant mode of the conditional $p( \boldsymbol{\theta} \mid \sigma_\theta^2, \mathcal{D})$. Empirically, for the tic-tac-toe problem studied here, values of $\beta \geq 100$ appear to meet this threshold, yielding both stable posterior behavior (see the top panel of Figure \ref{fig: inv-gammas xo}) and competitive in-sample performance. 
However, a practical consideration arises when selecting an appropriate value of $\beta$. As $\beta$ becomes too large, the likelihood, and consequently the conditional posterior, becomes exceedingly sharp, leading to steep gradients around high-valued objective regions. In such cases, the MCMC chain is prone to becoming effectively ``trapped'' in these narrow peaks, as proposed moves away from the current mode receive vanishingly small acceptance probabilities $\alpha_\theta$. This occurs because the prior no longer exerts sufficient regularizing influence to counterbalance the likelihood's dominance, unlike in regimes where $\beta$ is moderate and the posterior retains a broader structure. Hence, while increasing $\beta$ can give rise to solutions with improved in-sample performance, it must be done judiciously to avoid compromising the chain's ability to explore alternative dominant modes of the conditional.\\

Additionally, we note from Table \ref{table: xo mcmc betas}, that the maximum of the log of the conditional posterior, $\log\left(p(\boldsymbol{\theta} \mid \mathcal{D}, \sigma_\theta^2)\right)$, increases as $\beta$ increases. This behavior is substantiated by Equation \ref{eq: posterior xo exp}, where it follows directly that increasing $\beta$ increases the contribution of the likelihood to the conditional posterior, thereby sharpening the overall posterior landscape.\\

\begin{table}[H]
    \centering
    \begin{tabular}{ccccc}
       & \multicolumn{2}{c}{$\frac{100}{K} \sum_{k = 1}^K \mathbb{I}\left(\rho_{T_k}\left( \boldsymbol{\theta}\right) = +1 \right) $} \\
         \cmidrule(lr){2-3}  
        Sharpness $\beta$ &  In-Sample & Out-of-Sample& $\max \{ \log \left( p\left(\boldsymbol{\theta} \mid \sigma_\theta^2, \mathcal{D}\right) \right)\}$ &ESS \\
        \hline
        $0.1$ & $56$ & $50.08$ & -254.1965  &2332.9560 \\
        $1$ & $70$ & $59.86$ & -237.8046 & 2340.9900 \\
        $10$ & $63$ & $56.68$ & -243.4494 & 4222.4970\\
        $50$  & $86$ & $72.98$ & -200.5137 & 690.1510  \\
        $100$ & $98$ & $72.62$ & -162.3267 & 927.7915 \\
        $1000$ & $96$ & $68.58$ & 733.5609 & 927.4778 \\
        \hline
    \end{tabular}
    \caption{The normalized number of $O$ wins, as a percentage $\left(\frac{100}{K} \sum_{k = 1}^K \mathbb{I}\left(\rho_{T_k}\left( \boldsymbol{\theta}\right) = +1 \right)  \right)$ for in-sample ($K = 100$) and out-of-sample ($K = 10, 000$) sets across likelihood sharpness $\beta \in \mathbb{R}^+$ using $\overset{(\text{II})}{\mathbf{model}} \left(\mathbf{a}^0, \boldsymbol{\hat{\theta}}^{\text{GA},\text{(II)}}_\nu\right)$ and $\sigma_{\text{Init}}^2 = 10$ for $100, 000$ MCMC iterations and burn-in of $20, 000$ iterations.}
    \label{table: xo mcmc betas}
\end{table}

\iffoo
\begin{figure}[H]
    \centering
\animategraphics[controls,autoplay,loop,width=0.6\textwidth]{1}{Inv-Gamma/}{1}{6}
\caption{$ \lVert \boldsymbol{\theta}^{(j)} \rVert^2$ for $j  = 1, \ldots 80, 000$ (post burn-in) with distribution of marginal $\sigma_\theta^2 \mid \mathcal{D}$ for varying likelihood sharpness $\beta$ using $\overset{(\text{II})}{\mathbf{model}} \left(\mathbf{a}^0, \boldsymbol{\hat{\theta}}^{\text{GA},\text{(II)}}_\nu\right)$ and $\sigma_{\text{Init}}^2 = 10$.}
\label{fig: inv-gammas xo}
\end{figure}
\else\begin{figure}[H]
    \centering
\begin{subfigure}{0.32\textwidth}
    \centering
    \includegraphics[width=\linewidth]{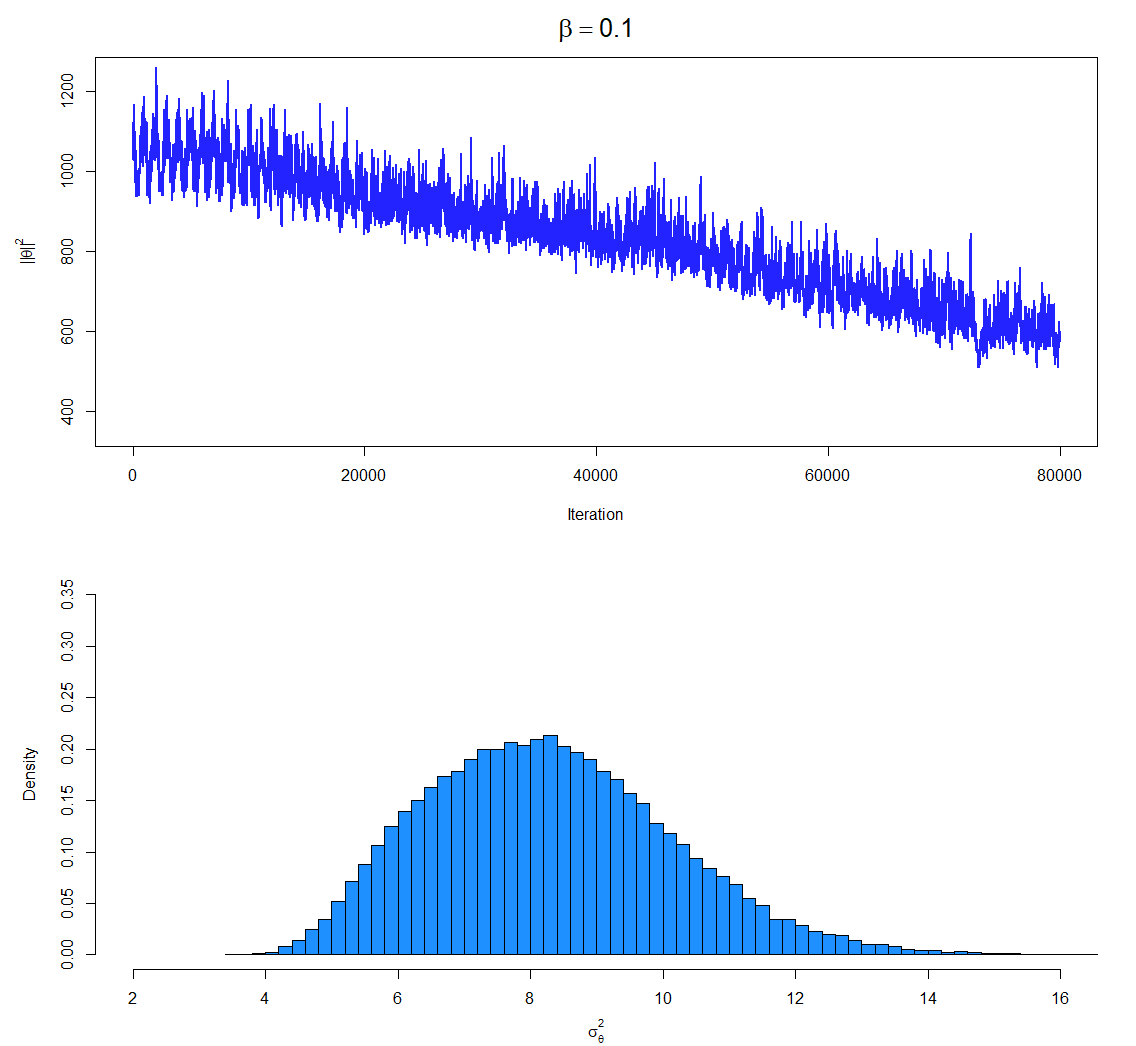}
\end{subfigure}\hfill
\begin{subfigure}{0.32\textwidth}
    \centering
    \includegraphics[width=\linewidth]{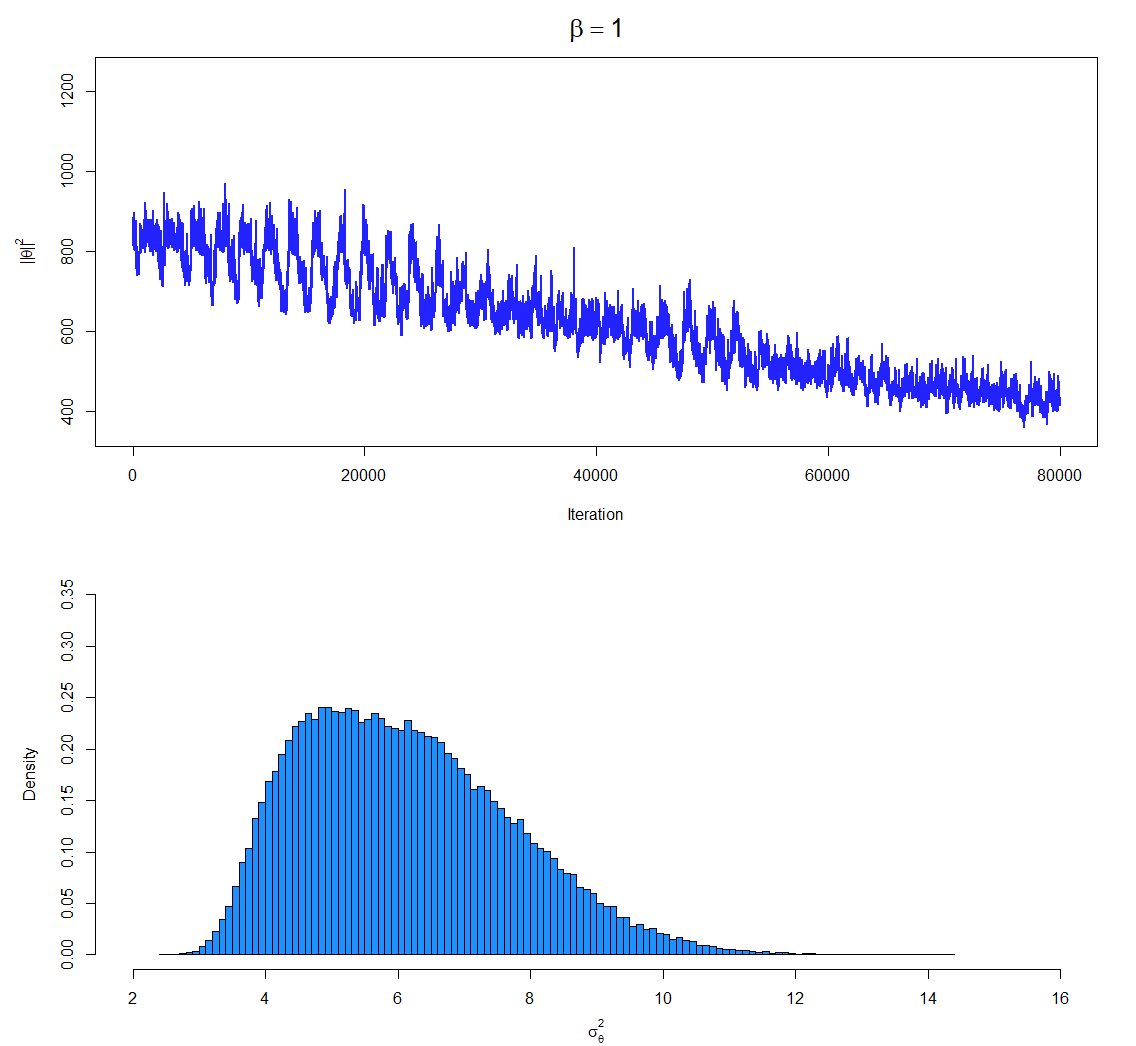}
\end{subfigure}\hfill
\begin{subfigure}{0.32\textwidth}
    \centering
    \includegraphics[width=\linewidth]{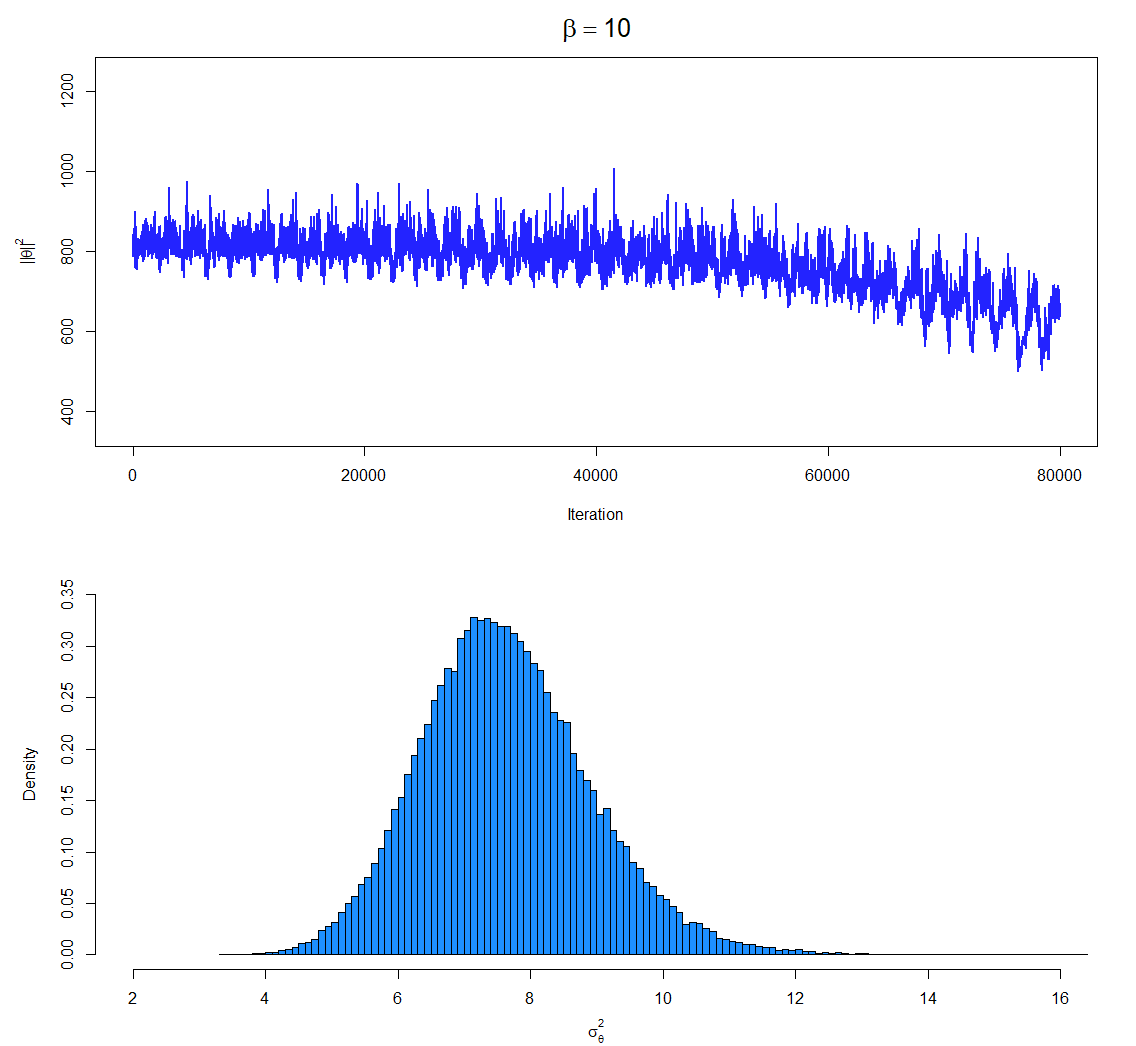}
\end{subfigure}

\vspace{0.7cm}

\begin{subfigure}{0.32\textwidth}
    \centering
    \includegraphics[width=\linewidth]{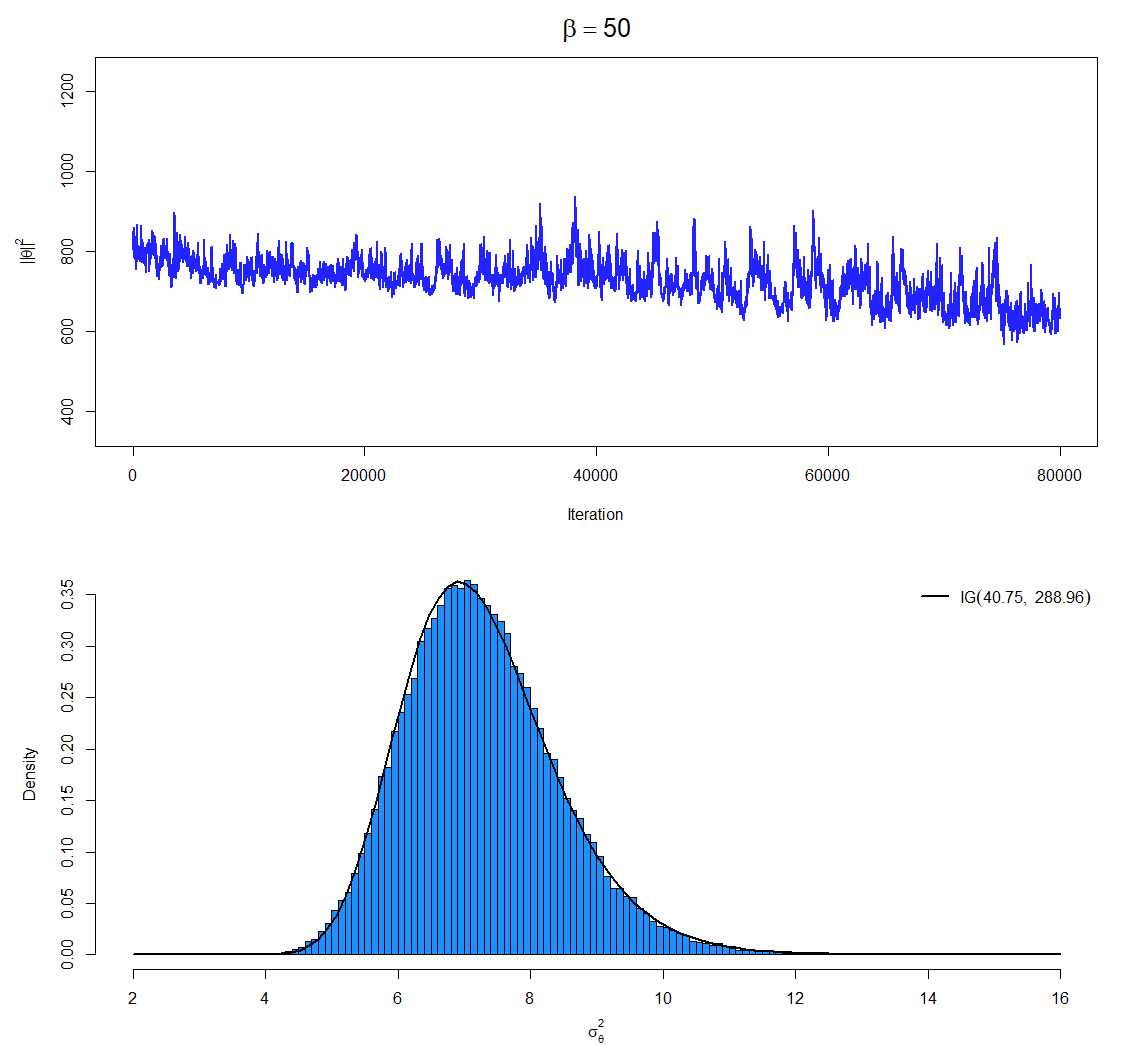}
\end{subfigure}\hfill
\begin{subfigure}{0.32\textwidth}
    \centering
    \includegraphics[width=\linewidth]{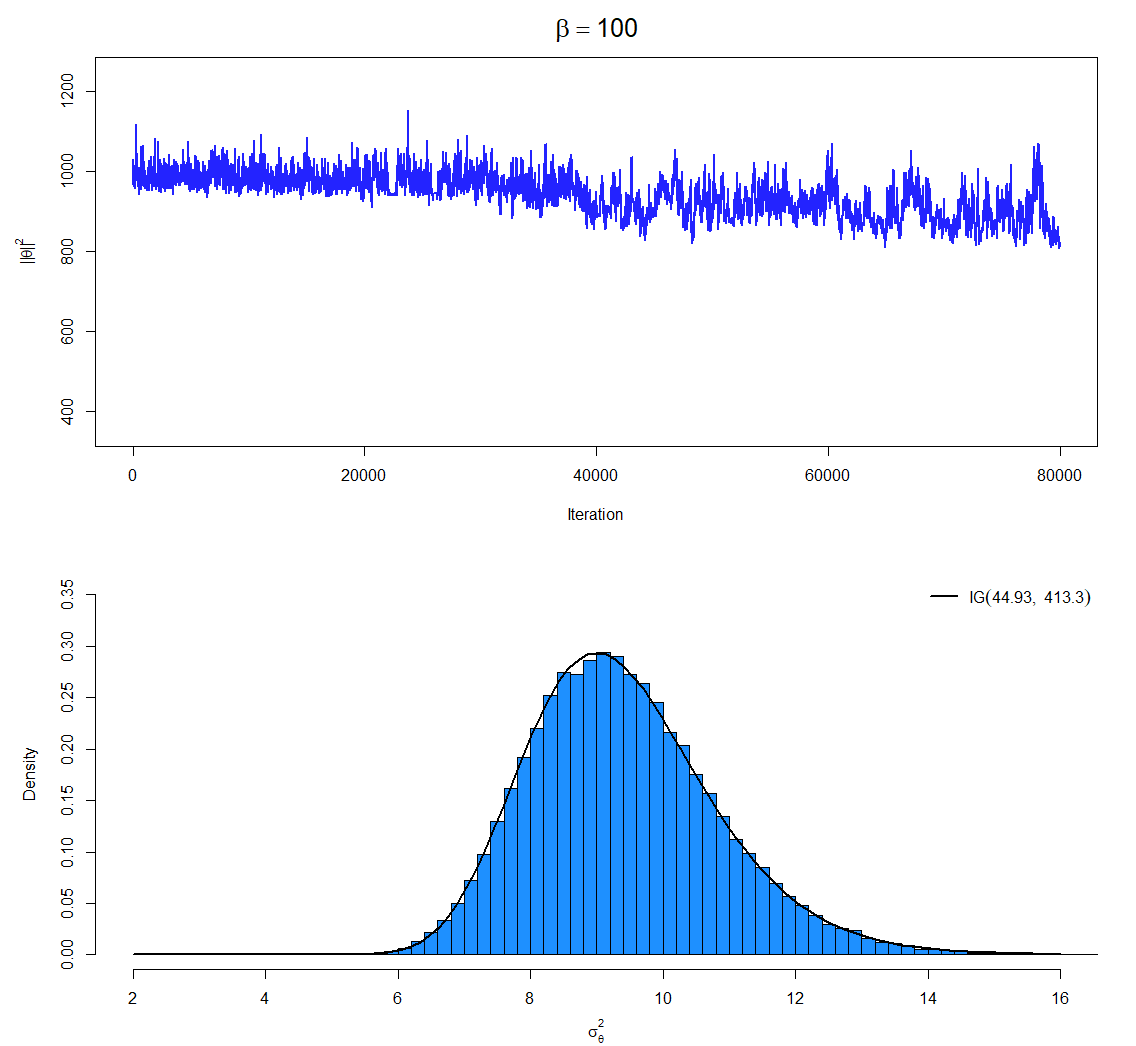}
\end{subfigure}\hfill
\begin{subfigure}{0.32\textwidth}
    \centering
    \includegraphics[width=\linewidth]{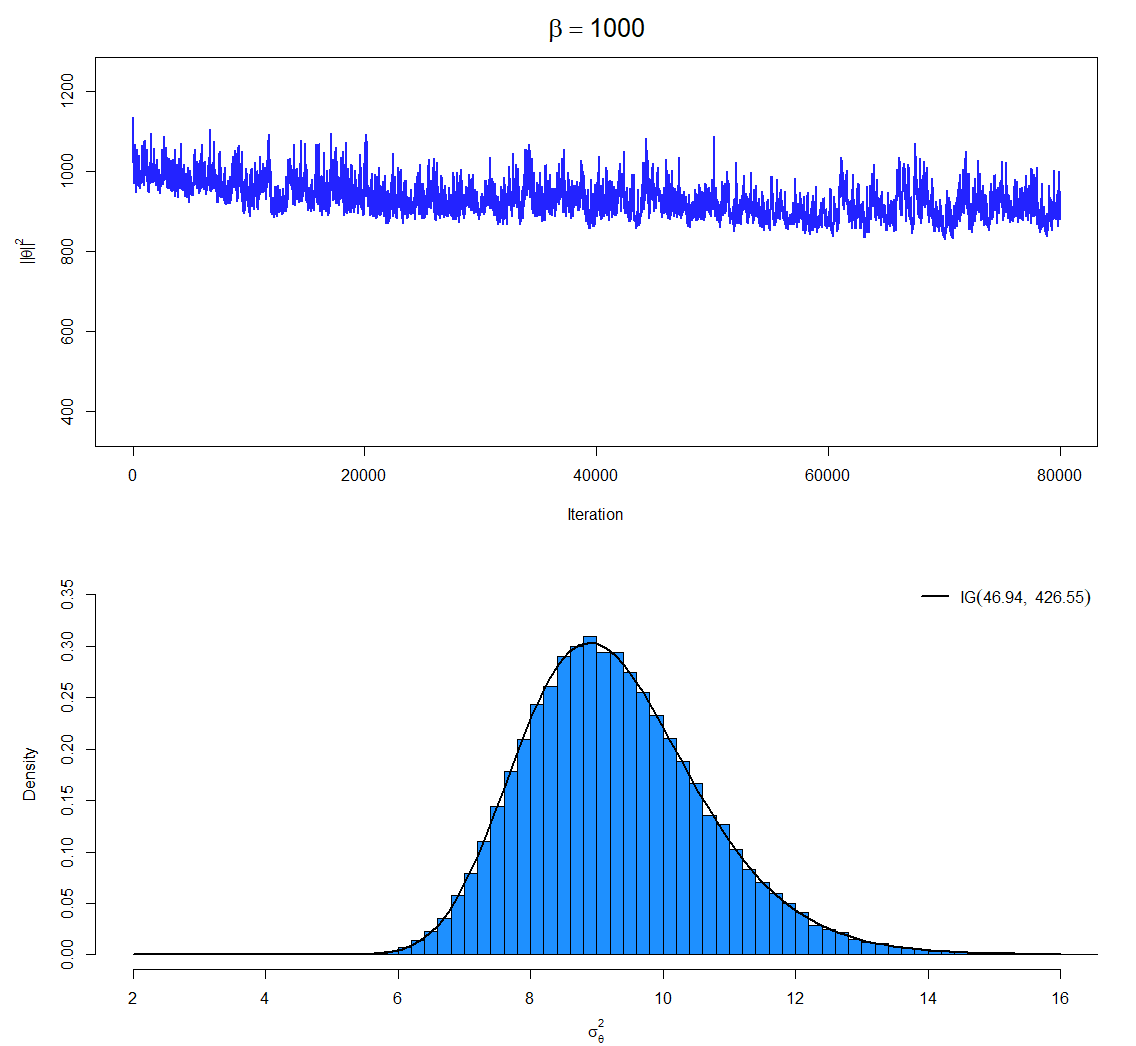}
\end{subfigure}

\caption{$ \lVert \boldsymbol{\theta}^{(j)} \rVert^2$ for $j  = 1, \ldots 80, 000$ (post burn-in) with distribution of marginal $\sigma_\theta^2 \mid \mathcal{D}$ for varying likelihood sharpness $\beta$ using $\overset{(\text{II})}{\mathbf{model}} \left(\mathbf{a}^0, \boldsymbol{\hat{\theta}}^{\text{GA},\text{(II)}}_\nu\right)$ and $\sigma_{\text{Init}}^2 = 10$.}
\label{fig: inv-gammas xo}
\end{figure}
\fi

With regard to the amount of regularization inferred from the training set, where regularization is represented by the dispersion parameter $\sigma_\theta^2 \propto \frac{1}{\nu}$, we may draw meaningful conjectures from the marginal distributions of $\sigma_\theta^2 \mid \mathcal{D}$ shown in Figure \ref{fig: inv-gammas xo}. Specifically, the variation in these distributions across different values of sharpness parameter $\beta$, suggests that different $\beta$ values inherently induce different degrees of regularization, which are in turn reflected in the MAP estimates of the parameters (as discussed in Section \ref{sec: 2-sample}, MAP estimates naturally encode a level of regularization inferred by the training set, since the marginal posterior $p(\theta_i \mid \mathcal{D})$ essentially integrates over both the remaining parameters $\boldsymbol{\theta}_{-i}$ and the dispersion parameter $\sigma_\theta^2$). That is, by examining the marginal distribution of $\sigma_\theta^2 \mid \mathcal{D}$, we gain insight into the implied concentration of the regularization strength $\nu$. Interestingly, we observe that beyond a certain threshold, approximately at $\beta = 100$ for our tic-tac-toe problem, these inverse-gamma marginals appear to converge in shape and scale, as evidenced by the similarity between the distributions for $\beta = 100$ and $\beta = 1000$ in Figure \ref{fig: inv-gammas xo}. This suggests that the strength of regularization inferred by the training set saturates beyond a certain level of likelihood sharpness.\\

Even more revealing with respect to the amount of regularization inferred by the training set, is the influence of the initial variance $\sigma_{\text{Init}}^2$ used to initialize the Markov chain. Recall that the initial proposal $\boldsymbol{\theta}^{(1)}$ is sampled from a multivariate normal distribution, $\mathcal{N}\left(\mathbf{0}_{S \times 1}, \sigma_{\text{Init}}^2 \mathbf{I}_S\right)$. Notably, when fixing $\beta = 100$ to ensure stability of $\lVert \boldsymbol{\theta}^{(j)} \rVert^2$, the resulting marginal distributions of $\sigma_\theta^2 \mid \mathcal{D}$ remain inverse-gamma distributed with approximately constant shape parameters but exhibit increasing rate parameters as $\sigma_{\text{Init}}^2$ increases, as shown in Table \ref{table: xo mcmc sigma_init}. This trend is visually supported in Figure \ref{fig: inv-gammas prop_var xo} (note that the axis scales vary across plots), which displays the inverse-gamma marginals $\sigma_\theta^2 \mid \mathcal{D}$ for increasing values of $\sigma_{\text{Init}}^2$. It is evident that both the mean and variance of the resulting distributions shift upward: the distributions move to the right and become more compressed. While this ``squashing” effect may not be immediately noticeable without paying attention to the axis scales, it highlights an important insight: since $\sigma_\theta^2 \propto \frac{1}{\nu}$, increasing the initial variance $\sigma_{\text{Init}}^2$ results in the inference of a weaker regularization strength from the training set. In other words, the amount of regularization implicitly inferred by the training set is not only a function of the likelihood sharpness $\beta$, but is also influenced by the choice of initial dispersion $\sigma_{\text{Init}}^2$.\\

\begin{table}[H]
    \centering
    \begin{tabular}{cccccc}
       & \multicolumn{2}{c}{$\frac{100}{K} \sum_{k = 1}^K \mathbb{I}\left(\rho_{T_k}\left( \boldsymbol{\theta}\right) = +1 \right) $}  \\
         \cmidrule(lr){2-3}  
        Initial Variance $\sigma_{\text{Init}}^2$ &  In-Sample & Out-of-Sample& Shape& Rate &ESS  \\ 
        \hline
        $0.1$ & $92$ & 71.18 & $41.7518$ & $3.1350$ & 414.6554 \\
        1 &  88 & 65.40 & 46.2323 &  41.3117 & 884.6408 \\
        $10$ & $98$ & $72.62$ & $44.9264$ & $413.2980$ & 927.7915\\
        $100$ & $92$ & $68.92$  & 47.28048 & 5704.8909 & 488.9163\\
        \hline
    \end{tabular}
    \caption{The normalized number of $O$ wins, as a percentage $\left(\frac{100}{K} \sum_{k = 1}^K \mathbb{I}\left(\rho_{T_k}\left( \boldsymbol{\theta}\right) = +1 \right)  \right)$ for in-sample ($K = 100$) and out-of-sample ($K = 10, 000$) sets across various initial variances $\sigma_{\text{Init}}^2$ accompanied by shape and rate parameters of marginal $\sigma_\theta^2 \mid \mathcal{D} \sim \text{Inv-Gamma}$ for likelihood sharpness $\beta = 100$.}
    \label{table: xo mcmc sigma_init}
\end{table}

\iffoo
\begin{figure}[H]
    \centering
\animategraphics[controls,autoplay,loop,width=0.6\textwidth]{1}{Inv-Gamma_prop-var/}{1}{4}
\caption{Distribution of marginal $\sigma_\theta^2 \mid \mathcal{D} \sim \text{Inv-Gamma(Shape, Rate)}$ for varying initial variance $\sigma_{\text{Init}}^2$ (plots are on different $x$ and $y$ scales).}
\label{fig: inv-gammas prop_var xo}
\end{figure}
\else
\begin{figure}[H]
    \centering
\includegraphics[width=0.49\textwidth]{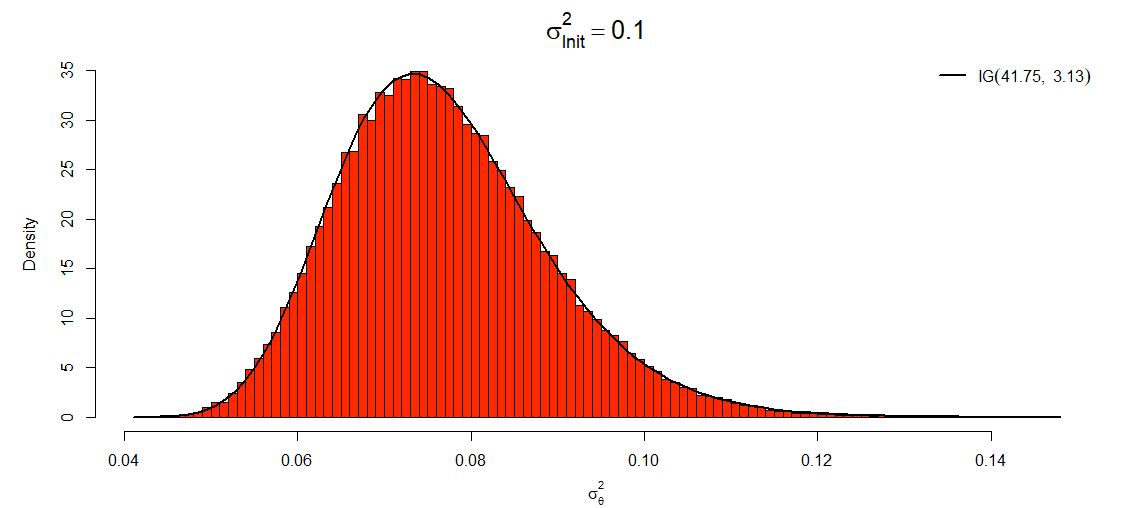}\hfill
\includegraphics[width=0.49\textwidth]{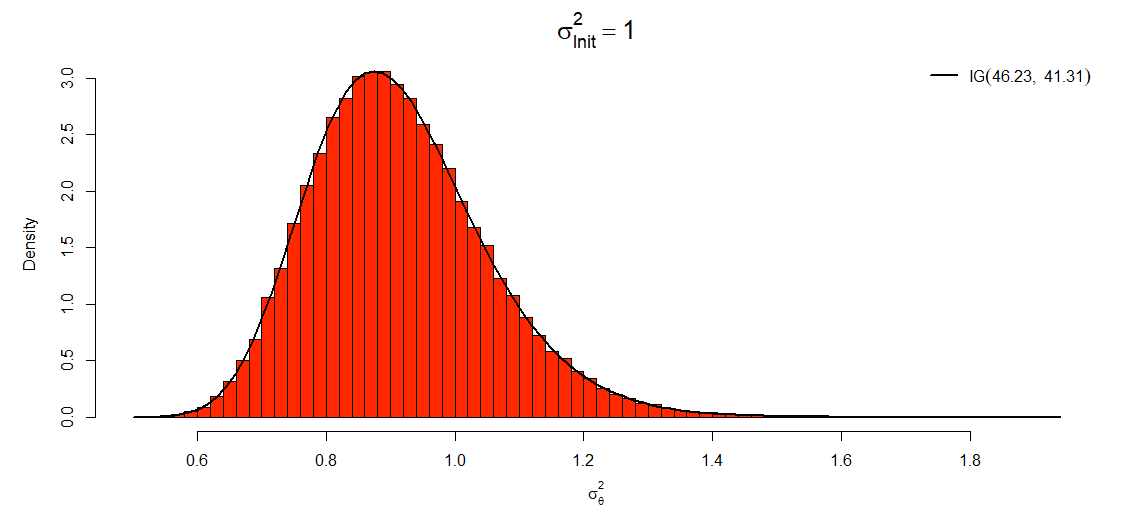}
\vspace{0.35cm}
\includegraphics[width=0.49\textwidth]{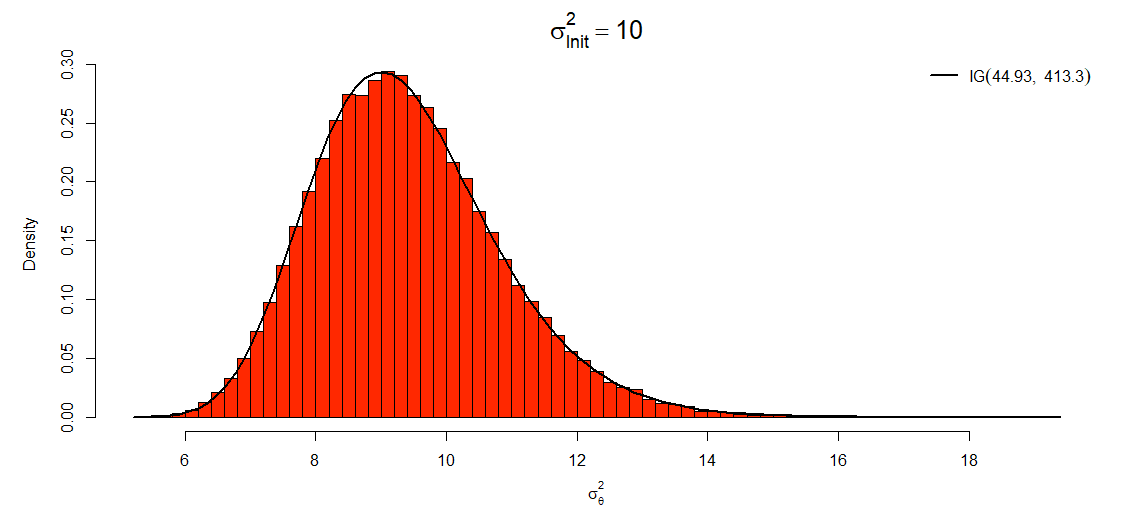}\hfill
\includegraphics[width=0.49\textwidth]{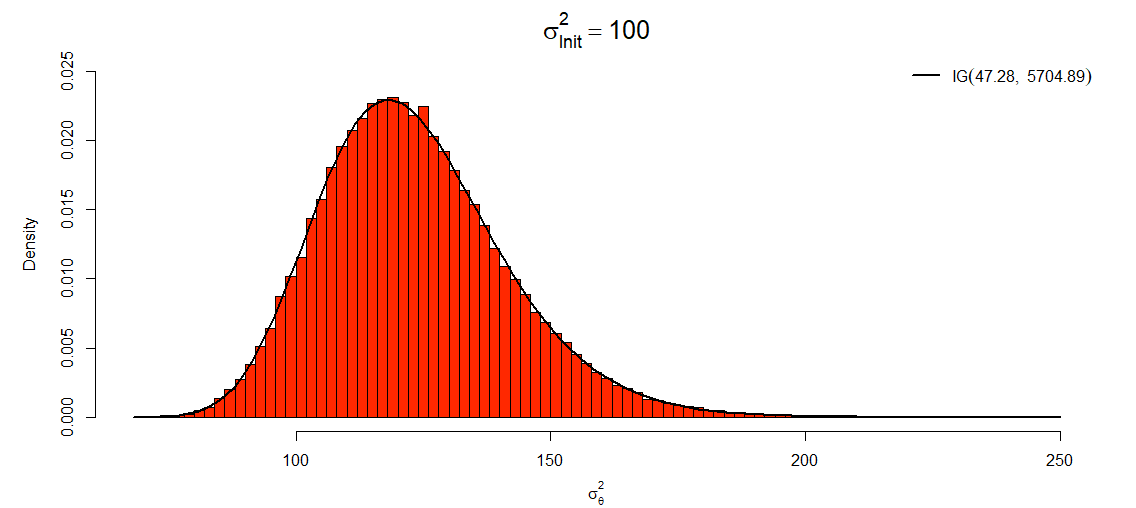}
\caption{Distribution of marginal $\sigma_\theta^2 \mid \mathcal{D} \sim \text{Inv-Gamma(Shape, Rate)}$ for varying initial variance $\sigma_{\text{Init}}^2$ (plots are on different $x$ and $y$ scales).}
\label{fig: inv-gammas prop_var xo}
\end{figure}
\fi

Now as previously argued in Section \ref{sec: 2-sample}, the motivation for adopting a hierarchical Bayesian framework, wherein a dispersion parameter $\sigma_\theta^2$ is introduced via a prior such that $\sigma_\theta^2 \propto \frac{1}{\nu}$, was to allow the training set to inform the degree of regularization. However, the preceding analysis including Section \ref{sec: drones mcmc} reveals a tension in this reasoning. Specifically, the psuedo-likelihood form, the likelihood sharpness $\beta$ and the initial dispersion $\sigma_{\text{Init}}^2$ are user-specified hyperparameters that exert a substantial influence on the marginal posterior distribution of $\sigma_\theta^2 \mid \mathcal{D}$. This dependence implies that the extent of regularization is not fully inferred by the training set, but is instead strongly shaped by likelihood and prior design choices: particularly the choices of $\beta$ and $\sigma_{\text{Init}}^2$. As such, we may question whether the hierarchical structure genuinely facilitates data-driven regularization or whether it merely reintroduces user-defined regularization through a more complex inferential route. From this perspective, the use of a Bayesian hierarchical model for the sole purpose of inferring $\sigma_\theta^2$ from the training set may appear unnecessary, especially when the same effect could be achieved by explicitly fixing $\sigma_\theta^2$ (and hence regularization strength $\nu$) to a chosen value. In this light, one might argue that the two-block MCMC procedure employed here, through which the Bayesian hierarchical model is implemented, is actually just the \emph{user} inferring a specific regularization strength but with ``extra steps''.\\

The remainder of the study further examines the implications of increased likelihood sharpness. To facilitate this analysis, the first block of the two-block MCMC scheme described in Section~\ref{sec: 2-sample} is replaced with an iterative optimization procedure, and this hybrid method is compared to the original algorithm. The comparison is illustrated using three arbitrary objective functions derived from the classic S-$17$ blackjack game. The environments and the corresponding role of the neural network as a control mechanism are described in the subsequent sections.

\subsection{The Blackjack Problem I: Controlling Player Decisions} \label{sec: bj problem 1}
In this section, we broaden the range of reinforcement learning problem types considered by introducing a stochastic, partially observable environment, specifically in the form of blackjack. Blackjack represents a substantially more complex reinforcement learning environment, characterized by uncertainty, delayed rewards, and a richer decision process than the previous two reinforcement learning tasks considered. As a result, effective learning requires greater model capacity, making the role of regularization both more necessary and more informative. Blackjack is therfore included to ensure that the analysis considers a task in which higher neural network capacity is required, highlighting how two-block MCMC behaves as model complexity increases. Furthermore, in this section, a neural network is used as the control policy to determine the player’s actions at each decision point.\\

Formally, blackjack is a card game in which a player competes directly against the dealer (rather than against other players at the table). In the S-$17$ variant, the dealer must stand on all soft $17$s (a hand-value of $17$ that includes an Ace counted as $11$) and may draw additional cards only until reaching this threshold. The player’s objective is to obtain a hand-value closer to $21$ than the dealer’s without exceeding $21$, in which case the player busts and loses immediately. A natural blackjack (an initial two-card hand-value of $21$) pays $3:2$, while the dealer wins whenever achieving a higher valid hand-value than the player. \iffoo An animated illustration of blackjack is displayed in Figure \ref{fig: bj move}.

\begin{figure}[H]
    \centering
      \animategraphics[autoplay,loop,width=0.62\linewidth]{4}{bj_move/blackjack_plot_}{1}{110}
\caption{Illustration of the game of blackjack (displaying player actions determined according to Basic Strategy).}
\label{fig: bj move}
\end{figure}
\fi
\subsubsection{Encoding}
We define a hand of blackjack as complete at time $k$, where the player's cards are dealt at times $\tau_k = 1, \ldots, T_k^{(p)}$ and recorded as the vector $\mathbf{c}_{k}^{(p)} = \left[c_{k,1}^{(p)}, \ldots, c_{k,T_k^{(p)}}^{(p)} \right]' \in \{1, 2 \ldots, 10 \}^{T_k^{(p)}}$. An Ace is encoded as $1$ (not to be confused with its potential hand-value of either $1$ or $11$), face cards ($J$, $Q$, and $K$) are encoded as $10$, and all other cards retain their nominal face values. The dealer's single upcard is denoted by the scalar $c_{k}^{(d,\mathrm{up})}$. The dealer's hole card (revealed only after the player's turn ends) at time $\tau_k = T_k^{(p)}+1$ is denoted by the scalar $c_{k}^{(d,\mathrm{hole})}$, and any additional dealer cards drawn at times $\tau_k = T_k^{(p)}+2, \ldots, T_k$ are recorded as the vector $\mathbf{c}_{k}^{(d)} = \left[ c_{k,T_k^{(p)}+2}^{(d)}, \ldots, c_{k,T_k}^{(d)} \right]' \in \{1, 2 \ldots, 10 \}^{T_k - T_k^{(p)} - 1}$.\\

We treat both the player's and the dealer's turns as two ordered sequences of events, and define the card history for the $k^{th}$ blackjack hand as
$\mathcal{H}_k = \left[  \left(\mathbf{c}_{k}^{(p)}\right)',c_{k}^{(d,\mathrm{up})},c_{k}^{(d,\mathrm{hole})}, \left(\mathbf{c}_{k}^{(d)}\right)' \right]' \in \{1, 2 \ldots, 10 \}^{T_k}$ where $\mathcal{H}_k$ represents all cards eventually observed by the player in the $k^{th}$ hand. Furthermore, we define the complete card history observed by the player at the end of the 
$k^{th}$ hand (from the $1^{st}$ hand of blackjack) as $\boldsymbol{\mathcal{H}}_k = \left[ {\mathcal{H}_1}', {\mathcal{H}_2}', \ldots, {\mathcal{H}_k}'\right]'$. Additionally, we define the complete set of the dealer's cards observed in the $k^{th}$ hand as $\mathbf{c}_k^{(d,\mathrm{all})} = \left[c_{k}^{(d,\mathrm{up})},\, c_{k}^{(d,\mathrm{hole})},\, \left(\mathbf{c}_{k}^{(d)}\right)' \right]' \in \{1, 2 \ldots, 10 \}^{T_k - T_k^{(p)}}$. Furthermore, we represent the partial player hand during the $k^{\text{th}}$ blackjack hand at time $\tau_k$, where $\tau_k \leq T_k^{(p)}$, as
$
\mathbf{c}_{k, 1:\tau_k}^{(p)} = \left[ c_{k,1}^{(p)}, \ldots, c_{k, \tau_k}^{(p)} \right]' \in \{1, 2, \ldots, 10\}^{\tau_k}
$. The corresponding partial card state including the dealer's upcard is then given by
$
\boldsymbol{c}_{k, \tau_k}^{\text{partial}} = \left[ \left( \mathbf{c}_{k, 1:\tau_k}^{(p)} \right)', \, c_k^{(d,\mathrm{up})} \right]' \in \{1, 2, \ldots, 10\}^{\tau_k + 1}.
$

\subsubsection{The hand's outcome}
We define the hand-value of a player's or dealer's cards, $\lvert\mathbf{c}\rvert  = R$ as such:
\begin{align}
    h(\mathbf{c}) &= 
    \begin{cases}
        \sum_{i = 1}^Rc_i + 10, & \text{if $c_i$ = 1} \text{ and } \sum_{i = 1}^Rc_i + 10 \leq 21 , \\    
        \sum_{i = 1}^Rc_i & \text{otherwise.} \nonumber
    \end{cases}
\end{align}
which implies that if an Ace is present ($c_i = 1$) and the total hand-value does not exceed $21$ when the Ace is treated as $11$, then the Ace is valued at $11$. Otherwise, it is valued as $1$. We record the bet multiplier at the conclusion of the $k^{th}$ hand, $s_k$, as such: 
\begin{align}
    s_k&= 
    \begin{cases} 
        +2, & \text{if } h\left(\mathbf{c_k}^{(p)}\right) > h\left(\mathbf{c}_k^{(d,\mathrm{all})} \right) \text{ and } h\left(\mathbf{c_k}^{(p)}\right) \leq 21 \text{ and player chose to  double-down where $\lvert \mathbf{c_k}^{(p)} \rvert  =3$},\\
      +2, & \text{if } h\left(\mathbf{c_k}^{(p)}\right) \leq 21 \text{ and } h\left(\mathbf{c}_k^{(d,\mathrm{all})} \right) > 21 \text{ and player chose to  double-down},\\
        +1.5, & \text{if } h\left(\mathbf{c_k}^{(p)}\right) = 21 \text{ for } \lvert \mathbf{c_k}^{(p)} \rvert = 2\text{ and } h\left(\left[c_{k}^{(d,\mathrm{up})},c_{k}^{(d,\mathrm{hole})}\right]'\right) \neq 21, \\
         +1, & \text{if } h\left(\mathbf{c_k}^{(p)}\right) > h\left(\mathbf{c}_k^{(d,\mathrm{all})} \right) \text{ and } h\left(\mathbf{c_k}^{(p)}\right) \leq 21 \text{ and player did not choose to  double-down},\\
         +1, & \text{if } h\left(\mathbf{c_k}^{(p)}\right) \leq 21 \text{ and } h\left(\mathbf{c}_k^{(d,\mathrm{all})} \right) > 21 \text{ and player did not choose to  double-down},\\
        -2, & \text{if } h\left(\mathbf{c_k}^{(p)}\right) > 21 \text{ and player chose to  double-down where $\lvert\mathbf{c_k}^{(p)} \rvert =3$},\\
        -1, & \text{if } h\left(\mathbf{c_k}^{(p)}\right) > 21 \text{ and player did not choose to  double-down,} \\
        -1, & \text{if } h\left(\mathbf{c_k}^{(p)}\right) < h\left(\mathbf{c}_k^{(d,\mathrm{all})} \right) \text{ and } h\left(\mathbf{c_k}^{(d,\mathrm{all})}\right) \leq 21,\\
        -0.5, & \text{if player surrendered hand}, \\
        0, &\text{otherwise}. \nonumber
    \end{cases}
\end{align}

\subsubsection{Control: Player decision}
We control player decisions/actions at time $\tau_k = 1, \ldots, T_k^{(p)}$ for the $k^{th}$ hand of blackjack through the means of the control vector $\mathbf{ct}\left(\boldsymbol{c}_{k, \tau_k-1}^{\text{partial}}, \boldsymbol{\theta}^{\text{Decision}}\right) \in \mathcal{A}_{\tau_k} \subseteq \{\text{Stay}, \text{Hit}, \text{Split}, \text{Surrender}, \text{Double-Down}\}$  for some parameter configuration $\boldsymbol{\theta}^{\text{Decision}}\in \mathbb{R}^R$. We note that $\mathcal{A}_{\tau_k}$ denotes the subset of admissible actions available at time $\tau_k \leq T_k^{(p)}$, immediately prior to the player's decision. Splitting is allowed only if the player's first two cards are identical $c_{k, 1}^{(p)} = c_{k, 2}^{(p)}$ and can only occur at $\tau_k =2$, and surrendering and doubling down can only occur at $\tau_k =2$. Moreover, the player's decision at time $\tau_k$ is conditioned on the cards observed up to time $\tau_k - 1$. Furthermore, the player action selection is probabilistic and derived from a softmax distribution over logits. Hence, for $\ell_a$ being the logit score for any valid action $a \in \mathcal{A}_{\tau_k}$, the probability of selecting that action is: 
\begin{align}
    \sigma_L(a \mid \boldsymbol{c}_{k, \tau_k-1}^{\text{partial}},\boldsymbol{\theta}^{\text{Decision}}) = \frac{\exp(\ell_a)}{\sum_{a' \in \mathcal{A}_{\tau_t}}\exp(\ell_{a'})} \nonumber.
\end{align}
The selected action $a^*$ corresponds to the action with the highest probability, that is,
$a^*=\operatorname*{argmax}_{a\in\mathcal{A}_{\tau_k}}
\sigma_L\bigl(a \mid \boldsymbol{c}_{k,\tau_k-1}^{\text{partial}},\,
\boldsymbol{\theta}\bigr)$.
Invalid actions (i.e., $a \notin \mathcal{A}_{\tau_t}$) are assigned $\ell_a = -\infty$, ensuring a zero probability is attributed to the invalid action. Now the interface between a model and the player action is undergone through this control vector for which $\mathbf{ct}: (\boldsymbol{c}_{k, \tau_k-1}^{\text{partial}}, \boldsymbol{\theta}^{\text{Decision}}) \rightarrow \mathbf{model} \left(\boldsymbol{\Omega}\left(\boldsymbol{c}_{k, \tau_k-1}^{\text{partial}}\right),  \boldsymbol{\theta}^{\text{Decision}} \right) \xrightarrow{\sigma_L(.)} a^* \in \mathcal{A}_{\tau_t} $ where $\boldsymbol{\theta}^{\text{Decision}}$ is fixed throughout all $k = 1, 2, \ldots, K$ blackjack hands for all time $\tau_k \leq T_k^{(p)}$, and all player decisions/actions are based on this fixed parameterization $\boldsymbol{\theta}^{\text{Decision}}$. In the framework of using a neural network as our model, we define $\boldsymbol{\Omega}: \boldsymbol{c}_{k, \tau_k-1}^{\text{partial}}\rightarrow \mathbf{a}^0 \in \mathbb{R}^{d_0}$ which signifies the vector of input nodes for time $\tau_k \leq T_k^{(p)}$. Furthermore, $\boldsymbol{\theta}^\text{Decision}$ are the weights and biases of the neural network, $\mathbf{w}^{\text{Decision}} \in \mathbb{R}^R$. 

\paragraph{Feature engineering}
We construct the feature vector using three inputs: the player's hand-value at time $\tau_k - 1$, the dealer's visible upcard for the $k^{th}$ hand, and a binary indicator denoting the presence of a usable ace in the player's hand (defined as an ace valued at $11$ rather than $1$). This representation aligns with the standard structure of established blackjack strategy tables, which prescribe optimal actions based on this triplet of information. By adopting this input configuration, we leverage a format that has been extensively validated through decades of empirical and theoretical research in blackjack literature. \\

Hence, our $1^{st}$ input node is defined as
$a_{1}^0 = \frac{h\left(\mathbf{c}_{k,1:(\tau_k-1)}^{(p)}\right)}{21}$ which represents the player’s hand-value at time $\tau_t - 1$, before the player makes a decision at time $\tau_k$, normalized by 21. The $2^{nd}$ input node is given by $a_{2}^0 = \frac{h\left(c_{k}^{(d,\mathrm{up})} \right)}{10}$ which encodes the dealer’s upcard value as a fraction of 10. Our $3^{rd}$ input node
is a binary indicator: $a_{3}^0 = \mathbb{I} \left(1 \in \mathbf{c}_{k,1:(\tau_k - 1)}^{(p)} \ \cap \ \left( \sum_{c \in {\mathbf{c}_{k,1:(\tau_t-1)}^{(p)}} }c + 10\right) \leq 21 \right)$ which denotes the presence of a usable ace in the player's hand up to time $\tau_k - 1$. 



\subsubsection{The arbitrary objective}
Consider an arbitrary objective where, for a given parameter configuration $\boldsymbol{\theta} = \boldsymbol{\theta}^{\text{Decision}}   \in  \mathbb{R}^{R}$ and after playing $K$ number of blackjack hands, we record the ROI, where the reward for the $k^{th}$ blackjack hand is denoted as $s_k\left(\boldsymbol{\theta} \right) \cdot bet_k$ ($bet_k = 1$ is the initial bet-size for each $k^{th}$ blackjack hand for simplicity, and scales accordingly for split and double-down actions). Hence:
\begin{align}
    \operatorname*{argmax}_{\boldsymbol{\theta}} \text{Obj}(\boldsymbol{\theta}) & = \operatorname*{argmax}_{\boldsymbol{\theta}} \frac{\sum_{k = 1}^K \left(s_k\left(\boldsymbol{\theta} \right) \cdot bet_k\right)} {\sum_{k = 1}^K 
 bet_k} \nonumber \\
 & = \operatorname*{argmax}_{\boldsymbol{\theta}}  \text{ROI}\left(\boldsymbol{\theta} \right). \nonumber
\end{align}
By including L2 regularization, our L2 penalized objective becomes:
\begin{align}
     \operatorname*{argmax}_{\boldsymbol{\theta}} \left(\text{ROI}\left(\boldsymbol{\theta} \right)  -  \nu \lVert\boldsymbol{\theta} \rVert^2 \right) . \nonumber
\end{align} 

\subsubsection{Results: Effects of regularization}
To train the blackjack agent, we simulate a sequence of $K = 1000$ hands of blackjack, termed a ``night''. The training set is fully determined by an initial seed $\omega_{0}^{\text{Train}}$ that governs the shuffle of the $D_0$-deck shoe at the start of the training night. Formally, we define a sequence of the $D_0$ deck shoe as $\mathcal{S}(\omega)$: indicating a specific permutation of the $D_0$-deck shoe, where $\omega$ is a random seed. The initial training shoe is then $\mathcal{S}(\omega_0^{\text{Train}})$. Our shoe utilizes a reshuffle threshold (deck penetration) of $50\%$, hence the shoe is reshuffled whenever the number of remaining cards falls below $\frac{1}{2} 52\cdot D_0$. Each reshuffle occurs by advancing the random seed deterministically using a known rule (for example, $\omega_{i+1} = f(\omega_i)$, for some deterministic function $f$), hence ensuring reproducibility across runs. Thus, while each reshuffle produces a distinct card ordering, the sequence of reshuffles is deterministic and reproducible given the initial seed $\omega_0$.\\

Now, we define our test set as $10, 000$ nights where each night is initialized with a distinct shoe shuffle $\mathcal{S}(\omega_{0, n}^{\text{Test}})$ for night $n$ where $\omega_{0, n}^{\text{Test}} \neq \omega_0^{\text{Train}}$ for all $n = \{1, \ldots, 10000\}$.  The blackjack rules still enforce a deck penetration of $50 \%$, with reshuffling triggered accordingly, hence we define ${\omega_i^{\text{Train}}}$ and ${\omega_{i, n}^{\text{Test}}}$ to be the seed values used for reshuffling in the training night and test nights respectively. Hence to ensure no shoe permutations in the test nights overlap with or is derived from those in training night, we ensure ${\omega_i^{\text{Train}}} \neq {\omega_{i, n}^{\text{Test}}}$ implying $\mathcal{S}(\omega_i^{\text{Train}}) \neq \mathcal{S}(\omega_{i, n}^{\text{Test}}) \ \forall i, n$, ensuring disjoint training and test environments. \\

Furthermore, to evaluate the impact of the regularization strength $\nu$, we apply the estimator $\boldsymbol{\hat{\theta}}^{\text{GA}}_\nu$ to the test set and assess both in-sample and out-of-sample hit-rates, as reported in Table \ref{table: blackjack reg table}, across various values of $\nu$. Additionally, the mean $\mu_{ROI \%}$ and standard deviation $\sigma_{ROI \%}$ of ROI distributions (normally distributed as illustrated in Figure \ref{fig: hists action}) are reported for the $10,000$ out-of-sample nights. The corresponding player decision tables are presented in Figure \ref{table: player decision tables}, from which we observe that the regularization strengths yielding the highest hit-rates, on the test set, tend to produce decision policies that closely resemble the S17 strategy, or more broadly, strategies anchored around a hand-value threshold of $17$. Nevertheless, none of the tested regularization strengths result in a hit-rate that surpasses those achieved by the standard S$17$, H$17$, or Basic Strategy policies reported in Table \ref{table: policy hit-rates}. Furthermore, no consistent or interpretable pattern emerges in the player decision tables as the regularization strength $\nu$ is varied; the resulting policies appear to change irregularly across different values of $\nu$. Notably, Table \ref{table: blackjack reg table} reveals a positive association between higher hit rates and improved $\mu_{ROI \%}$ (despite all policies exhibiting negative $\mu_{ROI \%}$). A notably striking observation is that the standard deviations of the ROI distributions, $\sigma_{{ROI}\%}$, for all solutions in Table \ref{table: blackjack reg table} are approximately the same, around $3.1\%$. This being the case except for $\boldsymbol{\hat{\theta}}^{\text{GA}}_{\nu = 0.0009}$, which exhibits a slightly higher $\sigma_{{ROI}\%}$ and, coincidentally, the lowest $\mu_{{ROI}\%}$. The decision tables suggests this may be due to it being the only solution that recommended doubling-down on certain hands.

\iffoo
\begin{figure}[H]
    \centering
\animategraphics[controls,autoplay,loop,width=0.8\textwidth]{1}{decision_tables/nu=0.000}{0}{9}
\caption{Player decision tables for varying $\nu$ (removal of surrender table seeing as, given the range of $\nu$, player never opts to surrender).}
\label{table: player decision tables}
\end{figure}
\else

\begin{figure}[H]
    \centering

\includegraphics[width=0.33\textwidth]{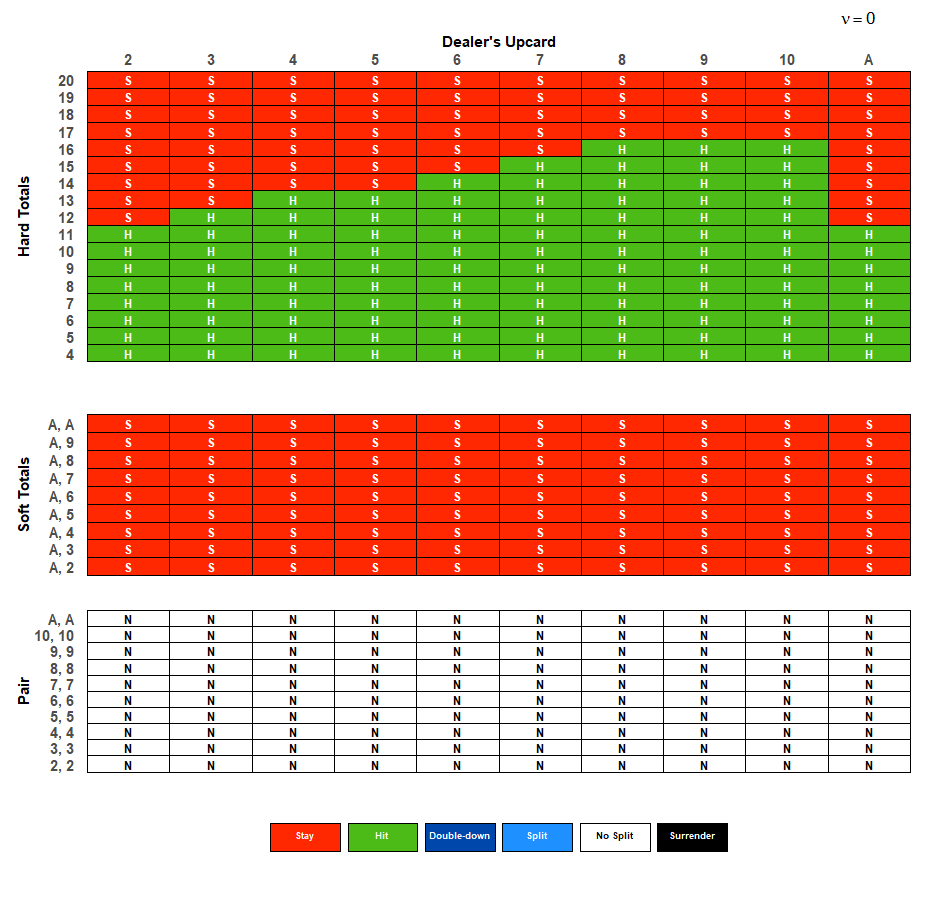}\hfill
\includegraphics[width=0.33\textwidth]{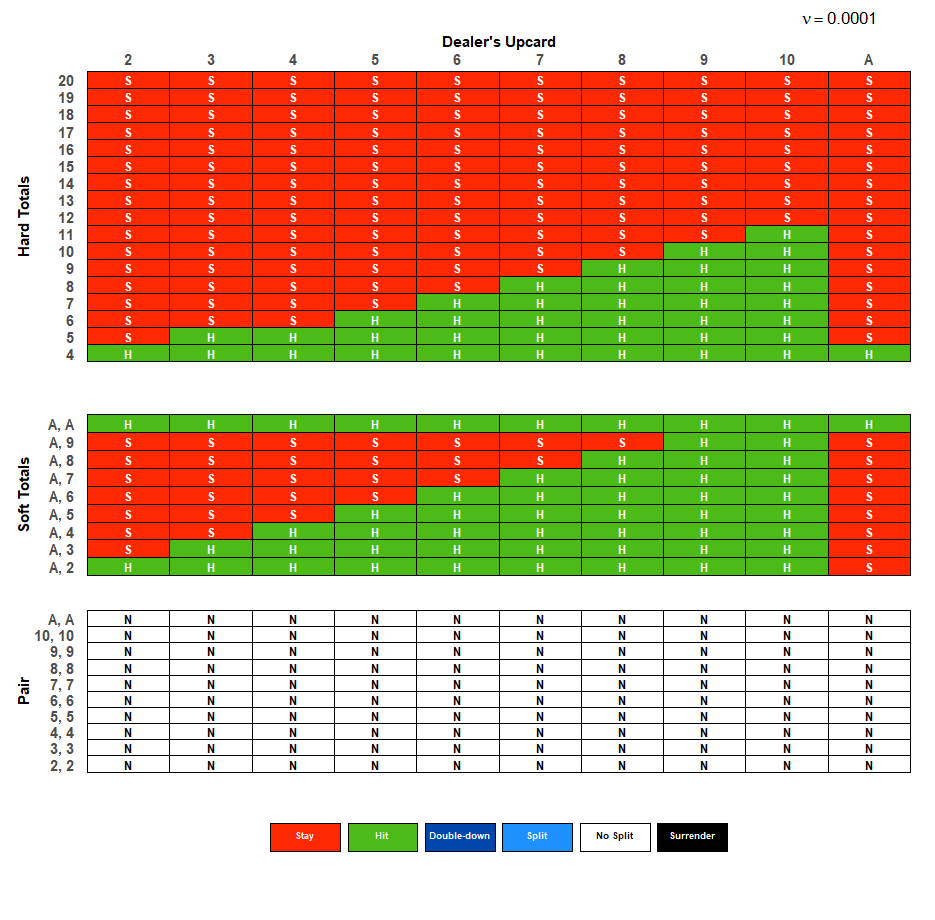}\hfill
\includegraphics[width=0.33\textwidth]{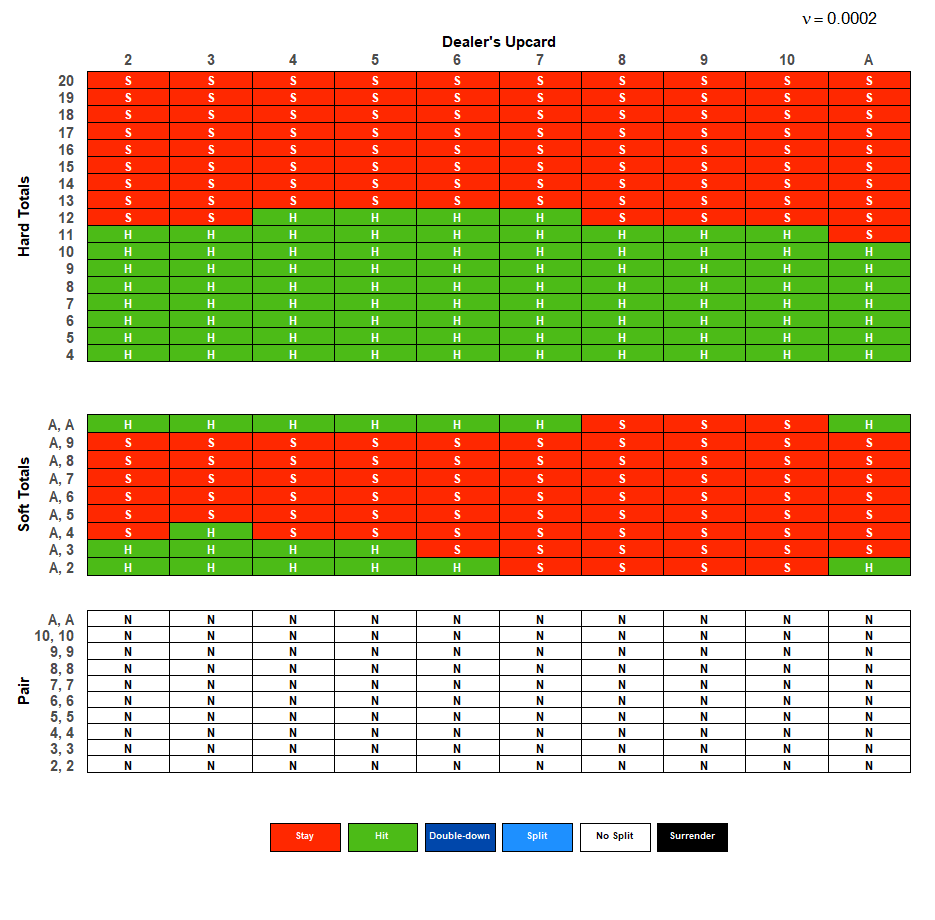}

\vspace{0.35cm}
\includegraphics[width=0.33\textwidth]{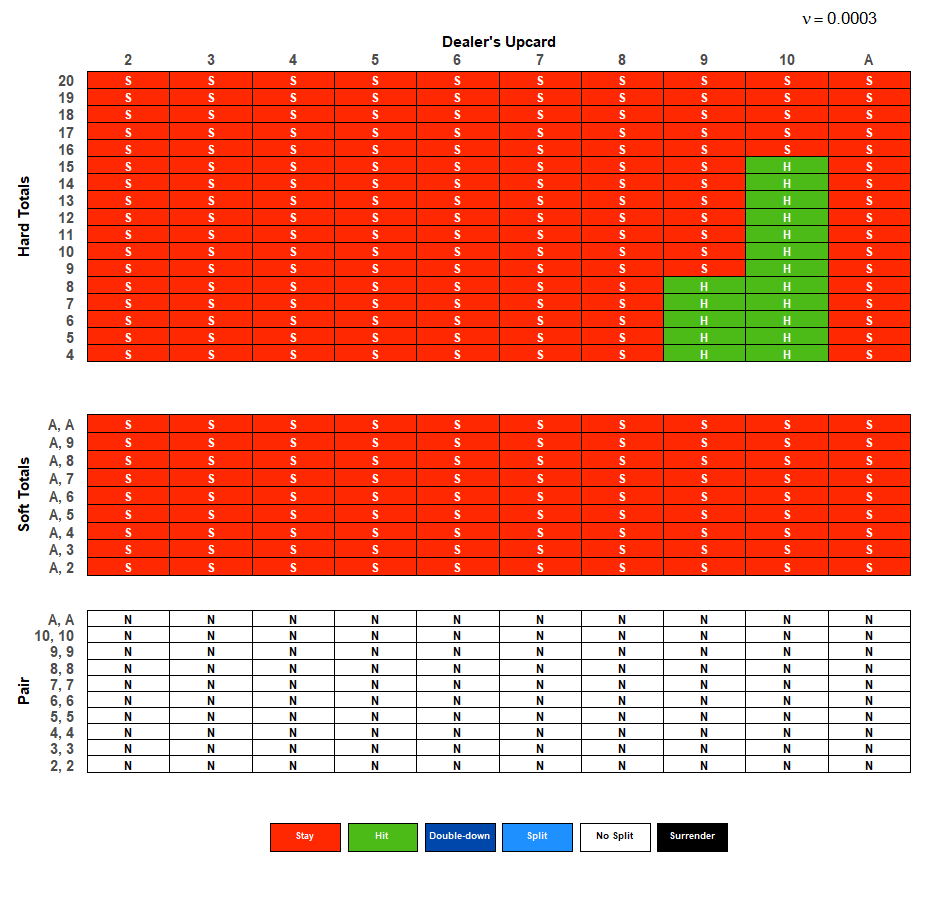}
\includegraphics[width=0.33\textwidth]{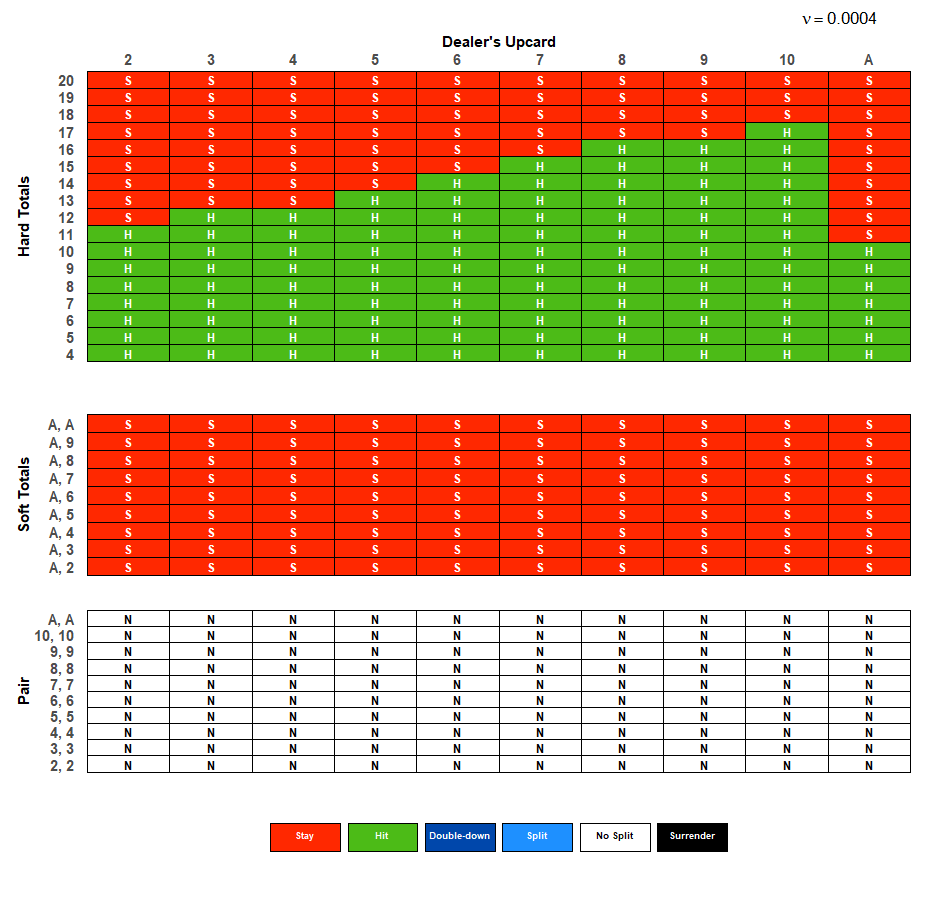}\hfill
\includegraphics[width=0.33\textwidth]{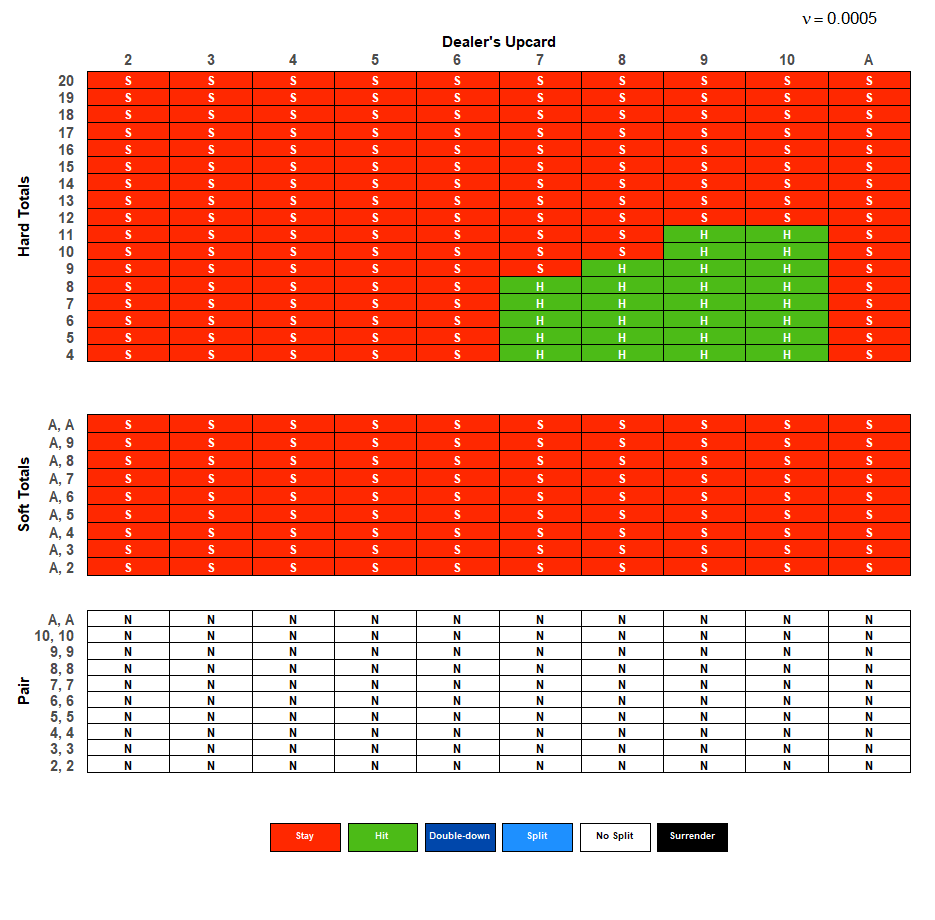}

\vspace{0.35cm}
\includegraphics[width=0.33\textwidth]{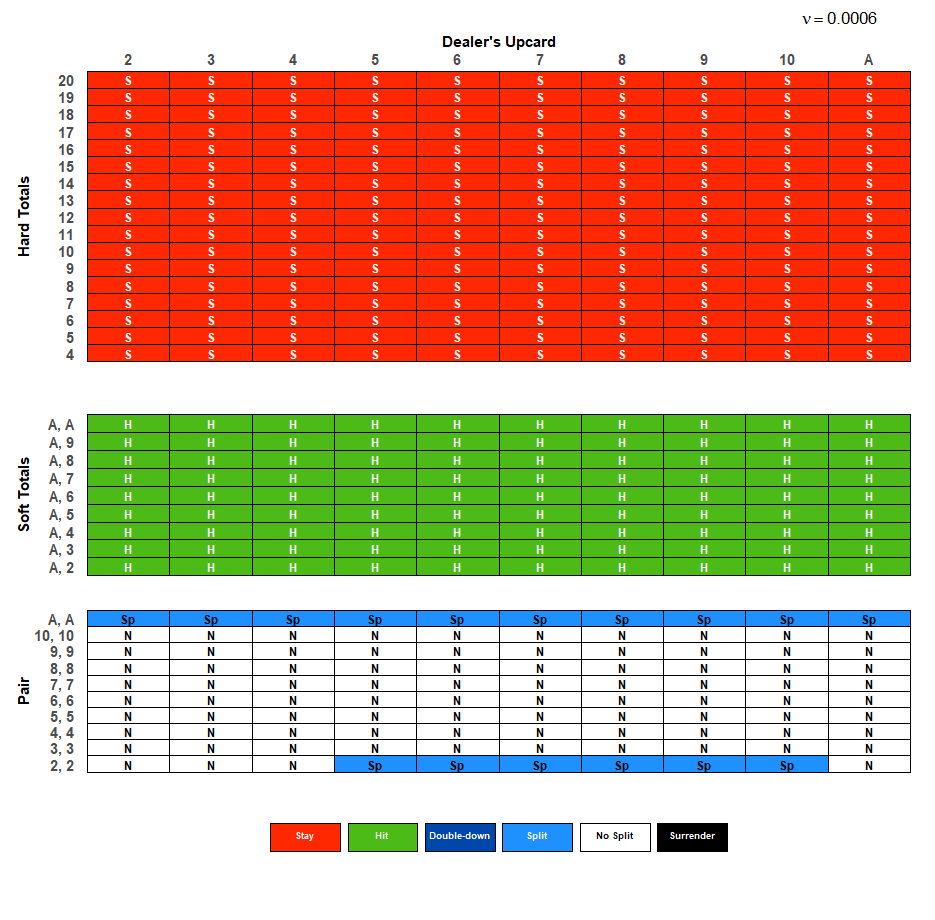}\hfill
\includegraphics[width=0.33\textwidth]{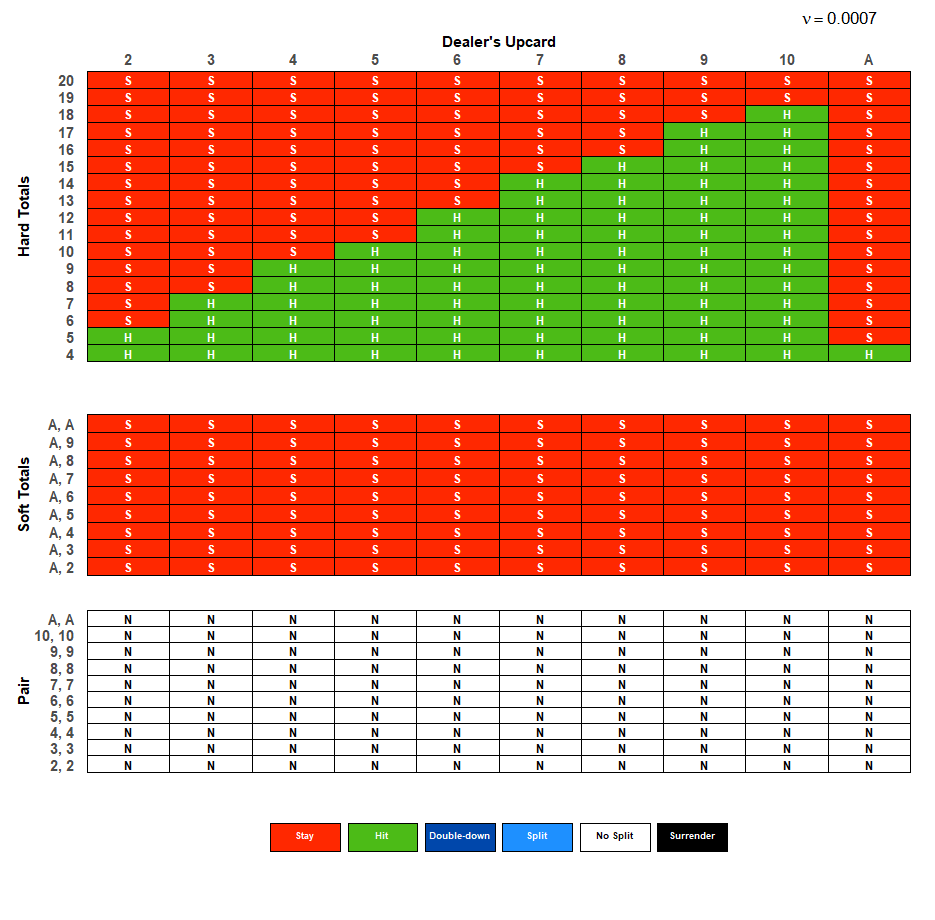}
\includegraphics[width=0.33\textwidth]{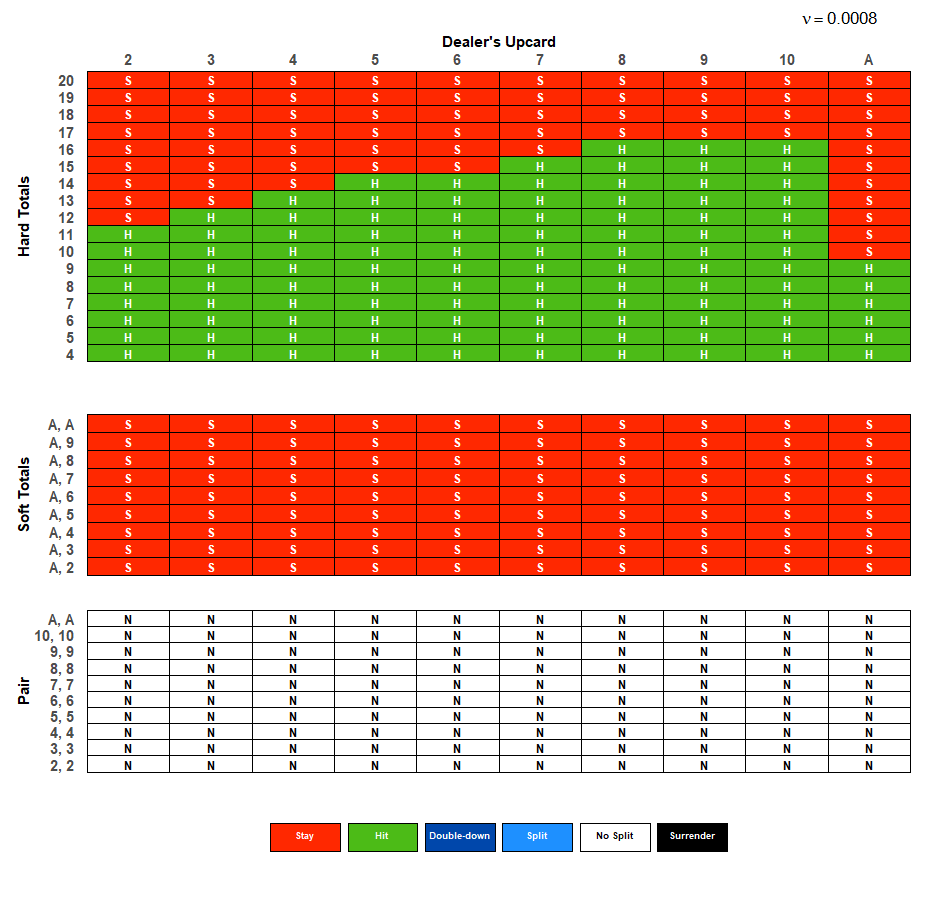}

\vspace{0.35cm}
\includegraphics[width=0.33\textwidth]{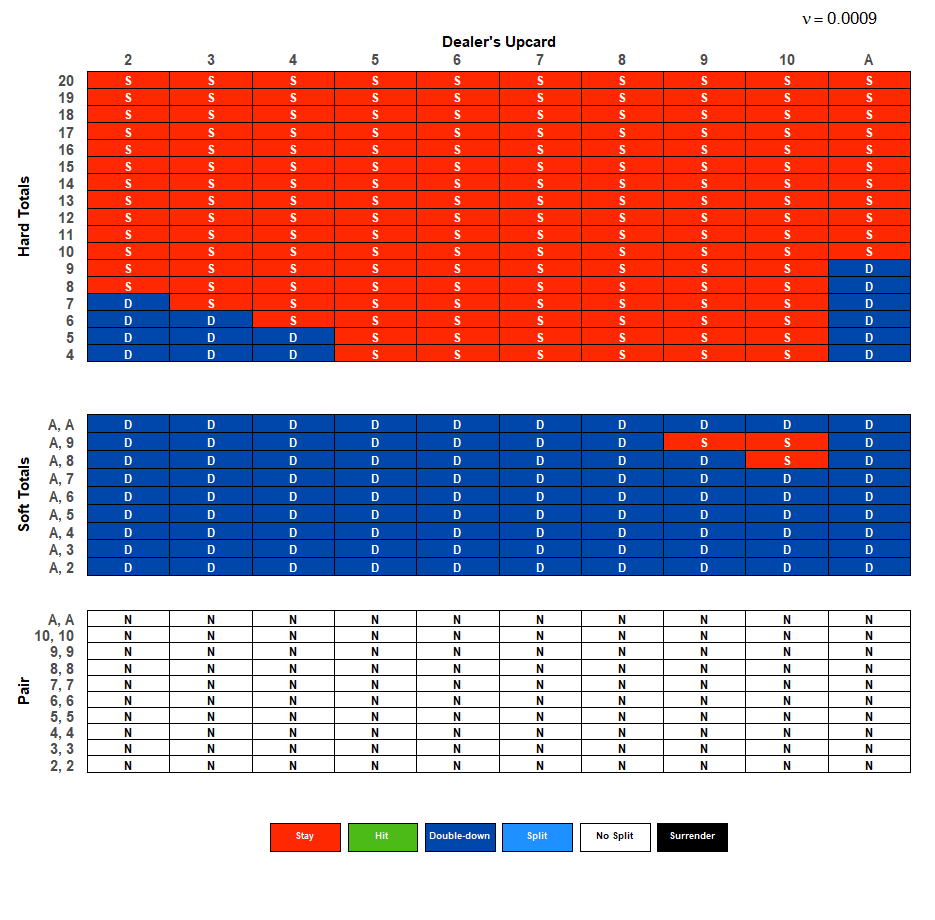}

\caption{Player decision tables for varying $\nu$ (removal of surrender table seeing as, given the range of $\nu$, player never opts to surrender).}
\label{table: player decision tables}
\end{figure}
\fi

\begin{table}[H]
    \centering
    \begin{tabular}{cccccc}
        & \multicolumn{2}{c}{In-Sample} & \multicolumn{3}{c}{Out-of-Sample} \\ \cmidrule(lr){2-3} \cmidrule(lr){4-6} 
     Regularization Strength $\nu$ & Hit-Rate & ROI $\%$  & Hit-Rate &  $\mu_{ROI \%}$ & $\sigma_{ROI \%}$   \\
     \hline
        $\textbf{0.0000}$ & ${0.4868}$ & $-0.1500$ & $\textbf{0.4575}$ & $\textbf{-5.5064}$ & $3.1067$\\
        $0.0001$ & $0.4708$ & $-2.8000$ & $ 0.4245$ & $-11.9999$ & $3.1051$\\
        $0.0002$ & $0.4840$ & $-0.6500$& $0.4464$ & $-7.7526$ & $3.1170$\\
        $0.0003$ & $0.4397$ & $-9.1000$ & $ 0.4222$ & $-12.3205$ & $3.0765$\\
        $\textbf{0.0004}$ & $0.4695$ & $-3.1500$ & $\textbf{0.4535}$ &$\textbf{-6.2797}$ & $3.1213$ \\
        $0.0005$ &$0.4542$ & $-5.9500$ & $0.4215$& $-12.5486$ & $ 3.1114$\\
        $0.0006$ & $0.4274$ & $-11.7500$& $  0.4058$ &$-15.8433$ & $3.1615$\\
        $0.0007$ &$0.4490$ & $ -7.5500 $& $0.4342$ & $-9.9631$ & $3.1131$\\
        $\textbf{0.0008}$ & ${0.4836}$ &$-0.6500$& $\textbf{0.4543}$ &$\textbf{-6.0970}$ & $3.1112$\\
        $0.0009$ & $0.4343$ & $-8.6500$& $0.4098$ & $-16.5464$ & $3.5682$ \\
        \hline
    \end{tabular}
    \caption{Regularization strength $\nu$ vs. hit-rate (in- and out-of-sample).}
    \label{table: blackjack reg table}
\end{table}

\begin{table}[H]
    \centering
    \begin{tabular}{cccccc}
        & \multicolumn{2}{c}{In-Sample} & \multicolumn{3}{c}{Out-of-Sample} \\ 
        \cmidrule(lr){2-3} \cmidrule(lr){4-6} 
        {Decision Policy} & {Hit-Rate} & ROI $\%$ & {Hit-Rate} & $\mu_{ROI \%}$ & $\sigma_{ROI \%}$\\ \hline
        Purely Random &0.2704 & -37.85 & $ 0.2597$ & $-41.8837$ & $3.1961$ \\ 
        Random Stay/Hit &0.3442 & -27.35 & $0.3331$ & $-29.6443$ & $3.0478$ \\ 
        S$17$  &0.4266 & -11.05& $0.4578$ & $-5.3242$ & $3.0711$\\ 
        H$17$ &0.4190 &-12.35 & $0.4596$ & $-5.0095$ & $3.0877$ \\ 
        Basic Strategy & 0.4430 & -4.25 & $0.4630$ & $-0.6716$ & $3.5657$\\ 
        \hline
    \end{tabular}
    \caption{Hit-rate and ROI performance of common blackjack decision policies.}
    \label{table: policy hit-rates}
\end{table}

\iffoo
\begin{figure}[H]
    \centering
\animategraphics[controls,autoplay,loop,width=0.8\textwidth]{2}{Hists_Action/}{0}{12}
\caption{Out-of-sample ROI distributions for varying $\nu$,  corresponding to Table \ref{table: blackjack reg table} and ROI distributions of common blackjack decision policies corresponding to Table \ref{table: policy hit-rates}.}
\label{fig: hists action}
\end{figure}
\else

\begin{figure}[H]
    \centering

\includegraphics[width=0.33\textwidth]{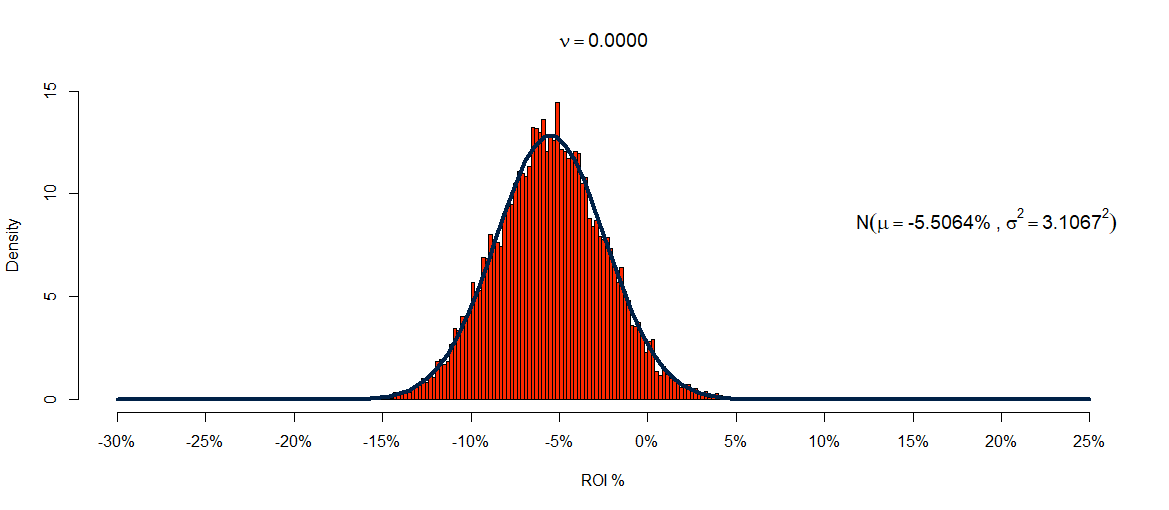}\hfill
\includegraphics[width=0.33\textwidth]{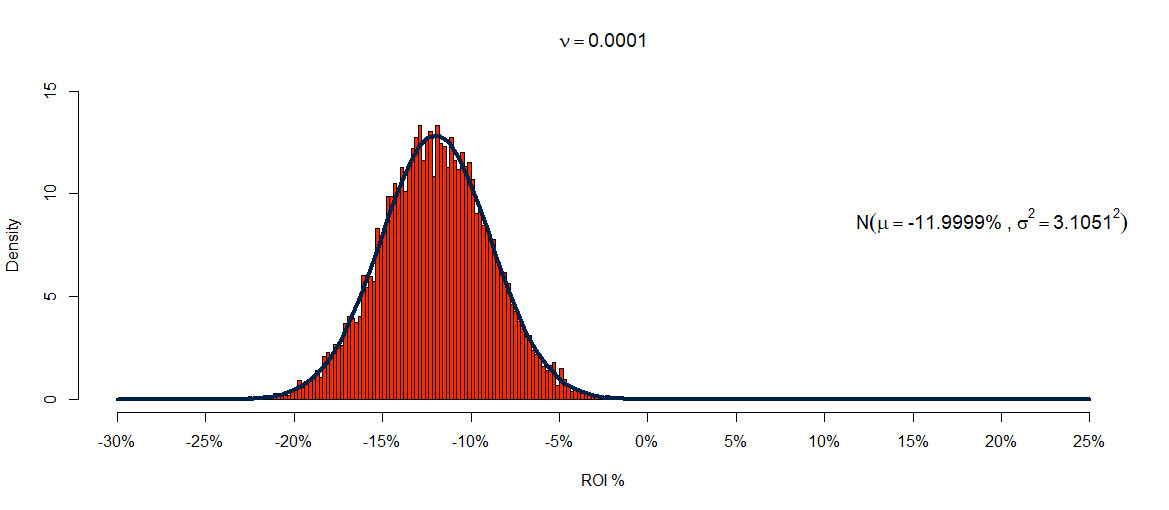}\hfill
\includegraphics[width=0.33\textwidth]{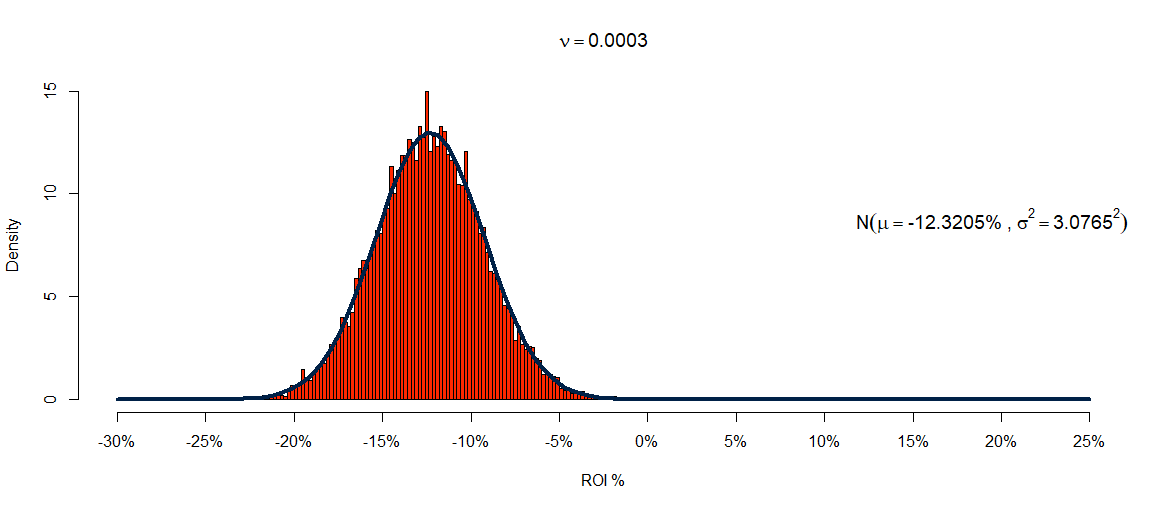}

\vspace{0.35cm}

\includegraphics[width=0.33\textwidth]{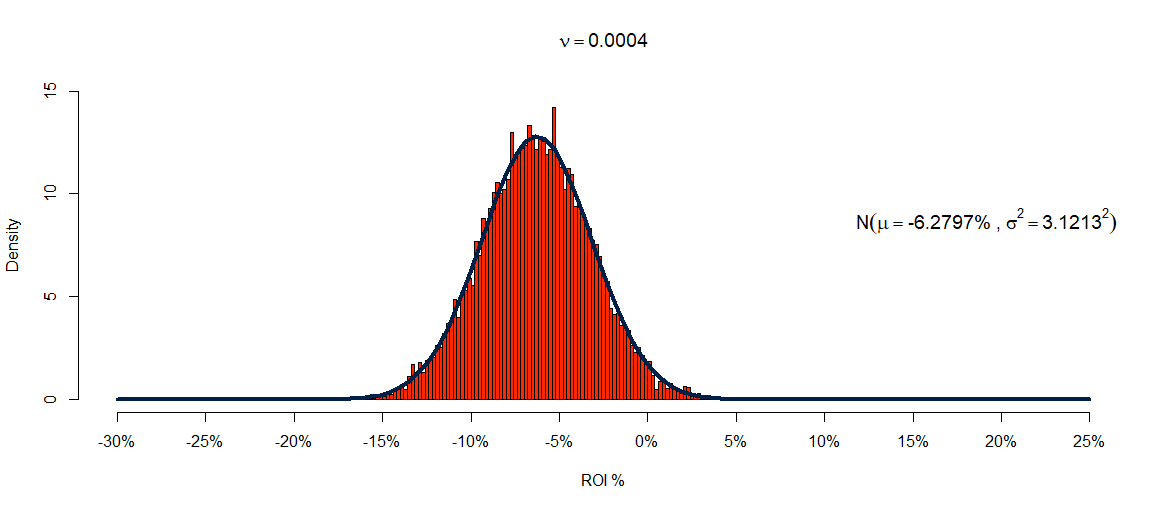}
\includegraphics[width=0.33\textwidth]{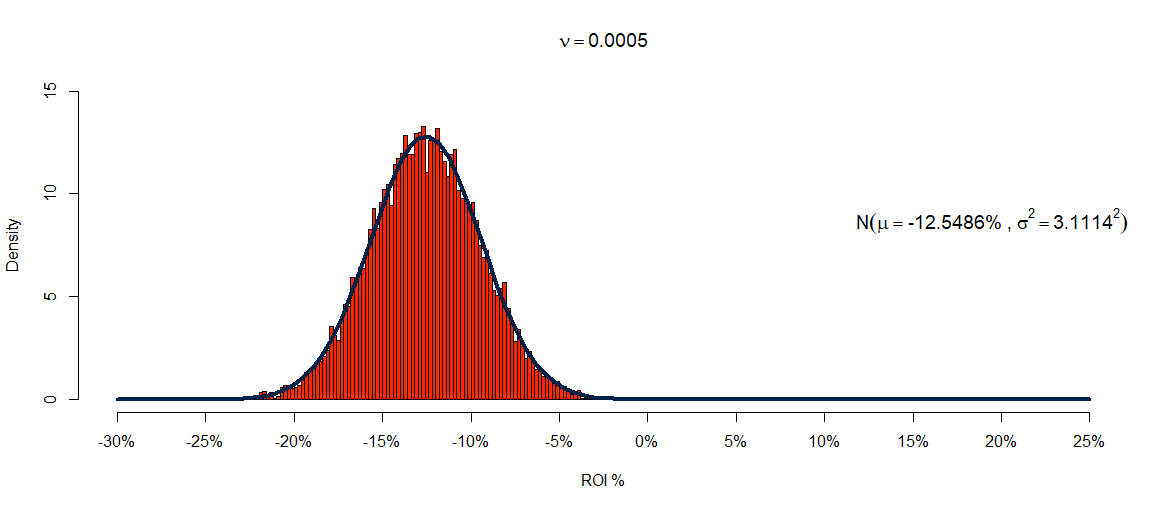}\hfill
\includegraphics[width=0.33\textwidth]{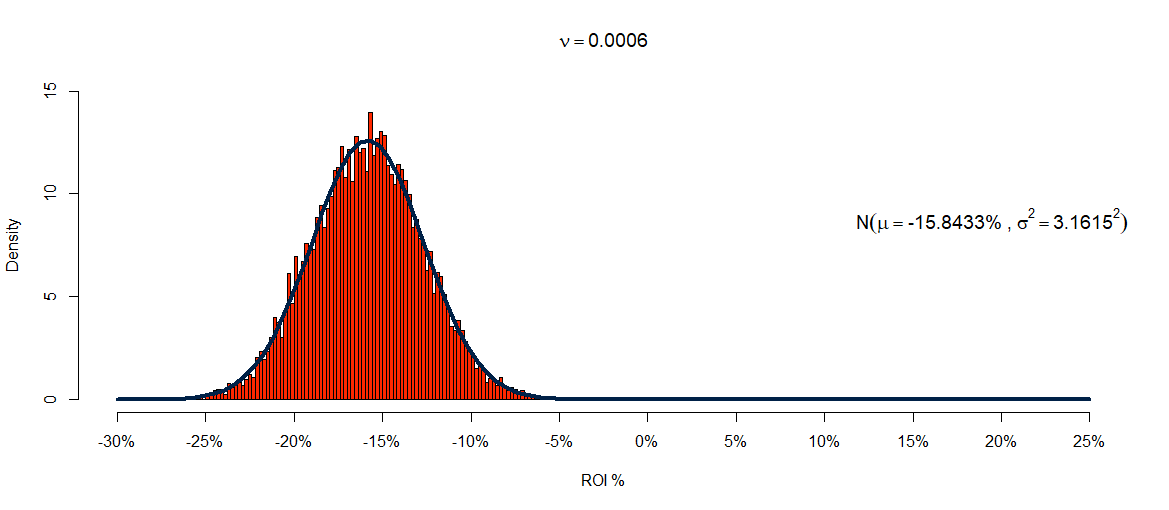}

\vspace{0.35cm}

\includegraphics[width=0.33\textwidth]{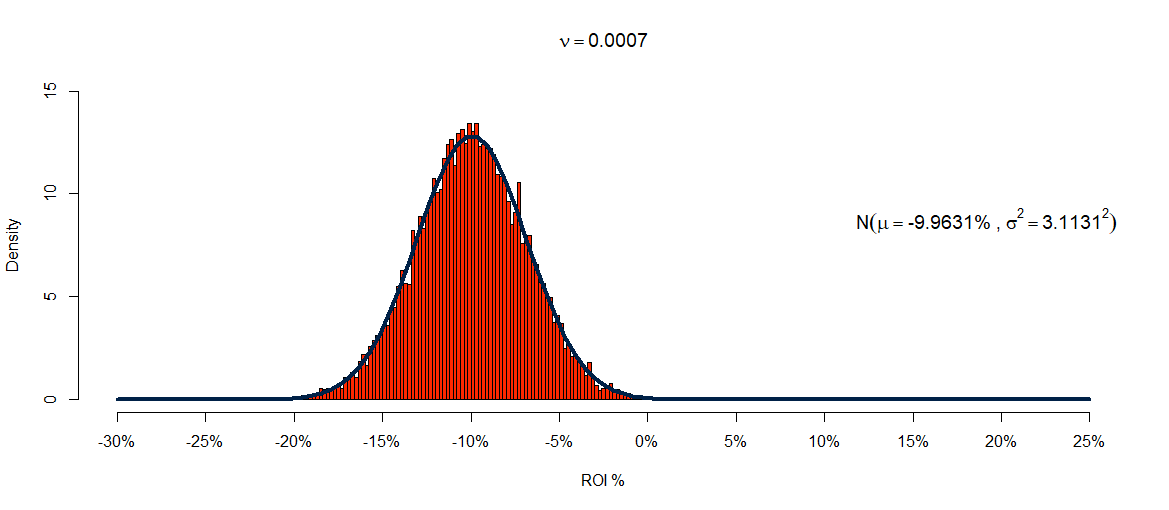}\hfill
\includegraphics[width=0.33\textwidth]{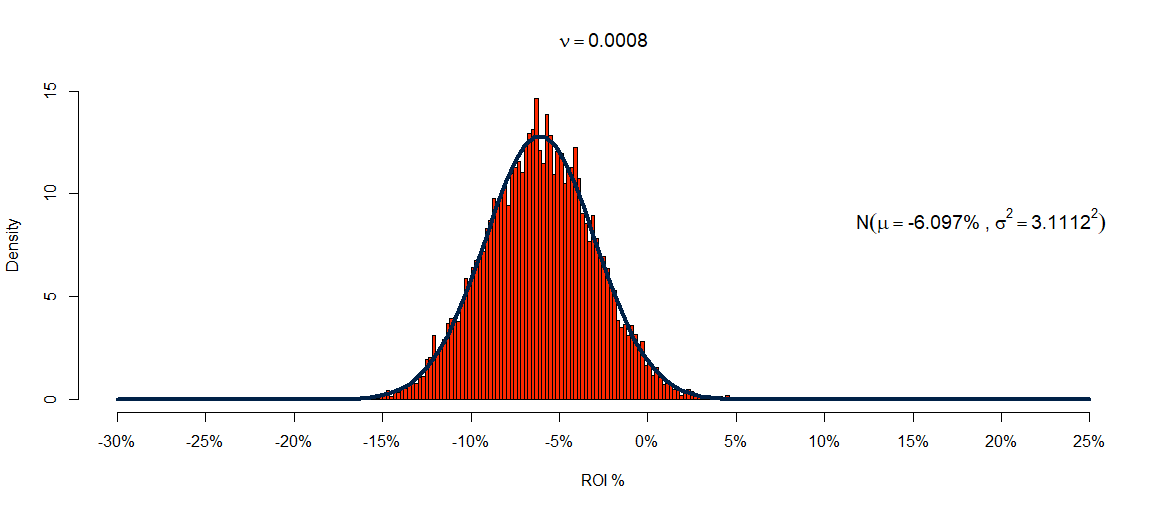}
\includegraphics[width=0.33\textwidth]{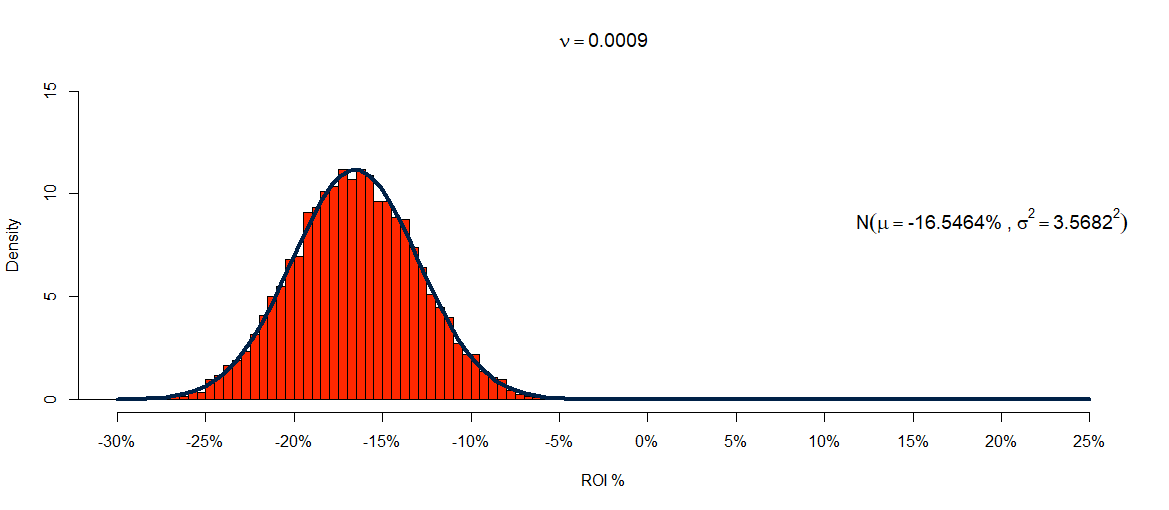}

\vspace{.35cm}
\includegraphics[width=0.33\textwidth]{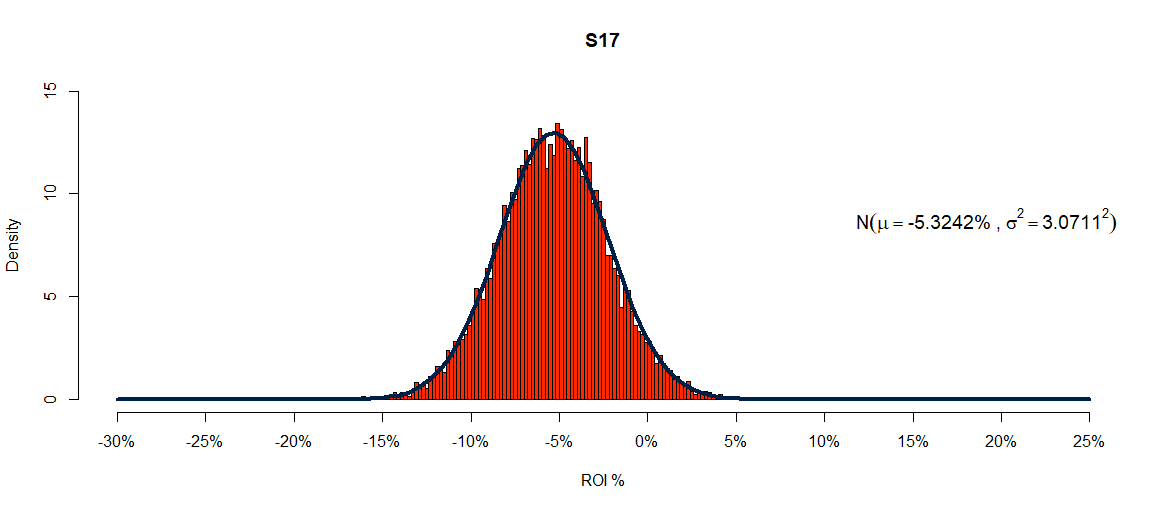}\hfill
\includegraphics[width=0.33\textwidth]{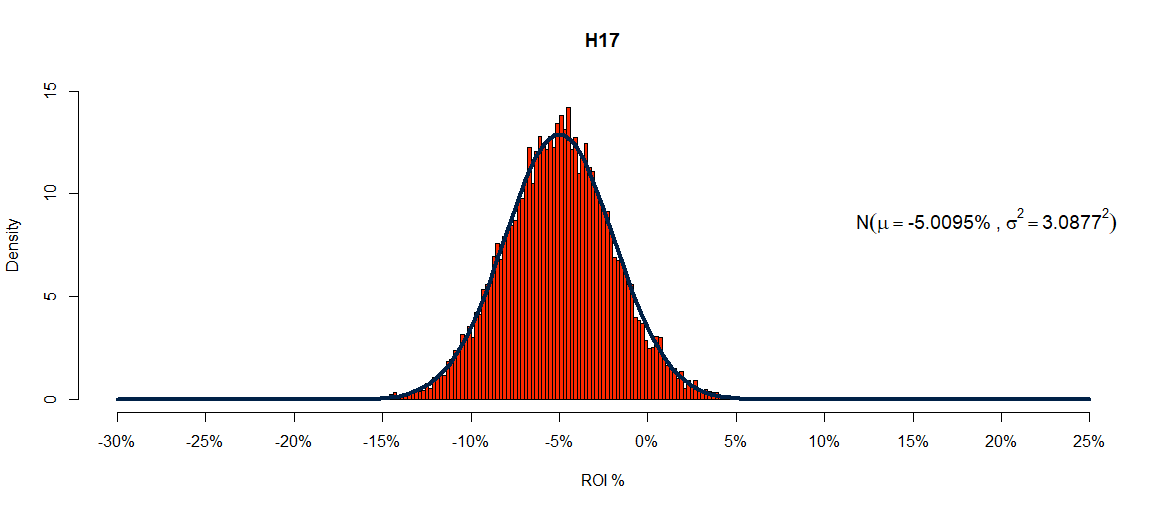}\hfill
\includegraphics[width=0.33\textwidth]{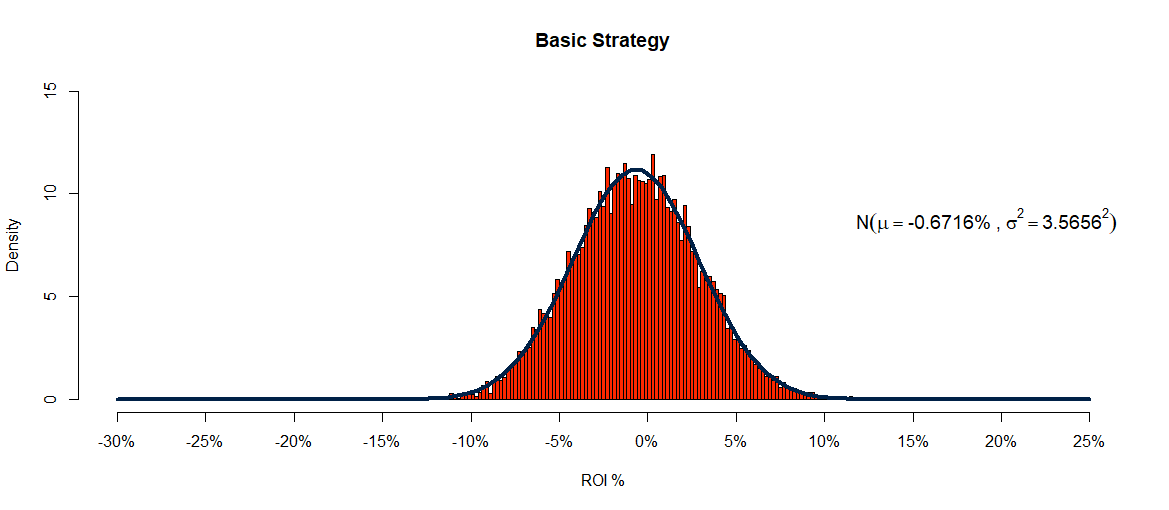}

\caption{Out-of-sample ROI distributions for varying $\nu$,  corresponding to Table \ref{table: blackjack reg table} and ROI distributions of common blackjack decision policies corresponding to Table \ref{table: policy hit-rates}.}
\label{fig: hists action}
\end{figure}
\fi

\subsection{The Blackjack Problem II: Controlling Bet Size}
In this section, we focus exclusively on controlling the bet size (placed by the player before any cards are drawn), without influencing the player's in-game decisions as in Section \ref{sec: bj problem 1}. That is to say, a neural network is used as a control mechanism to determine player bet sizing in the same stochastic, partially observable blackjack environment used previously. Accordingly, player decisions are assumed to follow Basic Strategy\footnote{See \ref{app: bj specs} for Basic Strategy rules.}, meaning the bet multiplier $s_k$ is determined solely by Basic Strategy recommendations, rather than being governed by $\boldsymbol{\theta}^{\text{Decision}}$ as in Section \ref{sec: bj problem 1}.

\subsubsection{Control: Bet size}
We control the amount to bet for the $k^{th}$ blackjack hand (which occurs before any cards are dealt for the $k^{th}$ hand), through the means of the control vector $\mathbf{ct}\left(\boldsymbol{\mathcal{H}}_{k-1}, \boldsymbol{\theta}^{\text{Bet}}\right) \in [0, 1]$ for some parameter configuration $\boldsymbol{\theta}^{\text{Bet}}\in \mathbb{R}^S$ (note the index for the card history vector $\boldsymbol{\mathcal{H}}$ is $k-1$ to indicate that betting occurs before any cards are observed for the $k^{th}$ blackjack hand). Now the interface between a model and the bet sizing is undergone through this control vector for which $\mathbf{ct}: (\boldsymbol{\mathcal{H}}_{k-1}, \boldsymbol{\theta}^{\text{Bet}}) \rightarrow \mathbf{model} \left(\boldsymbol{\Omega}\left(\boldsymbol{\mathcal{H}}_{k-1}\right), \boldsymbol{\theta}^{\text{Bet}} \right) \xrightarrow[]{\sigma_L(.)} [0, 1]$ where $\boldsymbol{\theta}^{\text{Bet}}$ is fixed throughout all $k = 1, 2, \ldots, K$ blackjack hands, and betting decisions at each $k^{th}$ hand are based on this fixed parameterization $\boldsymbol{\theta}^{\text{Bet}}$. In the framework of using a neural network as our model, we define $\boldsymbol{\Omega}: \boldsymbol{\mathcal{H}}_{k-1} \rightarrow \mathbf{a}^0 \in \mathbb{R}^{d_0}$ which signifies the vector of input nodes for the $t^{th}$ hand. Furthermore, $\boldsymbol{\theta}^{\text{Bet}}$ are the weights and biases of the neural network, $\mathbf{w}^{\text{Bet}} \in \mathbb{R}^S$, and $\sigma_L(.)$ denotes a sigmoid activation function. 

\paragraph{Feature engineering}
To encode the current state of the game in a manner conducive to learning, we propose a feature vector that captures both (i) the true count of the remaining deck as formally defined in \ref{app: TC} and (ii) a weighted summary of the distribution of unseen cards. The true count serves as a classical card-counting statistic reflecting the favourability of the shoe, while the weighted distribution provides fine-grained information about the residual card composition.\\

Specifically, we define the first input node of the neural network as the scaled true count $a_1^0 = \frac{TC_{k-1}}{3}$, where $TC_{k-1}$ is the true count at time $k-1$ normalized by a constant (here, $3$) to map the feature into a compact numerical range. To complement this, we construct additional features that reflect the remaining proportion of each card value in the shoe, weighted by their nominal face value. For cards $i = 2, \ldots 9$ and $i = 11$ we define $a_i^0 = \frac{4iD_0 - i\sum_{c \in \boldsymbol{\mathcal{H}}_{k-1}} \mathbb{I}(c = i)} {4iD_0 }$. This expression measures the remaining total ``value-weighted mass'' of card $i$, relative to its initial value-weighted mass across the full shoe. For ten-valued cards ($10$, J, Q, K), which occur with higher frequency ($16$ per deck), we define $a_{10}^0 = \frac{16 i D_0 - 10\sum_{c \in \boldsymbol{\mathcal{H}}_{k-1}} \mathbb{I}(c = 10)} {16 i  D_0 }$. Furthermore, we define two distinct models that differ in the number of input nodes utilized. Now $\overset{\text{(I)}}{\mathbf{model}} \left(a_1^0, {\boldsymbol{\theta}^{(\text{I})}}\right)$ relies solely on the true count to create the sole input node $a_1^0$. In contrast, $\overset{\text{(II)}}{\mathbf{model}} \left(\mathbf{a}^0, {\boldsymbol{\theta}^{(\text{II})}}\right) $ incorporates all $11$ previously specified input nodes, which offers a richer characterization of card composition than the true count alone.
\\

Hence, in the proposed neural network architecture, the input layer comprises of $d_0 = 11$ nodes, which collectively processes $\boldsymbol{\mathcal{H}}_{k-1}$ to produce a singular output, denoted as the betting propensity, $bet_k \in [0, 1]$. This represents the relative confidence in wagering, where the actual bet size is $\tilde{bet}_k  = 1+ 9\times bet_k \in [1, 10]$, ensuring that a positive, nonzero amount is wagered on each hand (where $1$ and $10$ may be viewed as the table minimum and maximum bets, respectively). Additionally, in scenarios where the true count satisfies $TC_{k-1} = 0$ and the card history is empty, $\lvert \boldsymbol{\mathcal{H}}_{k-1} \rvert =0$ (indicating a reshuffled shoe), the betting strategy defaults to $bet_k = 0$, hence $\tilde{bet}_k = 1$. In such cases, the historical card information $\boldsymbol{\mathcal{H}}_{k-1}$ is excluded from the control vector $\mathbf{ct}$, thereby bypassing its integration into the bet-size decision process.

\subsubsection{The arbitrary objective}
Consider an arbitrary objective where, for a given parameter configuration $\boldsymbol{\theta} = \boldsymbol{\theta}^{\text{Bet}} \in  \mathbb{R}^{S}$ and after playing $K$ number of blackjack hands, we record the ROI, where the reward for the $k^{th}$ blackjack hand is denoted as $s_k \cdot \tilde{bet}_k (\boldsymbol{\theta})$ (where $\tilde{bet}_k (\boldsymbol{\theta})$ scales accordingly to split and double-down actions). Hence:
\begin{align}
    \operatorname*{argmax}_{\boldsymbol{\theta}} \text{Obj}(\boldsymbol{\theta})  & = \operatorname*{argmax}_{\boldsymbol{\theta}} \frac{\sum_{k = 1}^K\left(s_k \cdot \tilde{bet_k} (\boldsymbol{\theta}) \right)} {\sum_{k = 1}^K 
 \tilde{bet_t} (\boldsymbol{\theta})} \nonumber \\
 & = \operatorname*{argmax}_{\boldsymbol{\theta}}  \text{ROI}\left(\boldsymbol{\theta} \right). \nonumber
\end{align} 
By including L2 regularization, our L2 penalized objective becomes:
\begin{align}
    \operatorname*{argmax}_{\boldsymbol{\theta}} \left(\text{ROI}\left(\boldsymbol{\theta} \right)  -  \nu \lVert\boldsymbol{\theta} \rVert^2 \right). \nonumber
\end{align} 

\subsubsection{Results: Effects of regularization}
\begin{figure}[H]
  \centering
    \includegraphics[width=0.85\textwidth]{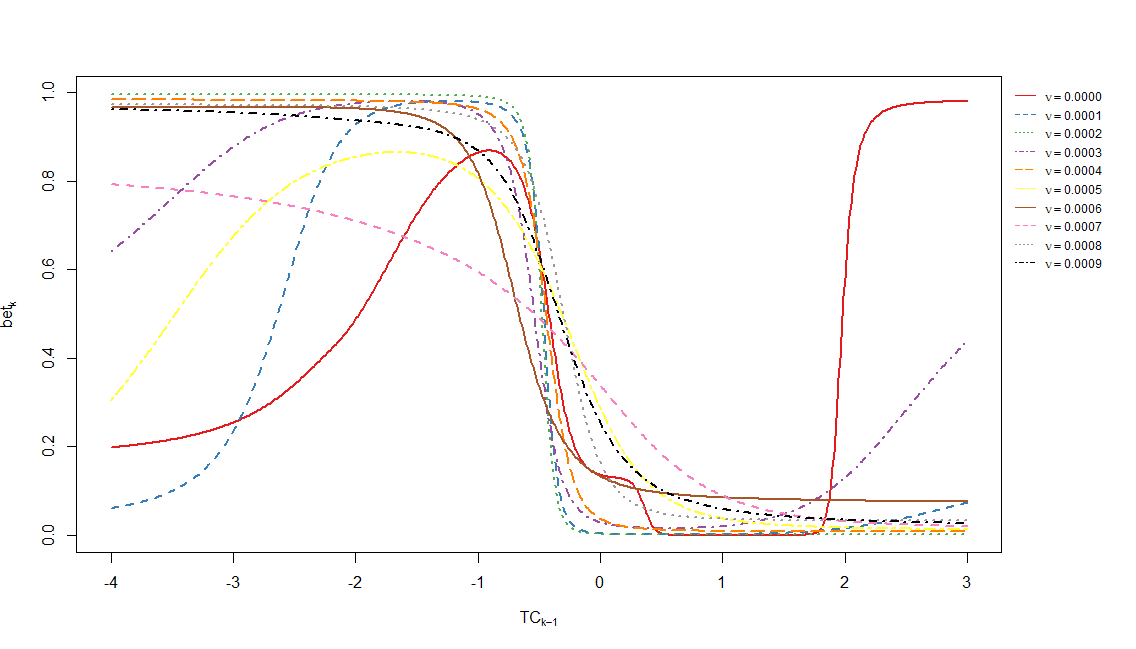}
    \caption{Betting propensity, $bet_k$, vs. the true count, $TC_{k-1}$, for various $\nu$  using $\overset{\text{(I)}}{\mathbf{model}} \left(a_1^0, {\boldsymbol{\hat\theta}^{\text{GA}, (\text{I})}_{\nu}} \right)$.}
    \label{fig: response bet}
\end{figure}

To evaluate the effect of varying regularization strengths $\nu$ on the learned betting behavior $bet_k$, we conduct a response curve analysis. For visualization purposes, we employ $\overset{\text{(I)}}{\mathbf{model}}$ that uses a single input node $a_1^0$, representing the scaled true count of the shoe denoted in the previous section. The resulting response function is given by $\overset{\text{(I)}}{\mathbf{model}} \left(a_1^0, {\boldsymbol{\hat\theta}^{\text{GA}, (\text{I})}_{\nu}} \right) \xrightarrow[]{} bet_k \in [0,1]$ and is depicted in Figure \ref{fig: response bet}. \\

As established in the blackjack literature (see also \ref{app: TC}), exploiting the game advantageously requires betting in proportion to the true count. Figure \ref{fig: response bet} reveals that the solutions $\boldsymbol{\hat\theta}^{\text{GA}, (\text{I})}_{\nu}$, across most $\nu$ values considered, tend to produce elevated betting propensities $bet_k$ predominantly at highly negative true counts. While a few values of $\nu$ do give rise to solutions which elicit high betting behavior at large positive true counts, as theoretically desirable, this behavior is not consistently observed for all solutions.\\

Now this behaviour is further demonstrated to be undesirable when examining Table \ref{table: bet policy roi}, where the notion of betting proportional to the true count of the shoe is illustrated. Notably, an inconsistency arises: in-sample ROI performance of these theoretical-based strategies is surprisingly poor (particularly when compared to the performance of the random betting policy). This observation suggests that the training set may not be conducive to the effective training of $\boldsymbol{\hat\theta}^{\text{GA}, (\text{I})}_{\nu}$. That is, this training set may not provide a sufficiently informative environment such that $\boldsymbol{\hat\theta}^{\text{GA}, (\text{I})}_{\nu}$ learns to bet in accordance to the true count. The theoretical betting strategy's superior performance is substantiated, however, when evaluating out-of-sample data, as presented in Table \ref{table: bet policy roi}. Specifically, the mean ROI ($\mu_{{ROI} \%}$) values for the theory-driven strategies demonstrate a marked improvement, approaching break-even levels with greater consistency compared to learnt betting in Table \ref{table: roi for a_1^0, a^0}.
\\

Table \ref{table: roi for a_1^0, a^0} reveals that both $\overset{\text{(I)}}{\mathbf{model}}$ and $\overset{\text{(II)}}{\mathbf{model}}$ exhibit substantial underperformance on the out-of-sample set. On a few occasions, the solutions $\boldsymbol{\hat\theta}^{\text{GA}}_{\nu}$ yield mean ROI values ($\mu_{{ROI}\%}$) that are only marginally higher than $\mu_{{ROI}\%}$ obtained from the purely random betting policy presented in Table~\ref{table: bet policy roi}. Unsurprisingly, none of the betting configurations derived from $\boldsymbol{\hat\theta}^{\text{GA}}_{\nu}$ solutions outperform any of the true count-based betting strategies on the out-of-sample set, as reported in Table \ref{table: bet policy roi}. These findings are consistent with the earlier observation that the training set may be ill-suited for facilitating effective learning based on established blackjack theory. Furthermore, we refrain from making general claims regarding comparative performance across varying values of $\nu$ between the two models, as the two models differ in complexity with respect to their number of input nodes: $\overset{(\text{I})}{\mathbf{model}}$ utilizing  ${a}_1^0$ as its feature vector and $\overset{(\text{II})}{\mathbf{model}}$ utilizing $\mathbf{a}^0$ as its feature vector.

\begin{table}[H]
    \centering
    \begin{tabular}{ccccccc}
       & \multicolumn{3}{c}{$\overset{\text{(I)}}{\mathbf{model}} \left(a_1^0, {\boldsymbol{\hat\theta}^{\text{GA}, (\text{I})}_{\nu}} \right)$} & \multicolumn{3}{c}{$\overset{\text{(II)}}{\mathbf{model}} \left(\mathbf{a}^0, {\boldsymbol{\hat\theta}^{\text{GA}, (\text{II})}_{\nu}} \right)$} \\
         \cmidrule(lr){2-4}  \cmidrule(lr){5-7} 
        & \multicolumn{1}{c}{In-Sample} & \multicolumn{2}{c}{Out-of-Sample}   & \multicolumn{1}{c}{In-Sample} & \multicolumn{2}{c}{Out-of-Sample} \\
        \hline
        Regularization Strength $\nu$ &  ROI $\%$& $\mu_{ROI \%}$ & $\sigma_{ROI \%}$ &  ROI $\%$& $\mu_{ROI \%}$ & $\sigma_{ROI \%}$  \\ 
        \hline
        $0.0000$ & $ 0.7440$ & $\textbf{-0.7650}$ &$4.4523$ & $ 14.6180$ &$-0.9887$ & $5.5638$ \\
        $0.0001$ & $0.7670$  &  $-1.1536$ & $ 5.1288$ & $11.4806$  &  $-0.9207$  &$5.2493$\\
        $0.0002$ & $-0.5578$ &  $-1.3047$  & $5.1894$ & $8.1637$ &  $-0.8404$  & $4.7393$  \\
        $0.0003$ & $-0.3340$ &  $-1.1145$  & $4.7978$ & $6.0955$ &  $-0.8418$  &$5.0195$   \\
        $0.0004$ & $-0.9778$ &  $-1.2743$ & $4.9373$ & $-2.0912$ &  $-0.8334$  & $3.9489$ \\
        $0.0005$ & $-0.9952$  &  $-1.1020$ & $4.3136$ & $-0.3641$ &  $-1.1610$  &$4.4662$ \\
        $0.0006$ & $-1.9658$  &  $-1.1819$ & $4.5306$ & $-3.9218$ &  $-0.6722$  &$3.6046$ \\
        $0.0007$ & $-2.6602$  &  $-1.0656$ & $4.0409$ & $-2.2319$ &  $\textbf{-0.4890}$  &$4.1832$ \\
        $0.0008$ & $-1.4639$  &  $-1.1926$ & $4.5497$ & $8.74736$ &  $-0.7584$  &$4.8677$ \\      
        $0.0009$ & $-1.8479$  &  $-1.1604$ & $4.3676$ & $-3.3625$ &  $-0.7375$  &$3.6303$ \\
        \hline
    \end{tabular}
    \caption{ROI in-sample and out-of-sample performance across regularization strengths $\nu$ using $\overset{\text{(I)}}{\mathbf{model}} \left(a_1^0, {\boldsymbol{\hat\theta}^{\text{GA}, (\text{I})}_{\nu}} \right)$ and $\overset{\text{(II)}}{\mathbf{model}} \left(\mathbf{a}^0, {\boldsymbol{\hat\theta}^{\text{GA}, (\text{II})}_{\nu}} \right)$.}
    \label{table: roi for a_1^0, a^0}
\end{table}

\begin{table}[H]
    \centering
    \begin{tabular}{ccccc}
        & \multicolumn{1}{c}{In-Sample} & \multicolumn{2}{c}{Out-of-Sample}   \\
        \cmidrule(lr){2-2}  \cmidrule(lr){3-4} 
        {Betting Policy}  &ROI $\%$ & $\mu_{ROI \%}$ & $\sigma_{ROI \%}$
        \\ \hline
        Purely Random & $-0.2202$ & $ -0.6909$ & $4.0185$  \\ 
        $TC_{k-1} > 0$ & $-5.8488$ & $ -0.0012$ & $5.0031$ \\ 
        $TC_{k-1} > 1$& $ -4.4912$& $ 0.0511$ & $5.5327$ \\ 
        $TC_{k-1} > 2$ & $-1.8258$ & $ -0.0233$ & $6.0224$ \\ 
        $TC_{k-1} > 3$ & $-5.9491$& $-0.2950$ & $5.7936$ \\ 
        \hline
    \end{tabular}
    \caption{ROI in-sample and out-of-sample performance of betting policies whose bet sizes are linearly proportional to $TC_{k-1}$ ($bet_k = \frac{TC_{k-1}}{3}\cdot \mathbb{I}(TC_{k-1} > x) \text{ for } x = 0, 1, 2, 3$).}
    \label{table: bet policy roi}
\end{table}

\iffoo
\begin{figure}[H]
    \centering
\animategraphics[controls,autoplay,loop,width=0.8\textwidth]{2}{Hists_Bet/}{0}{14}
\caption{Out-of-sample ROI distributions for varying $\nu$ using $\overset{\text{(II)}}{\mathbf{model}} \left(\mathbf{a}^0, {\boldsymbol{\hat\theta}^{\text{GA}, (\text{II})}_{\nu}} \right)$, corresponding to Table \ref{table: roi for a_1^0, a^0} and ROI distributions of common blackjack betting policies, corresponding to Table \ref{table: bet policy roi}.}
\label{table: hists bet}
\end{figure}
\else
\begin{figure}[H]
    \centering

\includegraphics[width=0.33\textwidth]{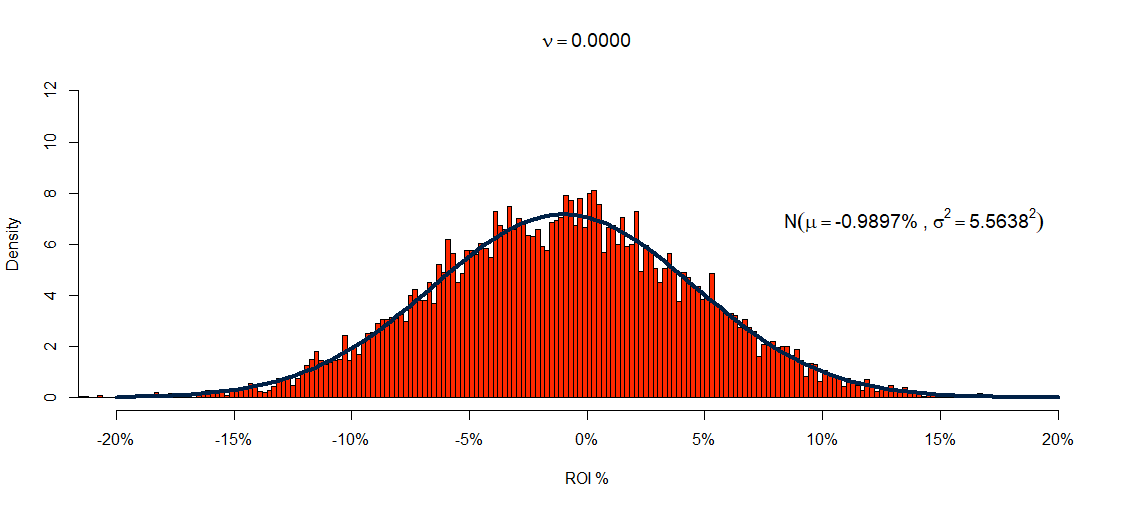}\hfill
\includegraphics[width=0.33\textwidth]{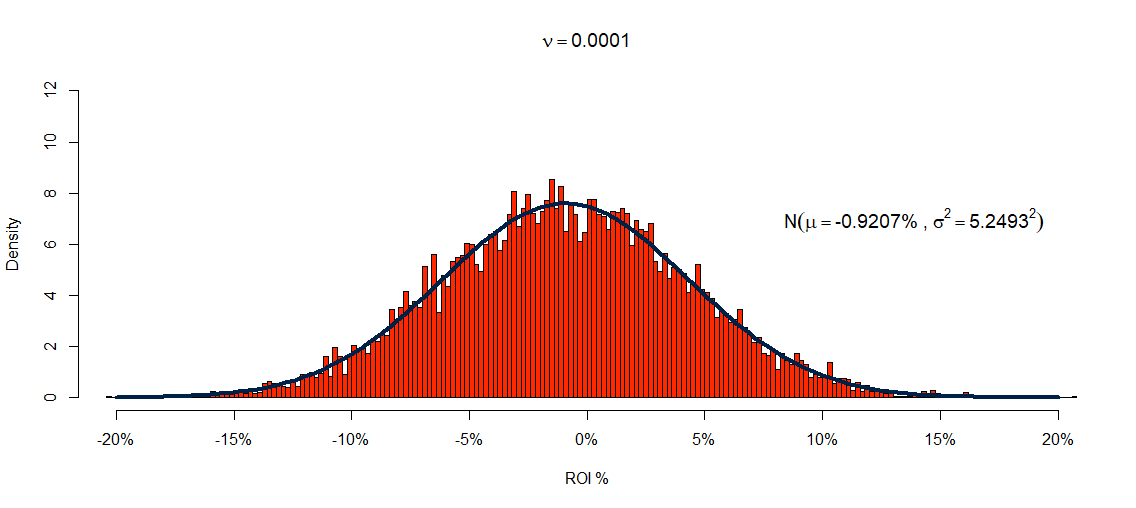}\hfill
\includegraphics[width=0.33\textwidth]{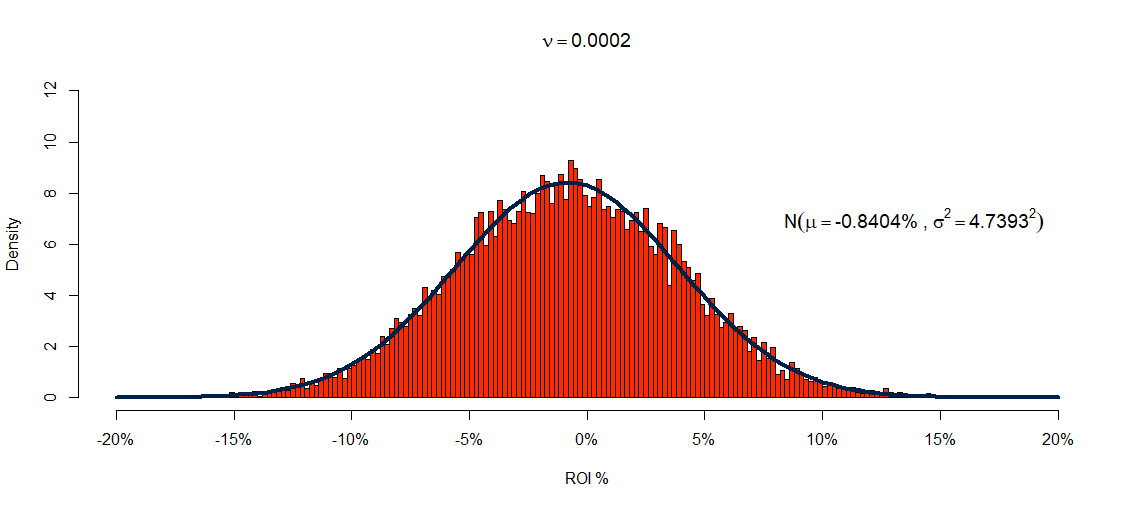}

\vspace{0.1cm}
\includegraphics[width=0.33\textwidth]{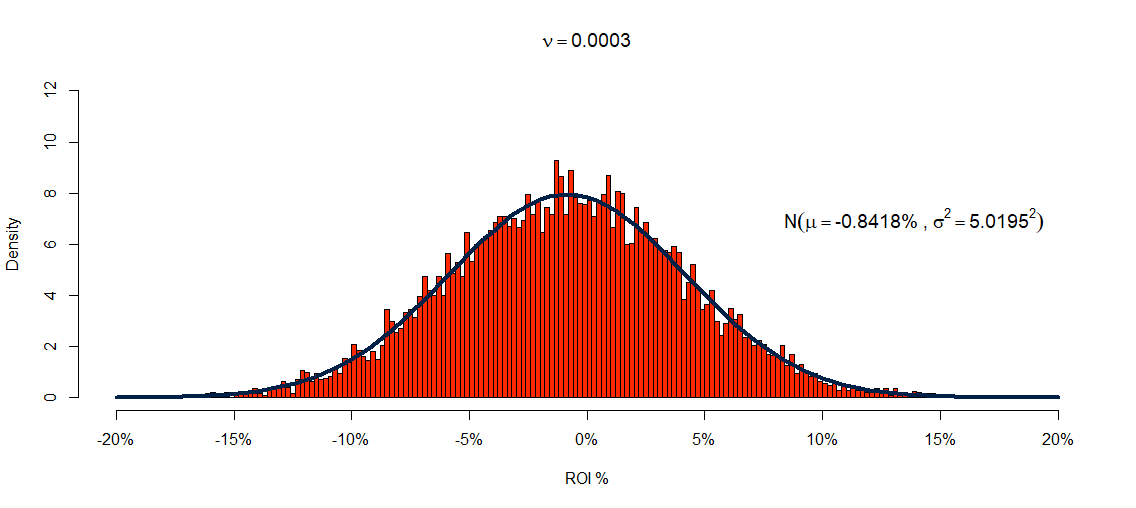}
\includegraphics[width=0.33\textwidth]{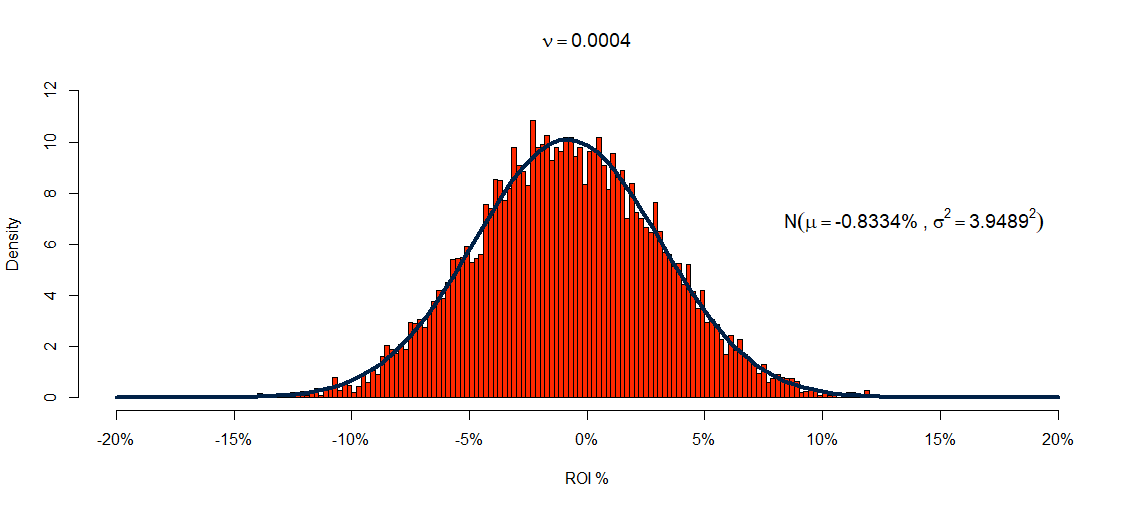}\hfill
\includegraphics[width=0.33\textwidth]{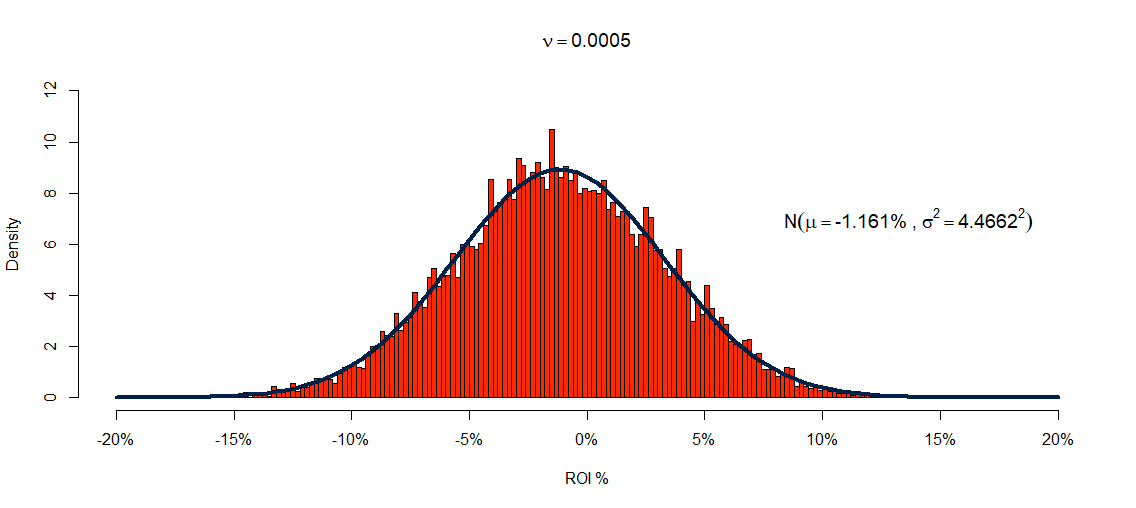}

\vspace{0.1cm}

\includegraphics[width=0.33\textwidth]{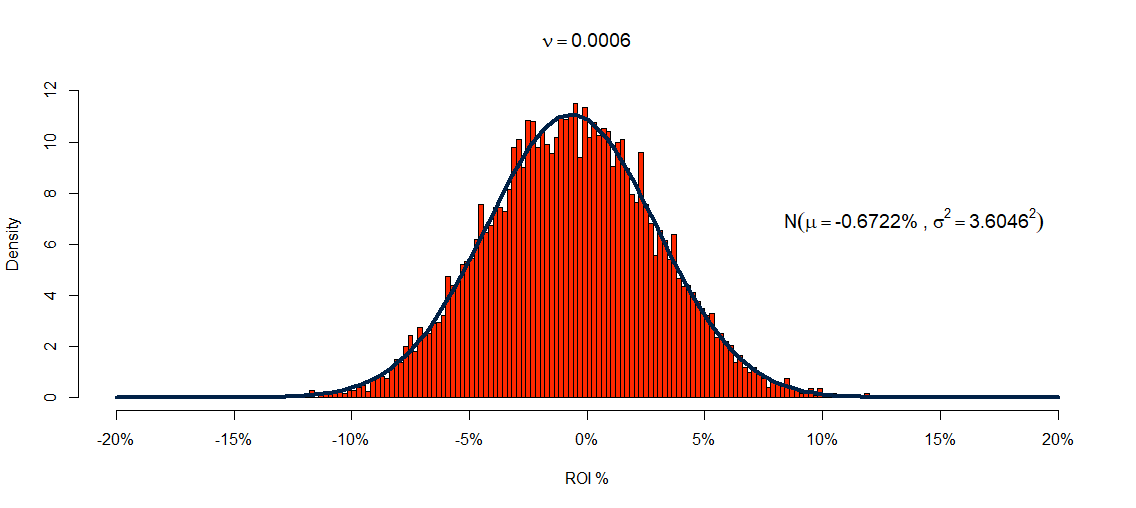}\hfill
\includegraphics[width=0.33\textwidth]{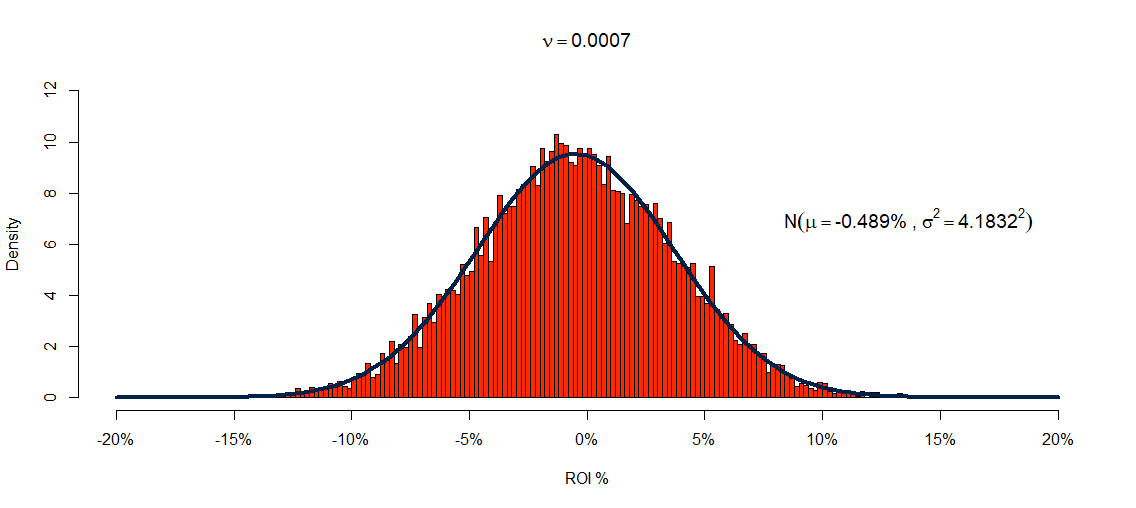}
\includegraphics[width=0.33\textwidth]{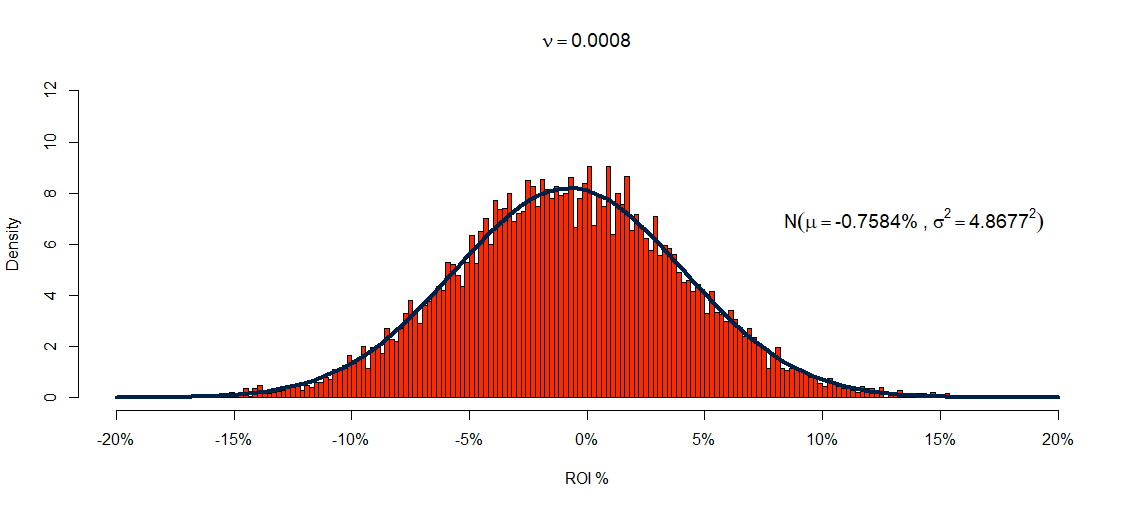}

\vspace{0.1cm}

\includegraphics[width=0.33\textwidth]{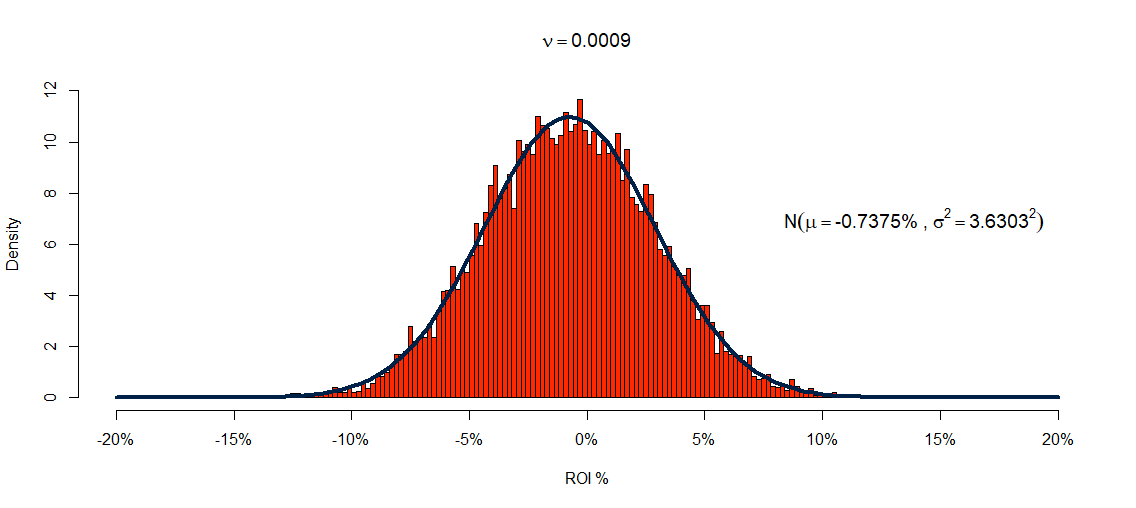}\hfill
\includegraphics[width=0.33\textwidth]{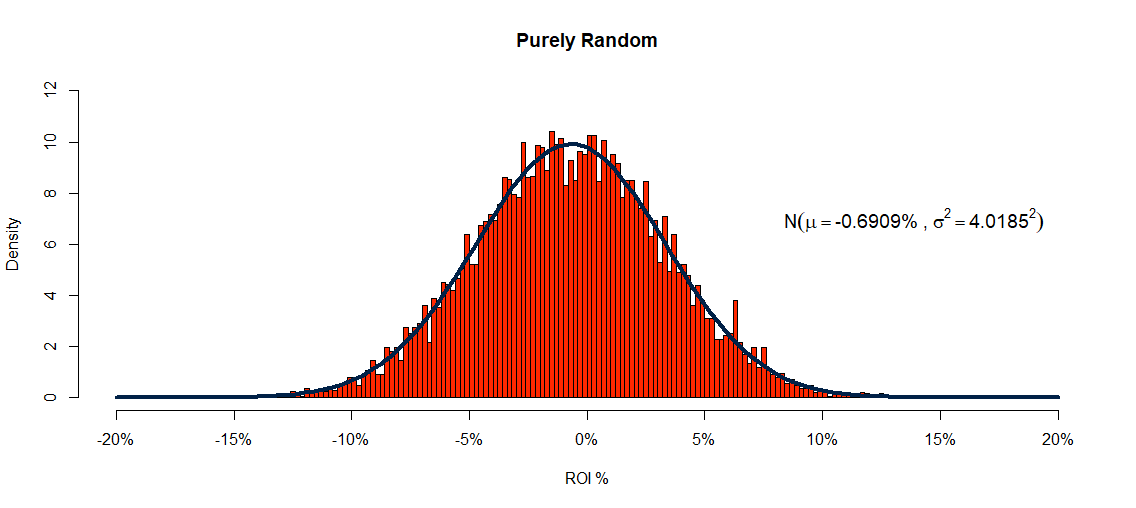}\hfill
\includegraphics[width=0.33\textwidth]{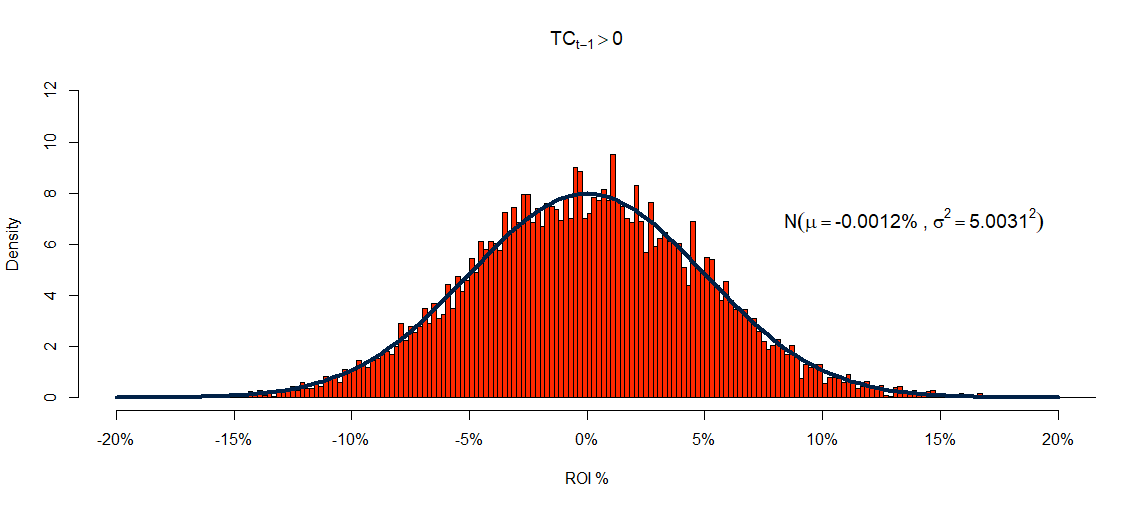}

\vspace{0.1cm}

\includegraphics[width=0.33\textwidth]{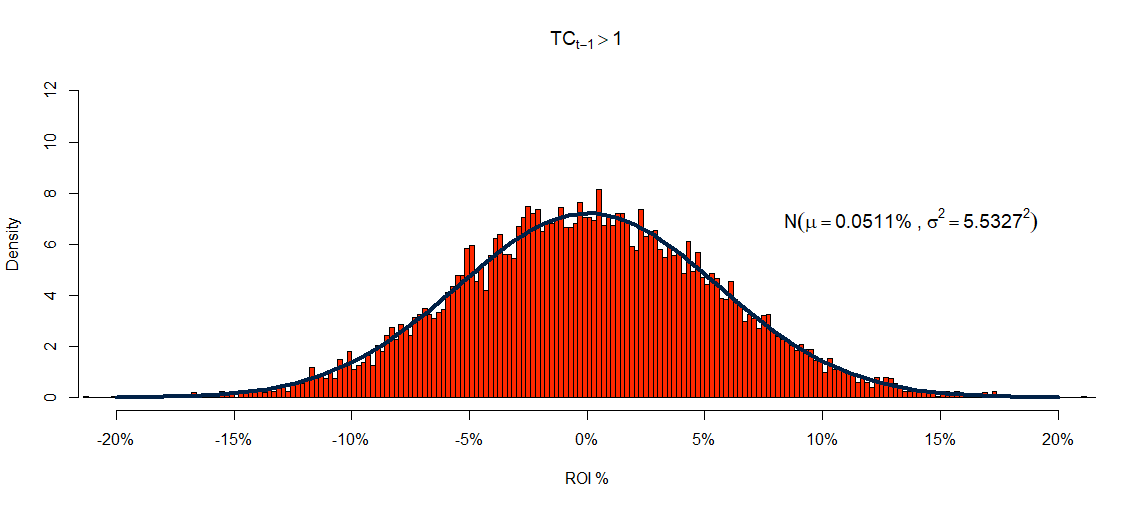}\hfill
\includegraphics[width=0.33\textwidth]{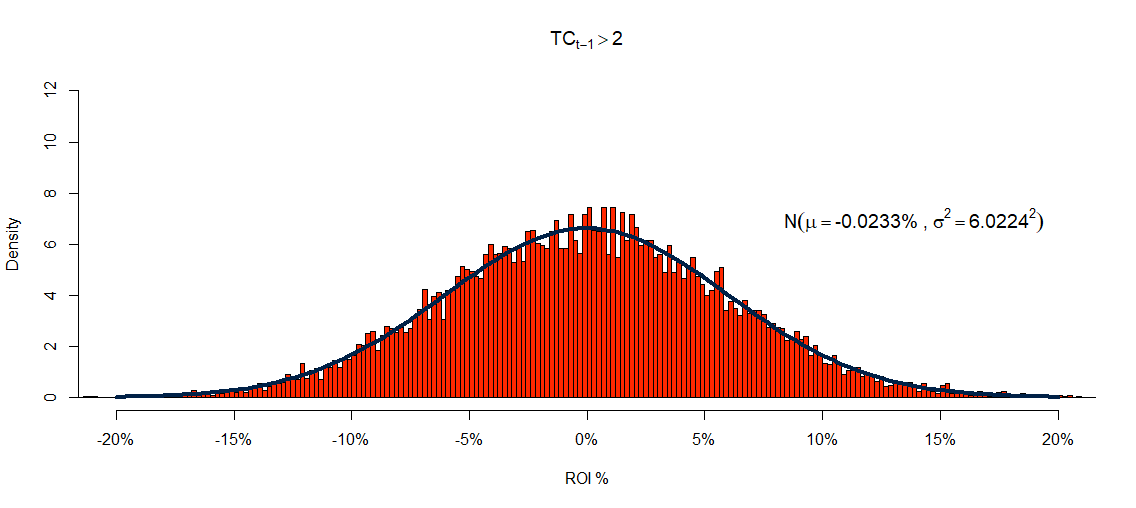}\hfill
\includegraphics[width=0.33\textwidth]{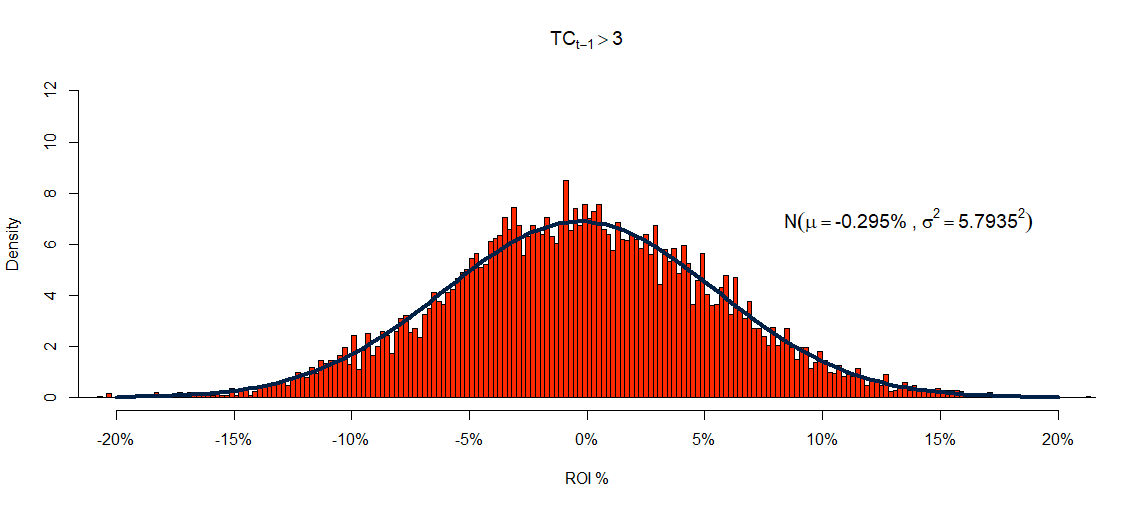}

\caption{Out-of-sample ROI distributions for varying $\nu$ using $\overset{\text{(II)}}{\mathbf{model}} \left(\mathbf{a}^0, {\boldsymbol{\hat\theta}^{\text{GA}, (\text{II})}_{\nu}} \right)$, corresponding to Table \ref{table: roi for a_1^0, a^0} and ROI distributions of common blackjack betting policies, corresponding to Table \ref{table: bet policy roi}.}
\label{table: hists bet}
\end{figure}
\fi

Now, Figure \ref{fig: Varying nu 100 nights} illustrates the distributions of bet sizes, $\tilde{bet}_k \in [1, 10]$, throughout a night using $\overset{\text{(II)}}{\mathbf{model}} \left(\mathbf{a}^0, {\boldsymbol{\hat\theta}^{\text{GA}, (\text{II})}_{\nu}} \right)$, for \iffoo $100$ \else $6$ \fi nights, where the net profit or loss is computed as the cumulative gain or loss at the end of each night (the $1,000$ hands of blackjack played). The optimization process consistently yields solutions that exhibit sparse betting patterns: the player seems to be consistently opted to bet close to the table minimum, given solutions with these specific $\nu$.

\iffoo
\begin{figure}[H]
    \centering
    \begin{minipage}{0.27\textwidth}
        \centering
\animategraphics[controls,autoplay,loop,width=0.8\textwidth]{3}{betsize/nu=0/bet_}{1}{100}
\caption*{$\nu = 0$}
    \end{minipage}
    \hfill
    \begin{minipage}{0.27\textwidth}
        \centering
\animategraphics[controls,autoplay,loop,width=0.8\textwidth]{3}{betsize/nu_=_0.0005/bet_}{1}{100}
        \caption*{$\nu = 0.0005$}
    \end{minipage}
    \hfill
    \begin{minipage}{0.27\textwidth}
        \centering
\animategraphics[controls,autoplay,loop,width=0.8\textwidth]{3}{betsize/nu_=_0.0007/bet_}{1}{100}
        \caption*{$\nu = 0.0007$}
    \end{minipage}
    \hfill
\caption{Distributions of bet size $\tilde{bet}_k$ over $100$ nights for various $\nu$.}
\label{fig: Varying nu 100 nights}
\end{figure}
\else

\begin{figure}[H]
\centering

\includegraphics[width=0.28\textwidth]{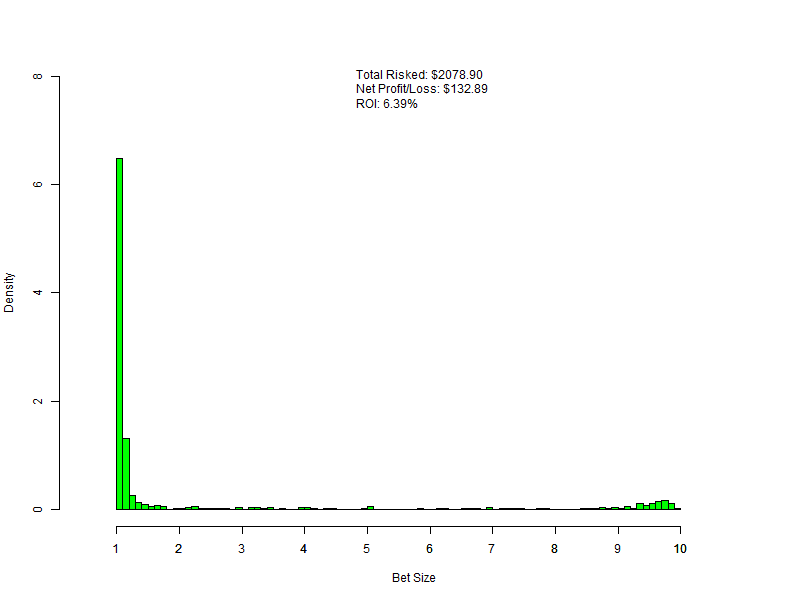}\hfill
\includegraphics[width=0.28\textwidth]{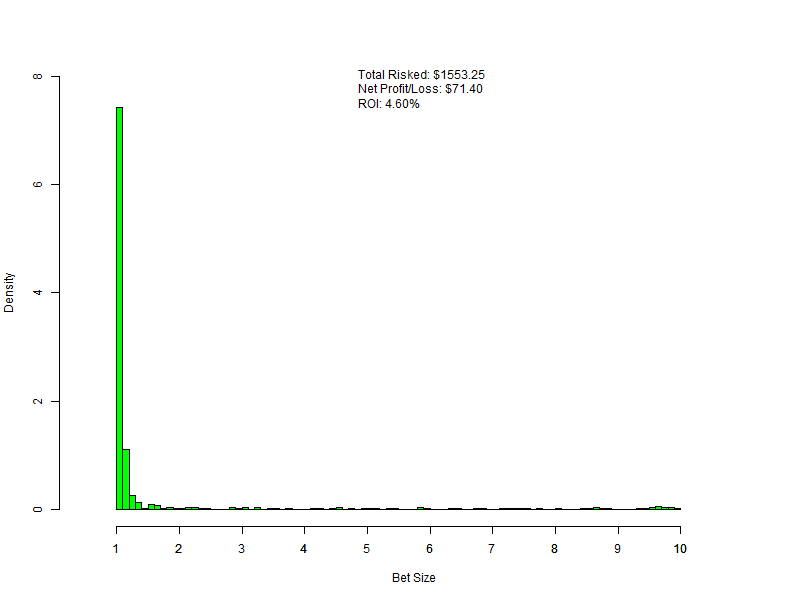}\hfill
\includegraphics[width=0.28\textwidth]{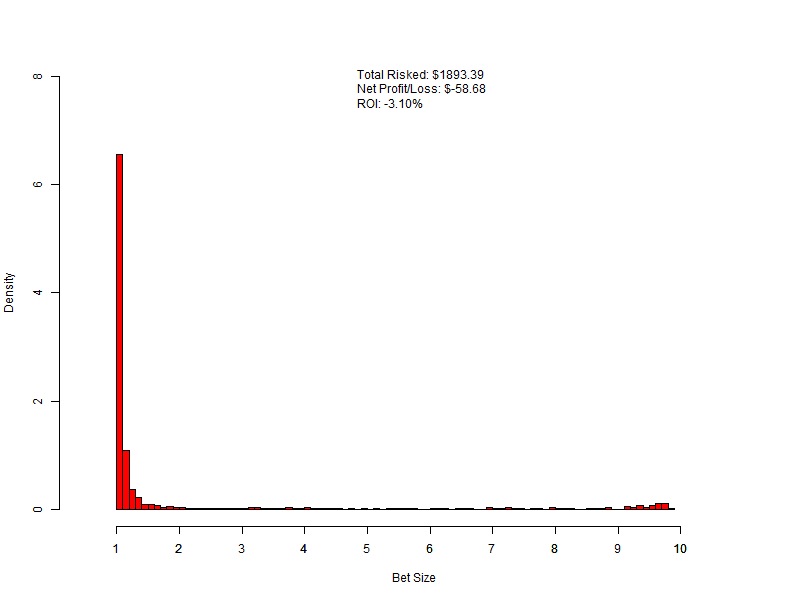}

\includegraphics[width=0.28\textwidth]{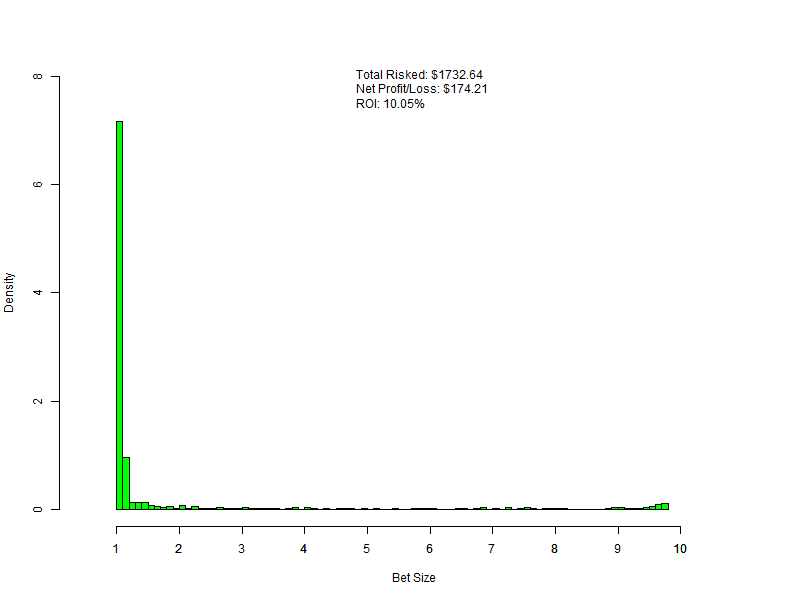}\hfill
\includegraphics[width=0.28\textwidth]{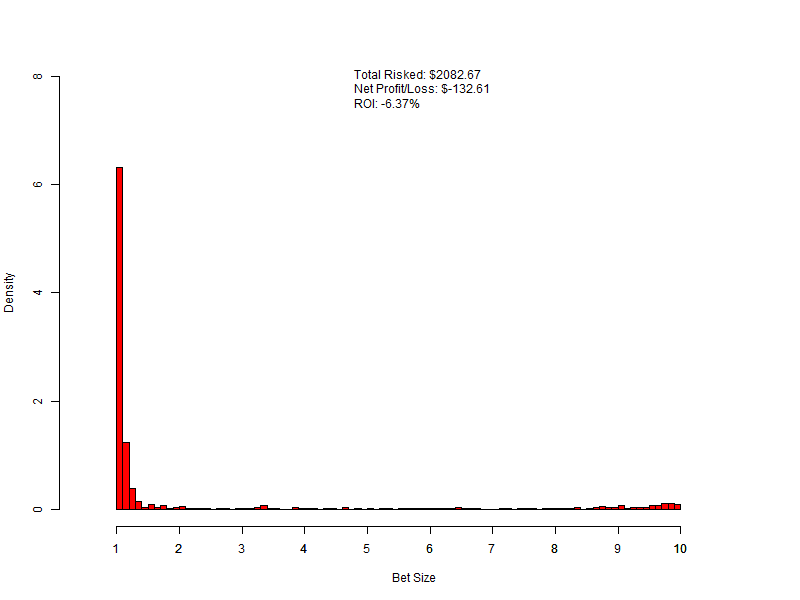}\hfill
\includegraphics[width=0.28\textwidth]{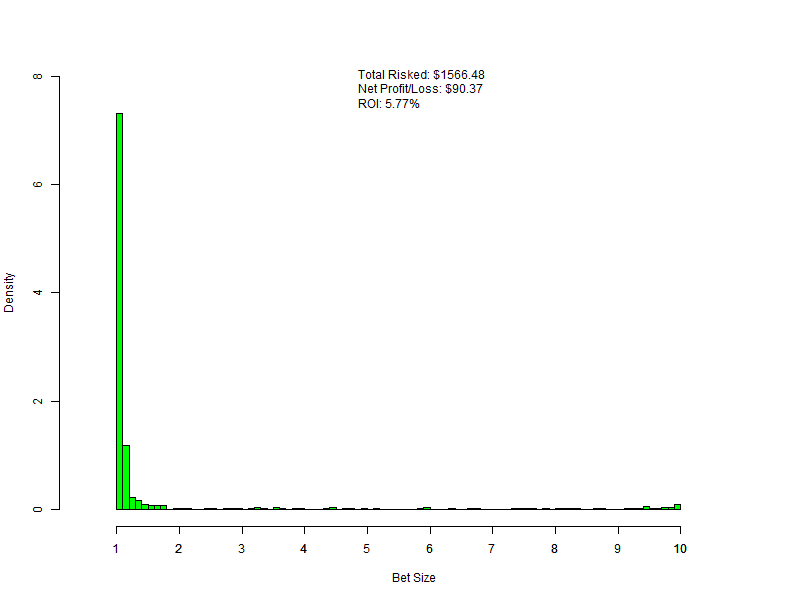}
\caption*{$\nu = 0.000$}

\includegraphics[width=0.28\textwidth]{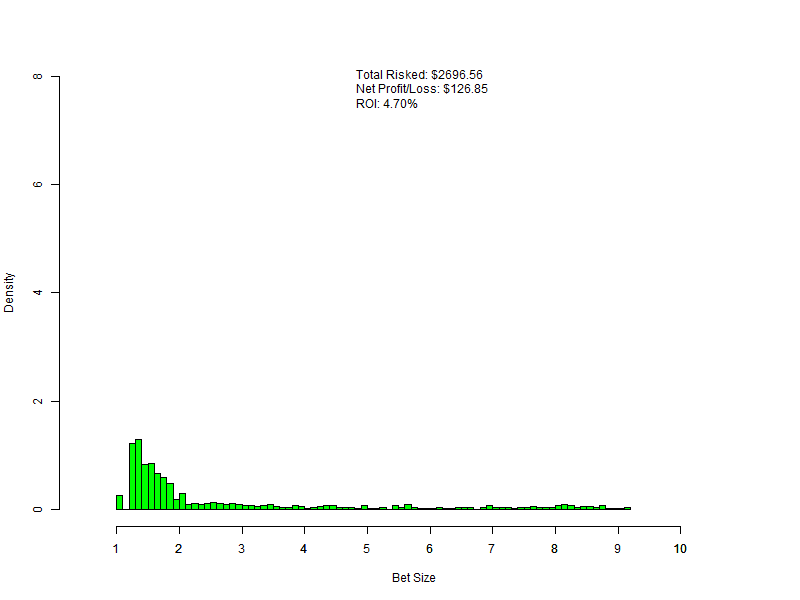}\hfill
\includegraphics[width=0.28\textwidth]{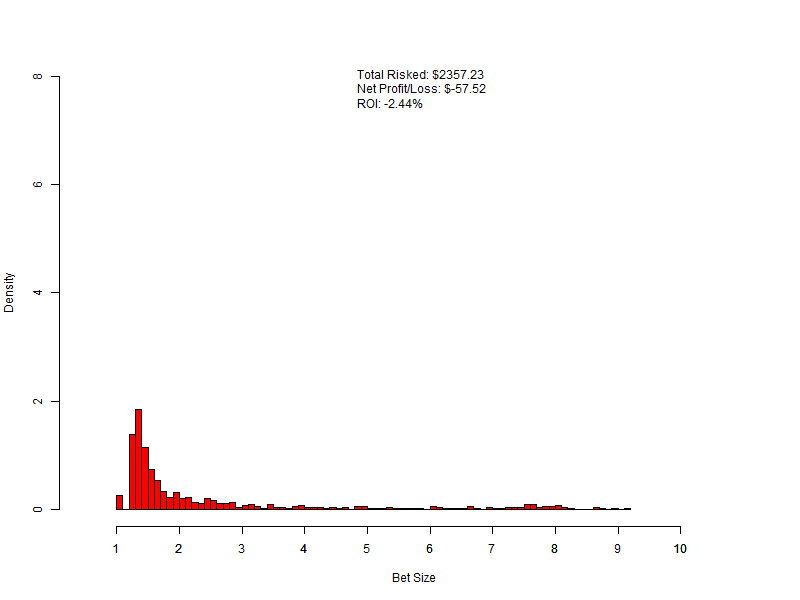}\hfill
\includegraphics[width=0.28\textwidth]{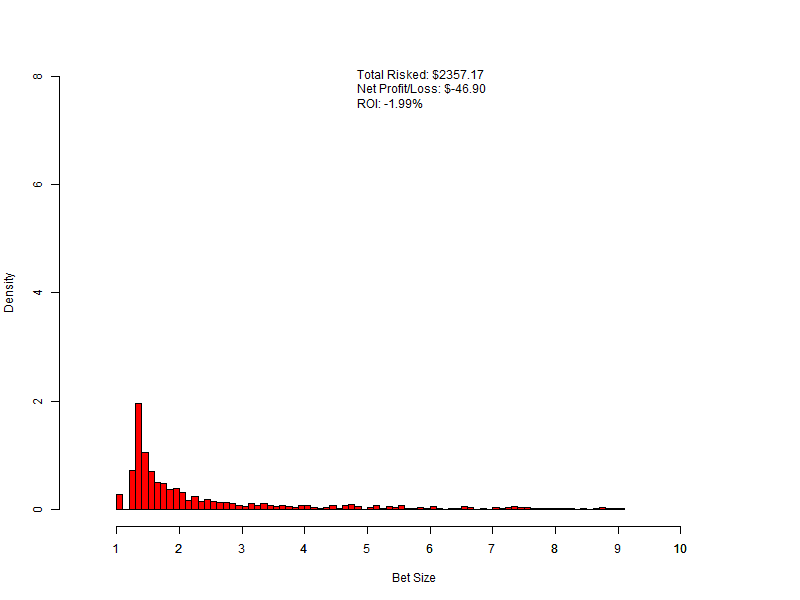}

\includegraphics[width=0.28\textwidth]{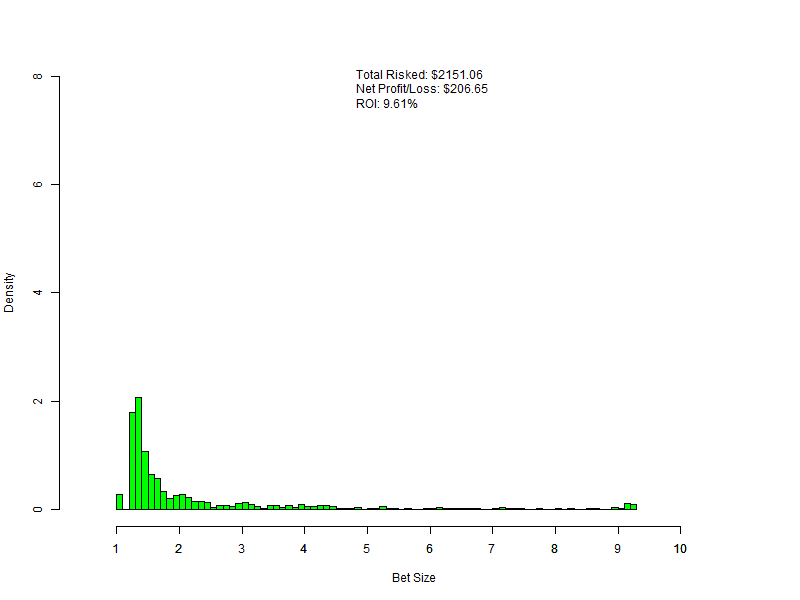}\hfill
\includegraphics[width=0.28\textwidth]{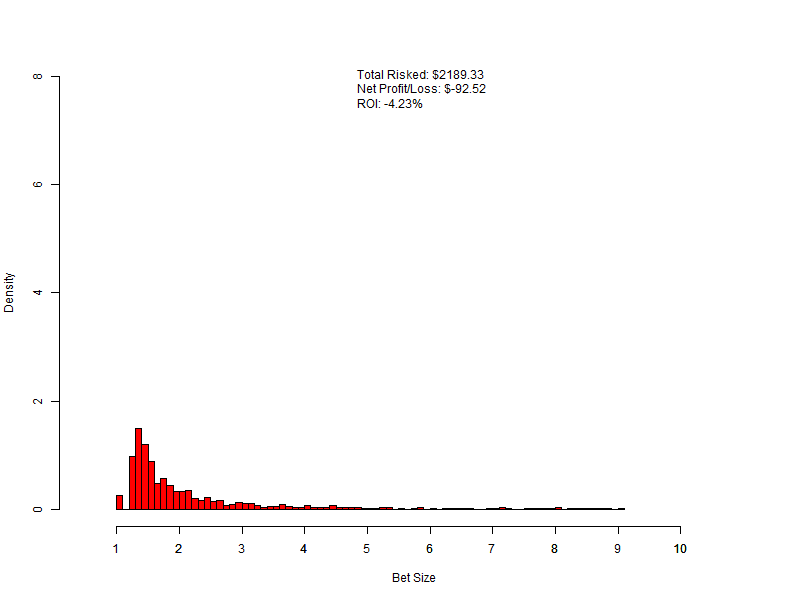}\hfill
\includegraphics[width=0.28\textwidth]{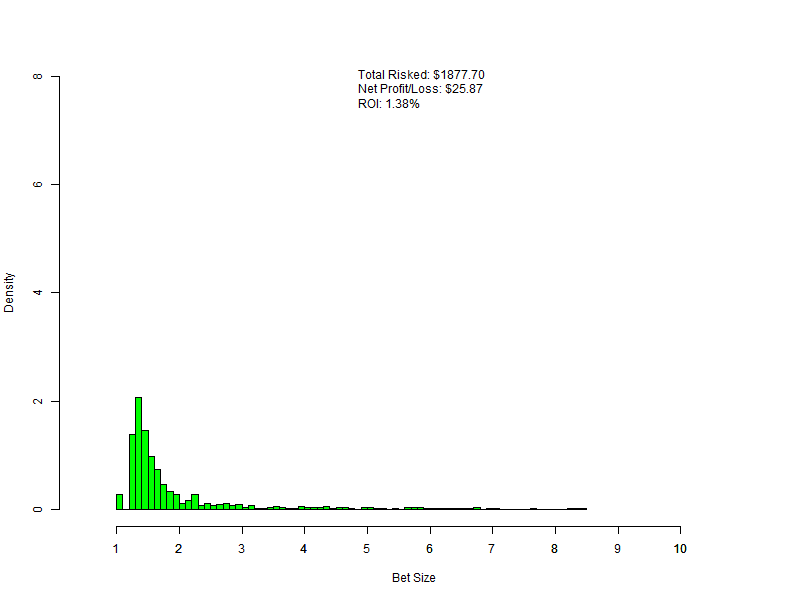}
\caption*{$\nu = 0.0005$}

\includegraphics[width=0.28\textwidth]{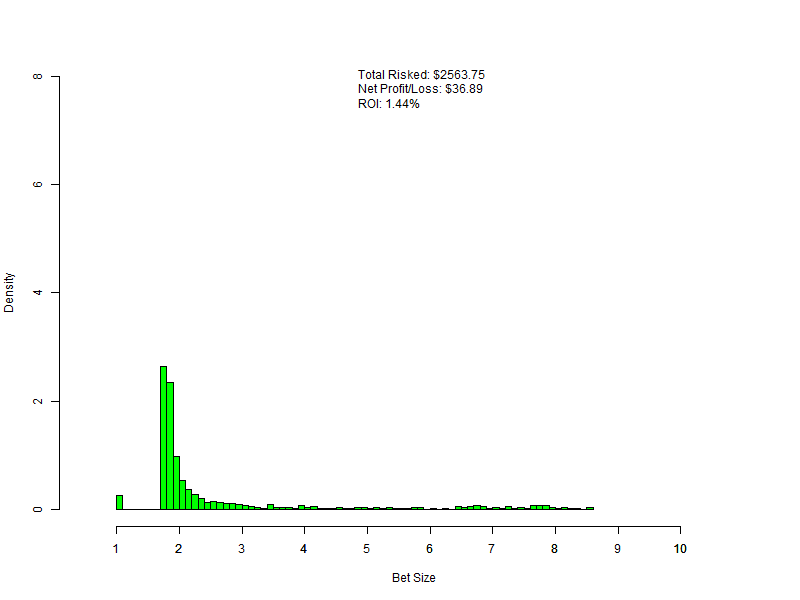}\hfill
\includegraphics[width=0.28\textwidth]{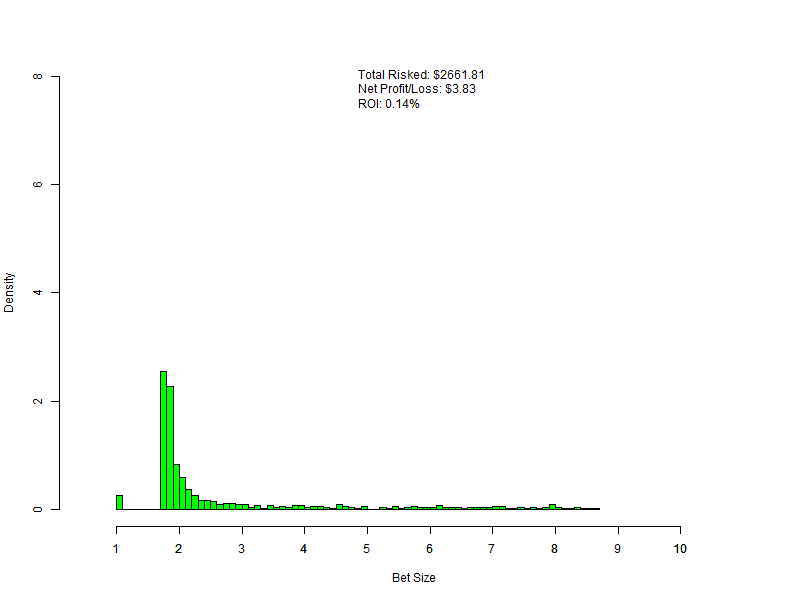}\hfill
\includegraphics[width=0.28\textwidth]{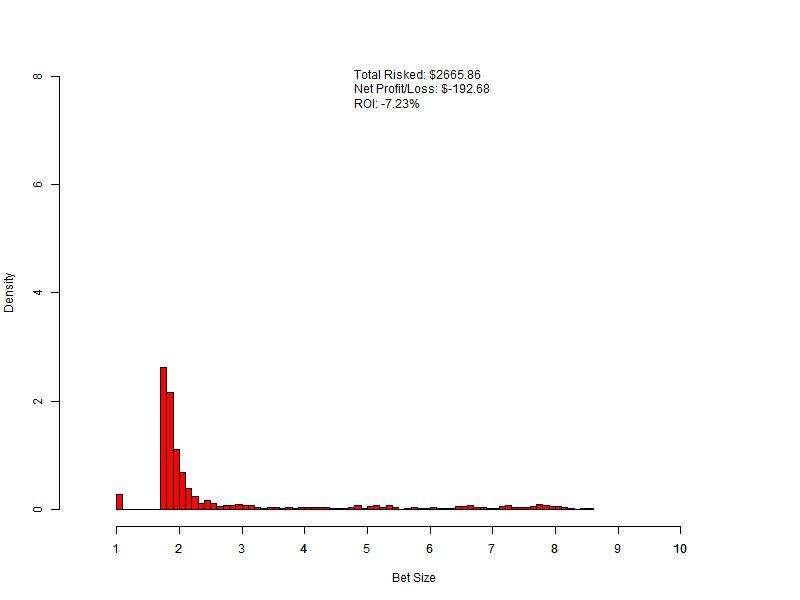}

\includegraphics[width=0.28\textwidth]{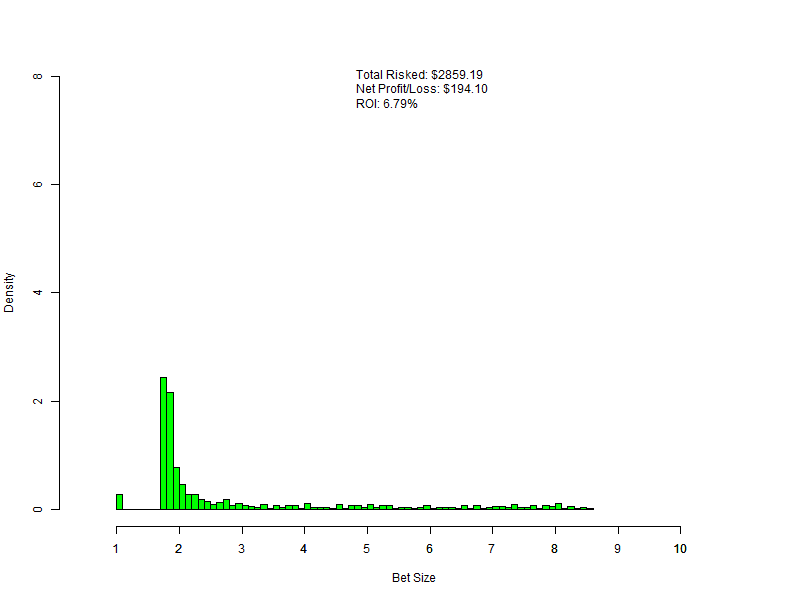}\hfill
\includegraphics[width=0.28\textwidth]{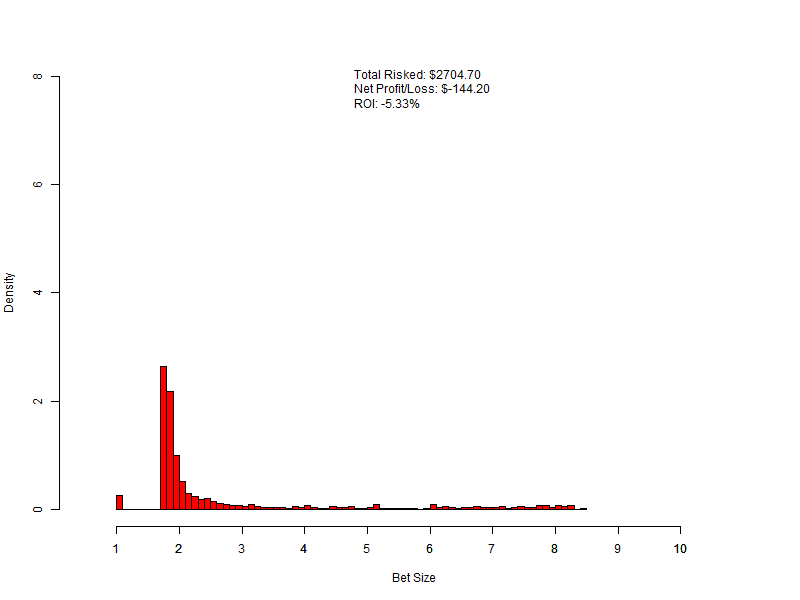}\hfill
\includegraphics[width=0.28\textwidth]{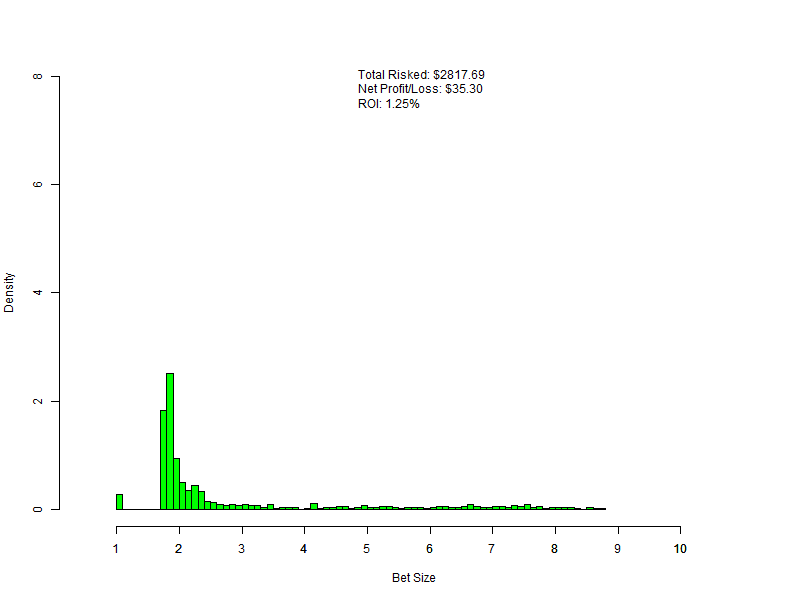}
\caption*{$\nu = 0.0007$}

\caption{Distributions of bet size $\tilde{bet}_k$ over $6$ nights for various $\nu$.}
\label{fig: Varying nu 100 nights}
\end{figure}
\fi

\section{Hybrid Algorithms}
Within the two-block MCMC framework described in Section \ref{sec: 2-sample}, the incorporation of pseudo-likelihoods introduces a sharpness parameter $\beta \in \mathbb{R}^+$ that governs the peakedness of the pseudo-likelihood. Consequently, this parameter also influences the shape of the conditional posterior $p(\boldsymbol{\theta} \mid \sigma_\theta^2, \mathcal{D})$, as increased values of $\beta$ lead the pseudo-likelihood to dominate the conditional, resulting in a more sharply peaked distribution (an effect discussed in greater detail in Section \ref{sec: beta}). An increase in $\beta$ renders the conditional posterior $p(\boldsymbol{\theta} \mid \sigma_\theta^2, \mathcal{D})$ more sensitive to variations in the objective function, thereby producing a sharply peaked distribution concentrated around a dominant mode. In other words, the heightened peakedness intensifies the concentration of samples in the vicinity of this dominant mode.\\

Now, to leverage this property, it is proposed to replace the MH sampling step in Section \ref{sec: 2-sample}, which samples $\boldsymbol{\theta} \mid \sigma_\theta^2, \mathcal{D}$, with an iterative optimization method. Given the sharply peaked nature of the conditional posterior, induced by a large $\beta$, the latter stages of an optimization method converge to the mode of $p(\boldsymbol{\theta} \mid \sigma_\theta^2, \mathcal{D})$, given the optimization method explicitly seeks $\operatorname*{argmax}_{\boldsymbol{\theta}} p( \boldsymbol{\theta} \mid \sigma_\theta^2,  \mathcal{D})$. These iterates can be regarded as samples from the approximate modal region of the conditional posterior, particularly when the posterior is highly concentrated around a dominant mode. \\

Additionally, Section \ref{sec: drones mcmc} and Section \ref{sec: xo mcmc} highlighted that variations in likelihood sharpness, the initial dispersion $\sigma_{\text{Init}}^2$, and the particular pseudo-likelihood formulation employed during MCMC, substantially influenced the dispersion parameter $\sigma_\theta^2$, and by extension, the degree of regularization inferred (given that $\sigma_\theta^2 \propto \frac{1}{\nu}$). This raised a critical concern: does the hierarchical Bayesian structure (implemented via two-block MCMC) truly enable data-driven regularization, or does it instead reintroduce user-specified regularization? After all, the user must still choose the pseudo-likelihood formulation (and its associated sharpness parameter $\beta$), as well as the initial dispersion $\sigma_{\text{Init}}^2$. In this light, it becomes necessary to reconsider the idea that the training data determines a meaningful level of regularization embedded within the MAP estimates.\\

Considering the iterative optimization procedure is to be employed rather than MH sampling in Block $1$ of the two-block MCMC framework, we are inclined to reinterpret the sampling of $\sigma_\theta^2 \mid \boldsymbol{\theta}, \mathcal{D}$ in Block $2$ as an auxiliary mechanism for enhancing exploration within the optimization process. Specifically, since the objective function being maximized is the conditional posterior $p(\boldsymbol{\theta} \mid \sigma_\theta^2, \mathcal{D})$, the act of sampling $\sigma_\theta^2 \mid \boldsymbol{\theta}, \mathcal{D}$ at each iteration induces a fluctuating optimization landscape. This variability results in a ``wobbly'' optimization trajectory, wherein the shape of the objective function changes slightly from one iteration to the next. Such non-static behaviour naturally encourages broader exploration of the parameter space before the algorithm converges toward a dominant mode of the conditional distribution. Furthermore, it must be noted that the sampling of $\sigma_\theta^2 \mid \boldsymbol{\theta}, \mathcal{D}$ per iteration continues to allow the training data to influence the inferred degree of regularization applied to the MAP estimates, although, as before, this influence remains substantially shaped by the user.

\subsection{Genetic algorithm hybrid} \label{sec: GA Hybrid}
Under the framework of a GA, where our fitness scores reflect that of maximizing the conditional  $p(\boldsymbol{\theta} \mid \sigma_\theta^2, \mathcal{D} )$, we can view each generation as a pursuit to refine $\boldsymbol{\theta} \in \mathbb{R}^S$ such that it closely reflects $\operatorname*{argmax}_{\boldsymbol{\theta}} p( \boldsymbol{\theta} \mid \sigma_\theta^2, \mathcal{D})$, that is, the mode of $p( \boldsymbol{\theta} \mid \mathcal{D}, \sigma_\theta^2)$. Furthermore, we may propose the notion that a GA can be viewed as a means of drawing $\boldsymbol{\theta}$ from an approximate modal region of $p( \boldsymbol{\theta} \mid \sigma_\theta^2, \mathcal{D})$, where each $n^{th}$ individual per $m^{th}$ generation, denoted as $\boldsymbol{\theta}^{(n, m)}$, may be viewed as a draw. \\

Now the fitness values for $\boldsymbol{\theta}^{(n, m)}$:
\begin{align*}
    f_{n, m} &= \operatorname*{max} \left\{p\left( \boldsymbol{\theta}^{(n, m-1)} \mid\sigma_{\theta^{(n, m-1)}}^2 ,  \mathcal{D}\right)\right\} \nonumber \\
    &= \operatorname*{max} \left\{ p \left(\mathcal{D} \mid\boldsymbol{\theta}^{(n, m-1)} \right) p\left(\boldsymbol{\theta}^{(n, m-1)} \mid \sigma_{\theta^{(n, m-1)}}^2\right) \right\} \nonumber \\
     & = \max \Bigg\{  p \left(\mathcal{D} \mid\boldsymbol{\theta}^{(n, m-1)} \right)
\cdot \frac{1}{\sqrt{\left(2\pi\sigma_{\theta^{(n, m-1)}}^2 \right)^S}} \exp \left( -\frac{1}{2\sigma_{\theta^{(n, m-1)}}^2}  \lVert \boldsymbol{\theta}^{(n, m-1)} \rVert^2  \right)\Bigg\} \nonumber \\
     & = \max \Bigg\{ \log\left( p \left(\mathcal{D} \mid\boldsymbol{\theta}^{(n, m-1)} \right) \right)
 -  \frac{1}{2\sigma_{\theta^{(n, m-1)}}^2}  \lVert \boldsymbol{\theta}^{(n, m-1)} \rVert^2   - \frac{S}{2} \log \left(2\pi\sigma_{\theta^{(n, m-1)}}^2 \right) \Bigg\}.
\end{align*}
Afterwhich we conduct the sampling step in Block 2 in Section~\ref{sec: 2-sample}, where
$\left(\sigma_\theta^2\right)^{(n, m)} \mid \boldsymbol{\theta}^{(n, m)}, \mathcal{D}
\sim \text{Inv-Gamma}\bigl(
  a + \tfrac{S}{2},\allowbreak\hspace{0pt}
  b + \tfrac{\lVert \boldsymbol{\theta}^{(n, m)} \rVert^2}{2}
\bigr)$.
\subsection{Gradient descent hybrid for classification} \label{sec: GD Hybrid}
Furthermore, for non-arbitrary objectives, that is for problems whose cost functions can be directly related to input-ouput pairs (for example, in supervised learning), which give rise to data-driven likelihoods (and not pseudo-likelihoods as we've been employing in this study), we may employ gradient-based optimization methods in order to maximize the conditional $p(\boldsymbol{\theta} \mid \sigma_\theta^2, \mathcal{D} )$. We propose that we may adopt the same approach as before, by postulating that each iteration of gradient descent (GD) may be viewed as sampling from the approximate modal region of the conditional $p(\boldsymbol{\theta} \mid \sigma_\theta^2, \mathcal{D} )$. Hence, the $m^{th}$ iteration of $\boldsymbol{\theta}$ for $m = 1, 2 \ldots, M$ with step-size $h$ is updated as:
\begin{align}
    \boldsymbol{\theta}^{(m)}  =\boldsymbol{\theta}^{(m-1)} - h \cdot \nabla \text{Obj} \left(\boldsymbol{\theta}^{(m-1)},  \left(\sigma_\theta^2\right)^{(m-1)} \right). \label{eq: gd}
\end{align}
As gradient descent seeks to minimize an objective function ($\text{Obj}$), we may express our optimization problem, $\operatorname*{argmax}_{\boldsymbol{\theta}} p( \boldsymbol{\theta} \mid \sigma_\theta^2,  \mathcal{D})$, as:
\[
\operatorname*{argmin}_{\boldsymbol{\theta}} \text{Obj}(\boldsymbol{\theta}) = \operatorname*{argmin}_{\boldsymbol{\theta}} \left\{ -\log \left[ p(\boldsymbol{\theta} \mid \sigma_\theta^2, \mathcal{D}) \right] \right\}.
\]
Now, 
\begin{align}
  - \log \left[p(\boldsymbol{\theta} \mid \sigma_\theta^2, \mathcal{D} )\right] & \propto  -\log \left[p \left(\mathcal{D} \mid\boldsymbol{\theta} \right) \right] - \log \left[p\left(\boldsymbol{\theta} \mid \sigma_{\theta}^2\right) \right] \nonumber \\ 
    & \propto -\log\left[p \left(\mathcal{D} \mid\boldsymbol{\theta} \right) \right]
 + \frac{1}{2\sigma_{\theta}^2}  \lVert \boldsymbol{\theta} \rVert^2  + \frac{S}{2} \log \left(2\pi\sigma_{\theta}^2 \right). \label{eq: gd half obj}
\end{align}
In the context of $K$-class classification, we know a neural network outputs logits $z_{i, 1}, z_{i, 2}, \ldots z_{i, d_L}$ for observations $i = 1, 2, \ldots, N$ and $k =1, 2,\ldots, d_L = K$ classes. These are transformed into probabilities using a softmax activation function. Hence, for a given observation $i$, the predicted probability of class $k$ is:
\begin{align*}
    p_{i, k} &= \sigma_L(z_{i, k}) = a_{k}(i)^L=\frac{\exp(z_{i, k})}{\sum_{j = 1}^{d_L} \exp(z_{i, j})}.
\end{align*}
So for a one-hot-encoded $y_i$, $p(y_i) = \prod_{k = 1}^{d_L} \left({a_k(i)^L}\right)^{y_{i, k}}$. Now for $N$ independent observations, the likelihood is:
\begin{align}
    p(\mathcal{D}\mid \boldsymbol{\theta}) &= \prod_{i = 1}^{N}\prod_{k = 1}^{d_L} \left({a_k(i)^L}\right)^{y_{i, k}} .\label{eq: lik classification}
\end{align}
Being such, we substitute this likelihood into Equation \ref{eq: gd half obj} to obtain our cost function:
\begin{align}
     \mathcal{L}\left(\boldsymbol{\theta}, \sigma_{\theta}^2 \right)  = -  \log \left[p(\boldsymbol{\theta} \mid \sigma_\theta^2, \mathcal{D} )\right] 
    &\propto 
    \underbrace{
        \left(
        -\sum_{i =1}^N \sum_{k = 1}^{d_L} y_{i, k} \log\left( a_k(i)^L \right)
        + \frac{1}{2\sigma_{\theta}^2}  \lVert \boldsymbol{\theta} \rVert^2
        \right)
    }_{ \mathcal{L}_{\text{cross-entropy}}(\boldsymbol{\theta})\text{ + L$2$ penalty}}
    + \frac{S}{2} \log\left(2\pi\sigma_{\theta}^2 \right) \nonumber \\
    & = \mathcal{L}^*_{\text{cross-entropy}}(\boldsymbol{\theta}, \sigma_{\theta}^2)  + \frac{S}{2} \log\left(2\pi\sigma_{\theta}^2 \right). \label{eq: gd obj}
\end{align}
Gradient descent proceeds via backpropagation, with the gradient of our cost 
function satisfying 
$\nabla \mathcal{L}(\boldsymbol{\theta},\allowbreak\hspace{0pt}\sigma_{\theta}^2)
 = 
 \nabla \mathcal{L}^*_{\text{cross-entropy}}
 (\boldsymbol{\theta},\allowbreak\hspace{0pt}\sigma_{\theta}^2)$ since the additional term in Equation \ref{eq: gd obj} does not depend on $\boldsymbol{\theta}$ seeing as $\nabla = \frac{\partial(\cdot)}{\partial \boldsymbol{\theta}}$. Now after sampling $\boldsymbol{\theta}^{(m)}$ from Equation \ref{eq: gd}, we sample our dispersion parameter as $(\sigma_\theta^2)^{(m)} \mid \boldsymbol{\theta}^{(m)}, \mathcal{D} 
\sim \text{Inv-Gamma}\allowbreak\left( a + \frac{S}{2},\ b + \frac{\lVert \boldsymbol{\theta}^{(m)} \rVert^2}{2} \right)$ as before. \\

Now under the framework of GD, for a fixed $\sigma_\theta^2$, our cost function (Equation \ref{eq: gd obj})  is static, and GD follows a smooth, deterministic trajectory toward a minimum. By sampling $\sigma_\theta^2$ per iteration, we create a ``wobbly'' optimization path: the gradient direction shifts not just due to the current $\boldsymbol{\theta}$, but also because the regularization term’s influence (that is, the influence of $\sigma_\theta^2$) fluctuates. With fixed $\sigma_\theta^2$, GD descends a static trough (a single, well-defined valley) to its lowest point. Varying $\sigma_\theta^2$ reshapes this trough each iteration: large $\sigma_\theta^2$ (low $\nu$) widens and shallows it, letting the algorithm wander; small $\sigma_\theta^2$ (high $\nu$) narrows and deepens it near the origin, tugging inward. The descent becomes a pursuit of a shifting bottom, possibly broadening the exploration before homing in. 
\subsection{\texorpdfstring{A toy example: $3$-class classification}{Toy example: 3-class classification}}

We apply the three previously described optimization methods: namely, two-block MCMC, GD Hybrid and GA Hybrid, to the 3-class classification problem introduced in \ref{sec: particle data}, using the likelihood formulation given in Equation \ref{eq: lik classification}. To ensure comparability across methods, we initialize all algorithms with the same starting solution, denoted  as $\boldsymbol{\theta}^{(1)}$. For the GA Hybrid approach, every individual in the initial population is set to this same value,  that is $\boldsymbol{\theta}^{(n, 1)} = \boldsymbol{\theta}^{(1)}$ for all $n$ individuals. Each method, however, comes with specific caveats that influence its behavior. For the GA Hybrid method, several hyperparameters significantly affect the optimization dynamics. In particular, the choice of lower and upper bounds constraining the search space directly impacts the magnitude of $\lVert \boldsymbol{\theta} \rVert^2$. Additionally, the mutation probability applied to a particular gene in a parent solution governs the extent of variability of $\lVert \boldsymbol{\theta} \rVert^2$. With respect to two-block MCMC, the effect of the initial dispersion parameter $\sigma_{\text{Init}}^2$ on performance has already been examined in Section \ref{sec: xo mcmc}. For the GD Hybrid approach, variation in the step size $h$ primarily influences the convergence rate.\\

Figure \ref{fig: hybrid sigma} illustrates that the different optimization methods give rise to varying concentrations in the marginal distributions of the dispersion parameter $\sigma_{\theta}^2$, which we attribute to differences in the degree of regularization implicitly induced by each method. This variation stems from the distinct convergence behaviors of  $\lVert \boldsymbol{\theta}^{(j)} \rVert^2$ across iterations. Recall that $\sigma_\theta^2 \mid \boldsymbol{\theta}, \mathcal{D} \sim \text{Inv-Gamma}\left(a + \tfrac{S}{2},\; b + \tfrac{\lVert \boldsymbol{\theta} \rVert^2}{2} \right), \ \text{with } a, b \approx 0$. Hence it follows that if $\lVert \boldsymbol{\theta}^{(j)} \rVert^2$ fluctuates around a constant value $c$, then the marginal distribution of $\sigma_\theta^2 \mid \mathcal{D}$ should also follow an inverse-gamma distribution with approximately constant shape and scale parameters as such: $\sigma_\theta^2 \mid \mathcal{D} \sim \text{Inv-Gamma}\left(a + \tfrac{S}{2},\; b + \tfrac{c}{2} \right)$. Notably, the GD Hybrid method exhibits the least variability in $\lVert \boldsymbol{\theta}^{(j)} \rVert^2$, followed by MCMC, both of which result in marginal distributions of $\sigma_{\theta}^2 \mid \mathcal{D}$ that closely follow an inverse-gamma form. Among the optimization methods considered, the GA Hybrid approach is the only one that results in a clearly non-inverse-gamma marginal of $\sigma_\theta^2 \mid \mathcal{D}$, which we attribute to the greater variability in $\lVert \boldsymbol{\theta}^{(j)} \rVert^2$ across iterations. This variation suggests that the GA explores more diverse regions of the parameter space, which in turn leads to a more dispersed marginal of $\sigma_\theta^2 \mid \mathcal{D}$ (as noted earlier, this variability can be modulated through the mutation probability).\\

Figure \ref{fig: hybrid theta2} presents the distribution of one of the estimated parameters, specifically $\hat{\theta}_2$. The top panel displays the results on a common scale to highlight the extent to which the different optimization methods produce distinct solutions, $\hat{\boldsymbol{\theta}}$, allowing for a direct comparison between them. In contrast, the bottom panel uses individual scales to better visualize the shape of each distribution of $\hat{\theta}_2$. This latter view illustrates that both hybrid methods can function as approximate sampling techniques, all yielding unimodal distributions for the estimated parameter, $\hat{\theta}_2$ (albeit with differing spreads). While MCMC serves as the baseline for comparison, since it directly samples from the conditional posterior $p( \boldsymbol{\theta} \mid  \sigma_\theta^2, \mathcal{D})$ rather than sampling from the approximate modal region of the posterior, multiple runs of the sampler often yield noticeably different parameter solutions. This behavior is also observed in the other two optimization methods as well. Nevertheless, these disparate solutions tend to produce comparable in-sample performance, suggesting the existence of a multi-modal posterior landscape. Accordingly, one cannot conclude that the methods are fundamentally dissimilar solely on the basis of differences in the estimated solutions $\hat{\boldsymbol{\theta}}$, as variation in solutions arises even within a single optimization method across multiple runs.
\\

Finally, Figure \ref{fig: hybrid classes} in conjunction with Table \ref{table: acc hybrid}, illustrates the performance of the three optimization methods. The response curves in Figure \ref{fig: hybrid classes} reveal substantial similarity across methods, with each successfully delineating distinct classification regions for the three particle types. Moreover, Table \ref{table: acc hybrid} reports comparable in- and out-of-sample performance across the three approaches, further supporting the notion that all methods achieve a similar level of predictive accuracy.

\iffoo
\begin{figure}[H]
    \centering
\animategraphics[controls,autoplay,loop,width=0.6\textwidth]{1}{Hybrid/sigma/}{1}{3}
\caption{$ \lVert \boldsymbol{\theta}^{(j)} \rVert^2$ for $j  = 1, \ldots 150, 000$ (post burn-in) with distribution of marginal of $\sigma_\theta^2 \mid \mathcal{D}$.}
\label{fig: hybrid sigma}
\end{figure}
\else
\begin{figure}[H]
    \centering
\includegraphics[width=0.32\textwidth]{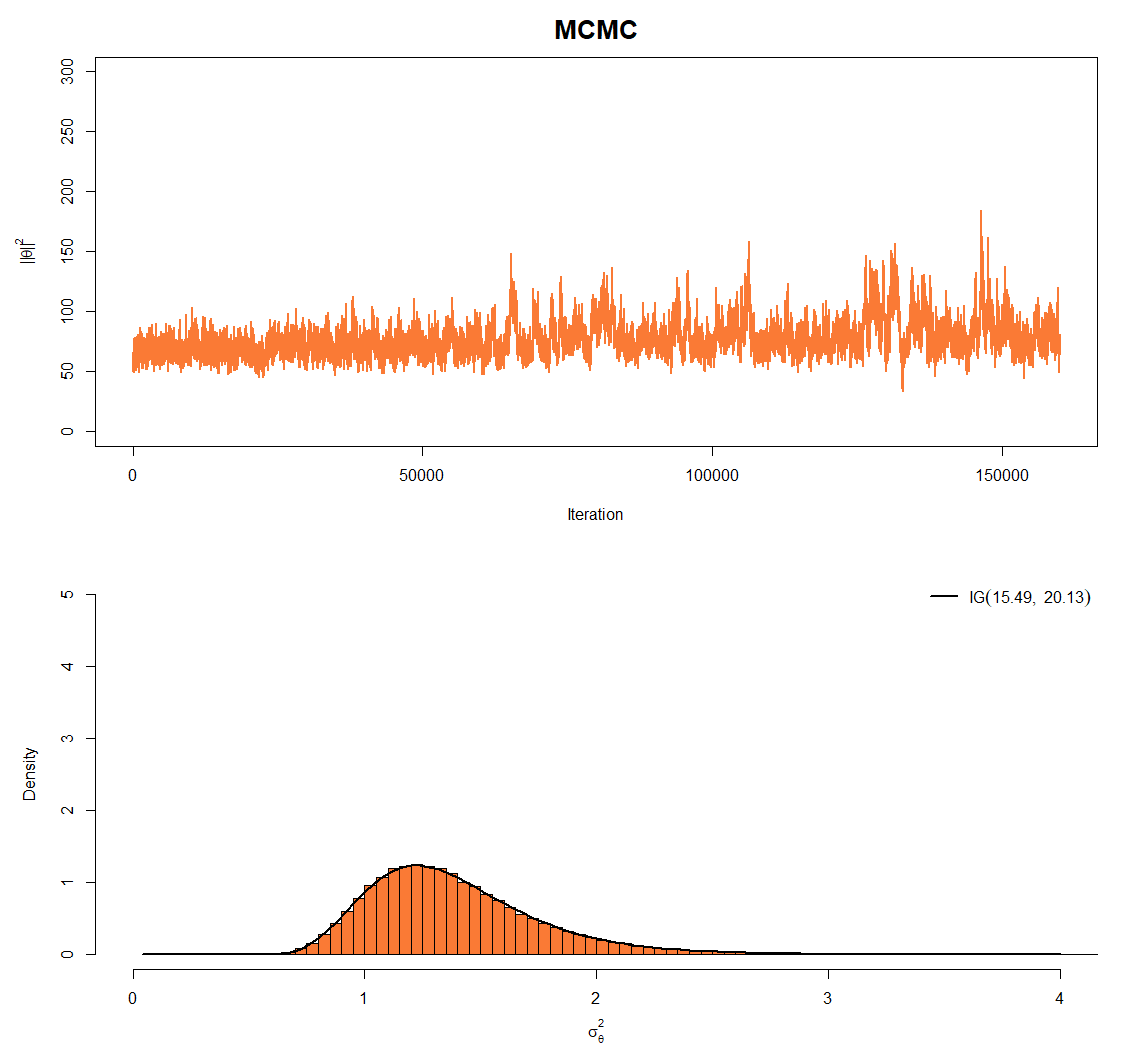}\hfill
\includegraphics[width=0.32\textwidth]{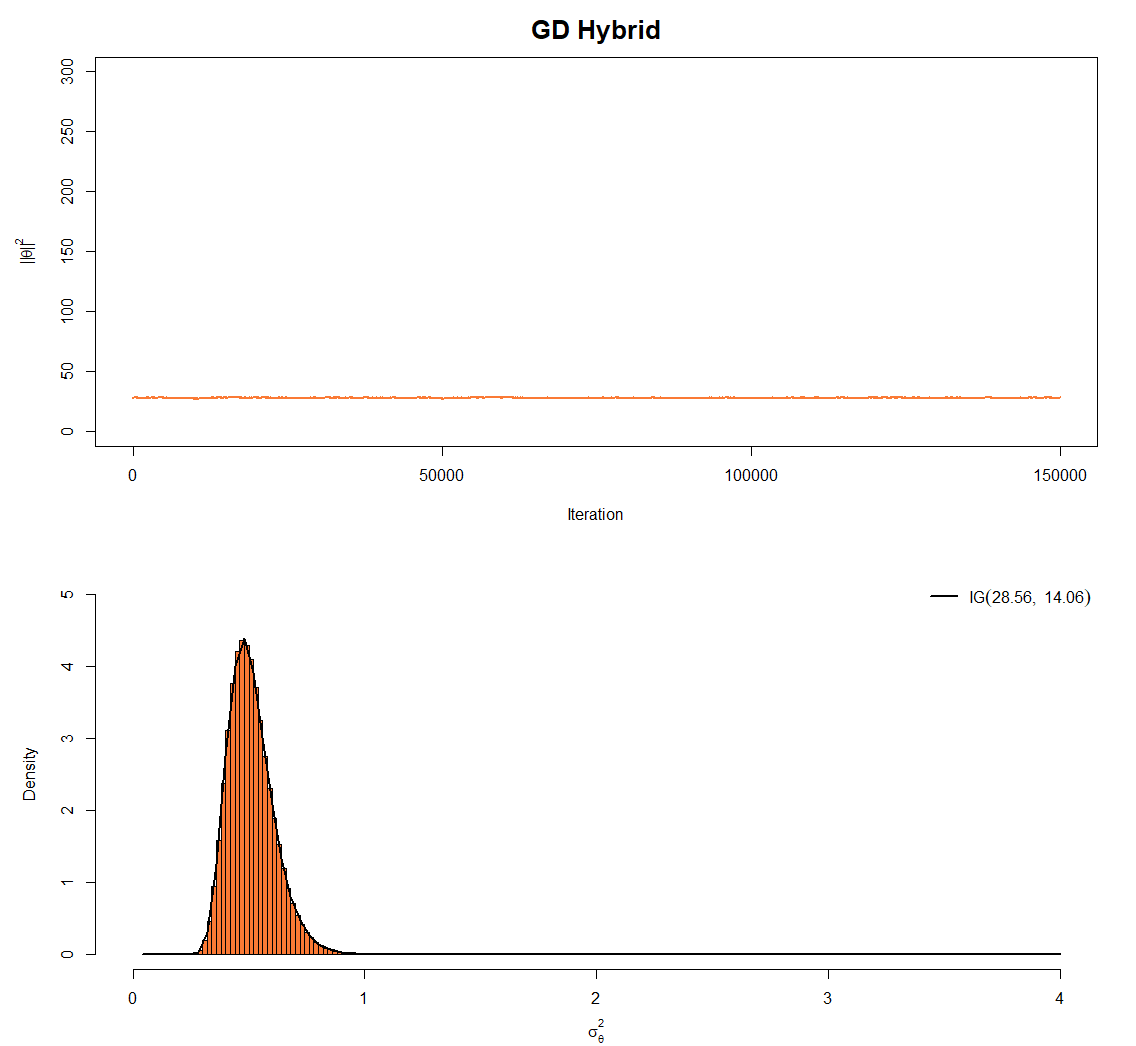}
\includegraphics[width=0.32\textwidth]{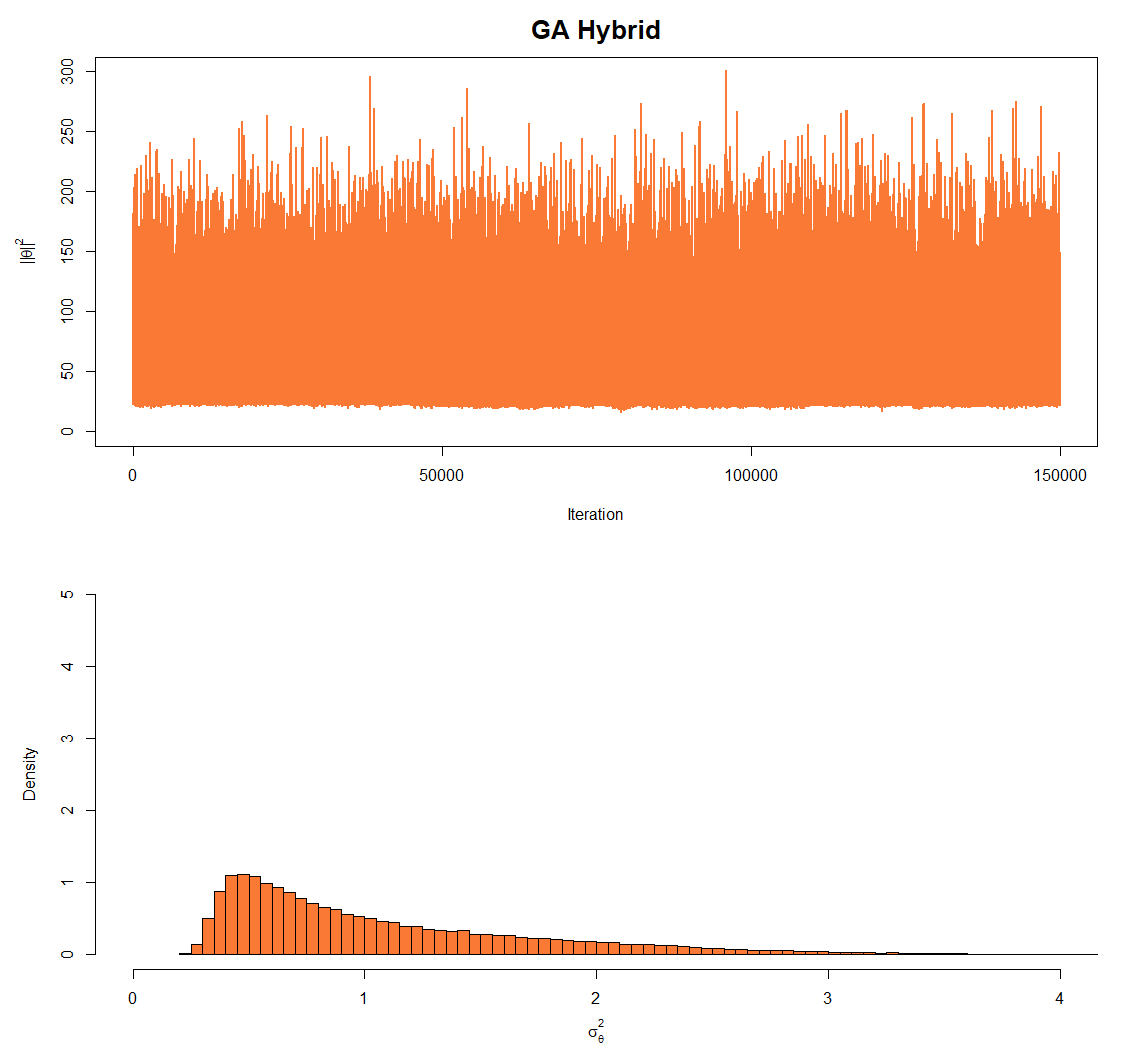}

\caption{$ \lVert \boldsymbol{\theta}^{(j)} \rVert^2$ for $j  = 1, \ldots 150, 000$ (post burn-in) with distribution of marginal of $\sigma_\theta^2 \mid \mathcal{D}$.}
\label{fig: hybrid sigma}
\end{figure}
\fi

\iffoo
\begin{figure}[H]
    \centering
\animategraphics[controls,autoplay,loop,width=0.6\textwidth]{1}{Hybrid/theta2/}{1}{3}
\caption{Distribution of $\hat{\theta}_2$. The top panel uses a common scale across all three optimization methods to facilitate direct comparison, while the bottom panel employs different scales across methods.}
\label{fig: hybrid theta2}
\end{figure}
\else

\begin{figure}[H]
    \centering
\includegraphics[width=0.32\textwidth]{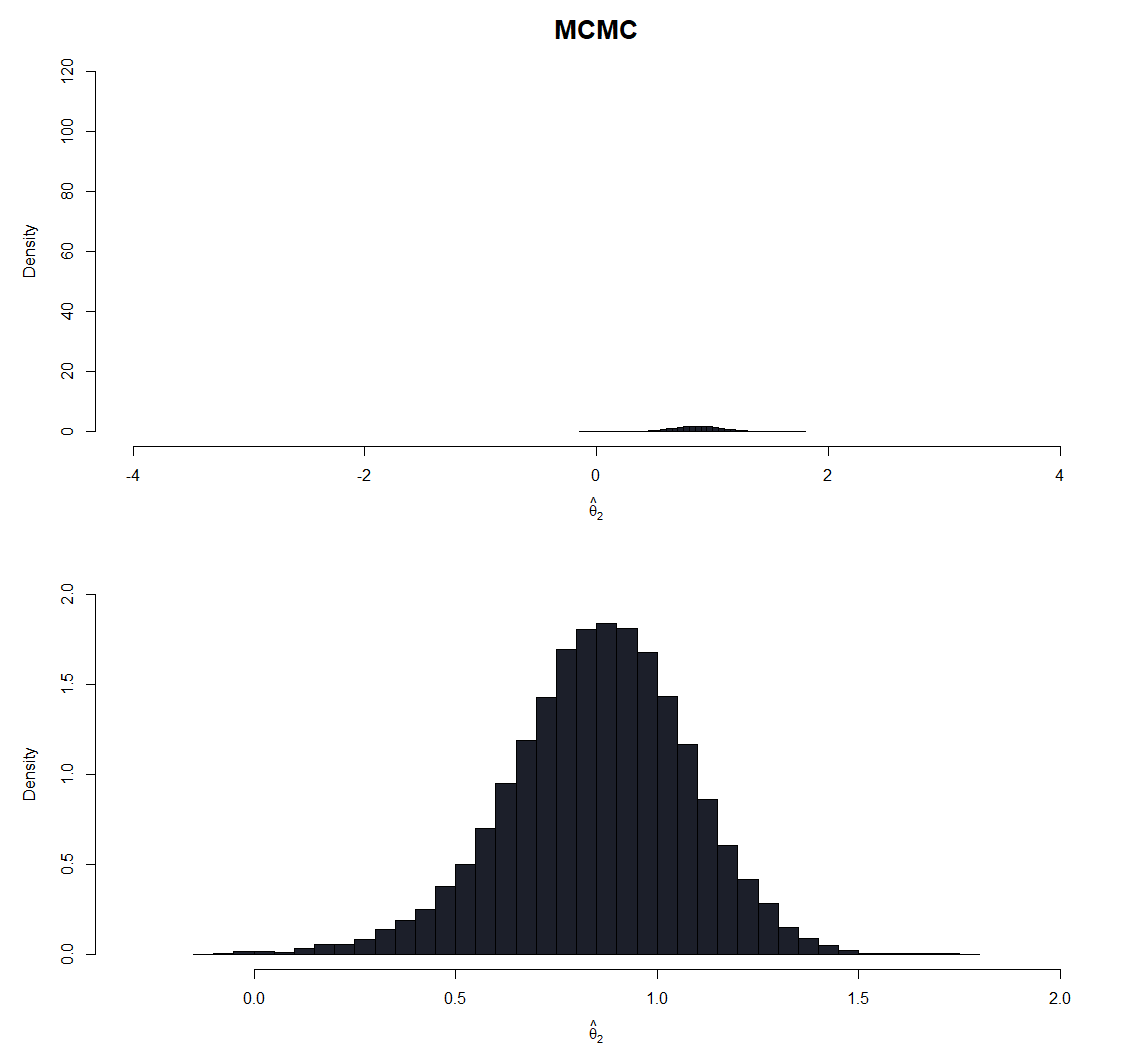}\hfill
\includegraphics[width=0.32\textwidth]{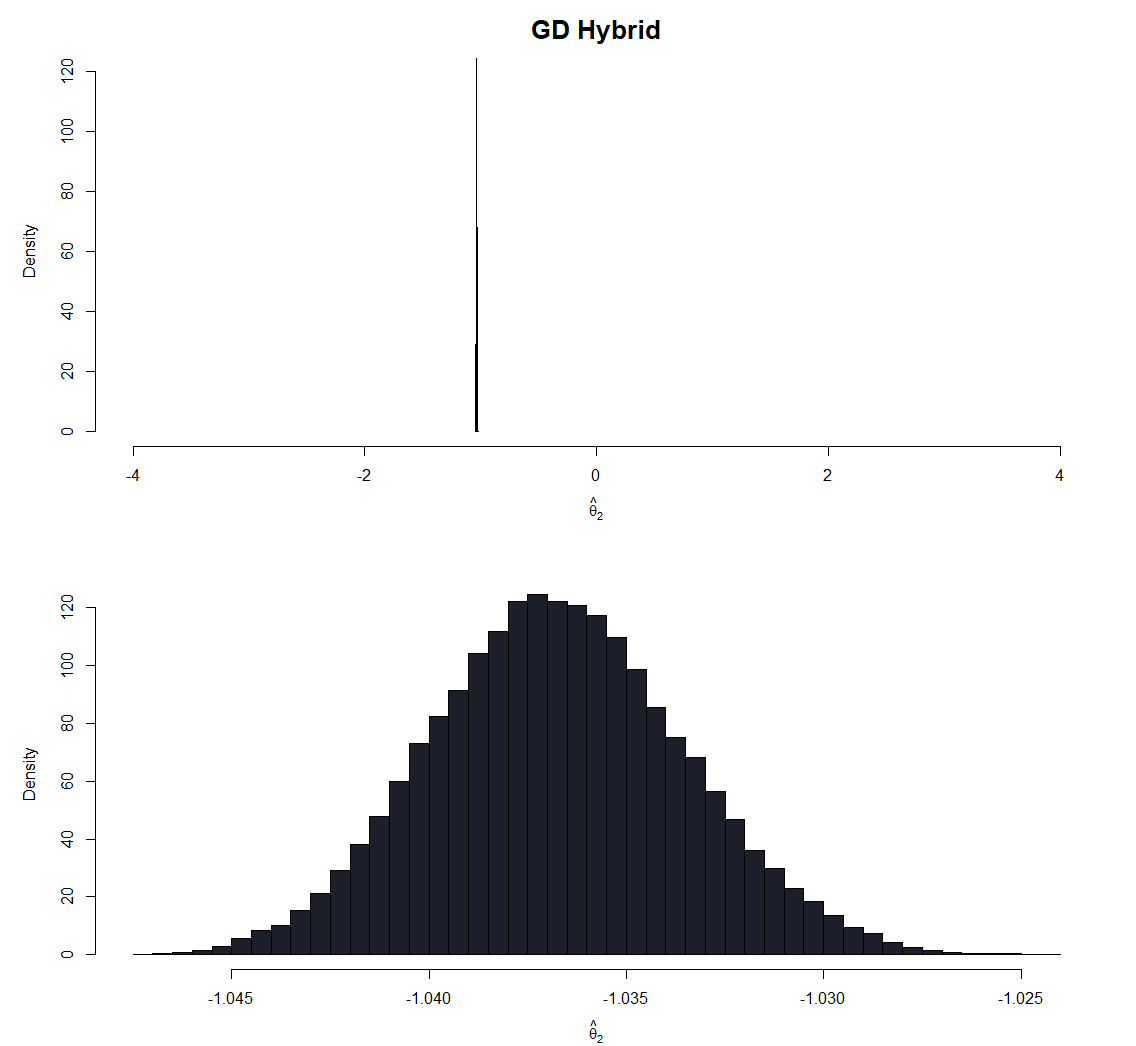}
\includegraphics[width=0.32\textwidth]{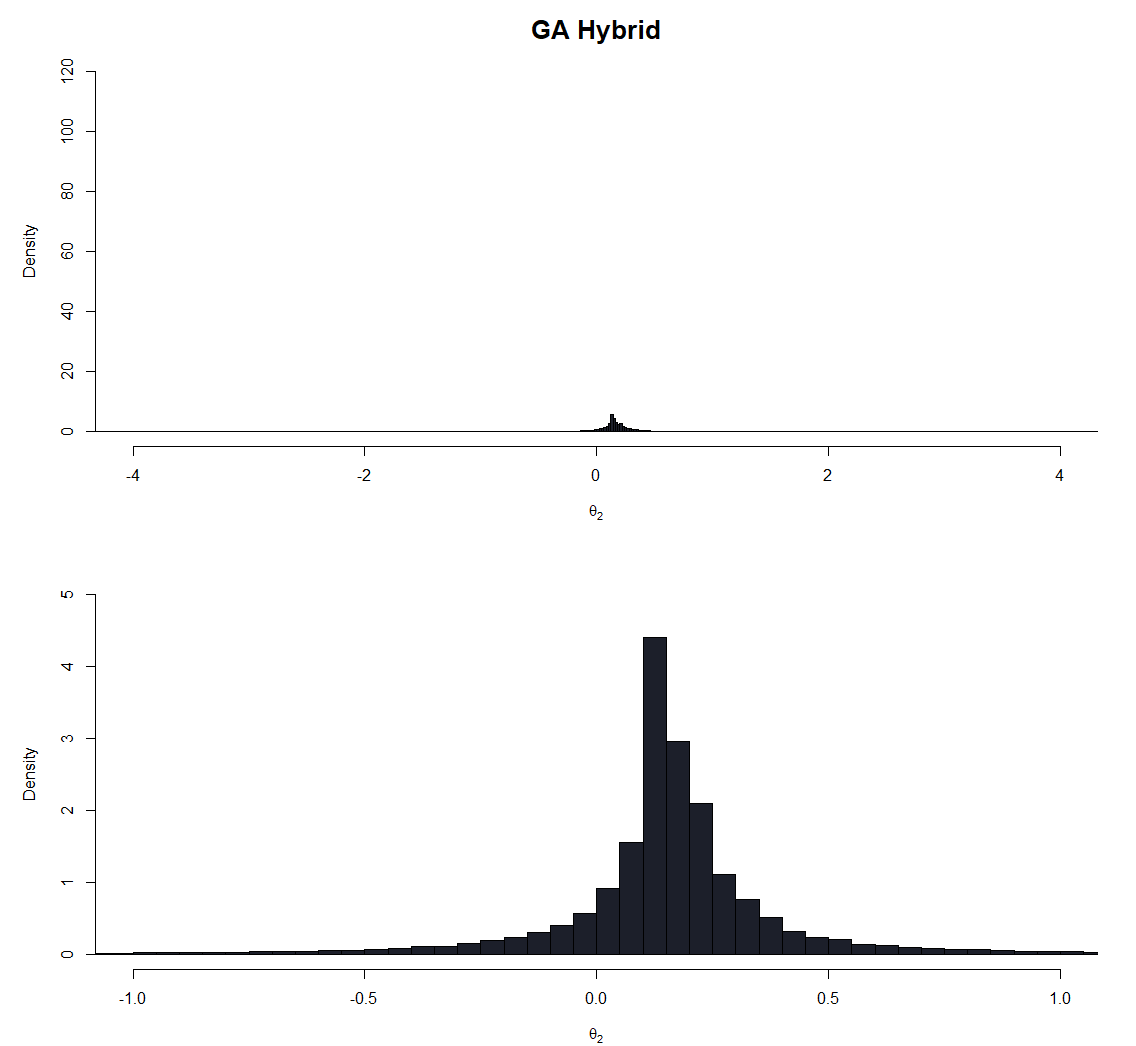}
\caption{Distribution of $\hat{\theta}_2$. The top panel uses a common scale across all three optimization methods to facilitate direct comparison, while the bottom panel employs different scales across methods.}
\label{fig: hybrid theta2}
\end{figure}
\fi

\iffoo
\begin{figure}[H]
    \centering
\animategraphics[controls,autoplay,loop,width=0.6\textwidth]{1}{Hybrid/classes/}{1}{3}
\caption{Response curve for different optimization methods, represented as a cross-sectional heat map of particle classification regions (circles represent the out-of-sample particles).}
\label{fig: hybrid classes} 
\end{figure}
\else

\begin{figure}[H]
    \centering
\includegraphics[width=0.32\textwidth]{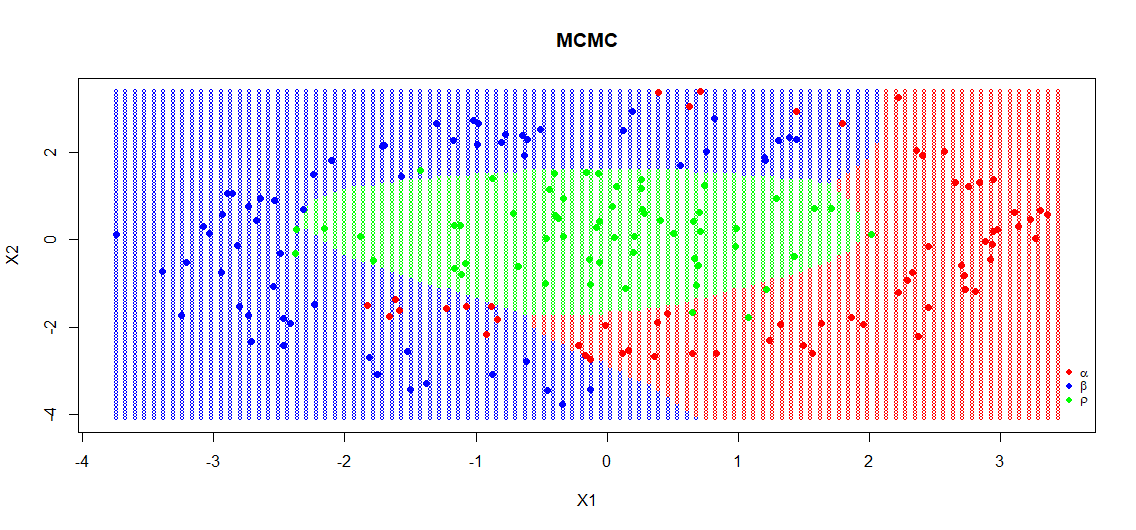}\hfill
\includegraphics[width=0.32\textwidth]{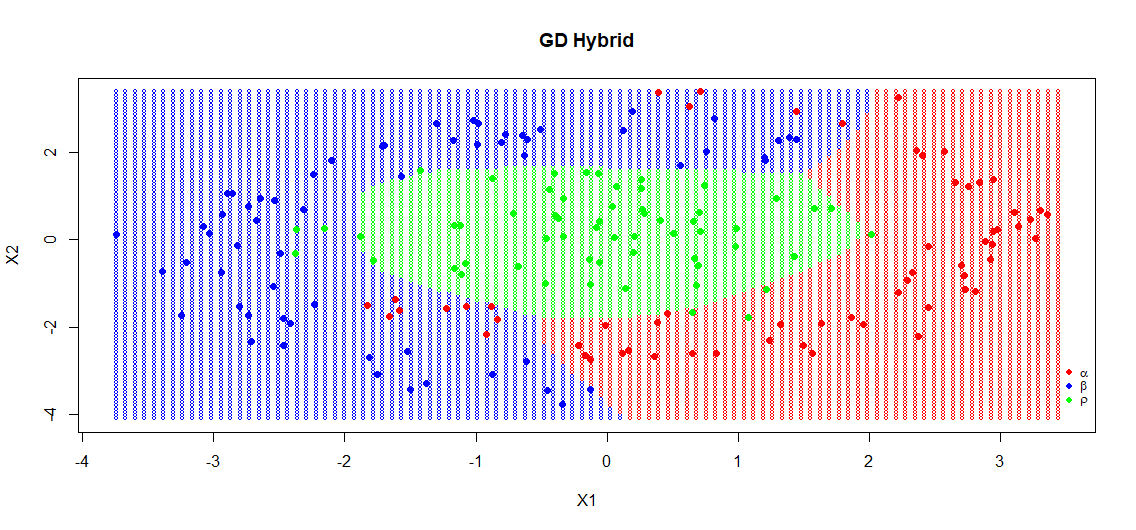}\hfill
\includegraphics[width=0.32\textwidth]{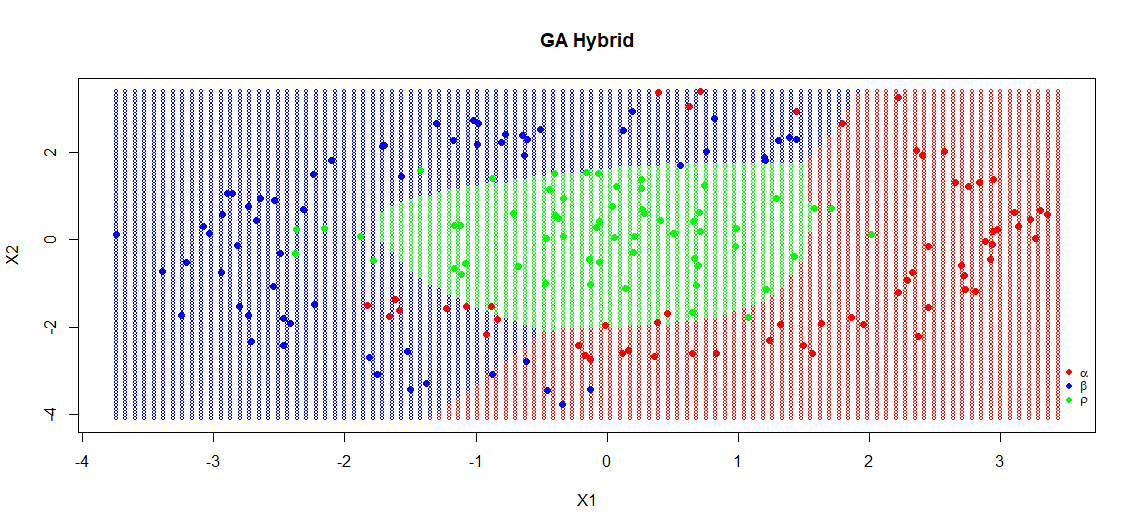}
\caption{Response curves for different optimization methods, represented as a cross-sectional heat map of particle classification regions (circles represent the out-of-sample particles).}
\label{fig: hybrid classes} 
\end{figure}
\fi

\begin{table}[H]
    \centering
    \begin{tabular}{ccccc}
        \hline
        {Optimization Method}  &In-sample & Out-of-sample 
        \\ \hline
        MCMC & $86.67$ & $86.67$   \\ 
        GD Hybrid & $85.55$ & $87.22$ \\ 
        GA Hybrid & 82.78 & $81.11$ \\ 
        \hline
    \end{tabular}
    \caption{Accuracy for in-sample and out-of-sample sets using two-block MCMC and two hybrid optimization techniques.}
    \label{table: acc hybrid}
\end{table}

It is important to recall that such hybrid optimization methods are comparable to that of two-block MCMC, only in settings where the conditional posterior $p(\boldsymbol{\theta} \mid \sigma_\theta^2, \mathcal{D})$ is highly peaked. The similarity in results across the three methods in this case may therefore indicate that, for the data under consideration, the conditional posterior is indeed sharply concentrated. However, this observation is likely data-specific and should not be taken as evidence that such comparability will generalize across all datasets. In the following section, we intentionally increase the sharpness of the likelihood function, and by extension the conditional posterior, to investigate the behaviour of the optimization methods under a sharpened conditional.
\subsection{Hybrid for Blackjack Problem I, II and III}
This section primarily examines the similarities and differences between solutions derived from the two-block MCMC framework (as outlined in Section \ref{sec: 2-sample}) and the GA Hybrid method previously introduced, when applied to Blackjack Problems I and II (using $\overset{(\text{II})}{\mathbf{model}}$ for II). In addition, we introduce a third variant, referred to as Blackjack Problem III, in which both the decision-making and bet-sizing parameter sets are embedded into a unified parameter vector: $
\boldsymbol{\theta} = \left[ {\boldsymbol{\theta}^{\text{Bet}}}', \, {\boldsymbol{\theta}^{\text{Decision}}}' \right] \in \mathbb{R}^{S+R}$. In this formulation, $\boldsymbol{\theta}$ contains the weights and biases governing two distinct neural networks: one dedicated to decision-making and the other to bet-sizing (noting we use $\overset{(\text{II})}{\mathbf{model}}$ for the bet-sizing network). The overarching objective for Blackjack Problem III remains unchanged, namely, to maximize the ROI achieved at the conclusion of the training night, such that:  
\begin{align}
    \operatorname*{argmax}_{\boldsymbol{\theta}} \text{Obj}(\boldsymbol{\theta})  & = \operatorname*{argmax}_{\boldsymbol{\theta}} \frac{\sum_{k = 1}^K\left(s_k\left(\boldsymbol{\theta}\right) \cdot \tilde{bet_k} (\boldsymbol{\theta}) \right)} {\sum_{k = 1}^K 
 \tilde{bet_t} (\boldsymbol{\theta})} \nonumber \\
 & = \operatorname*{argmax}_{\boldsymbol{\theta}}  \text{ROI}\left(\boldsymbol{\theta} \right) .\nonumber
\end{align} 
We employ the exponential likelihood across all three blackjack problems, albeit with differing sharpness parameters, $\beta$, selected to yield the most favourable MCMC convergence. For fairness in comparison, the corresponding GA Hybrid method implementations employ identical $\beta$ values. Accordingly, the pseudo-likelihood is given by $p(\mathcal{D} \mid  \boldsymbol{\theta}) = \exp\!\left(\beta \cdot \text{ROI}(\boldsymbol{\theta})\right)$. Now, given that our objective function is arbitrary, we restrict our analysis to the GA Hybrid method described in Section \ref{sec: GA Hybrid}, and do not employ the GD Hybrid method from Section \ref{sec: GD Hybrid}, as the objective function in question is non-differentiable with respect to $\boldsymbol{\theta}$. \\

Table \ref{table: hybrid bj} presents the results of the two-block MCMC optimization method alongside its corresponding GA Hybrid method for each of the three blackjack problems. In the case of Blackjack Problem I, both methods favour the stand action for all soft and hard totals, and never opt to surrender or split pairs, which explains why we obtain identical in- and out- of sample performance. \\

For Blackjack Problem II, we observe near-identical in-sample and out-of-sample performance. As illustrated in Figure \ref{fig: Hybrid hists II}, this similarity appears to stem from both methods producing agents with comparable betting behaviour: specifically, agents that concentrate their bet-size to approximately $6$ to $8$ units per night.\\

For Blackjack Problem III, two-block MCMC produces an agent that opts to stand on all hard totals but hit on all soft totals, whereas the Hybrid method selects to stand for both hard and soft totals. In both cases, neither method chooses to surrender or split pairs. Additionally, both methods yield betting agents whose bet-sizes are concentrated to approximately $3$ to $5$ units per night (as illustrated in Figure \ref{fig: Hybrid hists III}). Despite these differences, both methods produce solutions which achieve comparable in-sample ROI \%, suggesting the possibility that multiple combinations of decision-making strategies and bet-sizing schemes can lead to similar in-sample ROI $\%$.

\begin{table}[H]
    \centering
    \begin{tabular}{ccccccc}
       & \multicolumn{3}{c}{MCMC} & \multicolumn{3}{c}{Hybrid} \\
         \cmidrule(lr){2-4}  \cmidrule(lr){5-7} 
        & \multicolumn{1}{c}{In-Sample} & \multicolumn{2}{c}{Out-of-Sample}   & \multicolumn{1}{c}{In-Sample} & \multicolumn{2}{c}{Out-of-Sample} \\
        \hline
        Blackjack Problem &  ROI $\%$& $\mu_{ROI \%}$ & $\sigma_{ROI \%}$ &  ROI $\%$& $\mu_{ROI \%}$ & $\sigma_{ROI \%}$  \\ 
        \hline
        I & $-14.8500$ & $-16.0162$ &$3.0645$ & -14.8500  &$-16.0162$  &$3.0645$ \\
        II & $-3.9011$  & $-0.6765$ & $3.6044$ & -3.9154 & -0.6841 & 3.6069 \\
        III &-11.9441 & -15.7276 & 3.1629 & -12.1605 & -14.4514 & 2.8351   \\
        \hline
    \end{tabular}
    \caption{ROI in-sample and out-of-sample performance for the three blackjack problems using $\beta = 50$ for Problem I, $\beta = 250$ for Problem II and $\beta = 50$ for Problem III.}
    \label{table: hybrid bj}
\end{table}

\iffoo
\begin{figure}[H]
    \centering
    \begin{minipage}{0.45\textwidth}
        \centering
    \centering
\animategraphics[controls,autoplay,loop,width=0.8\textwidth]{3}{Hybrid_hists/II/mcmc/bet_}{1}{100}
\caption*{Bet-size ($\tilde{bet_k}(\boldsymbol{\hat{\theta}^\text{MCMC, (II)}})$)  distributions per night according to solution derived from MCMC.}
    \end{minipage}
    \hfill
    \begin{minipage}{0.45\textwidth}
        \centering
\animategraphics[controls,autoplay,loop,width=0.8\textwidth]{3}{Hybrid_hists/II/hybrid/bet_}{1}{100}
        \caption*{Bet-size ($\tilde{bet_k}(\boldsymbol{\hat{\theta}^\text{Hybrid, (II)}})$)  distributions per night according to solution derived from GA Hybrid.}
    \end{minipage}
    \hfill
\caption{Bet-size distributions per night (over $100$ nights) for Blackjack Problem II.}
\label{fig: Hybrid hists II}
\end{figure}
\else
\begin{figure}[H]
    \centering

\includegraphics[width=0.32\textwidth]{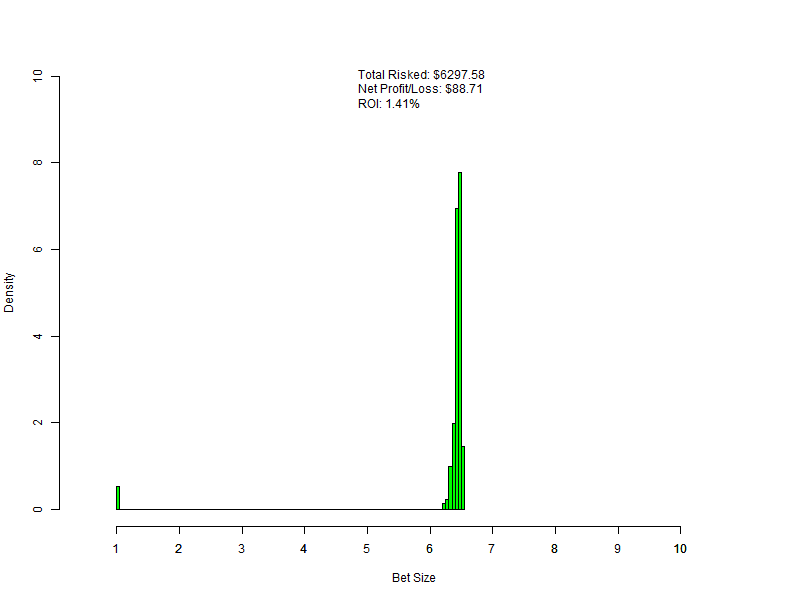}\hfill
\includegraphics[width=0.32\textwidth]{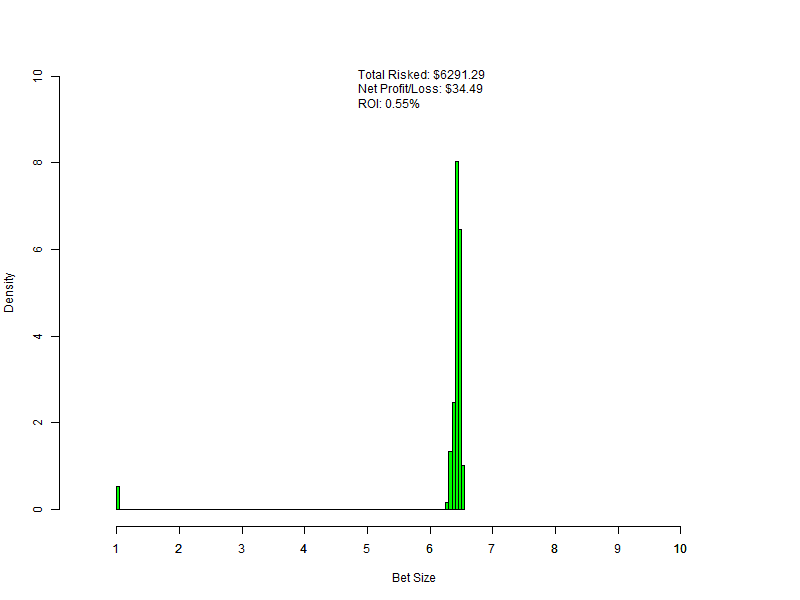}\hfill
\includegraphics[width=0.32\textwidth]{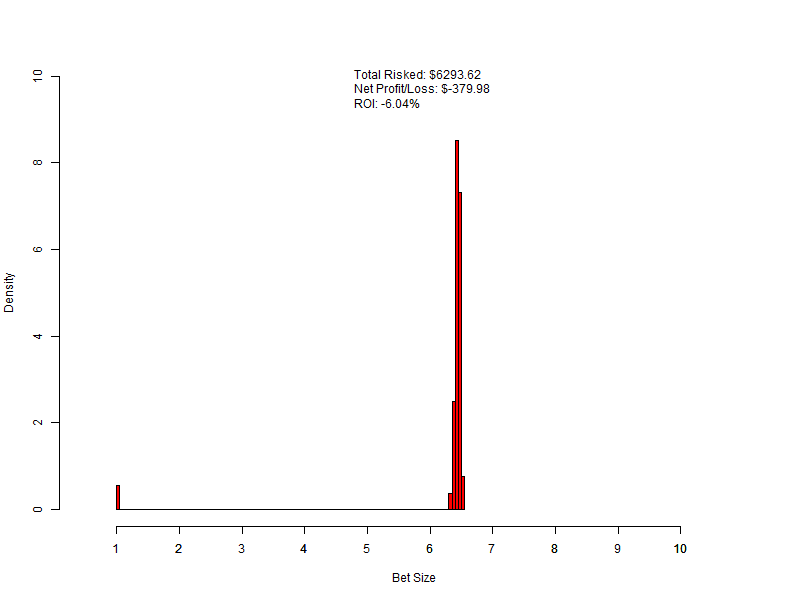}
\vspace{0.1cm}
\includegraphics[width=0.32\textwidth]{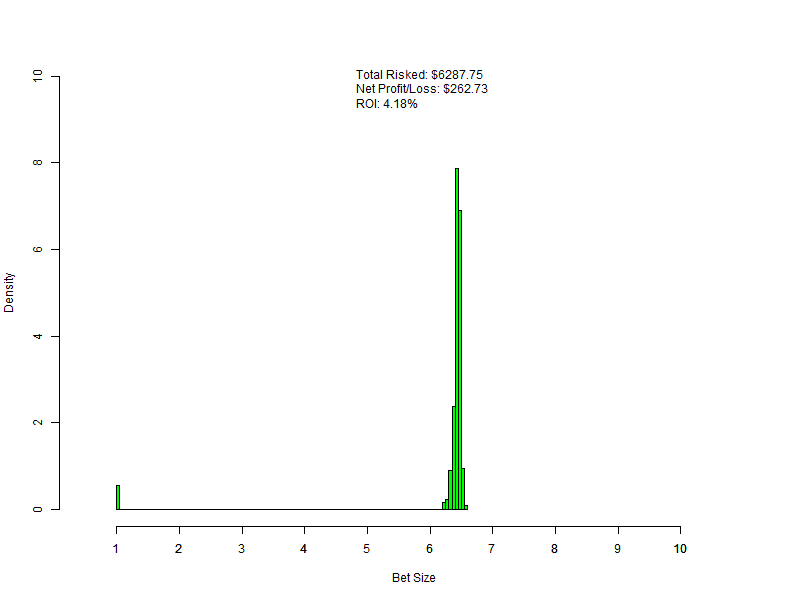}\hfill
\includegraphics[width=0.32\textwidth]{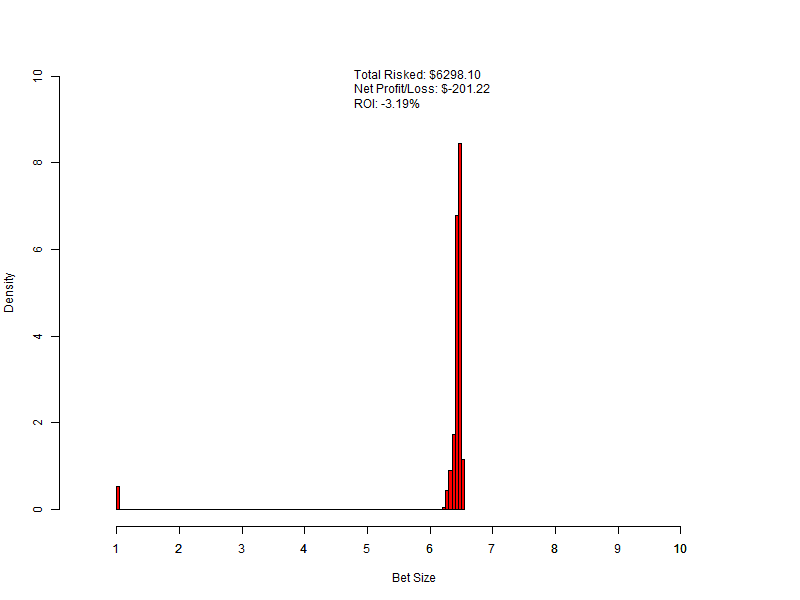}\hfill
\includegraphics[width=0.32\textwidth]{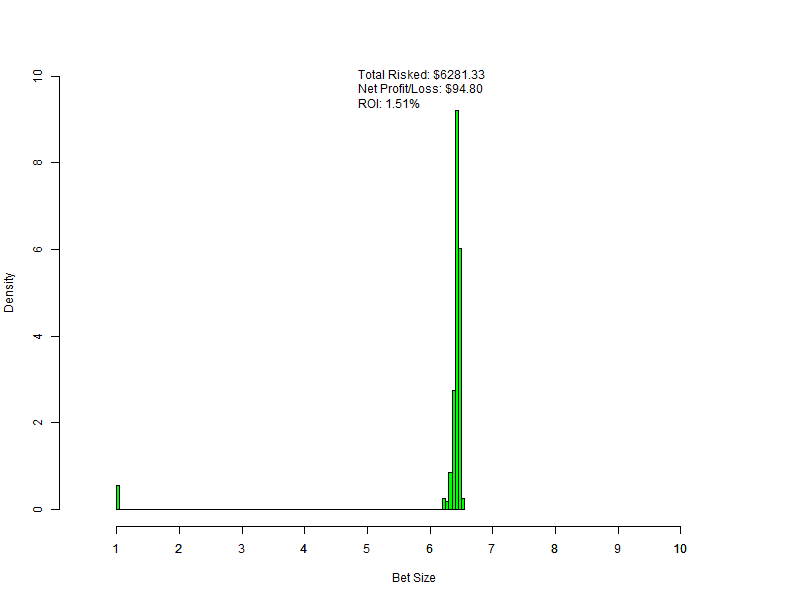}
 \caption*{Bet-size ($\tilde{bet_k}(\boldsymbol{\hat{\theta}^\text{MCMC, (II)}})$)  distributions per night according to solution derived from MCMC.}
\vspace{0.35cm}

\includegraphics[width=0.32\textwidth]{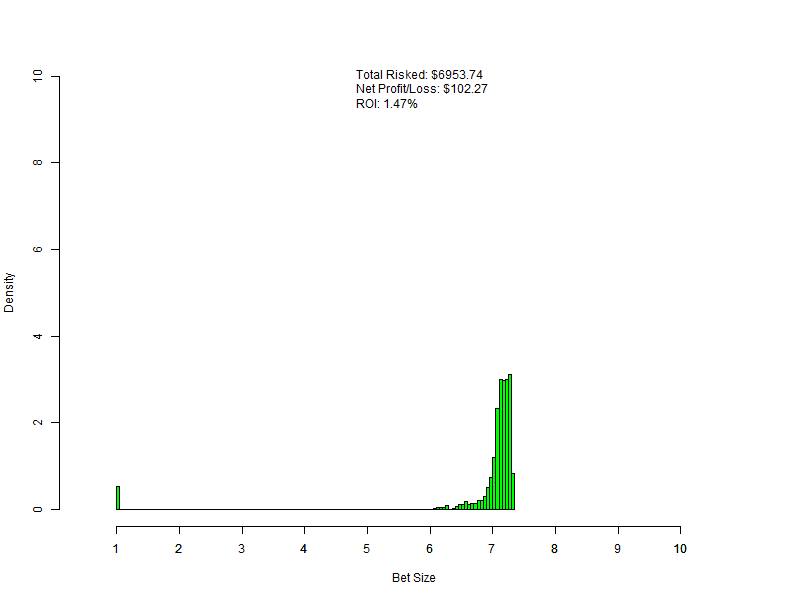}\hfill
\includegraphics[width=0.32\textwidth]{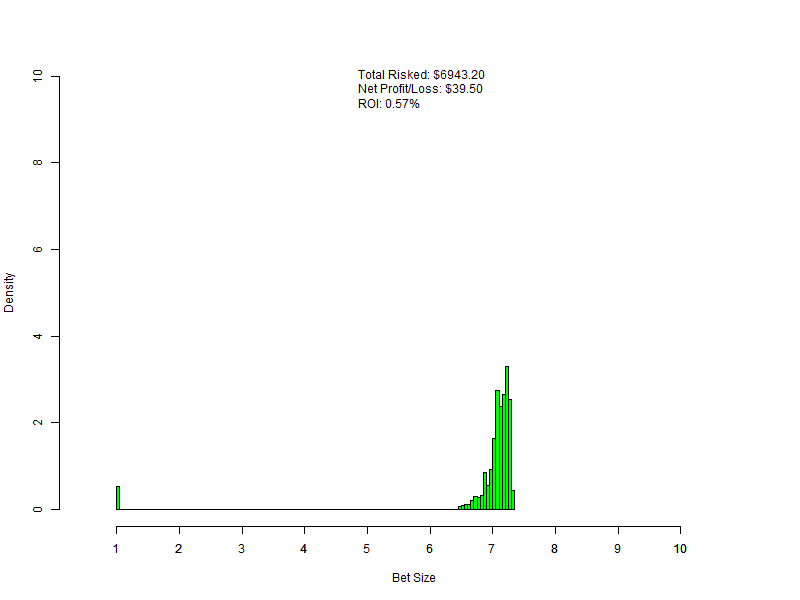}\hfill
\includegraphics[width=0.32\textwidth]{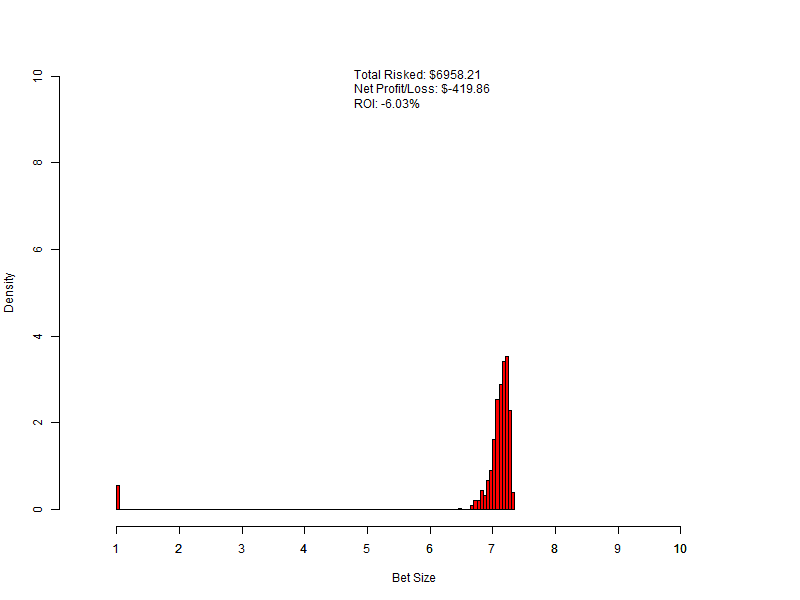}\hfill
\includegraphics[width=0.32\textwidth]{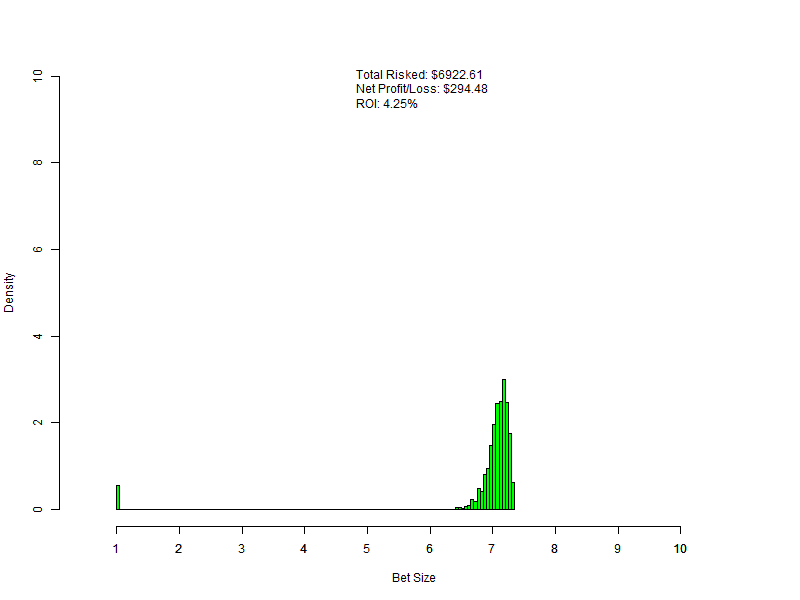}\hfill
\includegraphics[width=0.32\textwidth]{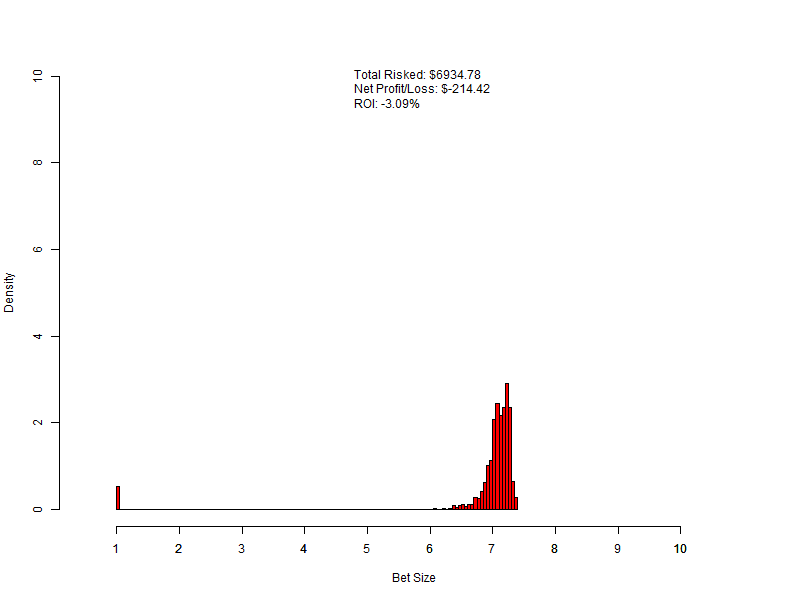}\hfill
\includegraphics[width=0.32\textwidth]{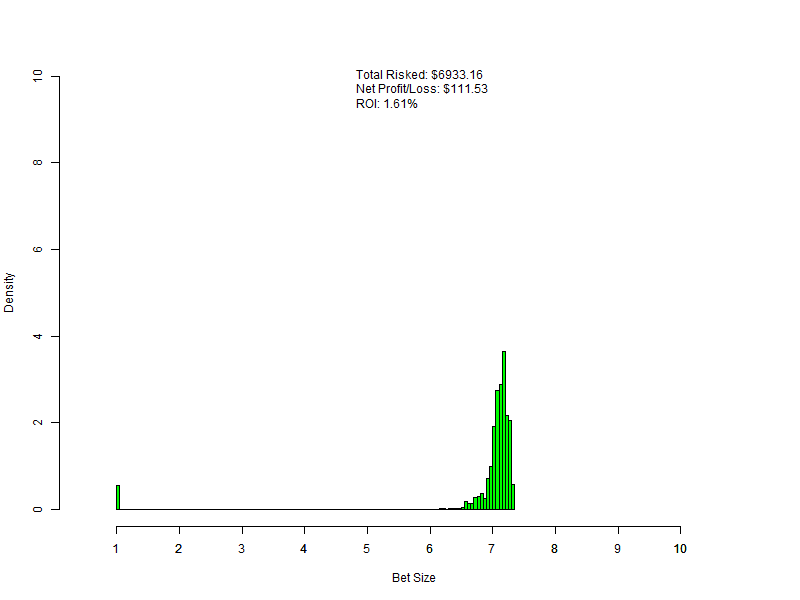}
\caption*{Bet-size ($\tilde{bet_k}(\boldsymbol{\hat{\theta}^\text{Hybrid, (II)}})$)  distributions per night according to solution derived from GA Hybrid.}
\caption{Bet-size distributions per night (over $6$ nights) for Blackjack Problem II.}
\label{fig: Hybrid hists II}
\end{figure}
\fi

\iffoo
\begin{figure}[H]
    \centering
    \begin{minipage}{0.45\textwidth}
        \centering
    \centering
\animategraphics[controls,autoplay,loop,width=0.8\textwidth]{3}{Hybrid_hists/III/mcmc/bet_}{1}{100}
\caption*{Bet-size ($\tilde{bet_k}(\boldsymbol{\hat{\theta}^\text{MCMC}})$)  distributions per night according to solution derived from MCMC.}
    \end{minipage}
    \hfill
    \begin{minipage}{0.45\textwidth}
        \centering
\animategraphics[controls,autoplay,loop,width=0.8\textwidth]{3}{Hybrid_hists/III/hybrid/bet_}{1}{100}
        \caption*{Bet-size ($\tilde{bet_k}(\boldsymbol{\hat{\theta}^\text{Hybrid}})$)  distributions per night according to solution derived from GA Hybrid.}
    \end{minipage}
    \hfill
\caption{Bet-size distributions per night (over $100$ nights) for Blackjack Problem III.}
\label{fig: Hybrid hists III}
\end{figure}
\else

\begin{figure}[H]
    \centering
\includegraphics[width=0.28\textwidth]{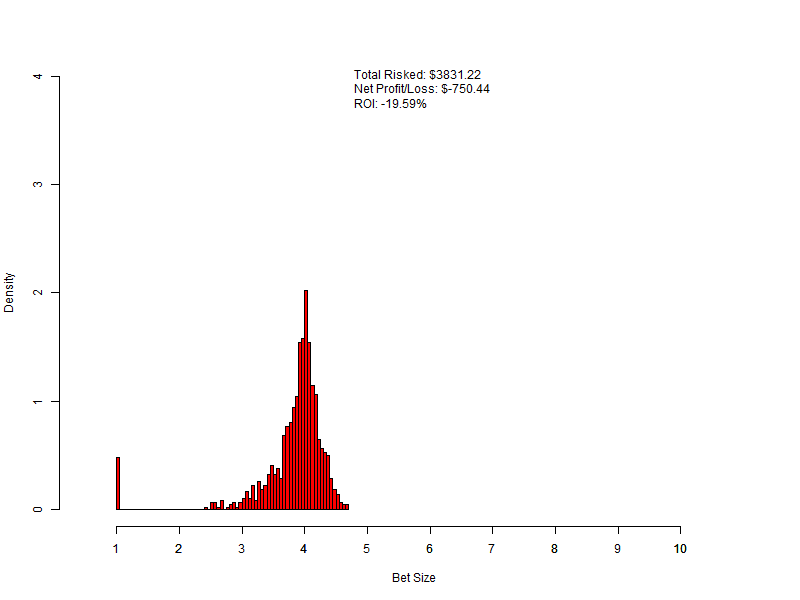}\hfill
\includegraphics[width=0.28\textwidth]{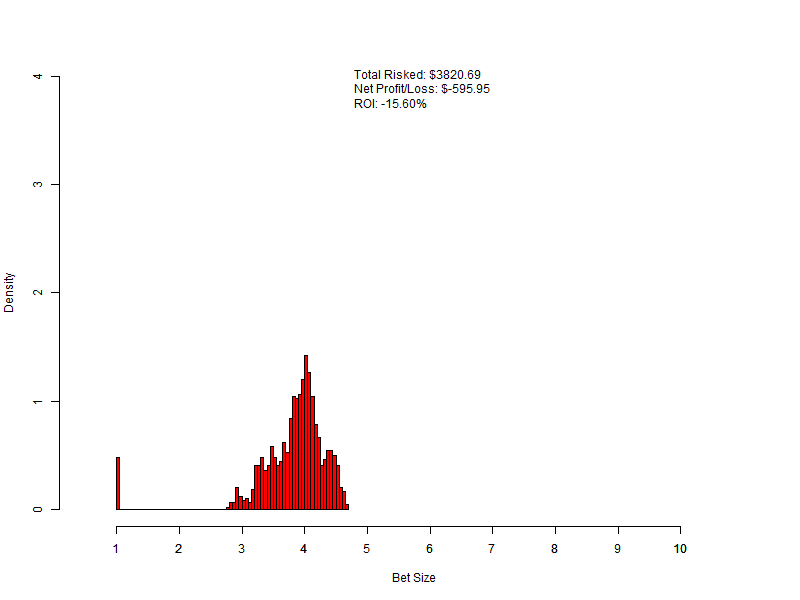}\hfill
\includegraphics[width=0.28\textwidth]{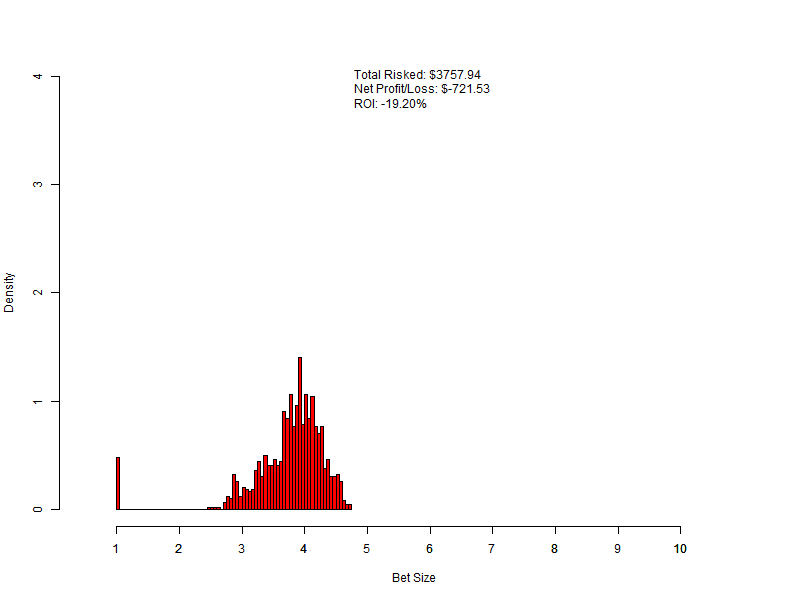}\hfill
\includegraphics[width=0.28\textwidth]{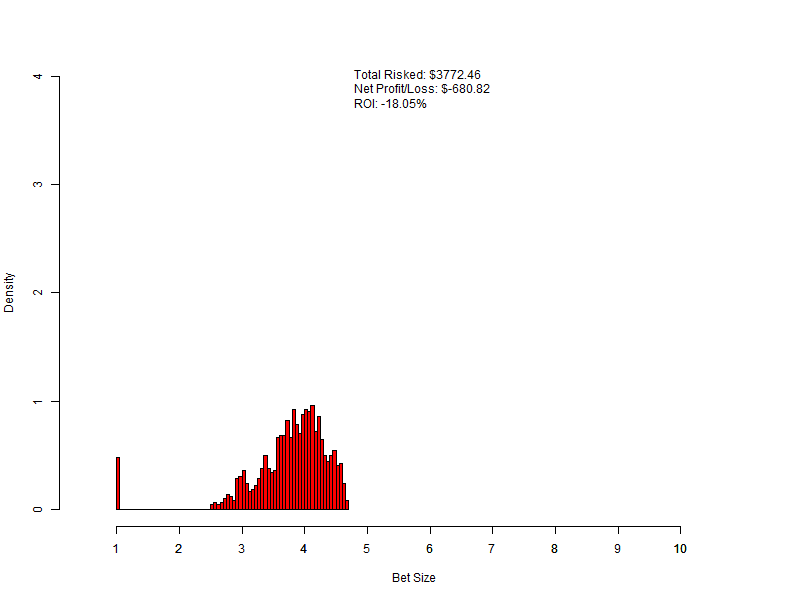}\hfill
\includegraphics[width=0.28\textwidth]{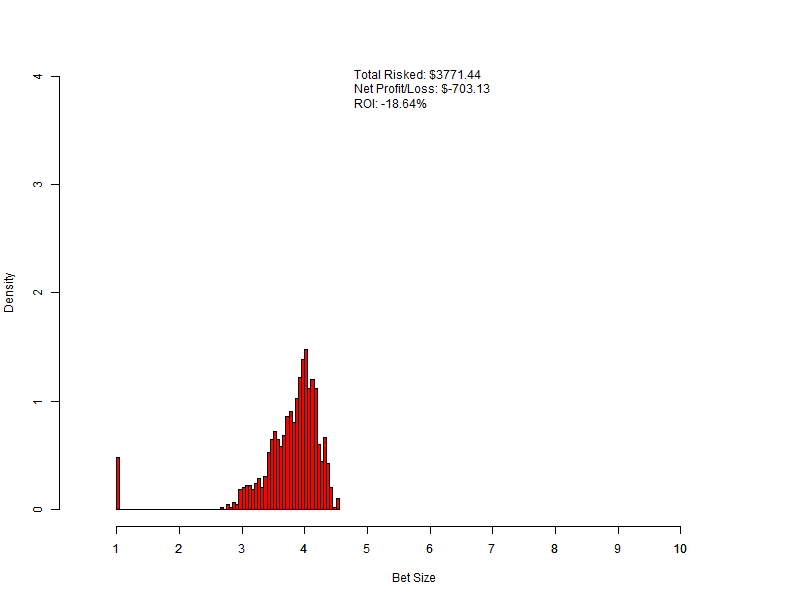}\hfill
\includegraphics[width=0.28\textwidth]{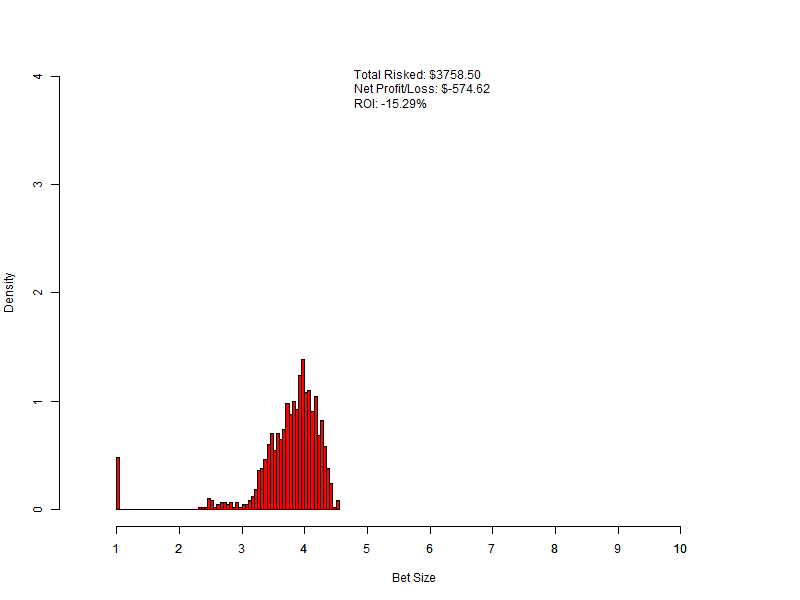}
\caption*{Bet-size ($\tilde{bet_k}(\boldsymbol{\hat{\theta}^\text{MCMC}})$)  distributions per night according to solution derived from MCMC.}

\vspace{0.1cm}
\includegraphics[width=0.28\textwidth]{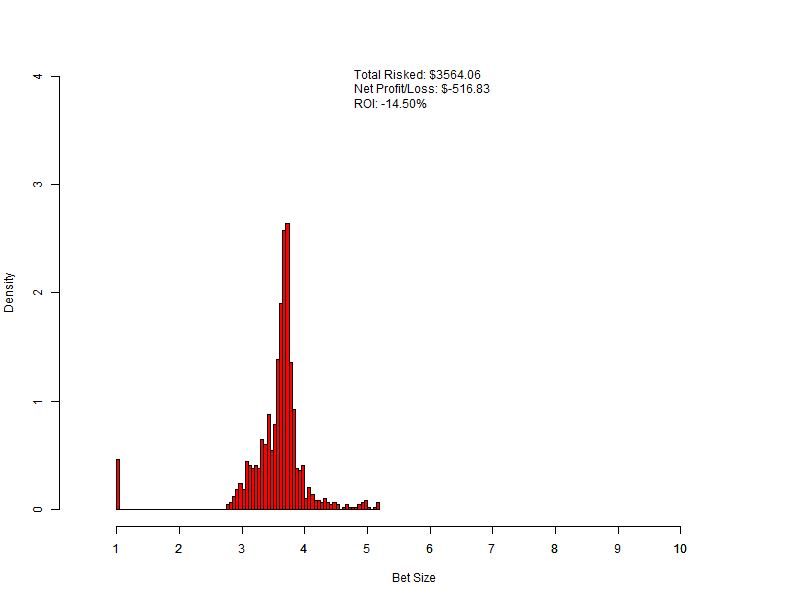}\hfill
\includegraphics[width=0.28\textwidth]{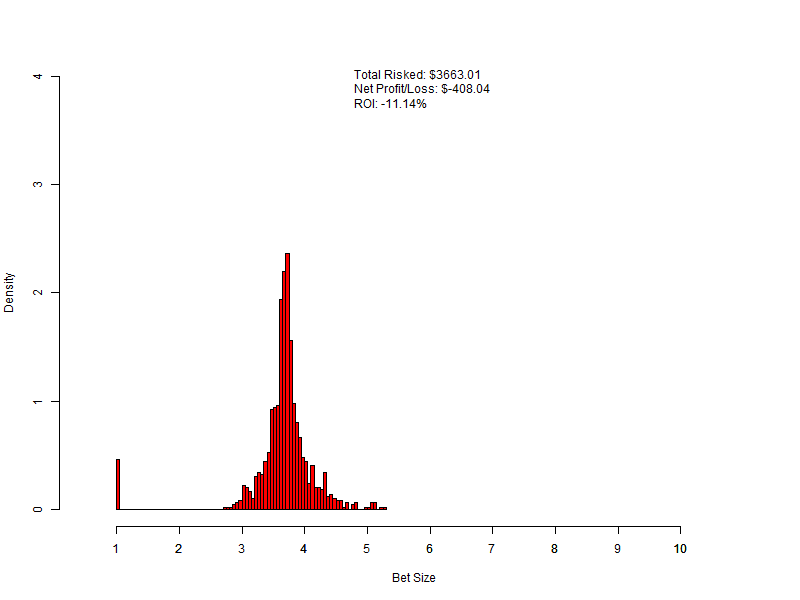}\hfill
\includegraphics[width=0.28\textwidth]{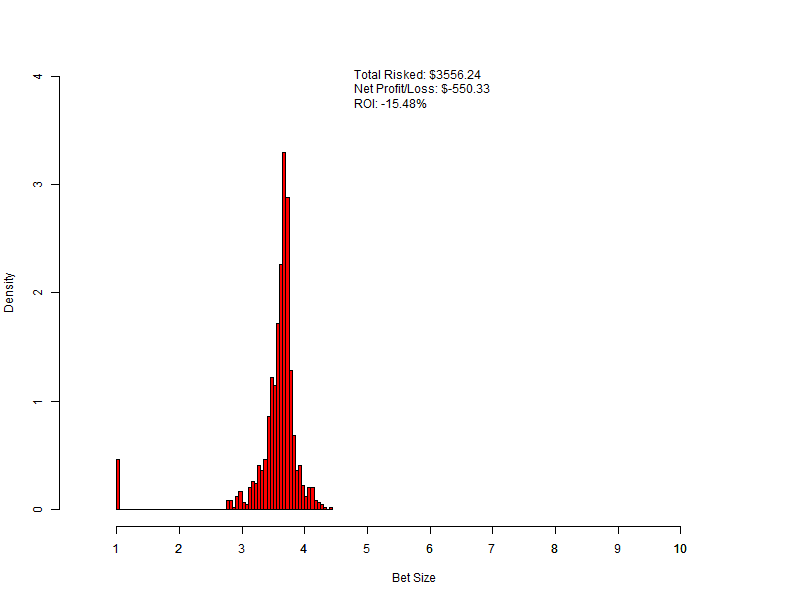}\hfill
\includegraphics[width=0.28\textwidth]{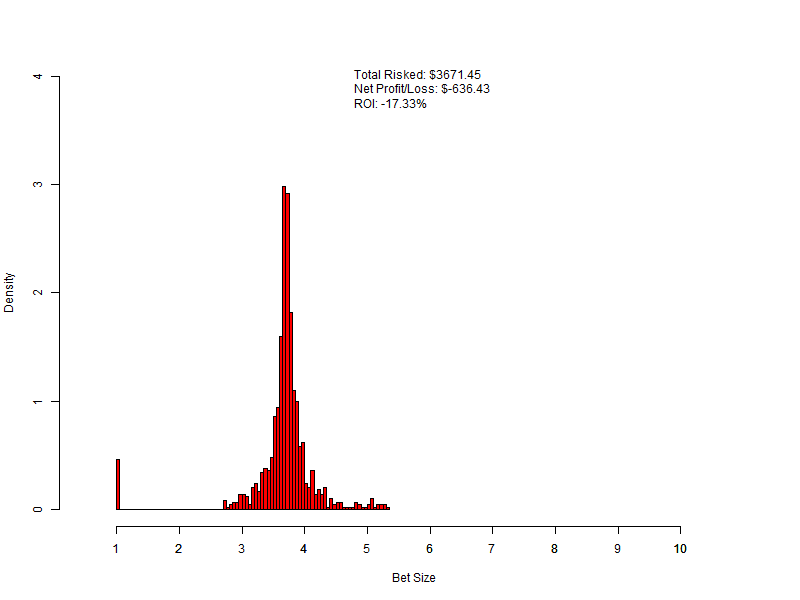}\hfill
\includegraphics[width=0.28\textwidth]{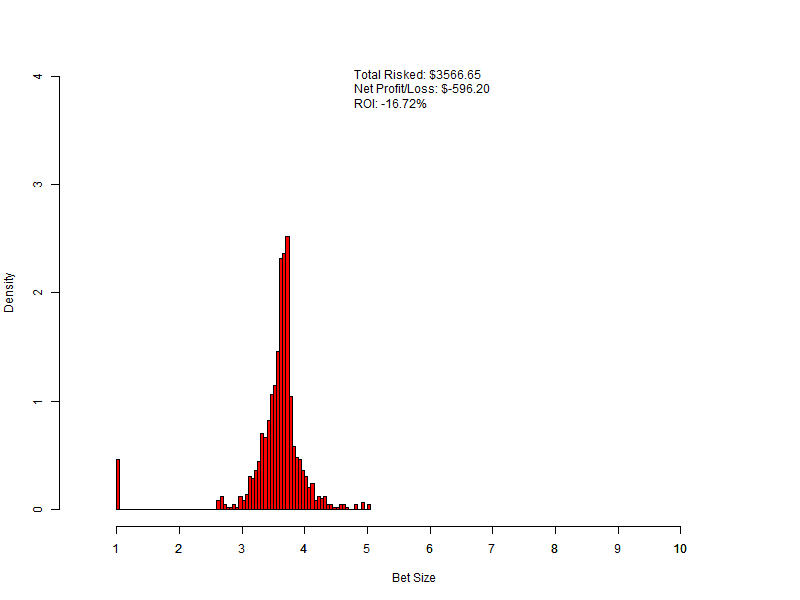}\hfill
\includegraphics[width=0.28\textwidth]{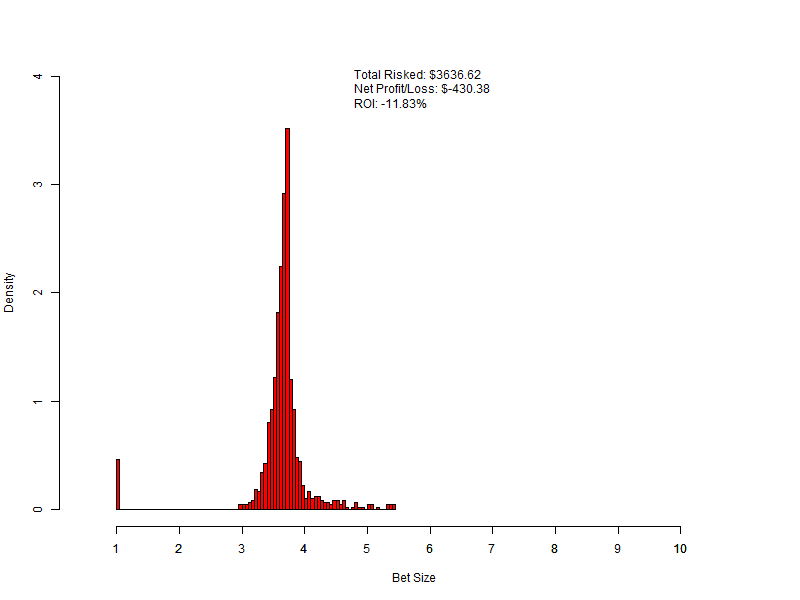}
 \caption*{Bet-size ($\tilde{bet_k}(\boldsymbol{\hat{\theta}^\text{Hybrid}})$)  distributions per night according to solution derived from GA Hybrid.}

\caption{Bet-size distributions per night (over $6$ nights) for Blackjack Problem III.}
\label{fig: Hybrid hists III}
\end{figure}
\fi

Furthermore, this section illustrates that, when the conditional posterior $p( \boldsymbol{\theta} \mid  \sigma_\theta^2, \mathcal{D})$ is sufficiently sharp (achieved via an increased $\beta$), both the two-block MCMC framework and the Hybrid method tend to optimize solutions which converge toward the same dominant mode of the conditional posterior (recalling from earlier that increasing $\beta$ suppresses minor modes and accentuates dominant modes). This is consistent to what was articulated by \cite{kirkpatrick1983optimization} with regards to SA; where a decreased temperature parameter (inverse to $\beta$) concentrates samples around global minima of the cost function. This results in comparable in-sample performance between the aforementioned methods. Such evidence supports the earlier postulate that an iterative optimization procedure could serve as a viable replacement for the MH sampling in Block $1$ of the two-block MCMC framework in Section \ref{sec: 2-sample}, provided the posterior is sufficiently sharp.

\section{Conclusion}
The study aimed to illustrate the shortcomings of two-block MCMC, which is often employed to allow the training data to infer a level of regularization by incorporating the sampling of the dispersion parameter, $\sigma_\theta^2$, into the algorithm. The study showed that the pseudo-likelihood form, the likelihood sharpness parameter $\beta$, and the initial dispersion $\sigma_{\text{Init}}^2$ are in fact user-specified hyperparameters that exert a substantial influence on the degree of regularization inferred. As such, the use of a Bayesian hierarchical model in this context does not genuinely infer regularization from the training data; rather, it is the \emph{user} who determines the effective strength of regularization, albeit with additional steps.\\

Furthermore, the study demonstrated that if one were to increase likelihood sharpness to an extreme, one may effectively reduce the two-block MCMC to a hybrid approach in which the first block is replaced by an iterative optimization procedure, yielding nearly identical in-sample performance to the original scheme (as solutions from both methods converge toward the same dominant mode of the conditional posterior). In this sense, the sampling of the dispersion parameter at each iteration functions primarily as a mechanism to add exploration to the search process, rather than as a means to infer regularization.\\

Taken together, the results of this study directly address a gap implicit in the existing literature on MCMC-based optimization. While prior work has cautioned that applying MCMC to arbitrary objectives blurs the distinction between posterior inference and stochastic optimization, these concerns have largely remained conceptual. By explicitly examining two-block MCMC in settings where the likelihood is deliberately constructed to be proportional to the arbitrary objective function, this study provides a concrete empirical demonstration of those concerns. In particular, it shows that claims of ``data-driven'' regularization in such frameworks are fragile, and that the effective strength of regularization is instead governed by user-specified choices, most notably likelihood sharpness and initial dispersion. In doing so, the study clarifies that two-block MCMC in arbitrary objective settings should generally be interpreted as a form of stochastic, temperature-controlled optimization rather than Bayesian posterior inference, thereby making explicit what has previously been only implicitly acknowledged in the literature.\\

Future work should explore treating the sharpness parameter in analogy to SA, whereby $\beta$ is increased according to a cooling schedule within the two-block MCMC framework, rather than being fixed \textit{a priori}. As expounded upon Section \ref{sec: mcmc for optimization}, parallel tempering, introduced by \cite{geyer1991markov}, operates by running multiple Markov chains in parallel at different temperatures, each targeting a tempered version of the objective-induced distribution. While this may appear to offer a natural extension of two-block MCMC, this study's results indicate that running a Markov Chain at different temperatures fundamentally alters the degree of regularization inferred by the sampler. Consequently, the direct application of tempering-based strategies to two-block MCMC is unlikely to yield meaningful results.\\

A worthwhile avenue for future research, however, would be to explore simulated tempering, introduced by \cite{marinari1992simulated} (see Section~\ref{sec: mcmc for optimization} of the current study), which treats the temperature, and hence the sharpness parameter $\beta$, as a stochastic variable rather than a fixed tuning parameter. Additionally, and more generally, one could investigate how pre-specified temperature schedules, as formalized in annealed MCMC schemes by \cite{geyer1995annealing} (again, see Section~\ref{sec: mcmc for optimization}), influence the behavior of two-block MCMC results. Nonetheless, caution is warranted with these approaches: continually increasing sharpness may substantially degrade the mixing properties of the Markov chain, and the mechanisms proposed in this study to mitigate such issues may not directly extend to these settings.

\pagebreak

\bibliographystyle{plainnat}

\bibliography{References}

\pagebreak

\begin{appendices}
\addtocontents{toc}{\protect\setlength{\protect\cftsecnumwidth}{7em}}
\renewcommand{\thesection}{Appendix~\Alph{section}}
\section{Joint Metropolis-Hastings} \label{app:JointMH}
 We define $\boldsymbol{\Lambda} = \left[\boldsymbol{\theta}', {\sigma}_\theta^2 \right]' \in \mathbb{R}^{S+1}$ and re-derive our MH algorithm to include $\sigma_\theta^2$. Hence, given the current state of $\boldsymbol{\Lambda}$, that is $\boldsymbol{\Lambda}^{(j)}$, the MH algorithm proposes a new value $\boldsymbol{\Lambda}^{*}$ obtained from $\boldsymbol{\Lambda}^{*} = \boldsymbol{\Lambda}^{(j)} + \mathbf{Q}$. Subsequently,  $\boldsymbol{\Lambda}^{*}$ is accepted as the new value in the Markov chain under the following acceptance criterion:
\begin{equation}
    \boldsymbol{\Lambda}^{(j+1)} =
    \bigg\{
\begin{array}{@{}rl@{}}
\boldsymbol{\Lambda}^*, & \text{if} \; U < \alpha,\\
\boldsymbol{\Lambda}^{(j)},  & \text{otherwise.} 
\end{array} \nonumber
\end{equation}
Where the vector $\mathbf{Q} =[{\mathbf{Q}_\theta}', Q_{\sigma_\theta^2}]' \in \mathbb{R}^{S+1}$ denotes drawn values from proposal densities: $\mathbf{Q}_\theta \sim \mathcal{N}(\mathbf{0}, \sigma_{Q_\theta}^2 \mathbf{I}_S)$ and ${Q}_{\sigma_\theta^2} \sim \text{Inv-Gamma}(a_Q, b_Q)$. Now, $\alpha$ is given by:
\begin{align}
    \alpha &= \min \left( 
    \frac{p(\boldsymbol{\Lambda}^{*}\vert \mathcal{D})}{p(\boldsymbol{\Lambda}^{(j)}\vert \mathcal{D})} \cdot 
    \frac{Q(\boldsymbol{\Lambda}^{(j)}\vert \boldsymbol{\Lambda}^{*} )}{Q(\boldsymbol{\Lambda}^{*}\vert \boldsymbol{\Lambda}^{(j)})}, 1 
    \right) \nonumber \\
    &= \min \left( 
    \frac{p(\mathcal{D} \vert \boldsymbol{\Lambda}^{*}) p(\boldsymbol{\Lambda}^{*})}{p(\mathcal{D} \vert 
    \boldsymbol{\Lambda}^{(j)}) p(\boldsymbol{\Lambda}^{(j)})} \cdot \frac{Q(\boldsymbol{\theta}^{(j)}, \sigma_{\theta^{(j)}}^2\vert \boldsymbol{\theta}^{*}, \sigma_{\theta^{*}}^2 )}{Q(\boldsymbol{\theta}^{*}, \sigma_{\theta^{*}}^2\vert \boldsymbol{\theta}^{(j)}, \sigma_{\theta^{(j)}}^2)}
 , 1 
    \right) \nonumber \\
    &= \min \left( 
    \frac{p(\mathcal{D} \vert \boldsymbol{\theta}^{*}) p(\boldsymbol{\theta}^{*}, \sigma_{\theta^{*}}^2)}{p(\mathcal{D} \vert 
    \boldsymbol{\theta}^{(j)}) p(\boldsymbol{\theta}^{(j)}, \sigma_{\theta^{(j)}}^2)} \cdot \frac{Q_\theta(\boldsymbol{\theta}^{(j)}\vert \boldsymbol{\theta}^{*}) Q_{\sigma_\theta^2}(\sigma_{\theta^{(j)}}^2\vert \sigma_{\theta^{*}}^2  ) }{Q_\theta(\boldsymbol{\theta}^{*} \vert\boldsymbol{\theta}^{(j)})Q_{\sigma_\theta^2}(\sigma_{\theta^{*}}^2\vert \sigma_{\theta^{(j)}}^2)}
 , 1 
    \right) & \text{Likelihood not dependent on $\sigma_\theta^2$} \nonumber \\
     & = \min \left( 
    \frac{p(\mathcal{D} \vert \boldsymbol{\theta}^{*}) p(\boldsymbol{\theta}^{*} \vert \sigma_{\theta^{*}}^2) p(\sigma_{\theta^{*}}^2)}{p(\mathcal{D} \vert 
    \boldsymbol{\theta}^{(j)}) p(\boldsymbol{\theta}^{(j)} \vert\sigma_{\theta^{(j)}}^2)p(\sigma_{\theta^{(j)}}^2)} \cdot \frac{Q_{\sigma_\theta^2}(\sigma_{\theta^{(j)}}^2\vert \sigma_{\theta^{*}}^2  ) }{Q_{\sigma_\theta^2}(\sigma_{\theta^{*}}^2\vert \sigma_{\theta^{(j)}}^2)} 
     , 1 
    \right). & \text{Symmetry of ${Q}_\theta$} \label{Eq: alpha hyperprior} 
\end{align}
Assuming ${\sigma}_\theta^2 \sim \text{Inv-Gamma}\left(a,b \right)$ and still assuming $\boldsymbol{\theta}\vert{\sigma}_\theta^2 \sim \mathcal{N}(\mathbf{0}, \sigma_\theta^2 \mathbf{I}_S)$, Equation \ref{Eq: alpha hyperprior} simplifies to:
\begin{align}
    \alpha &= \min \left( \frac{h\left( k(\boldsymbol{\theta}^*) \right) \cdot \frac{1}{\sqrt{\left(2\pi\sigma_{\theta^*}^2 \right)^S}} \exp \left( -\frac{1}{2\sigma_{\theta^*}^2}  \lVert \boldsymbol{\theta}^* \rVert^2  \right) \cdot \frac{1}{\left(\sigma_{\theta^{*}}^{2}\right)^{a+1}} \exp{\left(-\frac{b}{\sigma_{\theta^{*}}^2} \right)}  \cdot \frac{1}{\left({\sigma}_{\theta^{(j)}}^{2}\right)^{a_Q+1}} \exp{\left(-\frac{b_Q}{\sigma_{\theta^{(j)} }^2} \right)} }{h\left( k(\boldsymbol{\theta}^{(j)}) \right) \cdot \frac{1}{\sqrt{\left(2\pi\sigma_{\theta^{(j)}}^2 \right)^S}} \exp \left( -\frac{1}{2\sigma_{\theta^{(j)}}^2}  \lVert \boldsymbol{\theta}^{(j)} \rVert^2  \right) \cdot \frac{1}{\left({\sigma}_{\theta^{(j)}}^{2}\right)^{a+1}} \exp{\left(-\frac{b}{\sigma_{\theta^{(j)} }^2} \right)} \cdot \frac{1}{\left(\sigma_{\theta^{*}}^{2}\right)^{a_Q+1}} \exp{\left(-\frac{b_Q}{\sigma_{\theta^{*}}^2} \right)} }
     , 1 \right), \nonumber
\end{align}
and by taking the $\log$:
\begin{align}
    \log \left(\alpha\right) &= \min \Bigg( \log \left( h(k(\boldsymbol{\theta}^*)) \right) - \frac{S}{2} \log (2\pi \sigma_{\theta^*}^2) - \frac{\|\boldsymbol{\theta}^*\|^2}{2\sigma_{\theta^*}^2} - (a+1) \log \left( \sigma_{\theta^*}^2\right) - \frac{b}{\sigma_{\theta^*}^2}  \nonumber \\
    &\quad - (a_Q+1) \log \left( \sigma_{\theta^{(j)}}^2 \right) - \frac{b_Q}{\sigma_{\theta^{(j)}}^2} - \log \left( h(k(\boldsymbol{\theta}^{(j)})) \right) + \frac{S}{2} \log (2\pi \sigma_{\theta^{(j)}}^2) \nonumber \\
    &\quad + \frac{\|\boldsymbol{\theta}^{(j)}\|^2}{2\sigma_{\theta^{(j)}}^2} + (a+1) \log \left( \sigma_{\theta^{(j)}}^2 \right) + \frac{b}{\sigma_{\theta^{(j)}}^2} + (a_Q+1) \log \left( \sigma_{\theta^*}^2\right) + \frac{b_Q}{\sigma_{\theta^*}^2}, 0 \Bigg). \nonumber
\end{align}
\section{Neural Network Architecture} \label{app: NN}
We define the feed-forward recursive relation in scalar form, hence the $j^{th}$ node on the $l^{th}$ layer for the $i^{th}$ observation is given as:
\begin{align}
    a_j^l(i) = \sigma_l\left(\sum_{k =1}^{d_{l-1}}a_k^{l-1}(i)w_{kj}^l + b_j^l\right), \nonumber
\end{align}
for $l = 1, 2, \ldots L; j = 1, 2, \ldots d_l; i = 1, 2\ldots N$. Now, $\sigma_l(.)$ denotes the activation function on layer $l$, $d_{l-1}$ denotes the number of nodes in layer $l-1$, $w_{kj}^l$ denotes the $kj^{th}$ weight linking the $k^{th}$ node in layer $l-1$ and the $j^{th}$ node in layer $l$ with $b_j^l$ denoting the $j^{th}$ bias in layer $l$. The equation is evaluated subject to the initial conditions $a_j^{(0)} = x_{ij}$ for all $j$ at the $i^{th}$ training example.
\section{Genetic Algorithm}
Define $\boldsymbol{\theta}^{(n, m)} \in \mathbb{R}^p$ to be the $n^{th}$ solution (individual) from the $m^{th}$ generation such that $n = \{1, \ldots, N \}$ and $m = \{1, \ldots, M \}$. We present the $N$-size population as $\boldsymbol{\Theta}^{(m)} = \left[ \boldsymbol{\theta}^{(1, m)}, \boldsymbol{\theta}^{(2, m)}, \ldots, \boldsymbol{\theta}^{(N, m)}\right]_{p\times N}$ for the $m^{th}$ generation. We initialize by setting $\boldsymbol{\Theta}^{(0)}$ where $\boldsymbol{\theta}^{(n, 0)} \sim \mathcal{U}_p$ for $n = \{1, \ldots, N \}$. At the termination of the algorithm, we return the solution \(\boldsymbol{\theta}^{(n^\star, m^\star)}\) that achieved the highest objective value across all individuals and generations:
\[
\boldsymbol{\theta}^{(n^\star, m^\star)} = \operatorname*{arg\,max}_{n \in \{1,\ldots,N\},\; m \in \{1,\ldots,M\}} \text{Obj}\left( \boldsymbol{\theta}^{(n, m)} \right).
\]
Hence for generation $m = 1, \ldots, M$:
\subsection*{Fitness}
For $n = 1, \ldots, N$, we compute the fitness for each $n^{th}$ individual of the $m^{th}$ generation as $f_{n, m} = \text{Obj}\left(\boldsymbol{\theta}^{(n, m-1)} \right)$.
\subsection*{Selection (Roulette Wheel)}
\begin{enumerate}
  \item Compute selection probabilities:
  \[
    p_{n, m} = \frac{f_{n, m}}{\sum_{i=1}^N f_{i, m}}, \quad n=1,\ldots,N.
  \]
  
  \item Compute cumulative probabilities:
  \[
    C_{n, m} = \sum_{i=1}^n p_{i, m}, \quad n=1,\ldots,N.
  \]
  
  \item For each selection \(i=1,\ldots,N\):
  \begin{enumerate}
    \item Sample \(r \sim \mathcal{U}(0,1)\).
    \item Find the smallest \(n\) such that \(C_{n, m} \geq r\).
    \item Select parent $\boldsymbol{\theta}^{(n,m)}  = \tilde{\boldsymbol{\theta}}^{(n,m)}$. 
\end{enumerate}
\item To form the mating pool $\boldsymbol{\mathcal{M}}^{(m)} = \left[ \tilde{\boldsymbol{\theta}}^{(1, m)}, \tilde{\boldsymbol{\theta}}^{(2, m)}, \ldots, \tilde{\boldsymbol{\theta}}^{(N, m)}\right]_{p\times N}$.
  \end{enumerate}
\subsection*{Recombination (Blend-$\alpha$ crossover)}
 Select two parents $\tilde{\boldsymbol{\theta}}^{(i^*, m)}$ and $\tilde{\boldsymbol{\theta}}^{(j^*, m)}$ from the mating pool $\boldsymbol{\mathcal{M}}^{(m)}$, to create offspring (with a fixed $\alpha$):
\begin{enumerate}
  \item {Repeat} until \(N\) offspring are created:
  \begin{enumerate}
 For each gene \(r = 1, 2, \ldots, p\):
    \begin{enumerate}
      \item Sample \(u_r \sim \mathcal{U}(0,1)\).
      \item Compute blend weight:
      \[
      v_r = (1 + 2\alpha) u_r - \alpha.
      \]
      \item Generate offspring gene:
      \[
      \hat{\theta}^{(n,m)}_r = v_r \cdot \tilde{\theta}^{(i^\star, m)}_r + (1 - v_r) \cdot \tilde{\theta}^{(j^\star, m-1)}_r.
      \]
    \end{enumerate}
    to create offspring $\boldsymbol{\mathcal{O}}^{(m)} = \left[ \hat{\boldsymbol{\theta}}^{(1, m)}, \hat{\boldsymbol{\theta}}^{(2, m)}, \ldots, \hat{\boldsymbol{\theta}}^{(N, m)}\right]_{p\times N}$.
\end{enumerate}
\end{enumerate}
\subsection*{Mutation (Gaussian)}
Randomly perturb a subset of the newly created offspring in  $\boldsymbol{\mathcal{O}}^{(m)}$:
\begin{enumerate}
  \item Sample a subset of indices $\mathcal{K} \subset \{1, \ldots, N\}$ uniformly at random.
  \item For each $k \in \mathcal{K}$, apply a Gaussian perturbation:
  \[
  \hat{\boldsymbol{\theta}}^{(k,m)} \leftarrow \hat{\boldsymbol{\theta}}^{(k,m)} + \boldsymbol{\varepsilon}, \quad \boldsymbol{\varepsilon} \sim \mathcal{N}(\mathbf{0}, \sigma^2 \mathbf{I}).
  \]
\end{enumerate}
\subsection*{Replacement (Elitisim)}
Preserve the best-performing individuals from generation $m$ and newly created offspring:
\begin{enumerate}
  \item Compute $2N$ fitness values $f_{i, m} = \text{Obj} \left(\boldsymbol{\Theta}^{(m)}, \boldsymbol{\mathcal{O}}^{(m)} \right)$ for $i = 1, \ldots,2N$.
  \item 
  Identify the index set of the top $N$ individuals from  $\{ \boldsymbol{\Theta}^{(m)}, \boldsymbol{\mathcal{O}}^{(m)} \}$:
  \[
    \mathcal{E} = \left\{ i \in \{1, \ldots, 2N\} : f_{i, m} \text{ is among the top } N \text{ values in } \{f_{1,m}, \ldots, f_{2N,m} \} \right\}.
  \]
  
  \item The $(m+1)^{th}$ generation is defined as $\boldsymbol{\Theta}^{(m+1)} = \left[ \boldsymbol{\theta}^{(1, m+1)}, \boldsymbol{\theta}^{(2, m+1)}, \ldots, \boldsymbol{\theta}^{(N, m+1)}\right]_{p\times N}$ where every $\boldsymbol{\theta}^{(n, m+1)}$ is such that $n \in \mathcal{E}$.
\end{enumerate}

\section{Specifications}
\subsection*{A navigation problem} \label{app: nav specs}
\begin{table}[H]
    \centering
    \begin{tabular}{|c|c|}
        \hline
        \textbf{Parameter} & \textbf{Value} \\ \hline
        $R_{inner}$    & 0.25 \\\hline
        $R_{outer}$    & 1  \\ \hline
        $R_{crash}$  & 0.05 \\\hline
        $K$    & 250   \\\hline
        $J$    & 50   \\\hline
        $T$   & 100   \\ \hline
        $\delta$    & 0.01   \\\hline 
        $P_{lower}$   & $2K$   \\ \hline
        $P_{upper}$    & $3K$   \\ \hline
        $sf$    & 1  \\ \hline
        $\omega_0^{\text{Train}}$   & 2024  \\\hline
        $\{\omega_j^{\text{Test}}\}_{j = 1}^{1000}$ & $\{1, 2, \ldots, 1000 \}$ \\\hline
    \end{tabular}
    \caption{Specifications for the navigation problem.}
    \label{table: Drone specs}
\end{table}
\subsection*{The blackjack problems} \label{app: bj specs}
The simulations are conducted under standard S-$17$ blackjack rules, with an eight-deck shoe used for card dealing. The variant considered allows early surrender and double-after-split (DAS), but does not permit re-splitting or surrender after a split. The Basic Strategy rules used in this study follow the standard charts available in \cite{blackjackapprenticeship_strategy_charts}.
\subsection*{Overall specifications} \label{sec: overall specs}
\begin{table}[H]
    \centering
    \begin{tabular}{|c|c|}
        \hline
        \textbf{Parameter} & \textbf{Value} \\ \hline
        \multicolumn{2}{|c|}{\textbf{Neural Network}} \\\hline
        $L$ (\# layers in NN excluding input layer) & 3 \\\hline
        $d_1, d_2$ (\# nodes in each hidden layer) & 3 \\\hline
        $\sigma_1(\cdot), \sigma_2(\cdot)$ (activation functions for both hidden layers) & $\text{tanh}(\cdot)$ \\ \hline
        \multicolumn{2}{|c|}{\textbf{Metropolis-Hastings}} \\\hline
        $\sigma_{\text{Init}}^2$ (unless otherwise stated)  & 1 \\\hline
        $\delta$ & 1000 \\\hline
        $\Delta$ & 100 \\ \hline
        $(s^2)^{(0)}$ & 1 \\ \hline
        $\kappa$ & 0.6 \\ \hline
        \multicolumn{2}{|c|}{\textbf{GA $\&$ RS}} \\\hline
        $M$  & 1000 \\ \hline
        $N$ & 100 \\ \hline
        $\#$ RS iterations & MN \\ \hline
    \end{tabular}
    \caption{Overall specifications of the algorithms used across all three reinforcement learning problems.}
    \label{table: Overall specs}
\end{table}

\section{Random Search}
Define the objective function \(\text{Obj}(\boldsymbol{\theta})\), search space \(\boldsymbol{\theta} \in \Theta \subset \mathbb{R}^p\) and number of samples/iterations \(S\).
  
\begin{enumerate}
  \item Initialize by setting best score \(f^\star \leftarrow -\infty\), best parameter \(\boldsymbol{\theta}^\star \leftarrow \text{null}\).
  \item For  \(s = 1, \ldots, S\):
  \begin{enumerate}
    \item Sample \(\boldsymbol{\theta}^{(s)} \sim \mathcal{U}_p(\Theta)\).
    
    \item Evaluate objective:
    \[
      f_s = \text{Obj}\left( \boldsymbol{\theta}^{(s)} \right).
    \]
    
    \item If \(f_s > f^\star\), then update:
    \[
      f^\star \leftarrow f_s, \quad \boldsymbol{\theta}^\star \leftarrow \boldsymbol{\theta}^{(s)}.
    \]
  \end{enumerate}

  \item Our solution is the best found parameter \(\boldsymbol{\theta}^\star\) with corresponding score \(f^\star\).
\end{enumerate}


\section{The True Count} \label{app: TC}
In card counting (e.g., Hi-Lo system), the running count is the sum of values assigned to seen cards (e.g., +1 for 2-6, 0 for 7-9, -1 for 10-A). The true count adjusts this for the number of decks remaining. We define the running count at time $k$ as $RC_k = \sum_{c \in \boldsymbol{\mathcal{H}}_k} \rho(c)$, where $\rho(c)$ is the count value of card $c$ under the counting system. Assuming a $D_0 = 8$-deck shoe, the true count is defined as $TC_k = \frac{RC_k}{52D_0 - \lvert \boldsymbol{\mathcal{H}}_k\rvert}$. A high true count implies a greater proportion of high cards ($10$s and Aces) remaining in the shoe, which statistically favours the player by increasing the likelihood of blackjack or beating the dealer, and improving the effectiveness of doubling and splitting actions.

\section{3-Class Classification Particle Data} \label{sec: particle data}
Table \ref{tab: particle data} provides an overview of the variables used in the particle classification task. The dataset consists of a two-dimensional spatial cross-section, represented by $x$ and $y$ coordinates, of three distinct sub-atomic particle types, with a total of 360 observations. The dataset is available in the project files, and the corresponding training and test set allocations are defined within the accompanying codebase on \href{https://github.com/LKHJAR001/Optimization-and-Regularization-Under-Arbitrary-Objectives}{\faGithub}.

\begin{table}[H]
    \centering
    \begin{tabular}{cc}
        \hline
        Variable & Description \\
        \hline
        $X1$ & First coordinate in cross-section \\
        $Y2$ & Second coordinate in cross-section\\
        $Yi1$ & Response: $1$ if code-$\alpha$, $0$ otherwise. \\
        $Yi2$ & Response: $1$ if code-$\beta$, $0$ otherwise.  \\
        $Yi3$ & Response: $1$ if code-$\rho$, $0$ otherwise. \\
        \hline
    \end{tabular}
    \caption{Variable descriptions for the particle classification dataset.}
    \label{tab: particle data}
\end{table}

\end{appendices}


\end{document}